\documentclass{article}

 \usepackage[main, final]{neurips_2026}

\usepackage[utf8]{inputenc} %
\usepackage[T1]{fontenc}    %
\usepackage{url}            %
\usepackage{booktabs}       %
\usepackage{amsfonts}       %
\usepackage{nicefrac}       %
\usepackage{microtype}      %
\usepackage[table]{xcolor}
\usepackage{xcolor}         %
\usepackage{amsmath}
\usepackage{amssymb}
\usepackage{mathtools}
\usepackage{amsthm}
\usepackage{tcolorbox}
\usepackage{enumitem} %
\usepackage{fontawesome5} %

\usepackage{lipsum}
\theoremstyle{plain}
\newtheorem{theorem}{Theorem}[section]

\theoremstyle{definition}
\newtheorem{definition}[theorem]{Definition}

\theoremstyle{remark}

\newcommand{\ie}{\textit{i}.\textit{e}.}
\newcommand{\eg}{\textit{e}.\textit{g}.}
\usepackage[textsize=tiny]{todonotes}
\usepackage{bbm}
\usepackage{booktabs}
\usepackage{wrapfig}
\usepackage{subcaption}
\usepackage{graphicx}
\usepackage{colortbl}
\usepackage{multirow}
\usepackage{caption}
\usepackage{float}
\usepackage{caption}
\usepackage{threeparttable}
\usepackage[table]{xcolor}
\definecolor{mygreen}{HTML}{A9D18E}
\definecolor{myyellow}{HTML}{FFC000}
\usepackage{booktabs,multirow,array,xcolor,colortbl}
\usepackage{makecell}
\usepackage{tcolorbox}
\tcbuselibrary{breakable}
\usepackage{subcaption}
\usepackage[colorlinks, linkcolor=mydarkred, citecolor=myblue]{hyperref}
\usepackage{graphicx}
\DeclareGraphicsExtensions{.pdf,.png,.jpg,.jpeg}
\usepackage{array}
\usepackage{booktabs}
\usepackage{caption}
\usepackage{tabularx}
\usepackage{pgffor}
\usepackage{longtable}

\newcommand{\taskblock}[9]{%
\includegraphics[width=\linewidth]{#2} &
\includegraphics[width=\linewidth]{#3} &
\includegraphics[width=\linewidth]{#4} &
\includegraphics[width=\linewidth]{#5} &
\includegraphics[width=\linewidth]{#6} &
\includegraphics[width=\linewidth]{#7} &
\includegraphics[width=\linewidth]{#8} &
\includegraphics[width=\linewidth]{#9} \\
\multicolumn{8}{@{}p{\textwidth}@{}}{\textbf{#1}}\\[-0.2em]
\midrule
}

\usepackage{ifthen}
\newboolean{comments}
\setboolean{comments}{true} %

\ifthenelse{\boolean{comments}}{
  \setlength{\marginparwidth}{1.5cm}
  \newcommand{\zkn}[1]{\todo[color=orange!20, size=\tiny]{Zsolt: #1}}
  \newcommand{\zk}[1]{\textcolor{purple}{Zsolt: #1}}
}{
  \renewcommand{\zk}[1]{}
  \renewcommand{\zkn}[1]{}
}
\ifthenelse{\boolean{comments}}{
  \setlength{\marginparwidth}{1.5cm}
  \newcommand{\wendi}[1]{\todo[color=blue!20, size=\tiny]{Wendi: #1}}
  \newcommand{\wli}[1]{\textcolor{blue}{Wendi: #1}}
}{
  \renewcommand{\wendi}[1]{}
  \renewcommand{\wli}[1]{}
}

\definecolor{mydarkred}{rgb}{0.6,0,0}
\definecolor{myblue}{HTML}{268BD2}

\usepackage[capitalize,noabbrev]{cleveref}
\definecolor{mydarkblue}{rgb}{0,0.08,0.45}
\definecolor{thresholdgreen}{HTML}{7BC86C}
\definecolor{mygreen}{HTML}{658354}
\definecolor{blue}{HTML}{c9e9f6}   %
\definecolor{pink}{HTML}{F77189}   %
\definecolor{venuegray}{RGB}{120,120,120}
\newcommand{\venue}[1]{\textit{\textcolor{venuegray}{#1}}}

\title{Hide-and-Seek in Trajectories: Discovering Failure Signals for VLA Runtime Monitoring}

\author{%
  Seongheon Park\textsuperscript{1}\quad
  Wendi Li\textsuperscript{1}\quad
  Changdae Oh\textsuperscript{1}\quad
  Samuel Yeh\textsuperscript{1}\\[0.2em]
  \textbf{Zsolt Kira\textsuperscript{\textbf{2}}}\quad
  \textbf{Michael Hagenow\textsuperscript{\textbf{1}}}\quad
  \textbf{Sharon Li\textsuperscript{\textbf{1}}}\\[0.2em]
  \textsuperscript{1}University of Wisconsin--Madison \quad
  \textsuperscript{2}Georgia Institute of Technology \\ [0.2em]
  \texttt{\{seongheon\_park,sharonli\}@cs.wisc.edu}.
}

\begin{document}
\addtocontents{toc}{\protect\setcounter{tocdepth}{-1}}

\maketitle

\begin{abstract}
Vision-Language-Action (VLA) models enable robots to follow natural language instructions and generalize across diverse tasks, but they remain vulnerable to execution failures that compromise reliability in real-world deployment.
Detecting such failures during execution is therefore critical for the robust deployment of embodied systems.
Existing failure detection methods either rely on expensive action resampling or external models, while alternatives propagate trajectory-level labels uniformly across every timestep, obscuring localized failure signals.
In this paper, we propose \textbf{Hide-and-Seek}, a framework that formulates VLA failure detection as a coarsely supervised learning problem.
By combining inter-trajectory and intra-trajectory contrastive objectives, Hide-and-Seek localizes failure-indicative actions and induces temporally structured failure signals from trajectory-level supervision alone, without any step-level annotation.
We evaluate Hide-and-Seek on LIBERO, VLABench, and a real-world robotic platform across three representative VLA policies: OpenVLA, $\pi_0$, and $\pi_{0.5}$.
Our method achieves state-of-the-art multi-task failure detection performance with a practical accuracy--timeliness trade-off under conformal prediction, and generalizes well to both seen and unseen tasks. 
Code and videos are available at our \href{https://seongheon-96.github.io/hide_and_seek_site/}{\faRobot\ Project Page}.
\end{abstract}

\section{Introduction}

Vision-language-action (VLA) models~\cite{kim2024openvla,black2024pi_0,intelligence2025pi,bjorck2025gr00t, wu2026vlanext, zhang2026vlm4vla, torne2026mem, sridhar2025memer} enable robot policies to follow natural language instructions and learn from heterogeneous, multimodal demonstrations, showing promising generalization to novel environments and embodiments~\cite{xu2025anatomy}.
Despite these advances, VLAs remain susceptible to diverse execution failures
that undermine reliability
and increase the cost of real-world deployment~\cite{visinsky1994robotic, rahman2021run}.
Consider a household robot tasked with placing dishes into a cabinet: a subtle grasp failure, if undetected, can cascade into a dropped dish or costly recovery.
Detecting such failures during execution is therefore critical for the robust deployment of embodied systems~\cite{xu2025can, ali2025eve, ramrakhya2025grounding}.

Existing failure detection methods fall short along two axes that jointly define the challenge. The first is \emph{supervision cost}: obtaining step-level failure annotation is expensive and difficult to scale, requiring experts to identify the precise timestamps of error across long-horizon, stochastic trajectories.
Recent work SAFE~\cite{gu2025safe} sidesteps this issue by uniformly assigning trajectory-level failure labels to all timesteps,
 but this mislabels normal actions before failure onset, introducing substantial label noise that limits the efficacy of the learned detector. 
The second challenge is \emph{computational practicality}: 
approaches based on action resampling~\cite{romer2025failure,agia2024unpacking,jang2025verifier} or external VLM judges~\cite{yuan2026act,duan2024aha,qi2026self} incur substantial inference overhead that precludes real-time deployment.
Most prior methods are additionally designed for specialist policies on fixed tasks~\cite{xu2025can, farid2022failure,liu2024model,seo2025uncertainty}, limiting their applicability to the diverse, open-world settings that modern generalist VLAs encounter.
These limitations motivate a lightweight failure detector that localizes failure signals without step-level annotation.

\begin{figure*}[t!]
\centering
\includegraphics[width=\linewidth]{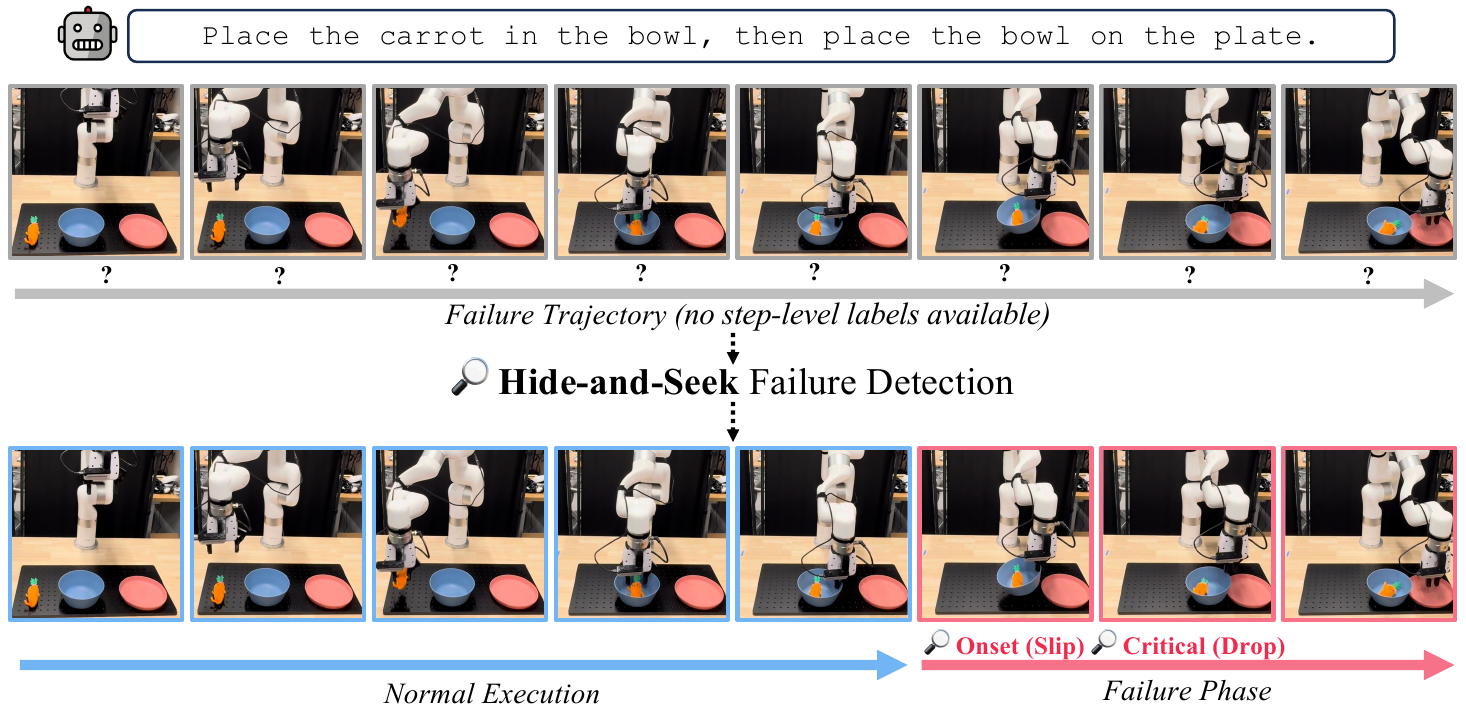}
\caption{
    \textbf{Hide-and-Seek Failure Detection.}
    Failure trajectories contain a substantial amount of normal actions before failure onset, yet only a trajectory-level label is available during training, leaving the temporal structure entirely unknown (top).
    From this coarse supervision, 
    Hide-and-Seek discovers the most failure-indicative actions (\eg, the failure onset and subsequent critical event) by contrasting scores across and within trajectories, naturally separating normal execution from the failure phase without any step-level annotation (bottom).
}
\label{fig:teaser}
\vspace{-0.5cm}
\end{figure*}

To address this challenge, we propose \textbf{{H}ide-and-{S}eek}, a framework that discovers failure-indicative actions from trajectory-level supervision alone (\Cref{fig:teaser}). 
We name our method to reflect the following metaphor: the failure signal is {hidden} among largely normal behavior within a trajectory, and the detector must {seek} it out using only the trajectory-level outcome as a clue. 
Rather than propagating the trajectory label uniformly to all time steps, Hide-and-Seek contrasts failure-indicative actions against non-failure actions at two complementary granularities.
First, we propose an \emph{inter-trajectory contrastive loss}, which enforces that the most failure-indicative step in a failure trajectory scores higher than the most failure-resembling step in a success trajectory. This guides the model to focus on discriminating the most salient failure signals. 
Moreover, we introduce an \emph{intra-trajectory contrastive loss} that shapes score dynamics within each failure trajectory by encouraging the average failure score to be higher after the failure begins than before, converting trajectory-level supervision into a temporally structured failure signal without requiring any temporal annotation.
Overall, the combination of the two losses imposes stronger structural regularity than uniform label propagation while adaptively discovering where the failure hides.

We evaluate Hide-and-Seek on two simulation benchmarks, LIBERO~\cite{liu2023libero} and VLABench~\cite{zhang2025vlabench}, as well as on a real-world robotic platform, using three representative VLA policies spanning both autoregressive and flow-matching-based paradigms: OpenVLA~\cite{kim2024openvla}, $\pi_0$~\cite{black2024pi_0}, and $\pi_{0.5}$~\cite{intelligence2025pi_}. 
Hide-and-Seek consistently outperforms all baselines on both seen and 
unseen tasks, 
surpassing the strongest classifier-based baseline by up to $+\textbf{11.7}\%$ bACC
while maintaining a practical accuracy--timeliness trade-off.
Notably, our method raises alarms near the annotated failure onset without any temporal supervision, and outperforms a VLM-based runtime monitor by $+\textbf{13.1}\%$ in accuracy while operating at over $\textbf{2{,}000}\times$ higher speed.

Our key contributions are summarized as follows:
\begin{itemize}

\item We propose {Hide-and-Seek}, a lightweight runtime failure detection framework that draws a novel connection between coarsely supervised learning and embodied failure detection, discovering failure-indicative actions from trajectory-level supervision alone.
    
    \item We design a contrastive objective that distinguishes failure-indicative actions from non-failure actions both across trajectories and within a trajectory, converting coarse trajectory-level supervision into a temporally structured failure signal.

    \item We comprehensively evaluate Hide-and-Seek on LIBERO and VLABench in simulation and on a real-world platform with OpenVLA, $\pi_0$, and $\pi_{0.5}$, achieving state-of-the-art performance on both seen and unseen tasks with a practical accuracy--timeliness trade-off.

\end{itemize}

\section{Related Works}

\paragraph{Vision-Language-Action models}predict actions conditioned on multimodal input such as visual observations, robot proprioceptive states, and natural language instructions, and fall into two broad paradigms.
The first paradigm treats action generation as an autoregressive sequence modeling problem~\cite{kim2024openvla,zheng2025tracevla,qu2025spatialvla,cen2025rynnvla,pertsch2025fast,liu2026faster,wang2025unified}, producing actions token-by-token in a manner analogous to language modeling~\cite{radford2019language,liu2023visual}.
The second paradigm instead frames policy learning as a generative process over continuous action spaces, leveraging diffusion or flow-matching to produce actions through iterative denoising or transport~\cite{black2024pi_0,intelligence2025pi_,intelligence2025pi,bjorck2025gr00t,barreiros2025careful,ye2026world, torne2026mem}.
Our method is architecture-agnostic and compatible with foundational VLA policies spanning both paradigms: the autoregressive model OpenVLA~\cite{kim2024openvla} and the flow-matching-based models $\pi_0$~\cite{black2024pi_0} and $\pi_{0.5}$~\cite{intelligence2025pi_}.
This ensures the generality and broad applicability of our approach.%

\vspace{-0.2cm}
\paragraph{Failure detection in robot manipulation} is critical for real-world deployment, as even minor errors can cascade into catastrophic outcomes~\cite{visinsky1994robotic,rahman2021run,antonante2023monitoring}. Existing approaches fall into four categories:
(1) {OOD detection-based methods}~\cite{liu2024multi,xu2025can,zhou2026rc,sinha2024real,he2024rediffuser} treat successful executions as in-distribution and flag deviations as failures, yet struggle to generalize to unseen distributions.
(2) {VLM-based methods}~\cite{yuan2026act,duan2024aha,qi2026self,agia2024unpacking} leverage vision-language models to identify and reason about failures from visual observations; however, they are limited by high inference latency and tend to detect failures only after they have fully manifested.
(3) {Multi-sampling-based methods}~\cite{romer2025failure,agia2024unpacking,jang2025verifier} quantify uncertainty per action via ensemble sampling, but incur prohibitive inference costs that limit practical deployment.
(4) {Classifier-based methods}~\cite{barreiros2025careful, seo2025uncertainty,gu2025safe, liu2024model,farid2022failure} train failure classifiers either by collecting costly step-level annotations that are difficult to scale, or by propagating the trajectory label to every timestep, which introduces substantial label noise. %
This motivates a new classifier-based method that robustly learns a failure detector from trajectory-level labels alone.

\vspace{-0.2cm}
\paragraph{Learning from coarse supervision} 
is a label-efficient paradigm that leverages coarse-grained annotations to localize fine-grained signals~\cite{jiang2023weakly,zhang2021weakly}. Often framed as weakly supervised learning, it has been successfully applied to object detection~\cite{tang2017multiple,wan2019c}, video temporal action localization~\cite{lee2021weakly,tang2023ddg}, and anomaly detection~\cite{sultani2018real,pang2023deep}.
For example, multiple instance learning~\cite{carbonneau2018multiple} assigns a single label to a bag of instances and learns to identify the responsible instances from bag-level supervision alone.
Despite the broad success across these domains, its connection to failure detection in embodied agents has not been explored.
We draw this novel connection, 
enabling localized failure signal discovery from coarse trajectory-level supervision without per-step annotations.

\section{Problem Setup}

At each timestep $t$, a VLA model receives an observation $o_t$, consisting of RGB images, a natural language instruction, and the current robot state, and outputs an action $a_t$.
We denote by $h_t \in \mathbb{R}^{d}$ the internal action embedding at time $t$, extracted from the model’s action token or action head.
To train and evaluate failure detectors, we execute the VLA across a set of tasks in simulation or the real world and collect rollout trajectories of embeddings, denoted as $\tau = (h_1, h_2, \dots, h_T)$, where $T$ is the episode horizon, and we write $\tau_{\le t} = (h_1, h_2, \dots, h_t)$ for the prefix up to timestep $t$.
Each trajectory is labeled with a trajectory-level outcome $y(\tau) \in \{0,1\}$, where $y(\tau)=1$ indicates failure and $y(\tau)=0$ indicates success.
Crucially, no step-level annotations are available, since obtaining them is costly and difficult to scale, particularly for long-horizon stochastic policies. We therefore address the \textit{coarsely supervised failure detection} problem, a more practical alternative to the fully supervised version.

\vspace{0.2cm}
\begin{definition}[\textbf{Coarsely Supervised Failure Detection}]
\emph{Given a set of successful trajectories $\mathcal{D}_s = \{\tau^{(k)} \mid y(\tau^{(k)}) = 0\}$ and a set of failure trajectories $\mathcal{D}_f = \{\tau^{(k)} \mid y(\tau^{(k)}) = 1\}$, where $y(\tau) \in \{0, 1\}$, the goal is to learn a fine-grained failure detector that discovers failure-indicative steps. At runtime, given a trajectory prefix $\tau_{\le t} = (h_1, h_2, \dots, h_t)$,
the detector produces a binary decision as follows:}
\begin{equation}
\label{eq:monitor}
G(\tau_{\leq t}) = \begin{cases} 1, & \text{if the trajectory prefix } \tau_{\leq t} \text{ is predicted to result in task failure,} \\ 0, & \text{otherwise.} \end{cases}
\end{equation}
\end{definition}
The central challenge is the mismatch between the {granularity of supervision} (trajectory-level) and the {granularity of detection} (a per-step decision). Our framework bridges this gap.

\section{Hide-and-Seek Failure Detection}
\label{sec:main_method}

\paragraph{Overview.}

In this section, we introduce our framework \textbf{Hide-and-Seek} for failure detection.
The name reflects the core challenge: 
the failure-inducing action is usually \textit{hidden} among largely normal behavior, and the detector must \textit{seek} it out from only trajectory-level supervision as a clue.
For instance, in a task of placing a carrot in a bowl, the robot may successfully grasp the carrot, yet fail only at the placement stage. 
Naively assigning the trajectory-level failure label to every timestep treats all actions as equally failure-indicative, 
mislabeling many normal actions as failures, thereby introducing substantial label noise (see \cref{sec:latent_vis}).

\begin{figure*}[t!]
\centering
\includegraphics[width=\linewidth]{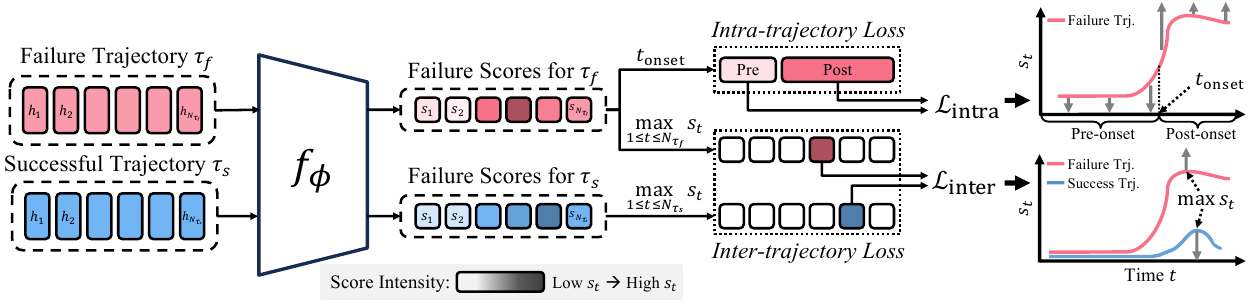}

\caption{
    \textbf{Overall framework.}
    Given a failure-success trajectory pair $(\tau_f, \tau_s)$,
    the detector $f_\phi$  produces per-step failure scores $s_t$ for both failure and successful trajectories.
$\mathcal{L}_{\mathrm{inter}}$ (Eq.~\ref{eq:inter}) enforces that the most failure-indicative action in the failure trajectory $\tau_f$ ranks higher than the hardest false-positive in the successful trajectory $\tau_s$.
$\mathcal{L}_{\mathrm{intra}}$ (Eq.~\ref{eq:onset}) defines a proxy failure onset as $t_{\mathrm{onset}} = \arg\max_t (s_t - s_{t-1})$, and enforces temporal separation within $\tau_f$ by encouraging average post-onset scores to exceed average pre-onset scores.
    The two losses jointly shape the score dynamics so that $s_t$ stays low during normal execution while sharply rising at $t_{\mathrm{onset}}$ in failure trajectories.
}
\vspace{-0.5cm}
\label{fig:2}
\end{figure*}

Rather than propagating the trajectory label uniformly, our method contrasts failure-indicative actions against non-failure actions at two complementary granularities: (1) an \emph{inter-trajectory contrastive loss} which contrasts the most salient failure-indicative action between failure and successful trajectories, 
(2) an \emph{intra-trajectory contrastive loss} that encourages temporal separation in failure scores between pre-onset (normal execution) and post-onset (failure phase) within a failure trajectory, without requiring step-level annotations.
The overall framework is illustrated in~\cref{fig:2}.

\subsection{ Training Objective}
\label{sec:hsfd}

We learn a sequential detector $f_\phi$ with a sigmoid output that maps each trajectory prefix to a scalar failure score\footnote{In practice, $f_\phi$ can be instantiated as a lightweight sequence model, such as an LSTM, GRU, or Transformer. We use a single-layer LSTM as the default choice and provide additional ablations in~\cref{tab:architecture,sec:arch_abl}.}.
Given a trajectory prefix $\tau_{\leq t} = (h_1, h_2, \dots, h_t)$ of VLA action embeddings, the detector produces $s_t = f_\phi(\tau_{\leq t}) \in [0,1]$. To optimize $f_\phi$, we propose the following losses.

\paragraph{Inter-trajectory contrastive loss.}

Rather than classifying every step independently, we shift our focus to the most failure-indicative action step.
The formulation of our loss rests on the key idea: for a failure trajectory $\tau_f \in \mathcal{D}_f$, at least one step carries a genuinely elevated failure score, and this score should be higher than the most failure-resembling action (the hardest false-positive, \emph{e.g.}, a momentary hesitation or awkward 
intermediate pose that the robot successfully recovers from)
in a successful trajectory $\tau_s \in \mathcal{D}_s$. This can be enforced with a margin-based contrastive loss:
\begin{equation}
\label{eq:inter}
  \mathcal{L}_{\mathrm{inter}}
  =
  \frac{1}{|\mathcal{D}_f|\,|\mathcal{D}_s|}
  \sum_{\tau_f \in \mathcal{D}_f}
  \sum_{\tau_s \in \mathcal{D}_s}
  \Big[\max\!\Big(0,\;
    m_r
    - \max_{1 \leq t \leq N_{\tau_f}} s_t
    + \max_{1 \leq t \leq N_{\tau_s}} s_t
  \Big)\Big],
\end{equation}
where $N_{\tau_f}$ and $N_{\tau_s}$ are the lengths of $\tau_f$ and $\tau_s$ respectively, and $m_r > 0$ is a margin hyperparameter.
The loss is computed over failure-success trajectory pairs $(\tau_f, \tau_s)$ within each mini-batch. 
This objective makes no assumption about \textit{where} the failure-indicative action occurs;
instead, it adaptively discovers the most salient failure signal using a coarser form of supervision. At the same time, it enforces stronger structural regularity compared to propagating the trajectory label uniformly. 

\paragraph{Intra-trajectory contrastive loss.}
Training with $\mathcal{L}_{\mathrm{inter}}$ alone optimizes only 
a single peak per trajectory, leaving the surrounding 
temporal structure unsupervised. 
Since successful task completion becomes less likely after failure onset~\citep{ross2011reduction}, we introduce a second contrastive objective that enforces temporal separation around a proxy failure onset 
$t_{\mathrm{onset}} = \arg\max_t (s_t - s_{t-1})$, defined as the point of the sharpest increase in failure score. 
Within each failure trajectory $\tau_f$, we encourage the average post-onset score to exceed the average pre-onset score:
\begin{equation}
\label{eq:onset}
    \mathcal{L}_{\mathrm{intra}}
    =\frac{1}{|\mathcal{D}_f|}
\sum_{\tau_f \in \mathcal{D}_f}
    \left[
        \max\!\left(0,\;
        m_o
        - \frac{1}{N_{\tau_f} - t_{\mathrm{onset}}+1}
          \sum_{t \geq t_{\mathrm{onset}}} s_t
        + \frac{1}{t_{\mathrm{onset}}-1}
          \sum_{t < t_{\mathrm{onset}}} s_t
        \right)
    \right],
\end{equation}
where $m_o > 0$ is a margin hyperparameter.
In~\cref{sec:onset_app}, we show that $t_{\mathrm{onset}}$ closely approximates the annotated failure onset and achieves comparable performance when used as the oracle reference, confirming that the sharpest score transition aligns with failure onset.

\paragraph{Overall objective.}
The complete training objective is a weighted combination of the inter-trajectory 
contrastive loss and the intra-trajectory contrastive loss:

\begin{equation}
    \mathcal{L}
    = \mathcal{L}_{\mathrm{inter}}
    + \lambda\, \mathcal{L}_{\mathrm{intra}},
    \label{eq:full}
\end{equation}

where $\lambda$ controls the loss weight.
The objective overall aligns with the hide-and-seek intuition: $\mathcal{L}_{\mathrm{inter}}$ seeks the hidden failure signal across trajectories, while $\mathcal{L}_{\mathrm{intra}}$ sharpens the temporal boundary between normal execution and failure.
Our method thus converts coarse trajectory-level supervision into a temporally structured failure signal without per-step annotation.

\subsection{Runtime Monitoring via Conformal Prediction}
\label{sec:monitor}
After training, we deploy the failure detector to monitor VLA execution at inference time.
At each timestep $t$, the detector produces a failure score $s_t = f_\phi(\tau_{\leq t}) \in [0,1]$ and raises a failure alarm whenever $s_t$ exceeds a threshold $\zeta_t$. 
This realizes the binary decision $G(\tau_{\le t})$ as a score-threshold rule:
\begin{equation}
\label{equ:failure_detector}
G(\tau_{\le t}) =
\begin{cases}
1, & \text{if } s_t \ge \zeta_t \quad \text{(failure declared)}, \\
0, & \text{otherwise}.
\end{cases}
\end{equation}

A key challenge in deployment is selecting $\zeta_t$ appropriately: a fixed threshold fails to account for the natural temporal evolution of failure scores, while a poorly calibrated threshold leads to either excessive false alarms or missed detections.
To address this, we adopt functional conformal prediction (CP)~\cite{diquigiovanni2021importance}, which constructs a time-varying prediction band from a held-out calibration set of successful trajectories.
Under the exchangeability assumption and a user-specified significance level $\alpha \in (0, 1)$, the band guarantees that for any new successful rollout, the trajectory-level false alarm rate is bounded by $\alpha$.
We employ the one-sided CP formulation~\cite{xu2025can,gu2025safe} where the threshold at each timestep is given by $\zeta_t = \mu_t + b_t$, with a time-varying mean $\mu_t$ and bandwidth $b_t$ calibrated on the empirical score distribution of $C$ successful calibration trajectories.
During deployment, a failure is declared at the earliest timestep $t$ for which $s_t \ge \zeta_t$, which we denote as the detection time.

\vspace{-0.2cm}
\section{Experiments}
\label{sec:experiments}
\subsection{Setup}
\label{sec:setup}

\vspace{-0.2cm}
\paragraph{Implementation details.}
\label{sec:tfa}

We extract per-timestep action embeddings from the action token or action head of the VLA.
For autoregressive policies such as OpenVLA, we average the embeddings across layers and degrees of freedom to obtain a compact representation $h_t \in \mathbb{R}^d$.
For flow-matching-based policies (e.g., $\pi_0$ and $\pi_{0.5}$), we extract hidden states before the velocity prediction head at the final denoising step, and align action chunks to obtain comparable per-timestep representations.
To mitigate temporal redundancy~\cite{zhao2023learning}, we apply non-overlapping sliding window aggregation with window size $w$, using average pooling within each window to produce $N_\tau = \lceil T/w \rceil$ window-level embeddings. 
Further implementation details are provided in~\cref{sec:implementation}.

\vspace{-0.2cm}
\paragraph{Benchmarks and models.}
We evaluate \textit{multi-task} failure detection on two simulation benchmarks, LIBERO-10~\cite{liu2023libero} and VLABench~\cite{zhang2025vlabench}, and a real-world UFactory xArm 6 platform, using representative VLA policies including OpenVLA~\cite{kim2024openvla}, $\pi_0$~\cite{black2024pi_0}, and $\pi_{0.5}$~\cite{intelligence2025pi_}.
We adopt a held-out task protocol, reserving a subset of tasks as unseen to evaluate generalization.
For all experiments, we report results averaged over 3 random seeds.
Details on benchmarks and task splits are provided in~\cref{sec:benchmark}.

\paragraph{Evaluation metrics.}

We evaluate runtime failure detection using three complementary metrics, treating failure trajectories as positives and successful trajectories as negatives.
(1) \textbf{Balanced Accuracy (bACC)} equally weights both classes, and serves as our primary class-balanced metric.
(2) \textbf{Weighted Accuracy (wACC)} weights each class by its empirical frequency, reflecting deployment scenarios with class imbalance.
(3) \textbf{Time-weighted Accuracy (TWA)}~\cite{romer2025failure} jointly captures correctness and timeliness by penalizing late detections based on their detection time.
All three metrics depend on the conformal threshold, set by a parameter $\alpha$ that controls the false alarm rate on successful trajectories.
We report results averaged over $\alpha \in \{0.15, 0.20, 0.25\}$ across all experiments.
Full definitions are provided in~\cref{sec:metrics}.

\vspace{-0.2cm}
\paragraph{Baselines.}
We comprehensively compare our approach against 12 action-based failure detection 
baselines:
\textbf{(1) OOD detection-based methods} model the in-distribution embedding space, extended with failure embeddings to match our pairwise supervision: Mahalanobis~\cite{lee2018simple}, Cosine $k$-NN~\cite{sun2022out}, PCA-KMeans~\cite{liu2024multi}, RND~\cite{he2024rediffuser}, and LogpZO~\cite{xu2025can}.
(2) \textbf{Multi-sampling-based methods}\footnote{We use $N=10$ samples per step to balance estimation 
quality and the sampling cost of large VLA models.} assess uncertainty through repeated action sampling: Cluster Entropy~\cite{kuhn2023semantic}, EigenScore~\cite{chen2024inside}, STAC~\cite{agia2024unpacking}, and ACE~\cite{romer2025failure}.
(3) \textbf{Classifier-based methods} train a failure classifier on frozen VLA representations: SAFE~\cite{gu2025safe}.
For OpenVLA, we additionally evaluate (4) \textbf{Token uncertainty-based methods}, which estimate failure likelihood from per-action token probabilities: Entropy~\cite{malinin2020uncertainty}, Negative log-likelihood (NLL)~\cite{ren2022out}.
Details about each baseline are provided in~\cref{sec:baselines}.

\subsection{Main Experiments}

\begin{table}[t]
\centering
\small
\caption{Comparison of failure detection performance across policies on LIBERO-10. We report balanced accuracy (\textbf{bACC}), weighted accuracy (\textbf{wACC}), and time-weighted accuracy (\textbf{TWA}), averaged over 3 random seeds with standard deviation. \textbf{Bold} and \underline{underline} denote the best and second-best results, respectively.}
\label{tab:vertical_models}
\setlength{\tabcolsep}{6pt}
\renewcommand{\arraystretch}{1.05}
\resizebox{\columnwidth}{!}{%
\begin{tabular}{@{} >{\centering\arraybackslash}p{10mm} l p{16mm}  c c c c c c @{}}
\toprule
\multicolumn{9}{c}{\textbf{Policy: OpenVLA (Success Rate: 51.0\%)}}\\
\midrule

& & &
\multicolumn{3}{c}{\textbf{Seen Tasks}} &
\multicolumn{3}{c}{\textbf{Unseen Tasks}} \\
\cmidrule(lr){4-6}\cmidrule(lr){7-9}

& \textbf{Method} &  &
\textbf{bACC}$\uparrow$ & \textbf{wACC}$\uparrow$ & \textbf{TWA}$\uparrow$ &
\textbf{bACC}$\uparrow$ & \textbf{wACC}$\uparrow$ & \textbf{TWA}$\uparrow$ \\
\midrule

\multirow{4}{*}{\rotatebox{90}{\textbf{Token Unc.}}} &
Max Entropy & \venue{ICLR'21} &
$0.514^{\pm 0.065}$ & $0.527^{\pm 0.063}$ & $0.455^{\pm 0.030}$ &
$0.521^{\pm 0.031}$ & $0.518^{\pm 0.072}$ & $0.488^{\pm 0.033}$ \\
& Avg. Entropy & \venue{ICLR'21} &
$0.513^{\pm 0.020}$ & $0.524^{\pm 0.038}$ & $0.478^{\pm 0.035}$ &
$0.491^{\pm 0.059}$ & $0.490^{\pm 0.049}$ & $0.456^{\pm 0.032}$ \\
& Max NLL & \venue{ICLR'23} &
$0.494^{\pm 0.054}$ & $0.493^{\pm 0.039}$ & $0.456^{\pm 0.050}$ &
$0.525^{\pm 0.033}$ & $0.528^{\pm 0.057}$ & $0.488^{\pm 0.022}$ \\
& Avg. NLL & \venue{ICLR'23} &
$0.506^{\pm 0.039}$ & $0.522^{\pm 0.028}$ & $0.475^{\pm 0.024}$ &
$0.495^{\pm 0.027}$ & $0.487^{\pm 0.045}$ & $0.470^{\pm 0.022}$ \\
\midrule

\multirow{5}{*}{\rotatebox{90}{\textbf{OOD Det.}}} &
Mahalanobis & \venue{ICML'18} &
$0.670^{\pm 0.038}$ & $0.667^{\pm 0.033}$ & $0.604^{\pm 0.039}$ &

$0.513^{\pm 0.064}$ & $0.495^{\pm 0.112}$ & $0.504^{\pm 0.044}$ \\
& Cosine k-NN & \venue{ICML'22} &
$0.758^{\pm 0.060}$ & $0.757^{\pm 0.057}$ & $0.632^{\pm 0.048}$ &
$0.639^{\pm 0.080}$ & $0.590^{\pm 0.130}$ & $0.522^{\pm 0.056}$ \\
& PCA KMeans & \venue{CoRL'24} &
$0.534^{\pm 0.031}$ & $0.544^{\pm 0.018}$ & $0.494^{\pm 0.024}$ &
$0.486^{\pm 0.019}$ & $0.494^{\pm 0.097}$ & $0.442^{\pm 0.045}$ \\
& RND & \venue{ICML'24} &
$0.526^{\pm 0.039}$ & $0.523^{\pm 0.034}$ & $0.493^{\pm 0.036}$ &
$0.509^{\pm 0.051}$ & $0.504^{\pm 0.078}$ & $0.478^{\pm 0.040}$ \\
& LogpZO & \venue{RSS'25} &
$0.658^{\pm 0.091}$ & $0.665^{\pm 0.089}$ & $0.594^{\pm 0.067}$ &
$0.545^{\pm 0.028}$ & $0.502^{\pm 0.049}$ & $0.505^{\pm 0.030}$ \\
\midrule

\multirow{4}{*}{\rotatebox{90}{\textbf{Multi.}}} &
Cluster Entropy & \venue{ICLR'23} &
$0.552^{\pm 0.030}$ & $0.524^{\pm 0.012}$ & $0.516^{\pm 0.023}$ &
$0.570^{\pm 0.064}$ & $0.578^{\pm 0.104}$ & $0.548^{\pm 0.034}$ \\
& EigenScore & \venue{ICLR'24} &
$0.663^{\pm 0.034}$ & $0.636^{\pm 0.039}$ & $0.536^{\pm 0.032}$ &
$0.671^{\pm 0.044}$ & $0.672^{\pm 0.035}$ & $0.524^{\pm 0.029}$ \\
& STAC$^{\dagger}$ & \venue{CoRL'24} &
$0.665^{\pm 0.057}$ & $0.651^{\pm 0.073}$ & $0.545^{\pm 0.051}$ &
$0.624^{\pm 0.062}$ & $0.642^{\pm 0.089}$ & $0.528^{\pm 0.042}$ \\
& ACE & \venue{NeurIPS'25} &
$0.607^{\pm 0.071}$ & $0.631^{\pm 0.083}$ & $0.516^{\pm 0.041}$ &
$0.604^{\pm 0.087}$ & $0.602^{\pm 0.072}$ & $0.542^{\pm 0.033}$ \\
\midrule

\multirow{3}{*}{\rotatebox{90}{\textbf{Class.}}} &
SAFE-LSTM & \venue{NeurIPS'25} &
$0.670^{\pm 0.100}$ & $0.665^{\pm 0.104}$ & $0.593^{\pm 0.039}$ & $0.686^{\pm 0.133}$ & $0.666^{\pm 0.152}$ & $0.553^{\pm 0.108}$ \\
& SAFE-MLP & \venue{NeurIPS'25} &
$\underline{0.823}^{\pm 0.036}$ & $\underline{0.819}^{\pm 0.034}$ & $\underline{0.630}^{\pm 0.056}$ & $\underline{0.775}^{\pm 0.062}$ & $\underline{0.741}^{\pm 0.082}$ & $\underline{0.590}^{\pm 0.039}$ \\
& \cellcolor{blue}\textbf{Ours} & \cellcolor{blue} &
\cellcolor{blue} $ \mathbf{0.852}^{\pm 0.051}$ & \cellcolor{blue} $ \mathbf{0.853}^{\pm 0.052}$ & \cellcolor{blue} $ \mathbf{0.660}^{\pm 0.035}$ & \cellcolor{blue} $ \mathbf{0.834}^{\pm 0.036}$ & \cellcolor{blue} $ \mathbf{0.828}^{\pm 0.034}$ & \cellcolor{blue} $ \mathbf{0.663}^{\pm 0.010}$ \\
\midrule\midrule

\multicolumn{9}{c}{\textbf{Policy: $\mathbf{\pi}_{\mathbf{0}}$ (Success Rate: 84.2\%)}}\\
\midrule

& & &
\multicolumn{3}{c}{\textbf{Seen Tasks}} &
\multicolumn{3}{c}{\textbf{Unseen Tasks}} \\
\cmidrule(lr){4-6}\cmidrule(lr){7-9}

& \textbf{Method} &  &
\textbf{bACC}$\uparrow$ & \textbf{wACC}$\uparrow$ & \textbf{TWA}$\uparrow$ &
\textbf{bACC}$\uparrow$ & \textbf{wACC}$\uparrow$ & \textbf{TWA}$\uparrow$ \\
\midrule

\multirow{5}{*}{\rotatebox{90}{\textbf{OOD Det.}}} &
Mahalanobis & \venue{ICML'18} &
$0.788^{\pm 0.043}$ & $0.840^{\pm 0.083}$ & $\mathbf{0.715}^{\pm 0.043}$ &
$0.634^{\pm 0.054}$ & $0.510^{\pm 0.057}$ & $0.543^{\pm 0.037}$ \\
& Cosine k-NN & \venue{ICML'22} &
$0.580^{\pm 0.040}$ & $0.841^{\pm 0.031}$ & $0.565^{\pm 0.032}$ &
$0.623^{\pm 0.035}$ & $0.696^{\pm 0.100}$ & $0.542^{\pm 0.025}$ \\
& RND & \venue{ICML'24} &
$0.706^{\pm 0.067}$ & $0.784^{\pm 0.131}$ & $0.629^{\pm 0.069}$ &
$0.520^{\pm 0.031}$ & $0.643^{\pm 0.052}$ & $0.505^{\pm 0.021}$ \\
& PCA KMeans & \venue{CoRL'24} &
$0.591^{\pm 0.140}$ & $0.595^{\pm 0.234}$ & $0.476^{\pm 0.048}$ &
$0.461^{\pm 0.093}$ & $0.471^{\pm 0.122}$ & $0.394^{\pm 0.089}$ \\
& LogpZO & \venue{RSS'25} & $0.786^{\pm 0.040}$ & $0.835^{\pm 0.078}$ & $0.691^{\pm 0.035}$ & $0.681^{\pm 0.050}$ & $0.679^{\pm 0.136}$ & $0.574^{\pm 0.028}$ \\
\midrule

\multirow{4}{*}{\rotatebox{90}{\textbf{Multi.}}} &
Cluster Entropy & \venue{ICLR'23} &
$0.548^{\pm 0.096}$ & $0.749^{\pm 0.163}$ & $0.454^{\pm 0.047}$ &
$0.732^{\pm 0.158}$ & $0.604^{\pm 0.201}$ & $0.552^{\pm 0.088}$ \\
& EigenScore & \venue{ICLR'24} &
$0.691^{\pm 0.057}$ & $0.761^{\pm 0.037}$ & $0.534^{\pm 0.030}$ &
$0.613^{\pm 0.143}$ & $0.668^{\pm 0.100}$ & $0.495^{\pm 0.071}$ \\
& STAC & \venue{CoRL'24} &
$0.619^{\pm 0.068}$ & $0.751^{\pm 0.057}$ & $0.501^{\pm 0.044}$ &
$0.490^{\pm 0.044}$ & $0.596^{\pm 0.093}$ & $0.440^{\pm 0.033}$ \\
& ACE & \venue{NeurIPS'25} &
$0.672^{\pm 0.074}$ & $0.764^{\pm 0.099}$ & $0.572^{\pm 0.049}$ &
$0.609^{\pm 0.066}$ & $0.630^{\pm 0.063}$ & $0.505^{\pm 0.029}$ \\
\midrule

\multirow{3}{*}{\rotatebox{90}{\textbf{Class.}}} &
SAFE-LSTM & \venue{NeurIPS'25} &
$0.774^{\pm 0.024}$ & $0.851^{\pm 0.041}$ & $0.663^{\pm 0.047}$ &
$0.609^{\pm 0.057}$ & $0.528^{\pm 0.176}$ & $0.524^{\pm 0.080}$ \\
& SAFE-MLP & \venue{NeurIPS'25} &
$\underline{0.870}^{\pm 0.038}$ & $\underline{0.915}^{\pm 0.047}$ & $0.688^{\pm 0.060}$ &
$\underline{0.801}^{\pm 0.083}$ & $\underline{0.867}^{\pm 0.091}$ & $\underline{0.648}^{\pm 0.088}$ \\
& \cellcolor{blue}\textbf{Ours} & \cellcolor{blue} &
\cellcolor{blue}  $\mathbf{0.885}^{\pm 0.041}$ & \cellcolor{blue}  $\mathbf{0.926}^{\pm 0.054}$ & \cellcolor{blue}  $\underline{0.693}^{\pm 0.021}$ &
\cellcolor{blue}  $\mathbf{0.892}^{\pm 0.011}$ & \cellcolor{blue}  $\mathbf{0.921}^{\pm 0.044}$ & \cellcolor{blue}  $\mathbf{0.705}^{\pm 0.043}$ \\
\bottomrule
\end{tabular}}
\begin{flushleft}
\footnotesize
$^{\dagger}$Adapted to OpenVLA's single-step output via MMD 
between adjacent timesteps (see~\cref{sec:multi_app}).
\end{flushleft}
\vspace{-1cm}
\end{table}

\begin{table}[t]
\centering
\small
\caption{Comparison of failure detection performance with $\pi_{0.5}$ on VLABench.}
\label{tab:vlabench}
\setlength{\tabcolsep}{6pt}
\renewcommand{\arraystretch}{1.05}
\resizebox{\columnwidth}{!}{%
\begin{tabular}{@{} >{\centering\arraybackslash}p{10mm} l p{16mm}  c c c c c c @{}}
\toprule
\multicolumn{9}{c}{\textbf{Policy: $\mathbf{\pi}_{\mathbf{0.5}}$ (Success Rate: 40.3\%)}}\\
\midrule

& & &
\multicolumn{3}{c}{\textbf{Seen Tasks}} &
\multicolumn{3}{c}{\textbf{Unseen Tasks}} \\
\cmidrule(lr){4-6}\cmidrule(lr){7-9}

& \textbf{Method} &  &
\textbf{bACC}$\uparrow$ & \textbf{wACC}$\uparrow$ & \textbf{TWA}$\uparrow$ &
\textbf{bACC}$\uparrow$ & \textbf{wACC}$\uparrow$ & \textbf{TWA}$\uparrow$ \\
\midrule
\multirow{5}{*}{\rotatebox{90}{\textbf{OOD Det.}}} 
& Mahalanobis & \venue{ICML'18} & $0.637^{\pm 0.076}$ & $0.519^{\pm 0.114}$ & $0.581^{\pm 0.079}$ & $0.611^{\pm 0.101}$ & $0.548^{\pm 0.092}$ & $0.545^{\pm 0.073}$ \\
& Cosine k-NN & \venue{ICML'22} & $0.535^{\pm 0.023}$ & $0.384^{\pm 0.044}$ & $0.485^{\pm 0.035}$ & $\underline{0.704}^{\pm 0.072}$ & $\underline{0.676}^{\pm 0.095}$ & $\underline{0.584}^{\pm 0.088}$ \\
& PCA KMeans & \venue{CoRL'24} & $0.642^{\pm 0.052}$ & $0.669^{\pm 0.085}$ & $0.529^{\pm 0.002}$ & $0.616^{\pm 0.065}$ & $0.618^{\pm 0.068}$ & $0.520^{\pm 0.053}$ \\
& RND & \venue{ICML'24} & $0.628^{\pm 0.007}$ & $0.618^{\pm 0.012}$ & $0.545^{\pm 0.013}$  & $0.551^{\pm 0.044}$ & $0.537^{\pm 0.014}$ & $0.507^{\pm 0.017}$   \\
& LogpZO & \venue{RSS'25} & $0.618^{\pm 0.067}$ & $0.503^{\pm 0.102}$ & $0.552^{\pm 0.071}$ & $0.674^{\pm 0.030}$ & $0.628^{\pm 0.028}$ & $0.581^{\pm 0.021}$ \\
\midrule

\multirow{4}{*}{\rotatebox{90}{\textbf{Multi.}}} 
& Cluster Entropy & \venue{ICLR'23} &
$0.629^{\pm 0.050}$ & $0.656^{\pm 0.069}$ & $\underline{0.602}^{\pm 0.029}$ &
$0.526^{\pm 0.055}$ & $0.506^{\pm 0.067}$ & $0.490^{\pm 0.050}$ \\
& EigenScore & \venue{ICLR'24} &
$0.553^{\pm 0.027}$ & $0.487^{\pm 0.030}$ & $0.532^{\pm 0.024}$ &
$0.566^{\pm 0.017}$ & $0.538^{\pm 0.025}$ & $0.554^{\pm 0.017}$ \\
& STAC & \venue{CoRL'24} &
$0.615^{\pm 0.058}$ & $0.586^{\pm 0.072}$ & $0.556^{\pm 0.068}$&
$0.591^{\pm 0.017}$ & $0.598^{\pm 0.048}$ & $0.576^{\pm 0.021}$ \\
& ACE & \venue{NeurIPS'25} &
$0.596^{\pm 0.032}$ & $0.545^{\pm 0.008}$ & $0.553^{\pm 0.033}$ &
$0.591^{\pm 0.018}$ & $0.595^{\pm 0.019}$ & $0.552^{\pm 0.006}$ \\
\midrule
\multirow{3}{*}{\rotatebox{90}{\textbf{Class.}}} 
& SAFE-LSTM & \venue{NeurIPS'25} & $0.629^{\pm 0.012}$ & $0.524^{\pm 0.030}$ & $0.587^{\pm 0.027}$ & $0.484^{\pm 0.063}$ & $0.533^{\pm 0.115}$ & $0.482^{\pm 0.065}$ \\
& 
SAFE-MLP & \venue{NeurIPS'25} & $\underline{0.788}^{\pm 0.111}$ & $\underline{0.766}^{\pm 0.106}$ & $0.581^{\pm 0.084}$ & $0.641^{\pm 0.029}$ & $0.634^{\pm 0.084}$ & $0.500^{\pm 0.038}$ \\
& \cellcolor{blue} \textbf{Ours} & \cellcolor{blue}  & \cellcolor{blue}  $\mathbf{0.856}^{\pm 0.075}$ & \cellcolor{blue} $\mathbf{0.827}^{\pm 0.098}$ & \cellcolor{blue} $\mathbf{0.662}^{\pm 0.100}$ & \cellcolor{blue} $\mathbf{0.713}^{\pm 0.035}$ & \cellcolor{blue} $\mathbf{0.709}^{\pm 0.023}$ & \cellcolor{blue} $\mathbf{0.608}^{\pm 0.017}$ \\
\bottomrule
\end{tabular}}
\vspace{-0.5cm}
\end{table}

\begin{table}[t]
\centering
\small
\caption{Real-world robot setup (left) and failure detection performance with $\pi_{0.5}$ (right).}
\label{tab:real_world}
\setlength{\tabcolsep}{4pt}
\renewcommand{\arraystretch}{1.0}
\begin{minipage}{0.29\columnwidth}
  \centering
\includegraphics[width=0.9\linewidth]{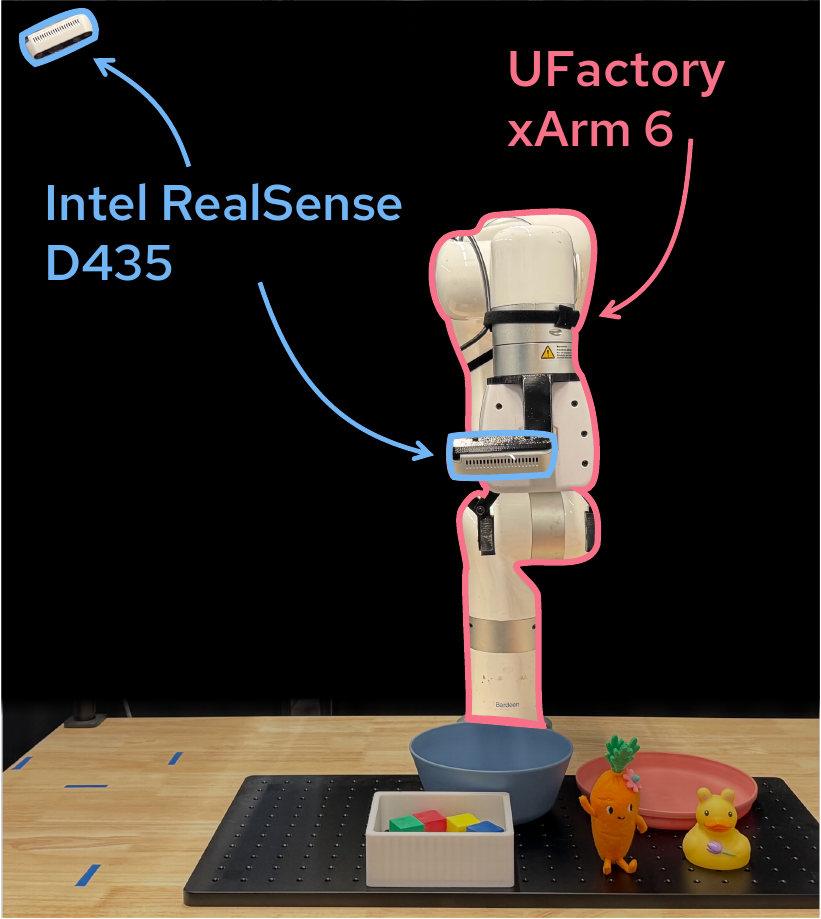}
\end{minipage}
\hfill
\hfill
\begin{minipage}{0.70\columnwidth}
\resizebox{\linewidth}{!}{%
\begin{tabular}{@{} l cccc cccc @{}}
\toprule
& \multicolumn{4}{c}{\textbf{CUBE (Success Rate: 61.2\%)}} & \multicolumn{4}{c}{\textbf{KITCHEN (Success Rate: 63.7\%)}} \\
\cmidrule(lr){2-5}\cmidrule(lr){6-9}
& \multicolumn{2}{c}{\textbf{Seen}} & \multicolumn{2}{c}{\textbf{Unseen}} 
& \multicolumn{2}{c}{\textbf{Seen}} & \multicolumn{2}{c}{\textbf{Unseen}} \\
\cmidrule(lr){2-3}\cmidrule(lr){4-5}\cmidrule(lr){6-7}\cmidrule(lr){8-9}
\textbf{Method}
& \textbf{bACC} & \textbf{TWA}
& \textbf{bACC} & \textbf{TWA}
& \textbf{bACC} & \textbf{TWA}
& \textbf{bACC} & \textbf{TWA} \\
\midrule
Mahalanobis & $0.683$ & $0.652$ & $0.681$ & $0.623$ & $0.779$ & $0.754$ & $0.630$ & $0.608$ \\
Cosine k-NN & $0.674$ & $0.617$ & $0.658$ & $0.647$ & $0.693$ & $0.612$ & $0.644$ & $0.592$ \\
PCA KMeans & $0.525$ & $0.509$ & $0.472$ & $0.463$ &  $0.541$ & $0.522$ & $0.566$ & $0.553$ \\
RND & $0.517$ & $0.505$ & $0.547$ & $0.538$ & $0.679$ & $0.661$ & $0.615$ & $0.605$ \\
LogpZO & $0.699$ & $0.666$ & $0.660$ & $0.639$  & $\underline{0.895}$ & $\underline{0.811}$ & $0.837$ & $\underline{0.794}$ \\
\midrule
SAFE-LSTM & $0.723$ & $0.597$ & $0.714$ & $0.623$ & $0.846$ & $0.780$ & $0.595$ & $0.577$ \\
SAFE-MLP &  $\mathbf{0.992}$ & $\underline{0.847}$  & $\underline{0.797}$ & $\underline{0.650}$ & $0.784$ & $0.748$ & $\underline{0.878}$ & $0.768$ \\
\cellcolor{blue} \textbf{Ours}
& \cellcolor{blue} $\underline{0.966}$ & \cellcolor{blue} $\mathbf{0.852}$ &\cellcolor{blue} $\mathbf{0.914}$ & \cellcolor{blue} $\mathbf{0.800}$ &  \cellcolor{blue} $\mathbf{0.968}$ & \cellcolor{blue} $\mathbf{0.863}$ & \cellcolor{blue} $\mathbf{0.972}$ & \cellcolor{blue} $\mathbf{0.876}$ \\
\bottomrule
\end{tabular}}
\end{minipage}
\end{table}

\paragraph{Simulation environment.}
\Cref{tab:vertical_models,tab:vlabench} report failure detection performance on LIBERO-10 and VLABench across three VLA policies: OpenVLA, $\pi_0$ and $\pi_{0.5}$.
\text{Hide-and-Seek} consistently achieves state-of-the-art performance, outperforming all baselines across models and benchmarks in terms of bACC, wACC, and TWA on both seen and unseen tasks.
OOD detection-based methods perform competitively on seen tasks but degrade substantially on unseen tasks, \eg, Mahalanobis drops from 
67.0\% to 51.3\% bACC on OpenVLA, revealing their limitation in generalizing beyond the training distribution.
Multi-sampling-based methods generally underperform classifier-based approaches, as uncertainty estimates derived from repeated action sampling do not reliably reflect task-level failure without access to any failure supervision.
Among classifier-based methods, both SAFE-LSTM and SAFE-MLP train on trajectory-level labels, diluting the decision boundary by treating all timesteps uniformly.
SAFE-MLP partially mitigates this with a linearly increasing temporal weight, but this fixed heuristic is agnostic to when failure actually manifests.
Our method instead adaptively discovers the most failure-indicative moments from trajectory-level supervision alone, outperforming SAFE-MLP by $+\textbf{6.8}\%$ bACC and $+\textbf{8.1}\%$ TWA on seen tasks, and $+\textbf{7.2}\%$ bACC and $+\textbf{10.8}\%$ TWA on unseen tasks on VLABench with $\pi_{0.5}$.
\emph{Qualitative results are provided in~{\cref{sec:qualitative}}}.

\paragraph{Real-world robot.}
\Cref{tab:real_world} reports failure detection performance on a UFactory xArm~6 platform with $\pi_{0.5}$ across two task suites, CUBE and KITCHEN.
Hide-and-Seek consistently outperforms all baselines on both seen and unseen tasks, achieving the best TWA across all four settings and gaining $+\textbf{11.7}\%$ bACC and $+\textbf{15.0}\%$ TWA on unseen CUBE over SAFE-MLP.
Baseline performance varies substantially across splits, \eg, SAFE-MLP achieves $99.2\%$ bACC on seen CUBE but drops to $79.7\%$ on unseen.
In contrast, Hide-and-Seek maintains strong and stable performance across all four settings, demonstrating that failure signals discovered from trajectory-level supervision transfer effectively to real-world execution and generalize beyond the simulator. 
Due to real-time deployment constraints, we exclude multi-sampling-based methods that incur significant inference overhead.
Experimental details are provided in~\cref{sec:real}.

\subsection{Ablation Studies}
\label{sec:abl}

\begin{figure*}[t]
\centering
\includegraphics[width=\linewidth]{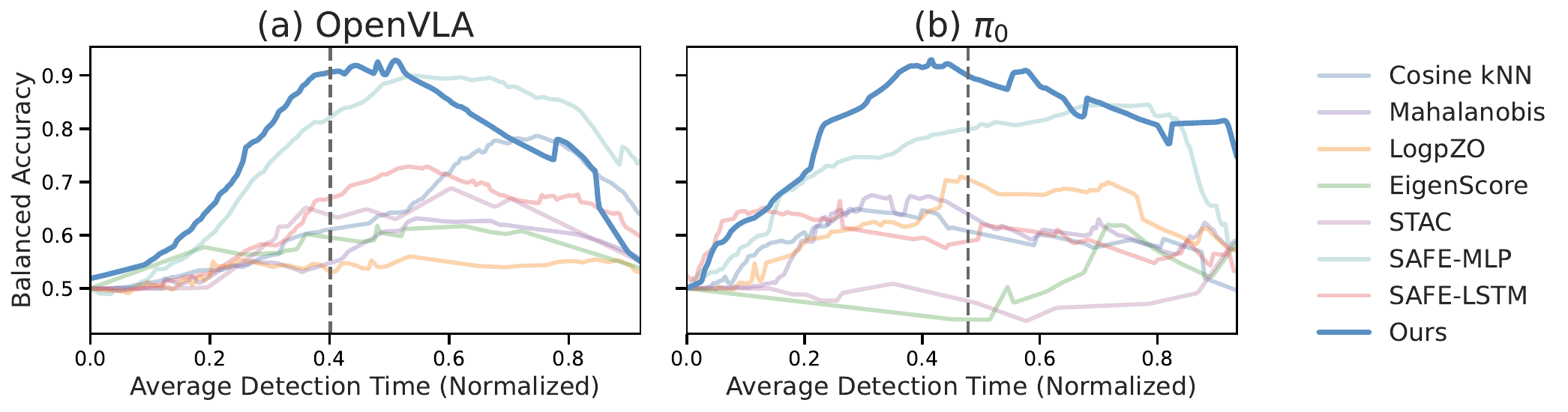}
\vspace{-0.5cm}
\caption{Detection accuracy--timeliness tradeoff on LIBERO-10 with OpenVLA and $\pi_{0}$. \textcolor{gray}{Gray dashed line} indicates the average normalized failure onset time across failure trajectories.}
\label{fig:tradeoff}
\vspace{-0.2cm}
\end{figure*}

\begin{table*}[t]
\centering
\begin{minipage}[t]{0.32\linewidth}
    \centering
    \small
    \caption{Architecture.}
    \label{tab:architecture}
    \resizebox{\linewidth}{!}{%
    \begin{tabular}{lcc}
        \toprule
        \textbf{Method} & 
        \textbf{OpenVLA} & $\mathbf{\pi_0}$ \\
        \midrule
        MLP & 0.765 & 0.743 \\
        GRU & 0.834 & 0.893 \\
        Transformer & 0.804 & 0.824 \\
                  \cellcolor{blue} LSTM   &   \cellcolor{blue} 0.843 &   \cellcolor{blue} 0.888 \\
        \bottomrule
    \end{tabular}}
\end{minipage}
\hfill
\begin{minipage}[t]{0.32\linewidth}
    \centering
    \small
    \caption{Component analysis.}
    \label{tab:component}
    \resizebox{\linewidth}{!}{%
    \begin{tabular}{cccc}
        \toprule
        $\mathcal{L}_{\mathrm{inter}}$ & $\mathcal{L}_{\mathrm{intra}}$ & \textbf{OpenVLA} & $\mathbf{\pi_0}$\\
        \midrule
        \checkmark & & 0.825 & 0.856 \\
         & \checkmark & 0.579 & 0.601 \\
        \cellcolor{blue} \checkmark & \cellcolor{blue} \checkmark & \cellcolor{blue}  0.843 & \cellcolor{blue}  0.888 \\
        \bottomrule
    \end{tabular}}
\end{minipage}
\hfill
\begin{minipage}[t]{0.32\linewidth}
    \centering
    \small
    \caption{Comparison with VLM-based runtime monitor.}
    \label{tab:vlm}
    \resizebox{\linewidth}{!}{%
    \begin{tabular}{lcc}
        \toprule
        \textbf{Method} & \textbf{bACC} $\uparrow$ & \textbf{Latency} $\downarrow$ \\
        \midrule
        VLM Monitor & 0.712 & 2.343s \\
        \cellcolor{blue}\textbf{Ours} & \cellcolor{blue} 0.843 & \cellcolor{blue} 0.001s \\
        \bottomrule
    \end{tabular}}
\end{minipage}
\vspace{-0.3cm}
\end{table*}

\paragraph{Detection accuracy--timeliness trade-off.}
A reliable failure detector must balance accuracy and promptness.
In our framework, the user-specified $\alpha$ controls this trade-off: stricter $\alpha$ yields more conservative thresholds with delayed detection, while looser $\alpha$ enables earlier detection at the cost of more false alarms. 
We sweep $\alpha \in [0.01, 1.0]$ to trace a curve of operating points for each method on unseen LIBERO-10 splits in~\cref{fig:tradeoff}.
For each $\alpha$, we compute the average detection time (normalized by trajectory length) and the corresponding bACC; curves closer to the top-left indicate better trade-off.
For temporal reference, we annotate the failure onset of each failed rollout using GPT-5.2~\cite{singh2025openai}
on recorded videos, shown as vertical dashed lines (see~\cref{sec:agree} for human agreement).
Hide-and-Seek achieves the best trade-off on LIBERO-10 with both OpenVLA and $\pi_0$, with curves consistently closer to the top-left than baselines, indicating earlier and more accurate failure detection.
We further quantify detection timeliness at the fixed $\alpha$ used in deployment in~\cref{sec:detection_time_abl}.

\paragraph{Detector architecture.}
We compare our LSTM-based~\cite{graves2012long} detector\footnote{Unless otherwise noted, the remaining ablations report bACC averaged over seen and unseen tasks.} against three alternatives in~\cref{tab:architecture}: an MLP, a GRU~\cite{chung2014empirical}, and a single-layer Transformer~\cite{vaswani2017attention}.
All variants share the same input features and training setup, differing only in the backbone.
Context-utilizing backbones (GRU, Transformer, LSTM) substantially outperform the per-step MLP, indicating that our formulation benefits from modeling temporal context when identifying failure-indicative actions. 
Among these, the LSTM and GRU outperform the Transformer.
We conjecture that recurrent backbones offer a more suitable inductive bias for sequential failure detection under limited training data.
We adopt the LSTM as our default backbone due to its consistently strong performance across settings.

\paragraph{Loss component analysis.}
We ablate the contribution of each training objective 
in~\cref{tab:component}.
Training with $\mathcal{L}_{\mathrm{inter}}$ alone already achieves strong performance, surpassing the best-performing baseline (SAFE-MLP in~\cref{tab:vertical_models}) by $+2.6\%$ on OpenVLA and $+2.1\%$ on $\pi_0$.
This result indicates that inter-trajectory contrastive loss effectively extracts failure-indicative signals even under weak trajectory-level supervision.
$\mathcal{L}_{\mathrm{intra}}$ alone yields degraded performance, as it relies on the failure onset proxy, which is only meaningful once $\mathcal{L}_{\mathrm{inter}}$ has shaped the score dynamics to reflect failure-indicative structure.
Combining both objectives yields consistent improvements on both 
OpenVLA ($+1.8\%$) and $\pi_0$ ($+3.2\%$), demonstrating that $\mathcal{L}_{\mathrm{intra}}$
effectively strengthens the onset signal that $\mathcal{L}_{\mathrm{inter}}$
discovers implicitly, converting it into explicit supervision 
without requiring any additional annotation.

\paragraph{Hide-and-Seek is more efficient and effective than VLM-based runtime monitoring.}
We compare our method against a VLM-based runtime monitor that prompts Qwen3-VL-8B-Instruct~\cite{bai2025qwen3} with online video question answering on historical frames up to the current timestep, outputting a discrete success/failure judgment per frame on LIBERO-10 with OpenVLA.
A trajectory is classified as a failure if at least one frame is judged as a failure.
As shown in~\cref{tab:vlm}, Hide-and-Seek outperforms the VLM-based monitor by a substantial margin ($+13.1\%$ bACC), while operating at over {$\textbf{2{,}000}\times$ \textbf{higher speed}} ($0.001$s vs.\ $2.343$s per step on a single NVIDIA A6000 GPU).
This demonstrates that VLA latent action embeddings provide a richer failure signal than visual observations alone and are considerably more practical for real-time deployment.

\begin{figure*}[t!]
    \centering
    \begin{subfigure}{0.32\textwidth}
        \centering
        \includegraphics[width=\linewidth]{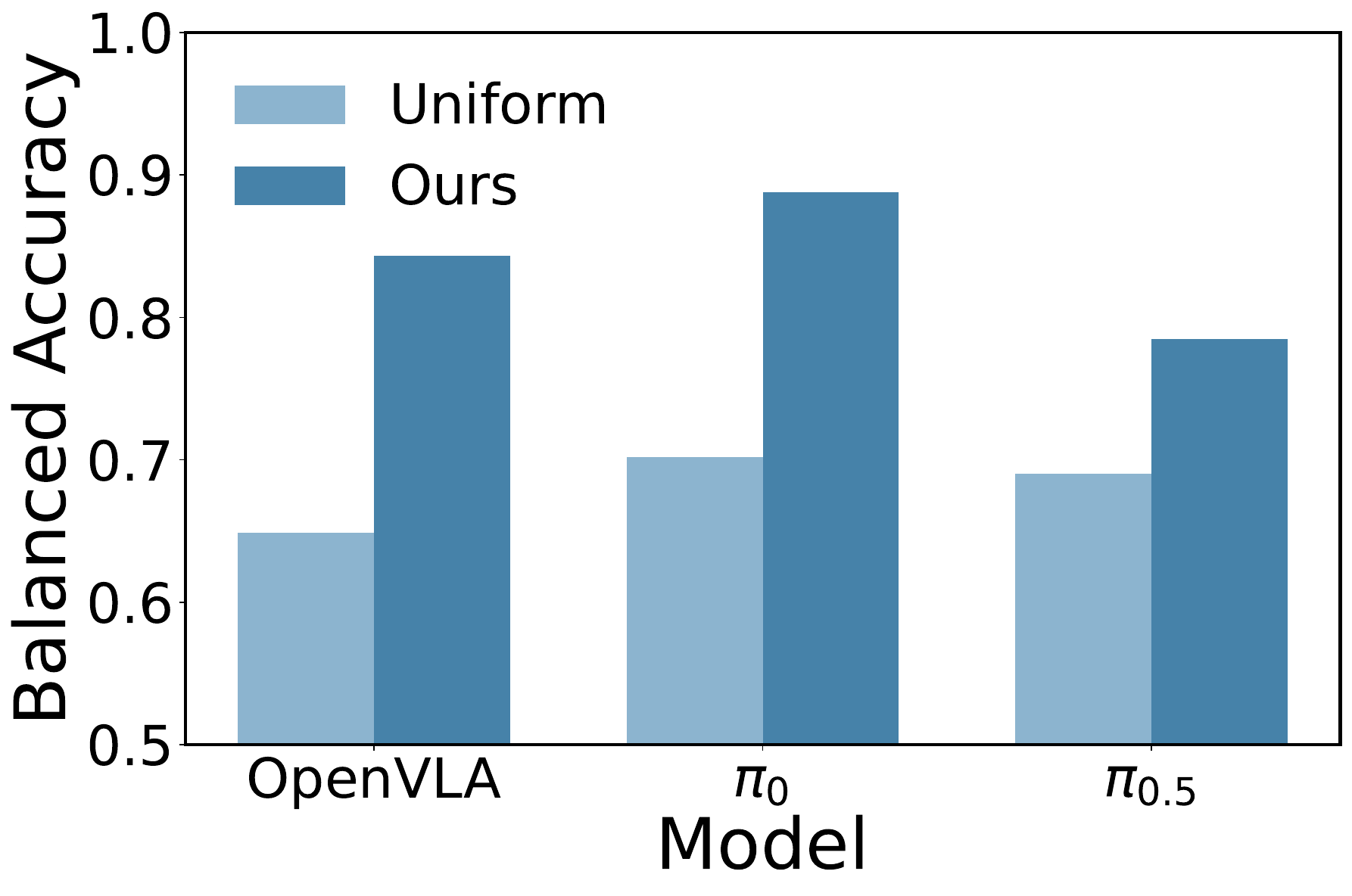}
        \caption{Uniform vs.\ ours.}
        \label{fig:uniform}
    \end{subfigure}
    \hfill
     \begin{subfigure}{0.32\textwidth}
        \centering
        \includegraphics[width=\linewidth]{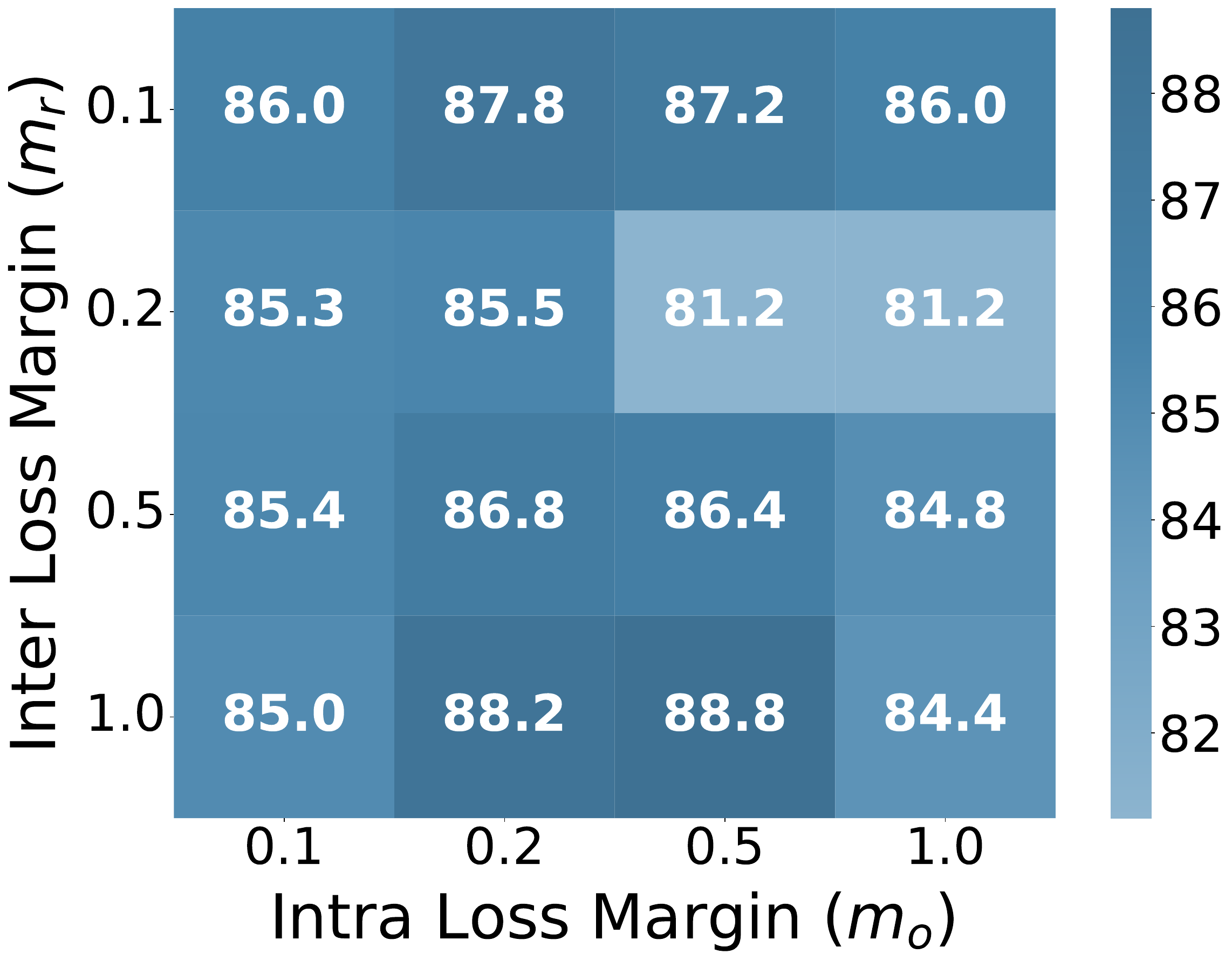}
        \caption{Effect of loss margin $m$.}
        \label{fig:margin}
    \end{subfigure}
    \hfill
     \begin{subfigure}{0.32\textwidth}
        \centering
        \includegraphics[width=\linewidth]{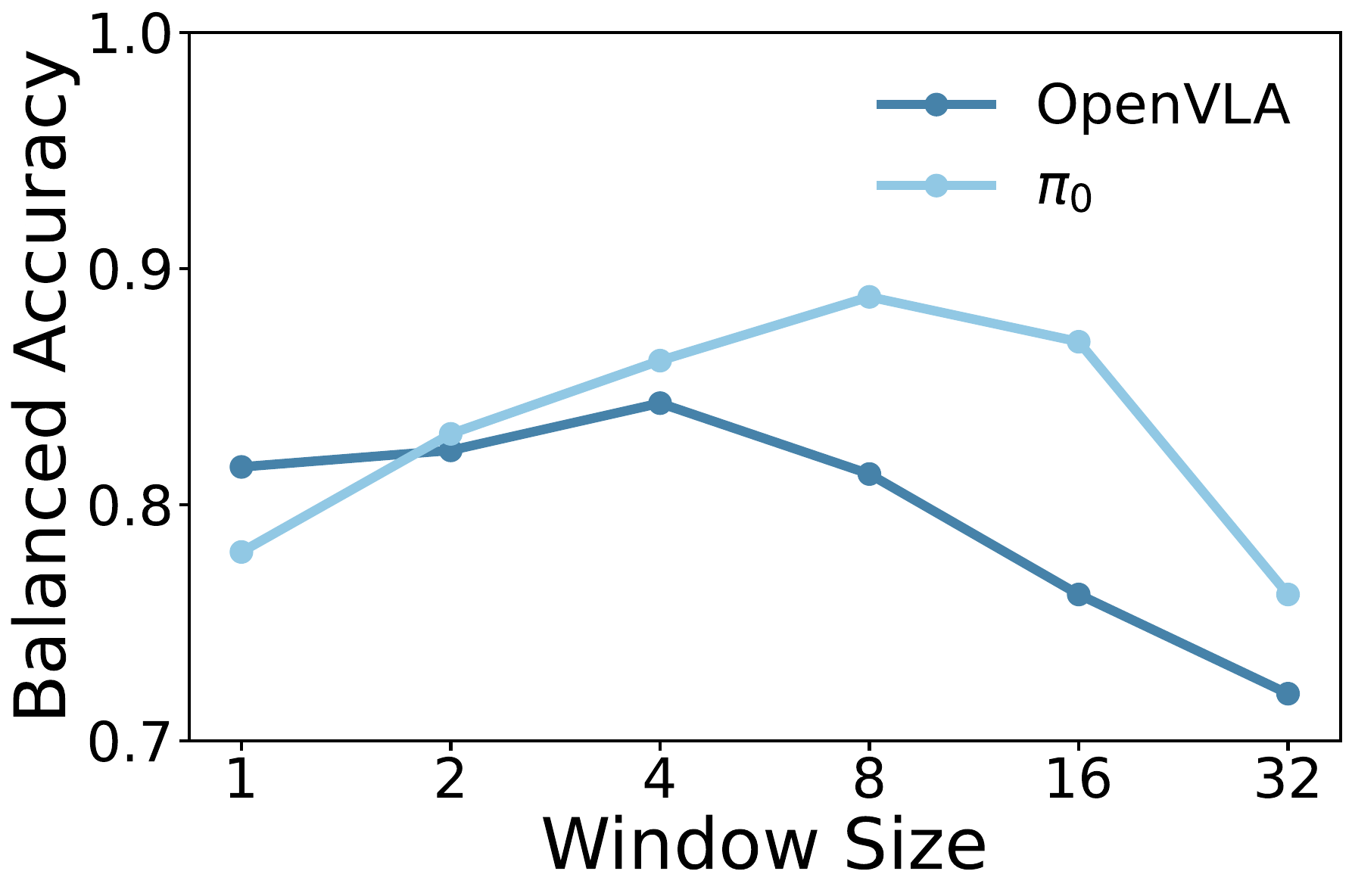}
        \caption{Effect of window size $w$.}
        \label{fig:window}
    \end{subfigure}
    \vspace{-0.1cm}
    \caption{
        (a) Comparison between uniform trajectory-level labeling 
        and our approach; 
        (b) effect of margins  $m_r$ and $m_o$ 
        on detection performance;
        (c) effect of window size $w$ on detection performance. All results report balanced accuracy averaged over seen 
        and unseen tasks on LIBERO-10.
    }
    \vspace{-0.4cm}
\end{figure*}

\paragraph{Comparison with uniform trajectory-level labeling.}

To validate the necessity of our formulation, we compare against a baseline that uniformly assigns the trajectory-level label to every timestep and trains with a margin loss on all steps under otherwise identical settings, including the window size.
As shown in~\cref{fig:uniform}, our approach substantially outperforms uniform labeling~\cite{gu2025safe}, achieving gains of $+\textbf{19.4}\%$ bACC on OpenVLA and $+\textbf{18.6}\%$ on $\pi_0$. These results indicate that uniform supervision introduces label noise by treating correct actions as failure-indicative (see~\cref{sec:latent_vis}) whereas our method isolates the critical failure signals necessary for reliable detection.

\paragraph{Effect of loss margin.}
We study the sensitivity of Hide-and-Seek to the margin hyperparameters $m_r$ in $\mathcal{L}_{\mathrm{inter}}$ and $m_o$ in $\mathcal{L}_{\mathrm{intra}}$ by sweeping both on LIBERO-10 with $\pi_0$.
As shown in~\cref{fig:margin},
performance remains stable with bACC varying by less than $4\%$ from the mean across all configurations, indicating that Hide-and-Seek is robust to margin choices rather than requiring precise tuning.
The best performance is achieved at $m_r=1.0, m_o=0.5$ 
($88.8\%$ bACC), and we adopt this setting as our default for $\pi_0$.
Insufficiently large margins ($m_r, m_o \leq 0.2$) provide weak separation between failure and successful actions, while an overly large intra loss margin ($m_o=1.0$) over-constrains the intra-trajectory score structure.

\paragraph{Effect of window size.}
We ablate the window size $w$ used for the sliding window aggregation in~\cref{sec:tfa}. As shown in~\cref{fig:window}, moderate window sizes consistently improve performance over small ones, confirming that temporal aggregation reduces noise in step-level representations while preserving informative dynamics. We observe that $w=4$ performs best for OpenVLA, whereas $w=8$ yields the highest accuracy for $\pi_0$.
This trend suggests that appropriate windowing captures the local temporal continuity of robot actions. Notably, for $\pi_0$, the optimal window size aligns with its action prediction horizon (replanning interval), indicating that matching the aggregation scale to the policy’s action chunking structure further enhances failure signal extraction.

\textbf{Additional ablations} on (1) failure onset proxies, (2) thresholding strategies, (3) layer-wise effects, and (4) inference-time policy steering are provided in~\cref{sec:ablation_app}.

\vspace{-0.2cm}
\section{Conclusion}
\label{sec:conclusion}

We present Hide-and-Seek, a lightweight runtime failure detector for VLA models that discovers failure-indicative actions from trajectory-level supervision alone.
By formulating failure detection as a coarsely supervised learning problem, Hide-and-Seek combines an inter-trajectory contrastive objective to identify critical failure-indicative actions with an intra-trajectory objective that sharpens score separation around the inferred failure onset.
Extensive experiments on LIBERO, VLABench, and a real-world robotic platform with OpenVLA, $\pi_0$, and $\pi_{0.5}$ show that Hide-and-Seek consistently outperforms existing methods on both seen and unseen tasks, achieving state-of-the-art multi-task failure detection accuracy with a practical accuracy--timeliness trade-off.

\section*{Acknowledgement}
We gratefully acknowledge the anonymous reviewers for their constructive feedback, as well as Yu Wang, Shawn Im, Heecheol Kim, and Dahye Kim for their valuable feedback.
Seongheon Park, Wendi Li, Changdae Oh, Samuel Yeh, and Sharon Li are supported in part by the AFOSR Young Investigator Program under award number FA9550-23-1-0184, the Office of Naval Research (ONR) under award number N000142612508, the National Science Foundation under awards IIS-2237037 and IIS-2331669, the Alfred P. Sloan Fellowship, Open Philanthropy (now Coefficient Giving), and Schmidt Sciences Foundation. Zsolt Kira was supported in part by the National Science Foundation under Grant No. IIS-2239292 and Schmidt Sciences Foundation. We would also like to thank Tutor Intelligence for donating the robots used in the hardware experiments.

{
\small
\bibliographystyle{unsrt}
\bibliography{main}
}

\newpage
\appendix

\begin{center}
    \LARGE \textbf{Appendix}
\end{center}

\tableofcontents
\addtocontents{toc}{\protect\setcounter{tocdepth}{2}}

\newpage

\section{Failure Score Visualization}
\label{sec:failure_vis}

\subsection{Failure Score Dynamics Across Splits}
\label{sec:score_dynamics}

\begin{figure}[h!]
\centering
\begin{subfigure}[b]{0.32\linewidth}
    \centering
    \includegraphics[width=\linewidth]{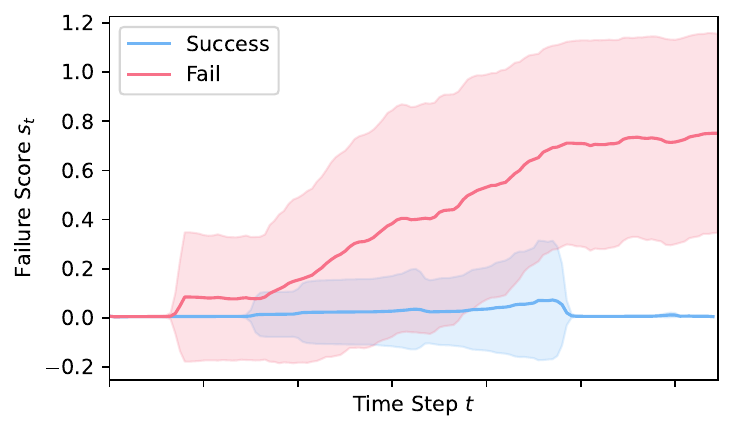}
    \caption{Train}
    \label{fig:score_trend_openvla_train}
\end{subfigure}
\hfill
\begin{subfigure}[b]{0.32\linewidth}
    \centering
    \includegraphics[width=\linewidth]{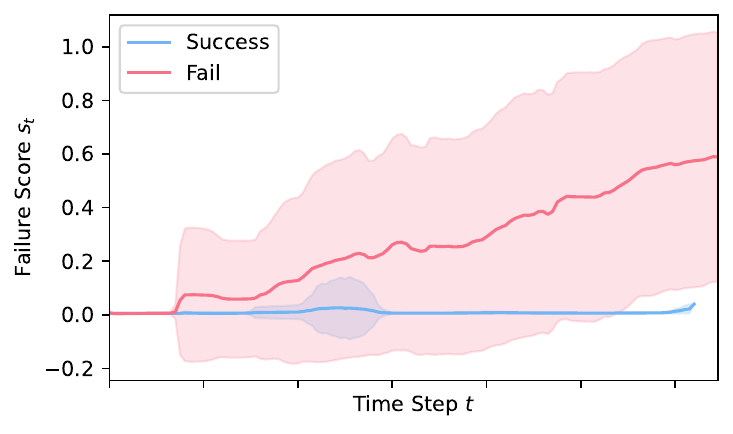}
    \caption{Seen}
    \label{fig:score_trend_openvla_seen}
\end{subfigure}
\hfill
\begin{subfigure}[b]{0.32\linewidth}
    \centering
    \includegraphics[width=\linewidth]{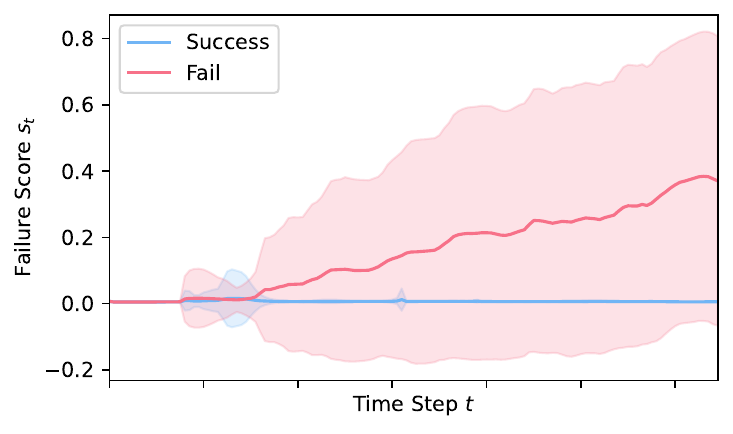}
    \caption{Unseen}
    \label{fig:score_trend_openvla_unseen}
\end{subfigure}
\caption{
    \textbf{Average failure score trend across episodes for 
    OpenVLA on LIBERO-10.} The per-timestep score is averaged 
    over successful (\textcolor{myblue}{blue}) and failure (\textcolor{pink}{red}) trajectories on 
    (a) the training set, (b) the seen evaluation split, and 
    (c) the unseen evaluation split. Shaded regions indicate standard deviation across episodes.}
\label{fig:score_trend_openvla}
\end{figure}

\begin{figure}[h!]
\centering
\begin{subfigure}[b]{0.32\linewidth}
    \centering
    \includegraphics[width=\linewidth]{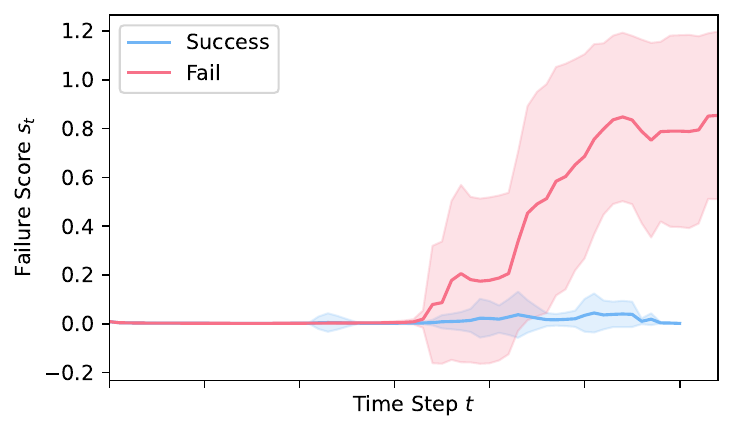}
    \caption{Train}
    \label{fig:score_trend_pi0_train}
\end{subfigure}
\hfill
\begin{subfigure}[b]{0.32\linewidth}
    \centering
    \includegraphics[width=\linewidth]{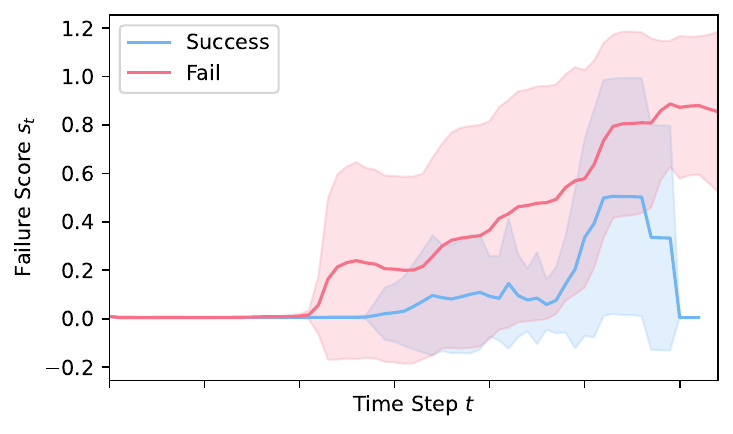}
    \caption{Seen}
    \label{fig:score_trend_pi0_seen}
\end{subfigure}
\hfill
\begin{subfigure}[b]{0.32\linewidth}
    \centering
    \includegraphics[width=\linewidth]{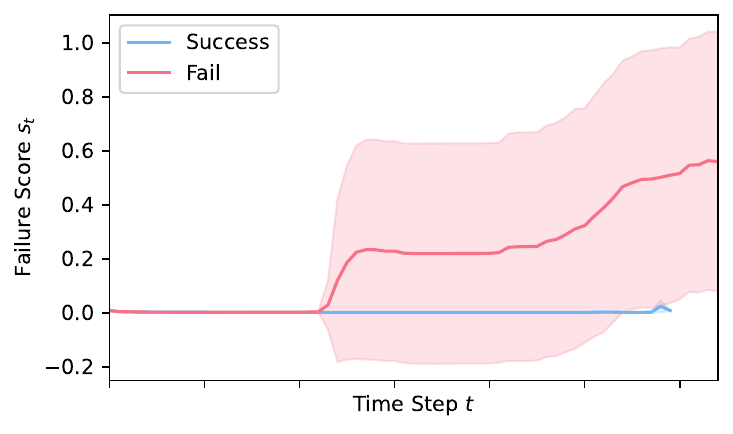}
    \caption{Unseen}
    \label{fig:score_trend_pi0_unseen}
\end{subfigure}
\caption{
    \textbf{Average failure score trend across episodes for 
    $\pi_0$ on LIBERO-10.}}
\label{fig:score_trend_pi0}
\end{figure}

We visualize the average per-timestep failure score $s_t$ across episodes, computed after training the detector, for the training set and the seen/unseen evaluation splits. Results are shown for OpenVLA (\Cref{fig:score_trend_openvla}) and $\pi_0$ (\Cref{fig:score_trend_pi0}) on LIBERO-10.
Across all splits, the failure score remains consistently low for successful trajectories, while exhibiting a clear upward trend for failed trajectories.
The consistent score dynamics across splits indicate that Hide-and-Seek learns transferable failure signals rather than overfitting to training-distribution cues.

Although \Cref{eq:inter} propagates gradients only through the maximum-scoring timestep of each successful trajectory, the identity of this argmax varies across iterations and minibatch pairings.
As a result, different timesteps receive supervision over the course of training.
In practice, this effectively distributes supervision across most timesteps of successful trajectories: scores on successful rollouts remain consistently low across all splits, while those on failure rollouts stay well separated.

\subsection{Qualitative Results}
\label{sec:qualitative}

Hide-and-Seek's failure scores correlate with observable failure indicators in the visual scene, despite the detector being trained without any step-level supervision.
The discovered onset point $t_{\mathrm{onset}}$ aligns with subtle early indicators of failure (\eg, a book starting to slip from the gripper), while the peak point $t_{\max}$ aligns with the most salient and critical failure event (\eg, the robot dropping the object).
These moments correspond to meaningful transitions in the physical state of the task and match human intuition about when failure occurs, even though the model receives only trajectory-level outcome labels during training. 

\begin{figure*}[p]
    \centering
    \setlength{\tabcolsep}{2pt}
    
    \begin{subfigure}{\textwidth}
        \includegraphics[page=1, width=\textwidth]{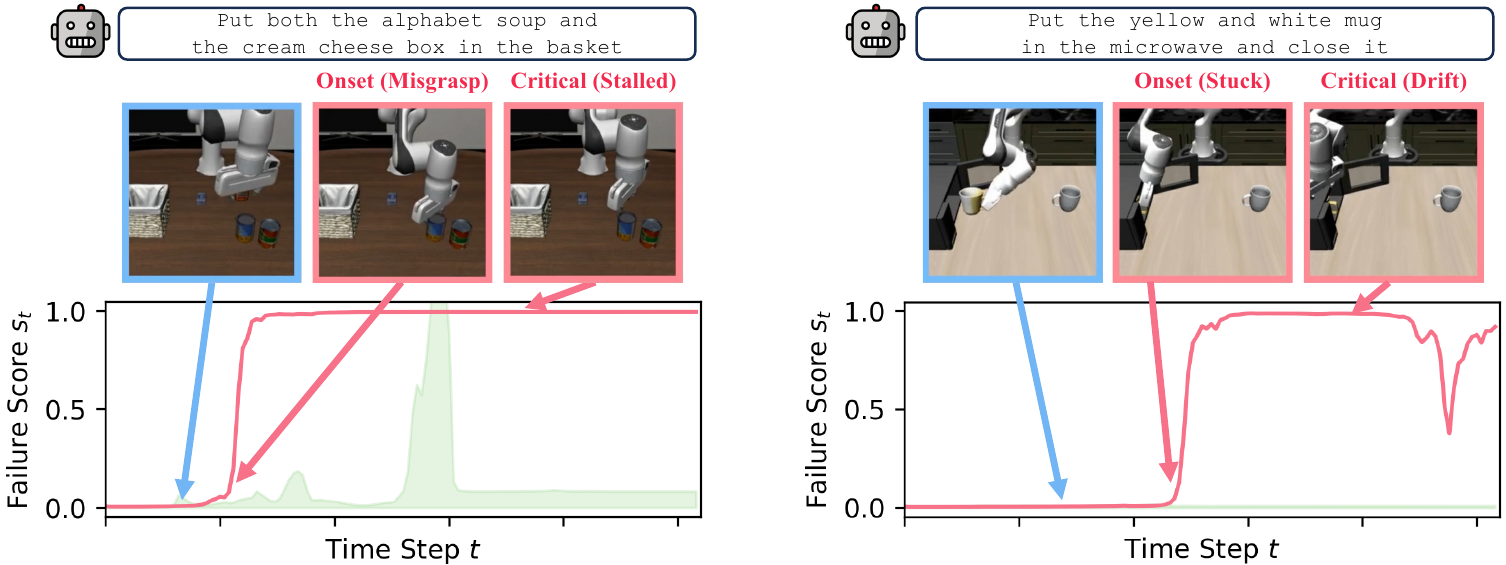}
        \caption{\textbf{Failure Trajectories.} (left) Robot fails to grasp the alphabet soup and stops moving.
(right) Robot gets stuck after placing the mug, then drifts away without closing the microwave.}
    \end{subfigure}
    
    \vspace{3pt}
    
    \begin{subfigure}{\textwidth}
        \includegraphics[ width=\textwidth]{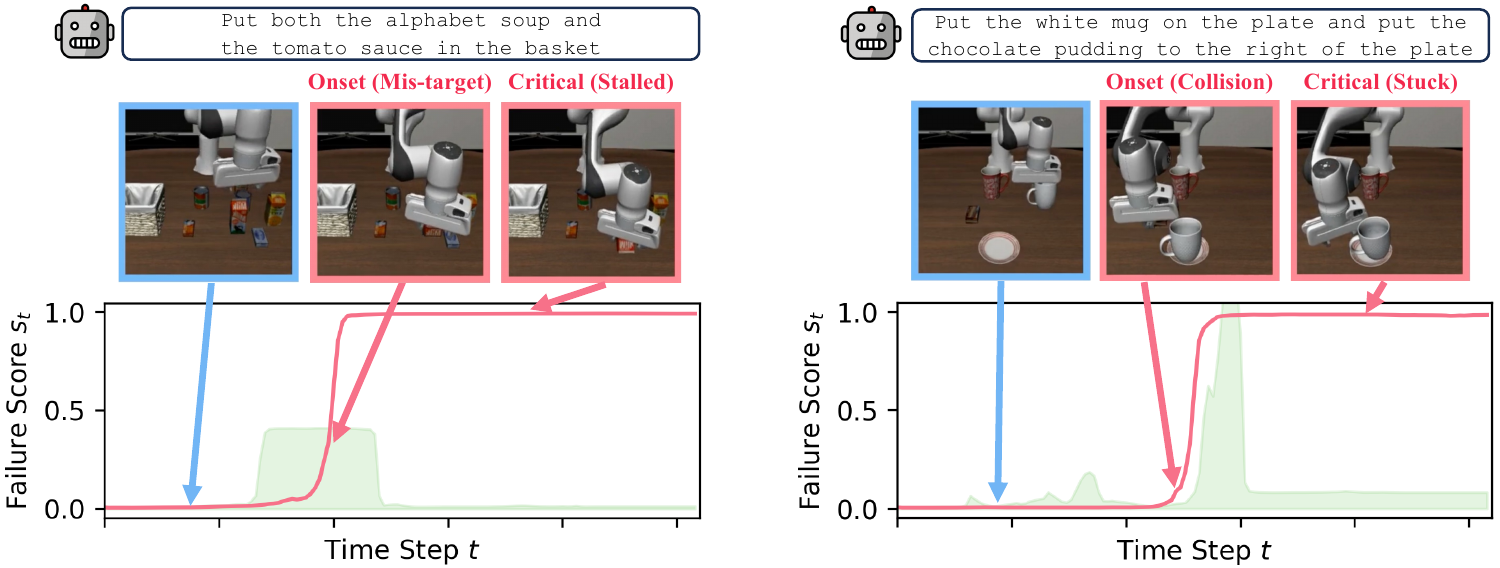}
\caption{\textbf{Failure Trajectories.} 
(left) Robot grasps the wrong object, drops it, and stalls. 
(right) Robot collides with the placed mug and gets stuck.}

    \end{subfigure}
    
    \vspace{3pt}
    
    \begin{subfigure}{\textwidth}
        \includegraphics[page=3, width=\textwidth]{figures/openvla_qual2.pdf}
        \caption{\textbf{Successful Trajectories.} The failure score remains consistently low throughout the trajectory.}
    \end{subfigure}
    
\caption{
    \textbf{Visualization of failure scores for the OpenVLA policy 
    across different tasks.} A failure is declared at the earliest timestep $t$ where the failure score $s_t$ (\textcolor{pink}{red curve}) exceeds the time-varying threshold $\zeta_t$ (\textcolor{thresholdgreen}{green region}) determined by conformal prediction.
}
        \label{fig:openvla_qualitative}
\end{figure*}

\begin{figure*}[p]
    \centering
    \setlength{\tabcolsep}{2pt}
    
    \begin{subfigure}{\textwidth}
        \includegraphics[page=1, width=\textwidth]{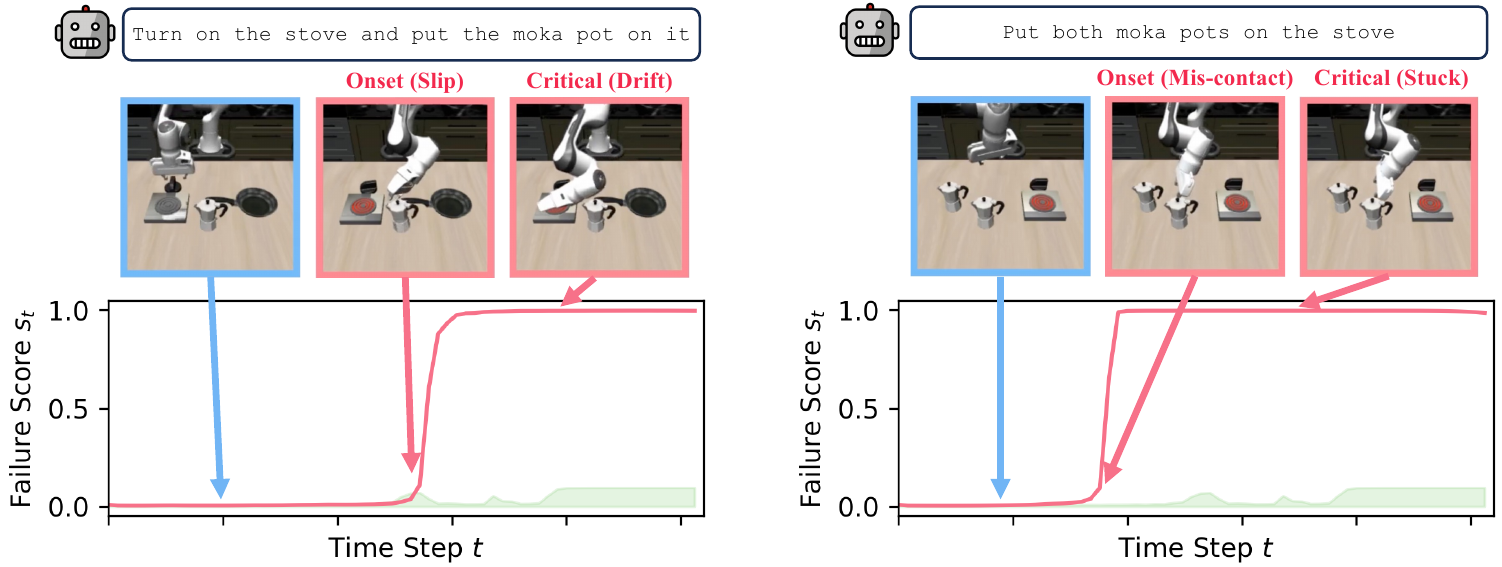}
        \caption{\textbf{Failure Trajectories.} (left) Robot slips the moka pot and drifts away. (right) Robot mis-contacts the top of the moka pot and gets stuck.}
    \end{subfigure}
    
    \vspace{3pt}
    
    \begin{subfigure}{\textwidth}
        \includegraphics[ width=\textwidth]{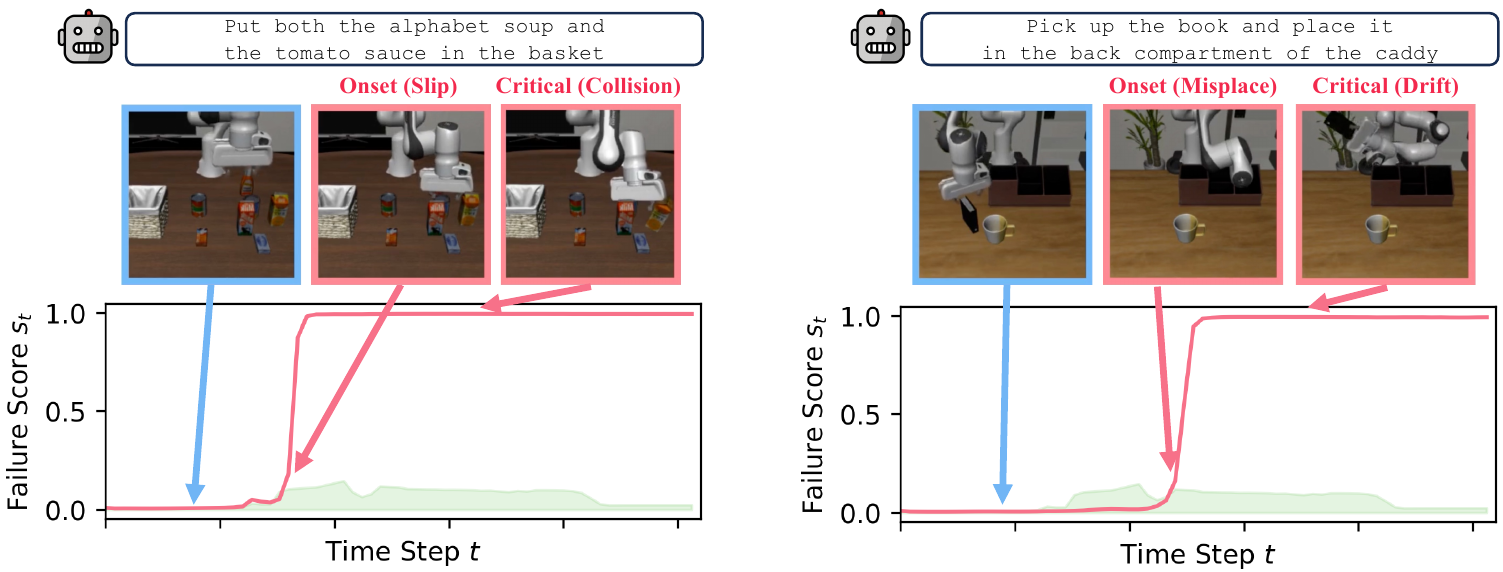}
        \caption{\textbf{Failure Trajectories.} (left) Robot slips the alphabet soup and collides with nearby objects. (right) Robot misplaces the book on the caddy's divider and drifts away.}
    \end{subfigure}
    
    \vspace{3pt}
    
    \begin{subfigure}{\textwidth}
        \includegraphics[page=3, width=\textwidth]{figures/pi0_qual3.pdf}
        \caption{\textbf{Successful Trajectories.} The failure score remains consistently low throughout the trajectory.}
    \end{subfigure}
    
\caption{
    \textbf{Visualization of failure scores for the $\pi_0$ policy across different tasks.}
    A failure is declared at the earliest timestep $t$ where the failure score $s_t$ (\textcolor{pink}{red curve}) exceeds the time-varying threshold $\zeta_t$ (\textcolor{thresholdgreen}{green region}) determined by conformal prediction.
}
    \label{fig:pi0_qualitative}
\end{figure*}

We visualize qualitative examples from OpenVLA (\Cref{fig:openvla_qualitative}) and $\pi_0$ (\Cref{fig:pi0_qualitative}) on LIBERO-10 tasks.
A failure is declared at the earliest timestep $t$ such that the failure score $s_t$ (\textcolor{pink}{red curve}) exceeds the threshold $\zeta_t$ (\textcolor{thresholdgreen}{green region}).
The CP threshold $\zeta_t$ adapts to score variability across the calibration set, becoming higher when successful trajectories exhibit greater variation (\eg, mid-execution) and lower when they are more consistent (\eg, initialization).
The failure score remains consistently low across successful trajectories and rises sharply at the onset of failure.

\clearpage

\clearpage

\section{Implementation Details} \label{sec:implementation}

We implement Hide-and-Seek in Python 3.10.19 with PyTorch 2.10.0~\cite{paszke2019pytorch} on a single NVIDIA A6000 GPU with 48GB of memory, where training typically completes in approximately 5 minutes.
We use the AdamW optimizer~\cite{loshchilov2017decoupled} with a learning rate of $5\times10^{-4}$ and L2 parameter regularization with weight $0.1$, applied uniformly across all simulation experiments.
The failure detector is a single-layer LSTM with hidden size 256, followed by a linear head and a sigmoid output layer.
We train for 300 epochs with a batch size of 64.
To address class imbalance, we use a balanced sampler that ensures each mini-batch contains an equal number of failure and success trajectories.

We adopt a warmup schedule where $\mathcal{L}_{\mathrm{inter}}$ is optimized alone for the first $N_{\mathrm{warm}}$ epochs until $t_{\mathrm{onset}}$ stabilizes as a reliable proxy for the failure onset, after which $\mathcal{L}_{\mathrm{intra}}$ is introduced.
We apply stop-gradient before computing $t_{\mathrm{onset}}$, \ie, $t_{\mathrm{onset}} = \arg\max_t \mathrm{sg}(s_t - s_{t-1})$ (implemented via \texttt{detach()}), so that gradients from $\mathcal{L}_{\mathrm{intra}}$ flow only through the score values $s_t$ in the pre/post-onset averages, not through the index selection.\footnote{We constrain $t_{\mathrm{onset}} \in [2, N_{\tau_f}]$ to ensure both summations in~\cref{eq:onset} are well-defined, although this boundary case barely occurs in practice.}
Hyperparameter search space and selection details are provided in~\cref{sec:hyperparameters}.
Importantly, we train a \textbf{single shared failure detector} jointly across multiple tasks and evaluate the same detector on the corresponding multi-task test set.

\subsection{Hyperparameters}
\label{sec:hyperparameters}

We adopt a two-stage hyperparameter tuning procedure for the simulation benchmarks.
We first perform a grid search over global hyperparameters shared across experiments—including optimizer settings and model architecture—on LIBERO-10 with OpenVLA, summarized in~\cref{tab:global_param}.
Given this configuration, we then tune benchmark- and policy-specific hyperparameters for each VLA model independently, as reported in~\cref{tab:model_param_sim}.
For real-robot experiments, we use a shared set of hyperparameters across both task categories, listed in~\cref{tab:model_param_real}.
All tuning is performed exclusively on seen tasks; the resulting models are deployed zero-shot on unseen tasks without further adaptation, and we report held-out evaluation results on both splits.
Baselines are tuned under the same protocol for fair comparison.

\begin{table}[h]
\centering
\caption{Hyperparameter search space applied uniformly across all experiments. The selected value is \textbf{bolded}.}
\label{tab:global_param}
\setlength{\tabcolsep}{6pt}
\renewcommand{\arraystretch}{1.05}
\begin{tabular}{@{} l c @{}}
\toprule
\textbf{Hyperparameter} & \textbf{Search Space} \\
\midrule
\multicolumn{2}{@{}l}{\textit{Optimizer}} \\
\quad Optimizer         & \{Adam, \textbf{AdamW}, SGD\} \\
\quad Learning rate         & \{1e-2, 5e-3, 1e-3, \textbf{5e-4}, 1e-4\} \\
\quad L2 regularization     & \{1, \textbf{1e-1}, 1e-2, 1e-3, 1e-4\} \\
\quad Training epoch  & \{100, \textbf{300}, 500, 1000\} \\
\quad Warmup epoch  & \{0, \textbf{50}, 100, 150\} \\
\midrule
\multicolumn{2}{@{}l}{\textit{Architecture}} \\
\quad LSTM hidden size       & \{128, \textbf{256}, 512\} \\
\quad Context length         & \{1, 4, 8, \textbf{16}, all\} \\

\bottomrule
\end{tabular}
\end{table}

\begin{table}[h!]
\centering
\caption{Hyperparameter search space and selected values per policy on simulation benchmarks.}
\label{tab:model_param_sim}
\setlength{\tabcolsep}{5pt}
\renewcommand{\arraystretch}{1.05}
\begin{threeparttable}
\resizebox{\columnwidth}{!}{%
\begin{tabular}{@{} l c ccc @{}}
\toprule
\textbf{Hyperparameter} & \textbf{Search Space} & \textbf{OpenVLA (LIBERO)} & $\boldsymbol{\pi_0}$ \textbf{(LIBERO)}  & $\boldsymbol{\pi_{0.5}}$ \textbf{(VLABench)} \\
\midrule
\multicolumn{5}{@{}l}{\textit{Temporal Aggregation}} \\
\quad Window size $w$          & \{1, 4, 5, 8, 16\}                              & 4 & 8$^\dagger$ & 5$^\dagger$ \\
\midrule
\multicolumn{5}{@{}l}{\textit{Loss}} \\
\quad Inter loss margin $m_r$        & \{0.1, 0.2, 0.3, 0.5, 1.0\}                 & 1.0 & 1.0 & 1.0 \\
\quad Intra loss margin $m_o$       & \{0.1, 0.2, 0.3, 0.5, 1.0\}                 & 0.2 & 0.5 & 0.2 \\
\quad Intra weight $\lambda$   & \{1e-3, 5e-3, 1e-2, 1e-1, 1\}               & 1e-1 & 5e-3 & 1e-1 \\
\bottomrule
\end{tabular}}
\begin{tablenotes}
    \small
    \item[$\dagger$] Set to match the policy's action chunk size (replanning horizon).
\end{tablenotes}
\end{threeparttable}
\end{table}

\begin{table}[h!]
\centering
\small
\caption{Hyperparameter search space for real-robot experiments. The selected values are \textbf{bolded}.}
\label{tab:model_param_real}
\setlength{\tabcolsep}{5pt}
\renewcommand{\arraystretch}{1.05}
\begin{tabular}{@{} l c @{}}
\toprule
\textbf{Hyperparameter} & \textbf{Search Space} \\
\midrule
\multicolumn{2}{@{}l}{\textit{Optimizer}} \\
\quad Optimizer         & \{Adam, \textbf{AdamW}, SGD\} \\
\quad Learning rate     &  \{1e-2,5e-3,1e-3,\textbf{5e-4},1e-4\} \\
\quad L2 regularization & \{1, 1e-1, \textbf{1e-2}, 1e-3, 1e-4\} \\
\quad Training epochs   & \{100, \textbf{300}, 500, 1000\} \\
\quad Warmup epochs     & \{0, \textbf{50}, 100, 150\} \\
\midrule
\multicolumn{2}{@{}l}{\textit{Architecture}} \\
\quad LSTM hidden size  & \{128, \textbf{256}, 512\} \\
\quad Context length    & \{1, 4, \textbf{8}, 16, all\} \\
\midrule
\multicolumn{2}{@{}l}{\textit{Temporal Aggregation}} \\
\quad Window size $w$   & \{1, 4, \textbf{8}, 16\} \\
\midrule
\multicolumn{2}{@{}l}{\textit{Loss}} \\
\quad Inter loss margin $m_r$  & \{0.1, 0.2, 0.3, 0.5, \textbf{1.0}\} \\
\quad Intra loss margin $m_o$  & \{0.1, \textbf{0.2}, 0.3, 0.5, 1.0\} \\
\quad Intra weight $\lambda$ & \{1e-3, \textbf{5e-3}, 1e-2, 1e-1, 1\} \\
\bottomrule
\end{tabular}
\end{table}

\subsection{Conformal Prediction}
\label{sec:conformal}
To convert per-step failure scores $s_t$ into binary alarms, we calibrate a time-varying threshold $\zeta_t$ using functional conformal prediction~\cite{diquigiovanni2021importance}.
Calibration is performed on a held-out set of $C$ successful trajectories $\mathcal{D}_{\text{cal}} = \{\tau^{(c)}\}_{c=1}^{C}$, padded to a common length $T$ by repeating the final score. Under exchangeability and a user-specified significance level $\alpha \in (0, 1)$, the resulting band guarantees that the trajectory-level false-alarm rate on new successful rollouts is bounded by $\alpha$.
We partition $\mathcal{D}_{\text{cal}}$ into two disjoint subsets $\mathcal{D}_{\text{cal}_A}$ and $\mathcal{D}_{\text{cal}_B}$ of sizes $N_1$ and $N_2$, respectively\footnote{In our implementation, $N_1$ and $N_2$ are 30\% and 70\% of the calibration set, respectively.}, where the first estimates the score profile of successful execution and the second calibrates the band width. On $\mathcal{D}_{\text{cal}_A}$, we compute the per-step mean failure score:
\begin{equation}
    \mu_t = \frac{1}{N_1} \sum_{i=1}^{N_1} s_t^{(i)}, \quad t = 1, \dots, T,
\end{equation}
and define a modulation function $\sigma(t)$ that captures the typical spread of successful trajectories around $\mu_t$. We adopt the adaptive variant from~\cite{diquigiovanni2021importance}:
\begin{equation}
    \sigma(t) = \max_{k \in \mathcal{H}} \left| s_t^{(k)} - \mu_t \right|,
\end{equation}
where $\mathcal{H} \subseteq [N_1]$ is the set of trajectories whose maximum residual falls within the $(1-\alpha)$-quantile of $\{\max_{t \in [T]} |s_t^{(m)} - \mu_t|\}_{m=1}^{N_1}$, excluding outliers; when $(N_1 + 1)(1-\alpha) > N_1$, we set $\mathcal{H} = [N_1]$.
For each calibration trajectory $j \in \mathcal{D}_{\text{cal}_B}$, we compute the maximum modulated deviation:
\begin{equation}
    D_j = \max_{t \in [T]} \frac{s_t^{(j)} - \mu_t}{\sigma(t)},
\end{equation}
using a one-sided formulation since failures manifest as upward deviations. Letting $h$ denote the $(1-\alpha)$-quantile of $\{D_j\}_{j=1}^{N_2}$, the time-varying threshold is:
\begin{equation}
    \zeta_t = \mu_t + h \cdot \sigma(t),
\end{equation}
which corresponds to $\zeta_t = \mu_t + b_t$ in~\cref{sec:monitor} with bandwidth $b_t = h \cdot \sigma(t)$. 
At runtime, a failure is declared at the earliest timestep $t$ for which $s_t \geq \zeta_t$.
As no single $\alpha$ is universally optimal, we report results averaged over $\alpha \in \{0.15, 0.20, 0.25\}$ in our main experiments.
For the accuracy--timeliness analysis in~\cref{fig:tradeoff}, we sweep $\alpha \in [0.01, 1.0]$ with a step size of $0.01$, plotting the corresponding average normalized detection time and balanced accuracy for each operating point, following~\cite{gu2025safe}. For visualization clarity, we truncate the x-axis at the maximum detection-time endpoint reached by our method.

We note that the false-alarm guarantee of functional conformal prediction holds under the assumption of exchangeability between calibration and test trajectories.
In the unseen-task setting, where calibration is performed using successful trajectories from seen tasks, this assumption holds approximately.
Since unseen tasks are unavailable at calibration time, this setup is the standard practice in prior runtime monitoring work~\cite{xu2025can,gu2025safe}. 
Empirically, we observe that failure score profiles on successful trajectories remain stable and consistent across seen and unseen tasks (see \cref{fig:score_trend_openvla,fig:score_trend_pi0}), suggesting that the calibrated thresholds transfer well in practice.
Consistent with this, our method maintains strong performance on unseen tasks (\cref{tab:vertical_models,tab:vlabench,tab:real_world}).

\subsection{Evaluation Metrics}
\label{sec:metrics}

We provide full definitions of the evaluation metrics for runtime failure detection. We treat failure trajectories as positives and successful trajectories as negatives; a true positive is recorded if the detector raises an alarm at any timestep within a failure trajectory.
An alarm is triggered when the detector score exceeds a time-varying threshold $\zeta_t$ obtained via functional conformal prediction.

\paragraph{Balanced Accuracy (bACC).}
Balanced accuracy equally weights positive and negative classes:
\begin{equation}
\text{bACC} = \frac{1}{2}(\text{TPR} + \text{TNR}),
\end{equation}
where TPR (true positive rate) is the fraction of failure trajectories correctly detected, and TNR (true negative rate) is the fraction of success trajectories with no false alarm. A true positive is counted if an alarm is raised at any timestep within a failure trajectory.

\paragraph{Weighted Accuracy (wACC).}
To account for class imbalance, weighted accuracy assigns weights based on the empirical class ratio:
\begin{equation}
\beta = \frac{\#\text{Successful Rollouts}}{\#\text{Rollouts}}, \quad
\text{wACC} = \beta \cdot \text{TNR} + (1-\beta) \cdot \text{TPR}.
\end{equation}
This formulation reflects deployment scenarios where class proportions may be imbalanced.

\paragraph{Time-weighted Accuracy (TWA).}
Time-weighted accuracy~\cite{romer2025failure} jointly captures correctness and timeliness of detection. 
Let  $t_i$ (i.e., detection time) denote the first alarm timestep for the $i$-th true positive trajectory, and let $T_i$ denote its trajectory length. Then:
\begin{equation}
    \text{TWA} = \frac{1}{2}\left(\frac{1}{|\mathcal{P}|}\sum_{i \in \mathcal{TP}}\left(1 - \frac{t_i}{T_i}\right) + \frac{|\mathcal{TN}|}{|\mathcal{N}|}\right).
\end{equation}
where $\mathcal{TP}$ and $\mathcal{TN}$ denote the sets of true positives and true negatives, respectively, and $\mathcal{P}, \mathcal{N}$ are the sets of positive and negative trajectories. Earlier detections yield higher scores, while delayed alarms are penalized proportionally to their detection time.

\section{Additional Ablation Studies} \label{sec:ablation_app}

\subsection{Failure Onset Proxy}
\label{sec:onset_app}

\begin{table}[h!]
\centering
\small
\caption{Effect of different proxy failure onset definitions in 
the intra-trajectory contrastive loss $\mathcal{L}_{\mathrm{intra}}$ 
on failure detection performance on LIBERO-10. Estimated Onset 
denotes the average normalized timestep of the selected proxy.}
\label{tab:onset_app}
\setlength{\tabcolsep}{5pt}
\renewcommand{\arraystretch}{1.05}
\resizebox{\columnwidth}{!}{%
\begin{tabular}{@{} l c c c c c c c @{}}
\toprule
\multicolumn{8}{c}{\textbf{Policy: OpenVLA (Success Rate: 51.0\%)}}\\
\midrule
& &
\multicolumn{3}{c}{\textbf{Seen Tasks}} &
\multicolumn{3}{c}{\textbf{Unseen Tasks}} \\
\cmidrule(lr){3-5}\cmidrule(lr){6-8}
\textbf{Proxy} & \textbf{Estimated Onset}
& bACC$\uparrow$ & wACC$\uparrow$ & TWA$\uparrow$
& bACC$\uparrow$ & wACC$\uparrow$ & TWA$\uparrow$ \\
\midrule
Oracle & 0.401
& $0.854^{\pm 0.027}$ & $0.872^{\pm 0.033}$ & $0.673^{\pm 0.028}$
& $0.836^{\pm 0.041}$ & $0.835^{\pm 0.068}$ & $0.668^{\pm 0.012}$ \\
$t_{\max}$ & 0.508
& $0.845^{\pm 0.022}$ & $0.839^{\pm 0.021}$ & $0.612^{\pm 0.020}$
& $0.803^{\pm 0.046}$ & $0.796^{\pm 0.040}$ & $0.591^{\pm 0.044}$ \\
\cellcolor{blue} $t_{\mathrm{onset}}$ & \cellcolor{blue} 0.417
& \cellcolor{blue} $0.852^{\pm 0.051}$
& \cellcolor{blue} $0.853^{\pm 0.052}$
& \cellcolor{blue} $0.660^{\pm 0.035}$
& \cellcolor{blue} $0.834^{\pm 0.036}$
& \cellcolor{blue} $0.828^{\pm 0.034}$
& \cellcolor{blue} $0.663^{\pm 0.010}$ \\
\midrule\midrule
\multicolumn{8}{c}{\textbf{Policy: $\mathbf{\pi_{0}}$ (Success Rate: 84.2\%)}}\\
\midrule
& &
\multicolumn{3}{c}{\textbf{Seen Tasks}} &
\multicolumn{3}{c}{\textbf{Unseen Tasks}} \\
\cmidrule(lr){3-5}\cmidrule(lr){6-8}
\textbf{Proxy} & \textbf{Estimated Onset}
& bACC$\uparrow$ & wACC$\uparrow$ & TWA$\uparrow$
& bACC$\uparrow$ & wACC$\uparrow$ & TWA$\uparrow$ \\
\midrule
Oracle & 0.478
& $0.853^{\pm 0.097}$ & $0.902^{\pm 0.093}$ & $0.686^{\pm 0.063}$
& $0.924^{\pm 0.022}$ & $0.981^{\pm 0.016}$ & $0.684^{\pm 0.046}$ \\
$t_{\max}$ & 0.546
& $0.859^{\pm 0.056}$ & $0.899^{\pm 0.076}$ & $0.613^{\pm 0.030}$
& $0.854^{\pm 0.092}$ & $0.848^{\pm 0.093}$ & $0.627^{\pm 0.062}$ \\
\cellcolor{blue} $t_{\mathrm{onset}}$ & \cellcolor{blue} 0.470
& \cellcolor{blue} $0.885^{\pm 0.041}$
& \cellcolor{blue} $0.926^{\pm 0.054}$
& \cellcolor{blue} $0.693^{\pm 0.021}$
& \cellcolor{blue} $0.892^{\pm 0.011}$
& \cellcolor{blue} $0.921^{\pm 0.044}$
& \cellcolor{blue} $0.705^{\pm 0.043}$ \\
\bottomrule
\end{tabular}}
\end{table}

A key empirical finding of our method is that training with $\mathcal{L}_{\mathrm{inter}}$ implicitly discovers the failure onset through $t_{\mathrm{onset}}$ without any temporal supervision.
To verify this, we compare three characteristic timesteps: the peak score timestep $t_{\mathrm{max}} = \arg\max_t s_t$, which is directly optimized by $\mathcal{L}_{\mathrm{inter}}$; the spike point $t_{\mathrm{onset}} = \arg\max_t (s_t - s_{t-1})$, which receives no direct gradient; and the GPT-5.2 annotated 
failure onset that serves as the oracle reference.
All timesteps are normalized by trajectory length.

As shown in~\cref{tab:onset_app}, $t_{\mathrm{max}}$ lags behind the oracle onset ($+0.107$ for OpenVLA, $+0.068$ for $\pi_0$), as it reflects the timestep of the most failure-indicative signal rather than the failure onset itself. Anchoring $\mathcal{L}_{\mathrm{intra}}$ at $t_{\mathrm{max}}$ therefore degrades both detection accuracy and timeliness: the loss mistakenly treats actions between the true onset and $t_{\mathrm{max}}$ as pre-onset references, suppressing the score rise that should occur at the actual onset and pushing detection later in the trajectory.

In contrast, $t_{\mathrm{onset}}$ closely approximates the oracle for both OpenVLA ($+0.016$) and $\pi_0$ ($-0.008$) despite receiving no direct gradient, and using it as the anchor achieves performance on par with the oracle itself and substantially better than $t_{\mathrm{max}}$. This confirms that the sharpest score transition aligns with the true failure onset, motivating our intra-trajectory onset contrast loss $\mathcal{L}_{\mathrm{intra}}$ to directly optimize the score separation around $t_{\mathrm{onset}}$.

\subsection{Thresholding Strategies}
\label{sec:threshold}

\begin{table}[h!]
\centering
\small
\caption{Effect of different thresholding strategies on failure detection performance on LIBERO-10. We report the mean and standard deviation over three random seeds.}
\label{tab:threshold_app}
\setlength{\tabcolsep}{5pt}
\renewcommand{\arraystretch}{1.05}
\resizebox{0.85\columnwidth}{!}{%
\begin{tabular}{@{} l c c c c c c @{}}
\toprule
\multicolumn{7}{c}{\textbf{Policy: OpenVLA (Success Rate: 51.0\%)}}\\
\midrule
& \multicolumn{3}{c}{\textbf{Seen Tasks}} &
\multicolumn{3}{c}{\textbf{Unseen Tasks}} \\
\cmidrule(lr){2-4}\cmidrule(lr){5-7}
\textbf{Thresholding} 
& bACC$\uparrow$ & wACC$\uparrow$ & TWA$\uparrow$
& bACC$\uparrow$ & wACC$\uparrow$ & TWA$\uparrow$ \\
\midrule
Fixed ($\zeta=0.5$) & $0.859^{\pm 0.035}$ & $0.864^{\pm 0.034}$ & $0.656^{\pm 0.024}$ & $0.792^{\pm 0.032}$ & $0.780^{\pm 0.026}$ & $0.625^{\pm 0.029}$ \\
Split CP        & $0.850^{\pm 0.030}$ & $0.842^{\pm 0.028}$ & $0.656^{\pm 0.011}$ & $0.833^{\pm 0.026}$ & $0.839^{\pm 0.022}$ & $0.658^{\pm 0.005}$ \\
\cellcolor{blue} Functional CP &\cellcolor{blue} $ 0.852^{\pm 0.051}$ & \cellcolor{blue} $ 0.853^{\pm 0.052}$ & \cellcolor{blue} $ 0.660^{\pm 0.035}$ & \cellcolor{blue} $ 0.834^{\pm 0.036}$ & \cellcolor{blue} $ 0.828^{\pm 0.034}$ & \cellcolor{blue} $ 0.663^{\pm 0.010}$ \\
\midrule\midrule
\multicolumn{7}{c}{\textbf{Policy: $\mathbf{\pi_{0}}$ (Success Rate: 84.2\%)}}\\
\midrule
& \multicolumn{3}{c}{\textbf{Seen Tasks}} &
\multicolumn{3}{c}{\textbf{Unseen Tasks}} \\
\cmidrule(lr){2-4}\cmidrule(lr){5-7}
\textbf{Thresholding} 
& bACC$\uparrow$ & wACC$\uparrow$ & TWA$\uparrow$
& bACC$\uparrow$ & wACC$\uparrow$ & TWA$\uparrow$ \\
\midrule
Fixed ($\zeta=0.5$) & $0.827^{\pm 0.101}$ & $0.730^{\pm 0.173}$ & $0.611^{\pm 0.056}$ & $0.707^{\pm 0.092}$ & $0.716^{\pm 0.108}$ & $0.609^{\pm 0.032}$ \\
Split CP & $0.729^{\pm 0.083}$ & $0.859^{\pm 0.070}$ & $0.605^{\pm 0.092}$ & $0.703^{\pm 0.097}$ & $0.890^{\pm 0.107}$ & $0.620^{\pm 0.093}$ \\
\cellcolor{blue} Functional CP & \cellcolor{blue}  $0.885^{\pm 0.041}$ & \cellcolor{blue}  $ 0.926^{\pm 0.054}$ & \cellcolor{blue}  $0.693^{\pm 0.021}$ &
\cellcolor{blue}  $0.892^{\pm 0.011}$ & \cellcolor{blue}  $0.921^{\pm 0.044}$ & \cellcolor{blue}  $0.705^{\pm 0.043}$ \\
\bottomrule
\end{tabular}}
\end{table}

We compare three thresholding strategies: (1)~a fixed threshold of $0.5$ requiring no calibration set, (2)~a time-invariant scalar threshold calibrated via split conformal prediction (Split CP)~\cite{angelopoulos2023conformal}, and (3)~a time-varying threshold from functional conformal prediction (Functional CP), which we adopt as our primary strategy. We report the mean across $\alpha \in \{0.15, 0.20, 0.25\}$.

We observe that our method is relatively robust to the choice of thresholding strategy under the OpenVLA policy, with only minor performance variations across methods.
In contrast, both fixed and Split~CP thresholds degrade noticeably under the $\pi_0$ policy, indicating higher sensitivity to threshold selection in this setting. 
Functional~CP consistently delivers more stable performance across both policies, suggesting improved robustness to policy-specific score distributions and better alignment with the sequential nature of the decision process.

\subsection{Layer-wise Ablation}

\begin{figure}[h!]
    \centering
    \begin{subfigure}[h]{0.48\linewidth}
        \centering
        \includegraphics[width=\linewidth]{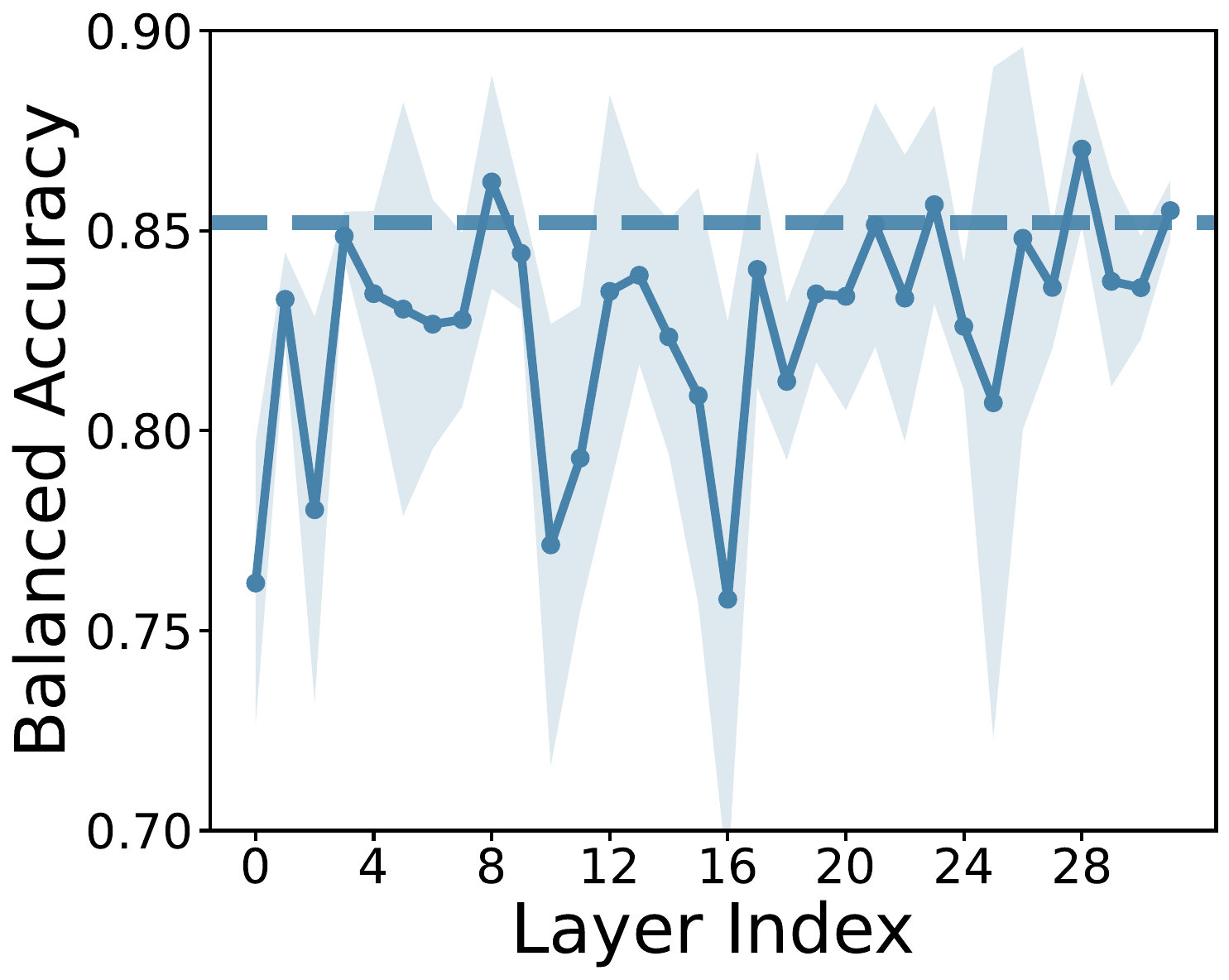}
        \caption{Seen tasks}
        \label{fig:layer_seen}
    \end{subfigure}
    \hfill
    \begin{subfigure}[h]{0.48\linewidth}
        \centering
        \includegraphics[width=\linewidth]{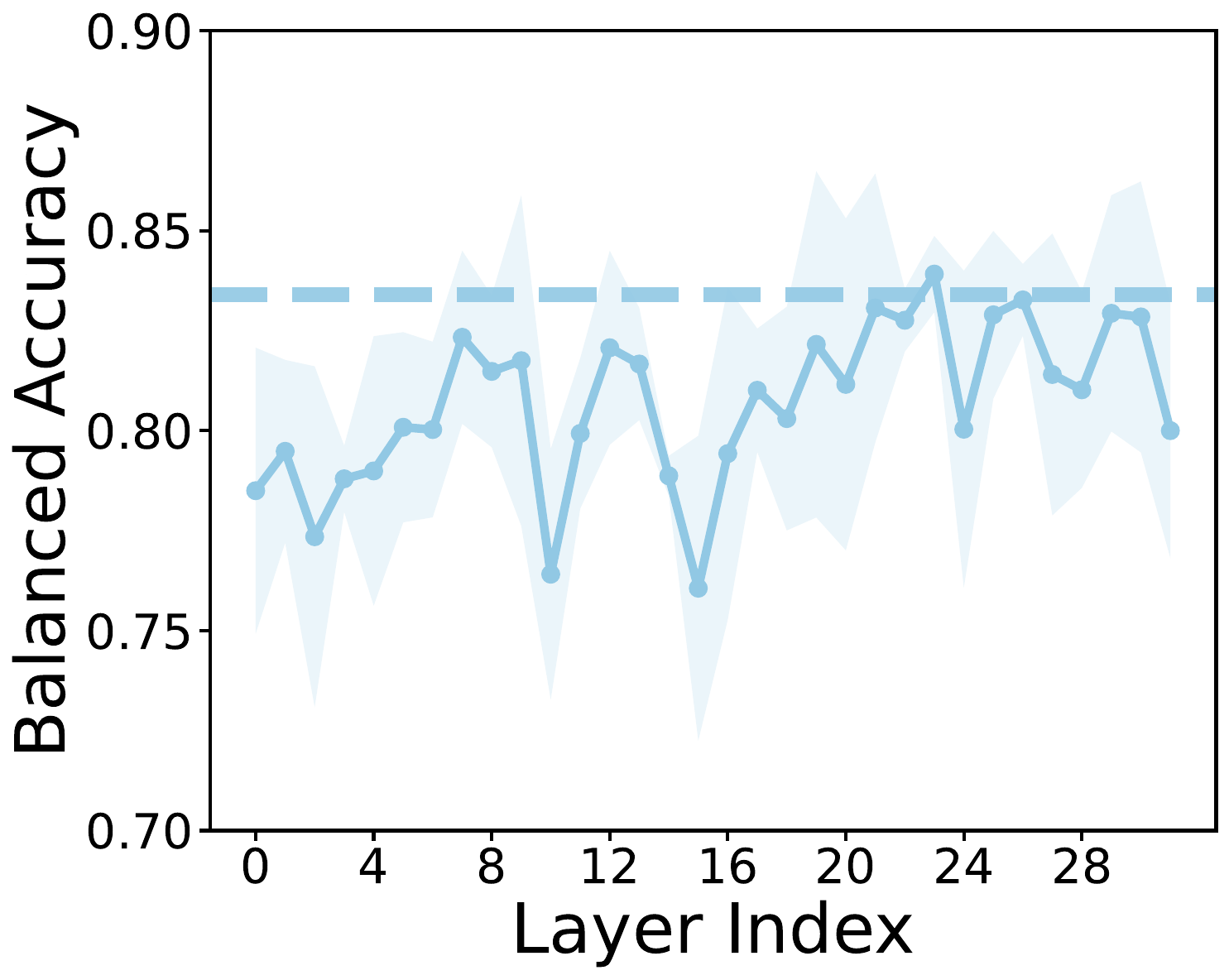}
        \caption{Unseen tasks}
        \label{fig:layer_unseen}
    \end{subfigure}
    \caption{
    Layer-wise ablation of OpenVLA internal representations for failure detection on LIBERO-10.
    We report balanced accuracy (bACC) across layer indices.
    The dashed line corresponds to using layer-averaged representations.
    }
    \label{fig:layer_ablation}
\end{figure}

We ablate which transformer layer of OpenVLA provides the most informative representation for failure detection by training Hide-and-Seek on each layer's embedding 
$h_t^{(\ell)} \in \mathbb{R}^d$ for $\ell \in \{0, 1, \ldots, 31\}$, 
and compare against our default of averaging across all layers.

As shown in~\cref{fig:layer_ablation}, performance varies across layers, but does not exhibit a sharp peak at any single layer.
Mid-to-late layers (e.g., layers $20$--$31$) generally yield stronger and more stable performance than early layers, consistent with the intuition that deeper representations capture more task-relevant abstractions.
However, the performance gap among well-performing layers is small (within $\sim\!3\%$ bACC), and the layer-averaged representation (dashed line) matches or exceeds the best individual layer on both seen ($85.2\%$) and unseen ($83.4\%$) tasks.
We therefore adopt layer averaging as our default since it eliminates the need for per-policy layer selection while providing consistent performance across both task splits.

It is worth noting that recent work in LLM/LVLM interpretability has reported that intermediate layers often encode the most semantically rich features for downstream tasks~\cite{skean2025layer, jiang2025devils}.
While our results show a similar mild advantage for mid-to-late layers, a more principled investigation of which layers encode failure-discriminative information across different VLA architectures and policy types is an interesting direction for future work.

\subsection{Inference-Time Policy Steering} \label{sec:policy_steering} 

\begin{wraptable}{r}{0.35\columnwidth} \centering \vspace{-16pt} \caption{ Success rate (\%) with verifier-guided action selection on the held-out LIBERO-10 test split. } \label{tab:policy_steering} \setlength{\tabcolsep}{6pt} \renewcommand{\arraystretch}{1.0} \begin{tabular}{lc} \toprule \textbf{Method} & \textbf{SR (\%)} \\ \midrule $\pi_0$ & 83.4 \\ \cellcolor{blue} $\pi_0$ + Hide-and-Seek & \cellcolor{blue}  \textbf{87.4} \\ \bottomrule \end{tabular} \vspace{-8pt} \end{wraptable}

Beyond runtime failure monitoring, we investigate whether Hide-and-Seek can also serve as an inference-time verifier for closed-loop policy improvement.
Specifically, on the held-out LIBERO-10 test split, we sample \(N\) candidate action chunks from the frozen $\pi_0$ policy at each decision step, score each candidate using the failure score predicted by Hide-and-Seek, and execute the candidate with the lowest predicted failure likelihood.
This procedure requires neither policy fine-tuning nor additional robot data.
As shown in~\Cref{tab:policy_steering}, verifier-guided action selection improves the task success rate from \(83.4\%\) to \(87.4\%\), corresponding to a \(+4.0\) percentage-point improvement over the base policy.
These results suggest that the failure signals discovered by Hide-and-Seek are useful not only for detecting impending failures, but also for distinguishing among candidate actions during execution. Thus, Hide-and-Seek can naturally support inference-time policy steering and may provide a useful signal for more general closed-loop mechanisms such as recovery, replanning, or human intervention.

\section{Additional Analysis}
\label{sec:analysis_app}

\subsection{VLA Latent Visualization} \label{sec:latent_vis}

\begin{figure}[h!]
\centering
\begin{subfigure}[h]{0.48\linewidth}
    \centering
    \includegraphics[width=\linewidth]{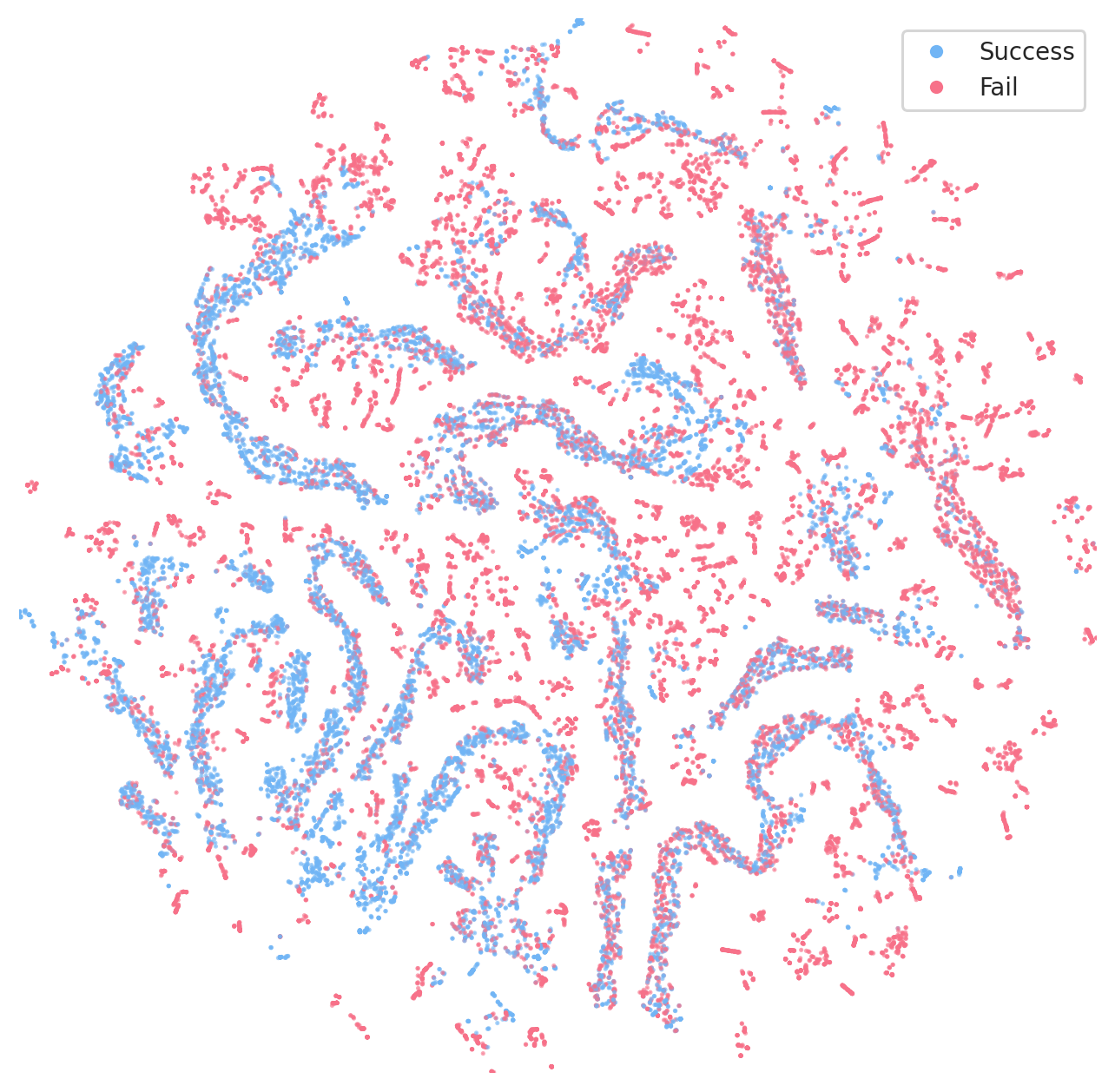}
    \caption{Uniform trajectory-level labeling}
    \label{fig:latent_uniform_openvla}
\end{subfigure}
\hfill
\begin{subfigure}[h]{0.48\linewidth}
    \centering
    \includegraphics[width=\linewidth]{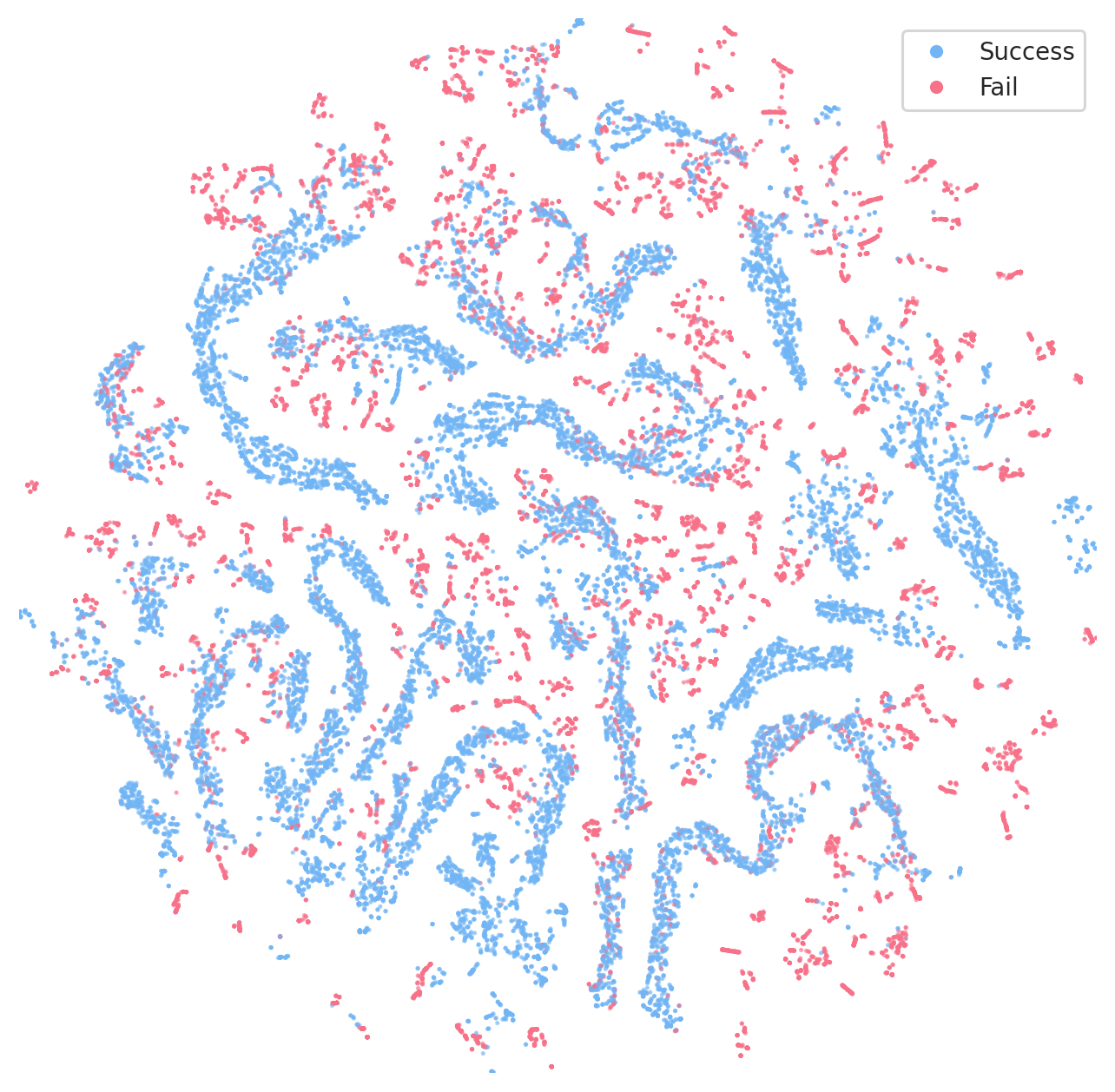}
    \caption{Onset-aware relabeling}
    \label{fig:latent_onset_openvla}
\end{subfigure}
\caption{
    \textbf{Visualization of OpenVLA action embeddings on 
    LIBERO-10.} 
    (a) Under uniform trajectory-level labeling, all timesteps in failure trajectories are assigned the failure label, mislabeling a substantial portion of normal actions as failures, resulting in significant overlap between success (blue) and failure (red) embeddings.
    (b) Relabeling pre-onset timesteps as success yields a 
    clearer separation between the two classes.
}
\label{fig:latent_vis}
\end{figure}

\begin{figure}[h!]
\centering
\begin{subfigure}[h]{0.48\linewidth}
    \centering
    \includegraphics[width=\linewidth]{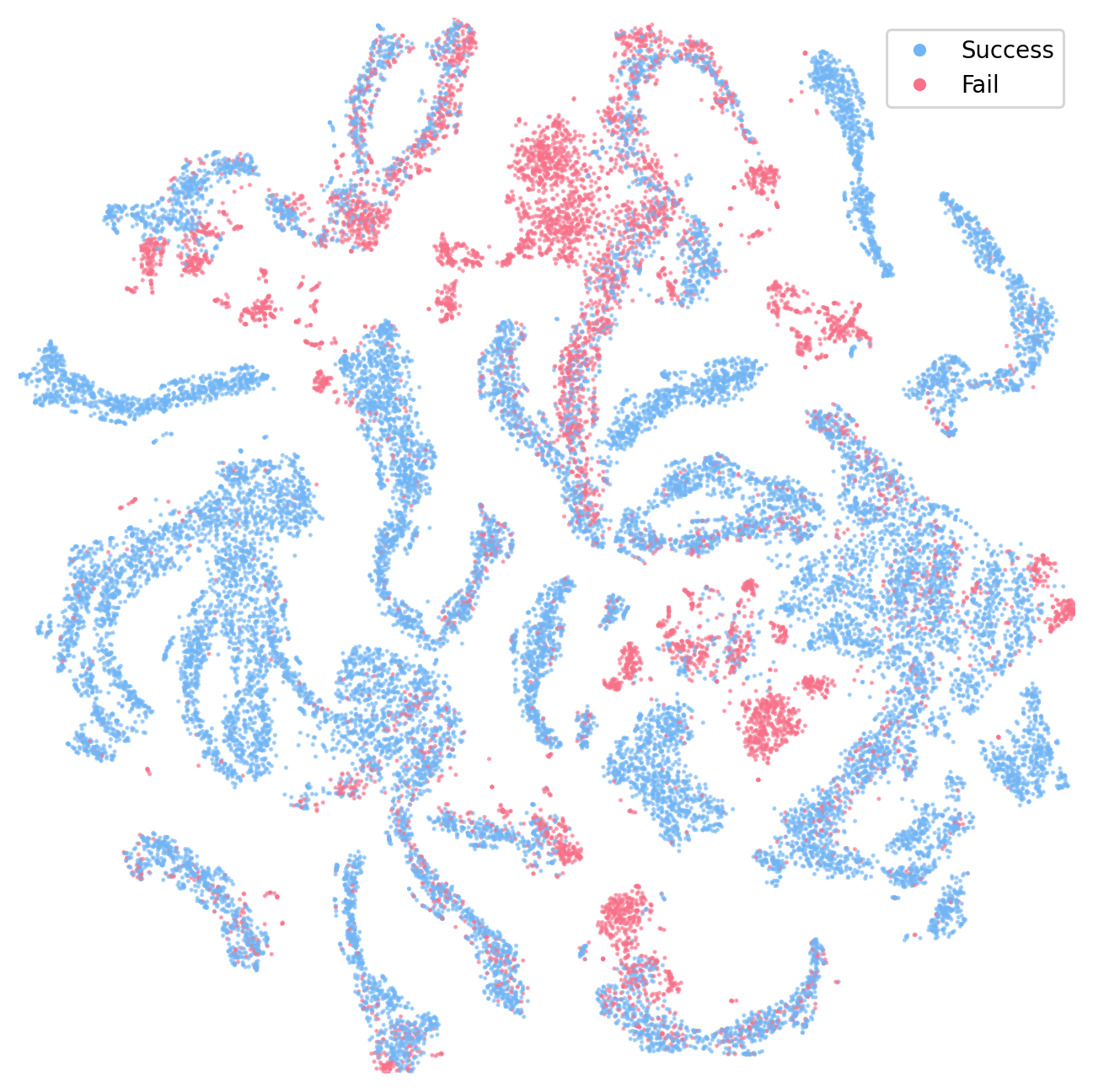}
    \caption{Uniform trajectory-level labeling}
    \label{fig:latent_uniform_pi0}
\end{subfigure}
\hfill
\begin{subfigure}[h]{0.48\linewidth}
    \centering
    \includegraphics[width=\linewidth]{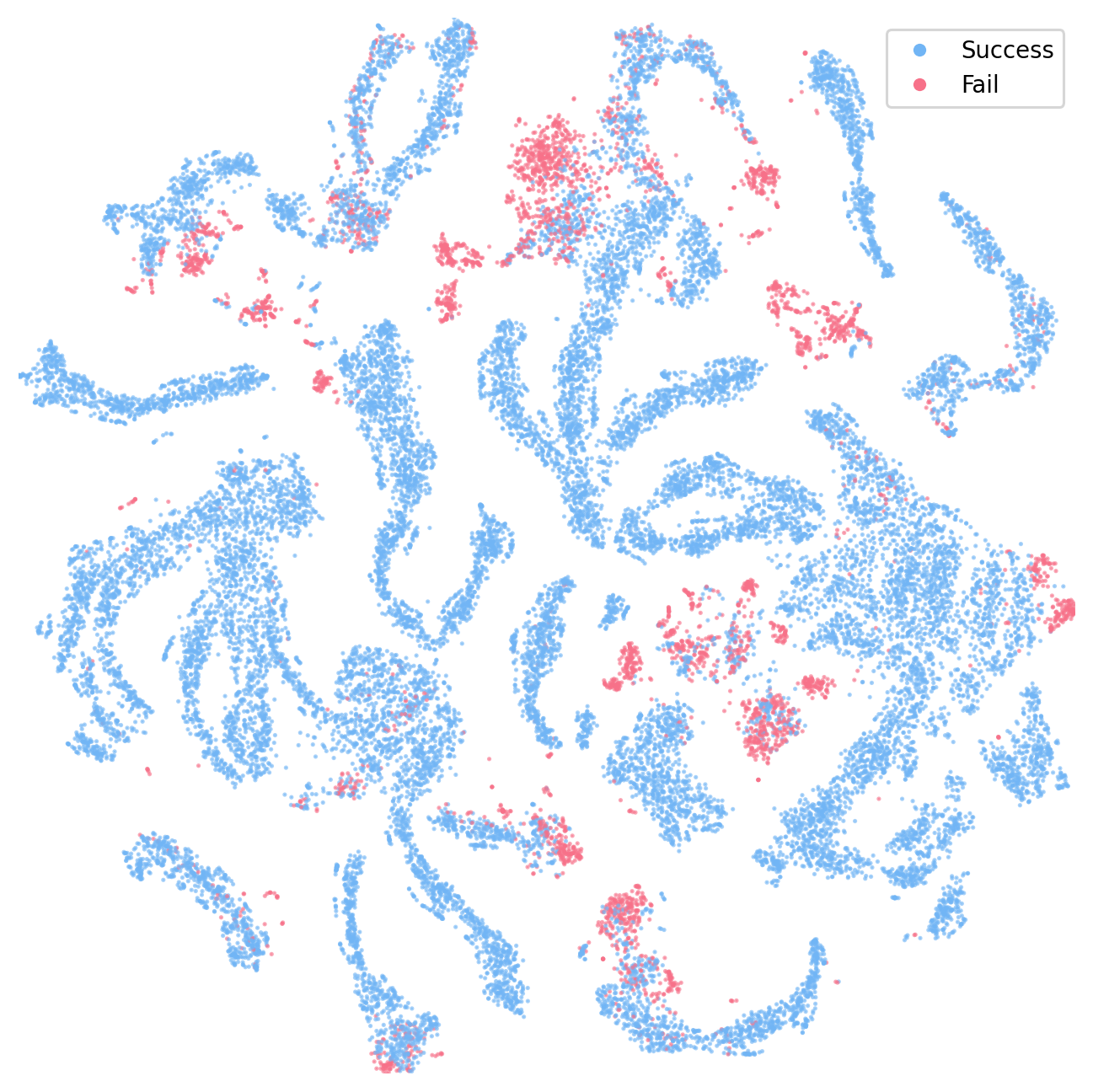}
    \caption{Onset-aware relabeling}
    \label{fig:latent_onset_pi0}
\end{subfigure}
\caption{
 \textbf{Visualization of $\pi_0$ action embeddings on 
    LIBERO-10.} 
}
\label{fig:latent_vis_pi0}
\end{figure}

To better understand the impact of supervision granularity, we visualize the internal VLA action embeddings described in~\cref{sec:tfa}.
We collect rollouts from all tasks in LIBERO-10 using OpenVLA and $\pi_0$, resulting in 500 episodes per policy (50 per task).
At each timestep, we extract a compact representation $h_t \in \mathbb{R}^d$: for OpenVLA, we average the action embedding across layer and DoF dimensions; for $\pi_0$, we use the hidden state before the velocity prediction head at the final denoising step.
We then project the embeddings to 2D using t-SNE~\cite{van2008visualizing} for visualization.

\Cref{fig:latent_uniform_openvla,fig:latent_uniform_pi0}
visualize the embeddings under the standard trajectory-level supervision, where all timesteps in a failure trajectory are assigned the same failure label.
We observe substantial overlap between success and failure embeddings, indicating that many normal actions are mislabeled as failure.
This suggests that uniformly propagating trajectory-level labels introduces significant noise and blurs the decision boundary.
To isolate failure-relevant signals, we refine the labels within failure trajectories using the failure onset annotations described in~\cref{sec:gpt5}, relabeling timesteps preceding the onset as success. As shown in~\cref{fig:latent_onset_openvla,fig:latent_onset_pi0}, this results in a much clearer separation between success and failure embeddings.

These observations provide empirical support for our formulation: uniformly propagating trajectory-level labels dilutes discriminative signals by assigning failure labels to predominantly correct actions, whereas isolating the failure phase yields a more separable representation.
This motivates our objective, which automatically discovers failure-indicative actions without requiring step-level annotations.

\subsection{Detection Timeliness Relative to Annotated Failure Onset}
\label{sec:detection_time_abl}

\begin{table}[h!]
\centering
\small
\caption{\textbf{Detection lag and balanced accuracy on LIBERO-10.} Detection lag $\Delta t = \frac{1}{|\mathcal{TP}|}\sum_i (t^{*}_i - t_{\text{det},i})/T_i$ is computed over true-positive failure trajectories; $\Delta t>0$ indicates early alarms and $\Delta t<0$ indicates delayed alarms. 
Results are averaged over $\alpha \in \{0.15, 0.20, 0.25\}$ and three random seeds.}
\label{tab:openvla_pi0_lag_bacc}
\setlength{\tabcolsep}{5pt}
\renewcommand{\arraystretch}{1.08}
\resizebox{\columnwidth}{!}{%
\begin{tabular}{lcccccccc}
\toprule
& \multicolumn{4}{c}{\textbf{OpenVLA}} & \multicolumn{4}{c}{$\boldsymbol{\pi_0}$} \\
\cmidrule(lr){2-5}\cmidrule(lr){6-9}
& \multicolumn{2}{c}{\textbf{Seen Tasks}} & \multicolumn{2}{c}{\textbf{Unseen Tasks}}
& \multicolumn{2}{c}{\textbf{Seen Tasks}} & \multicolumn{2}{c}{\textbf{Unseen Tasks}} \\
\cmidrule(lr){2-3}\cmidrule(lr){4-5}\cmidrule(lr){6-7}\cmidrule(lr){8-9}
\textbf{Method}
& \textbf{Lag} $\Delta t$ & \textbf{bACC}$\uparrow$
& \textbf{Lag} $\Delta t$ & \textbf{bACC}$\uparrow$
& \textbf{Lag} $\Delta t$ & \textbf{bACC}$\uparrow$
& \textbf{Lag} $\Delta t$ & \textbf{bACC}$\uparrow$ \\
\midrule
SAFE-LSTM
& $+0.035^{\pm 0.054}$ & $0.670^{\pm 0.100}$
& $-0.147^{\pm 0.087}$ & $0.686^{\pm 0.133}$
& $+0.196^{\pm 0.091}$ & $0.774^{\pm 0.024}$
& $+0.037^{\pm 0.102}$ & $0.609^{\pm 0.057}$ \\
SAFE-MLP
& $-0.057^{\pm 0.045}$ & $0.823^{\pm 0.036}$
& $-0.272^{\pm 0.189}$ & $0.775^{\pm 0.062}$
& $+0.121^{\pm 0.102}$ & $0.870^{\pm 0.038}$
& $-0.052^{\pm 0.154}$ & $0.801^{\pm 0.083}$ \\
\cellcolor{blue}  \textbf{Ours}
& \cellcolor{blue}  ${+0.042}^{\pm 0.043}$ &  \cellcolor{blue} ${0.852}^{\pm 0.051}$
& \cellcolor{blue}  ${+0.019}^{\pm 0.016}$ &  \cellcolor{blue} ${0.834}^{\pm 0.036}$
& \cellcolor{blue}  ${+0.078}^{\pm 0.030}$ &  \cellcolor{blue} ${0.885}^{\pm 0.041}$
& \cellcolor{blue}  ${+0.052}^{\pm 0.093}$ &  \cellcolor{blue} ${0.892}^{\pm 0.011}$ \\
\bottomrule
\end{tabular}}
\end{table}

A reliable failure detector should raise alarms close to the actual moment of failure (neither too early nor too late), enabling timely intervention without overly aggressive interruptions.
We quantify this via the normalized detection lag relative to the GPT-5.2 annotated failure onset $t^{*}$ (Section~\ref{sec:gpt5}):
\[
\Delta t = \frac{1}{|\mathcal{TP}|} \sum_{i \in \mathcal{TP}} \frac{t^{*}_i - t_{\text{det},i}}{T_i},
\]
where the detection time $t_{\text{det},i}$ is the first timestep at which $s_t \geq \zeta_t$, and $T_i$ is the trajectory length.
A positive $\Delta t>0$ indicates \textit{early} alarms, while a negative $\Delta t<0$ indicates \textit{delayed} alarms.

As shown in Table~\ref{tab:openvla_pi0_lag_bacc}, we compare against the strongest classifier-based baseline SAFE.
\textbf{Hide-and-Seek consistently raises alarms slightly before the annotated failure onset across all settings, despite being trained without step-level supervision.}
This enables timely intervention---such as pausing execution or requesting human assistance---before the policy becomes irrecoverably stuck or causes harm in real-world deployment.

In contrast, SAFE-MLP exhibits negative lag in three of four settings (e.g., $-0.272$ on OpenVLA unseen), indicating delayed detection after failure onset. We attribute this to its cumulative score formulation with a linearly increasing target weight (Eq.~\ref{eq:safe_mlp}), which biases scores toward later timesteps regardless of when failure occurs.
SAFE-LSTM produces more timely alarms on average, but with higher variance and lower accuracy.
Overall, Hide-and-Seek achieves both the highest bACC and consistently timely detection across all settings.

\subsection{Detector Architecture}
\label{sec:arch_abl}

\begin{table}[h!]
\centering
\small
\caption{Architecture ablation of the failure detector on LIBERO-10. All variants share the same training objective and differ only in backbone architecture. LR denotes the learning rate, and Reg denotes the L2 regularization weight.}
\label{tab:arch_ablation}
\setlength{\tabcolsep}{5pt}
\renewcommand{\arraystretch}{1.05}
\resizebox{\columnwidth}{!}{%
\begin{tabular}{@{} l c c c c c c c c @{}}
\toprule
\multicolumn{9}{c}{\textbf{Policy: OpenVLA (Success Rate: 51.0\%)}}\\
\midrule

& \multicolumn{2}{c}{\textbf{Hyperparameters}} &
\multicolumn{3}{c}{\textbf{Seen Tasks}} &
\multicolumn{3}{c}{\textbf{Unseen Tasks}} \\
\cmidrule(lr){2-3}\cmidrule(lr){4-6}\cmidrule(lr){7-9}

\textbf{Arch} &
\textbf{LR} & \textbf{Reg} &
\textbf{bACC}$\uparrow$ & \textbf{wACC}$\uparrow$ & \textbf{TWA}$\uparrow$ &
\textbf{bACC}$\uparrow$ & \textbf{wACC}$\uparrow$ & \textbf{TWA}$\uparrow$ \\
\midrule

MLP  & 1e-4 & 1e-1 &
$0.757^{\pm 0.027}$ & $0.757^{\pm 0.028}$ & $0.578^{\pm 0.027}$ &
$0.772^{\pm 0.064}$ & $0.742^{\pm 0.084}$ & $0.588^{\pm 0.020}$ \\

GRU  & 5e-4 & 1e-1 &
$0.844^{\pm 0.043}$ & $0.830^{\pm 0.049}$ & $0.639^{\pm 0.031}$ &
$0.822^{\pm 0.068}$ & $0.796^{\pm 0.072}$ & $0.623^{\pm 0.028}$ \\

Transformer & 5e-4 & 1e-4 &
$0.818^{\pm 0.021}$ & $0.818^{\pm 0.022}$ & $0.627^{\pm 0.021}$ &
$0.788^{\pm 0.077}$ & $0.753^{\pm 0.104}$ & $0.621^{\pm 0.030}$ \\

LSTM & 5e-4 & 1e-1 &
$\mathbf{0.852}^{\pm 0.051}$ & $\mathbf{0.853}^{\pm 0.052}$ & $\mathbf{0.660}^{\pm 0.035}$ &
$\mathbf{0.834}^{\pm 0.036}$ & $\mathbf{0.828}^{\pm 0.034}$ & $\mathbf{0.663}^{\pm 0.010}$ \\

\midrule\midrule

\multicolumn{9}{c}{\textbf{Policy: $\mathbf{\pi_{0}}$ (Success Rate: 84.2\%)}}\\
\midrule

& \multicolumn{2}{c}{\textbf{Hyperparameters}} &
\multicolumn{3}{c}{\textbf{Seen Tasks}} &
\multicolumn{3}{c}{\textbf{Unseen Tasks}} \\
\cmidrule(lr){2-3}\cmidrule(lr){4-6}\cmidrule(lr){7-9}

\textbf{Arch} &
\textbf{LR} & \textbf{Reg} &
\textbf{bACC}$\uparrow$ & \textbf{wACC}$\uparrow$ & \textbf{TWA}$\uparrow$ &
\textbf{bACC}$\uparrow$ & \textbf{wACC}$\uparrow$ & \textbf{TWA}$\uparrow$ \\
\midrule

MLP  & 5e-4 & 1e-3 &
$0.770^{\pm 0.087}$ & $0.842^{\pm 0.093}$ & $0.576^{\pm 0.079}$ &
$0.714^{\pm 0.041}$ & $0.729^{\pm 0.201}$ & $0.563^{\pm 0.017}$ \\

GRU  & 1e-4 & 1e-1 &
$0.868^{\pm 0.090}$ & $0.892^{\pm 0.061}$ & $0.686^{\pm 0.059}$ &
$\mathbf{0.916}^{\pm 0.024}$ & $\mathbf{0.972}^{\pm 0.027}$ & $0.696^{\pm 0.033}$ \\

Transformer & 1e-4 & 1e-4 &
$0.836^{\pm 0.067}$ & $0.914^{\pm 0.043}$ & $0.638^{\pm 0.085}$ &
$0.811^{\pm 0.034}$ & $0.853^{\pm 0.120}$ & $0.646^{\pm 0.016}$ \\

LSTM & 5e-4 & 1e-1 &
$\mathbf{0.885}^{\pm 0.041}$ & $\mathbf{0.926}^{\pm 0.054}$ & $\mathbf{0.693}^{\pm 0.021}$ &
$0.892^{\pm 0.011}$ & $0.921^{\pm 0.044}$ & $\mathbf{0.705}^{\pm 0.043}$ \\

\bottomrule
\end{tabular}}
\end{table}

We provide detailed results and hyperparameter settings for the backbone architecture ablation (summarized in~\cref{tab:architecture}) in~\cref{tab:arch_ablation}.
We compare our LSTM-based design against three common alternatives: an MLP, a GRU~\cite{chung2014empirical}, and a single-layer Transformer~\cite{vaswani2017attention}. 
All variants share the same input features, training objective, and schedule, differing only in the backbone architecture.
The MLP performs noticeably worse, highlighting the importance of temporal context for detecting failures in VLA policies. 

In contrast, all context-aware architectures (GRU, Transformer, LSTM) achieve substantially stronger performance, indicating that our method generalizes across sequence models rather than relying on a specific backbone.
Among them, the LSTM performs best, achieving TWA gains of $2.1$ and $3.3$ points over GRU and Transformer on seen tasks, and $4.0$ and $4.2$ points on unseen tasks, respectively, on OpenVLA. 
Overall, these results suggest that while our approach is broadly compatible with context-aware architectures, the LSTM provides the best performance in our setting.

\section{Benchmark Details}
\label{sec:benchmark}

\paragraph{Training and evaluation protocol.}
For both simulation (\cref{sec:sim}) and real-robot (\cref{sec:real}) experiments, we train the failure detector on the training set and evaluate on a held-out evaluation set.
For conformal prediction, we calibrate thresholds using held-out successful trajectories: (1) for seen-task evaluation, we use a calibration set from seen tasks; (2) for unseen-task evaluation, we use successful trajectories from the seen-task evaluation split as the calibration set.
All tuning is performed on seen tasks; \textbf{models are deployed zero-shot on unseen tasks} without further adaptation, and we report held-out results on both splits. 
Results are averaged over three random seeds, each with a different random partition of tasks into seen and unseen subsets. 
Dataset statistics are provided in~\cref{tab:benchmark_stats}.

\begin{table}[h!]
\centering
\small
\caption{Benchmark statistics for each split.}
\label{tab:benchmark_stats}
\setlength{\tabcolsep}{5pt}
\renewcommand{\arraystretch}{1.05}
\begin{tabular}{@{} l ccc cccc @{}}
\toprule
& \multicolumn{3}{c}{\textbf{Number of Tasks}} & \multicolumn{4}{c}{\textbf{Number of Rollouts}} \\
\cmidrule(lr){2-4}\cmidrule(lr){5-8}
\textbf{Benchmark} & \textbf{Seen} & \textbf{Unseen} & \textbf{Total} & \textbf{Train} & \textbf{Calib Seen} & \textbf{Eval Seen} & \textbf{Eval Unseen} \\
\midrule
LIBERO       & 7 & 3 & 10 & 210 & 35 & 105  & 150 \\
VLABench     & 7 & 3 & 10 & 210 & 35 & 105  & 150 \\
Real Cube    & 3 & 1 & 4 & 72 & 12 & 36  & 40 \\
Real Kitchen & 3 & 1 & 4 & 72 & 12 & 36  & 40 \\
\bottomrule
\end{tabular}
\end{table}

\subsection{Simulation Setup}
\label{sec:sim}

We evaluate on two simulation benchmarks.
\textbf{LIBERO}~\cite{liu2023libero} is a widely adopted benchmark for VLA evaluation; we use the LIBERO-10 suite, which consists of long-horizon tasks with diverse objects, layouts, and language instructions.
We evaluate OpenVLA~\cite{kim2024openvla} and $\pi_0$~\cite{black2024pi_0} using fine-tuned checkpoints released by the original authors. 
\textbf{VLABench}~\cite{zhang2025vlabench} is a recent benchmark covering diverse manipulation tasks; we evaluate $\pi_{0.5}$~\cite{intelligence2025pi_} fine-tuned on its primary tasks.
For both simulation benchmarks, we randomly hold out 3 out of 10 tasks as unseen for evaluation.

\subsection{Real-Robot Setup}
\label{sec:real}

Real-robot experiments are conducted on a \textbf{UFactory xArm~6} with 8 tasks grouped into two categories (\textbf{CUBE} and \textbf{KITCHEN}) of 4 tasks each, listed in~\cref{tab:real_tasks}, with sequential execution snapshots visualized in~\cref{tab:real_robot_sequences}.
We collect 25 demonstrations per task to fine-tune $\pi_{0.5}$~\cite{intelligence2025pi_} using LoRA~\cite{hu2022lora}.
For each category, we collect diverse rollouts by randomizing robot initial states and object configurations, and \textit{all failure trajectories arise naturally from the policy's own execution}—we apply no human-induced perturbations or adversarial interventions, ensuring the failure distribution reflects the policy's genuine deployment behavior.
This yields 98 successful and 62 failure trajectories for CUBE, and 102 successful and 58 failure trajectories for KITCHEN.
Within each category, we designate 3 tasks as seen and 1 task as unseen for evaluating generalization.

\begin{table}[h!]
\centering
\small
\caption{List of tasks used in the real-world experiments.}
\label{tab:real_tasks}
\setlength{\tabcolsep}{8pt}
\renewcommand{\arraystretch}{1.1}
\begin{tabular}{@{} c l l @{}}
\toprule
\textbf{Task} & \textbf{Category} & \textbf{Instruction} \\
\midrule
1 & \multirow{4}{*}{Cube}    & Place the yellow cube into the container. \\
2 &                          & Place the green cube into the container. \\
3 &                          & Place both the blue cube and the green cube into the container. \\
4 &                          & Stack the red cube on top of the blue cube. \\
\midrule
5 & \multirow{4}{*}{Kitchen} & Place the duck doll in the bowl. \\
6 &                          & Place the carrot in the bowl. \\
7 &                          & Place both the carrot and the duck in the bowl. \\
8 &                          & Place the carrot in the bowl, then place the bowl on the plate. \\
\bottomrule
\end{tabular}
\end{table}

{
\setlength{\tabcolsep}{2pt}
\begin{longtable}{@{}*{8}{>{\centering\arraybackslash}p{0.115\textwidth}}@{}}
\caption{Sequential execution snapshots for real-robot tasks. 
Each task is shown as 8 ordered frames (left-to-right). Tasks are grouped into two categories: CUBE (Tasks 1--4) and KITCHEN (Tasks 5--8).}
\label{tab:real_robot_sequences} \\

\toprule
\rowcolor{gray!20}
\multicolumn{8}{@{}c@{}}{%
  \rule{0pt}{1.4em}\textsc{\textbf{Category 1: Cube}}\rule[-0.6em]{0pt}{0pt}%
} \\
\midrule
\endfirsthead

\toprule
\midrule
\endhead

\bottomrule
\endfoot

\taskblock{Task 1. Place the yellow cube into the container.}
{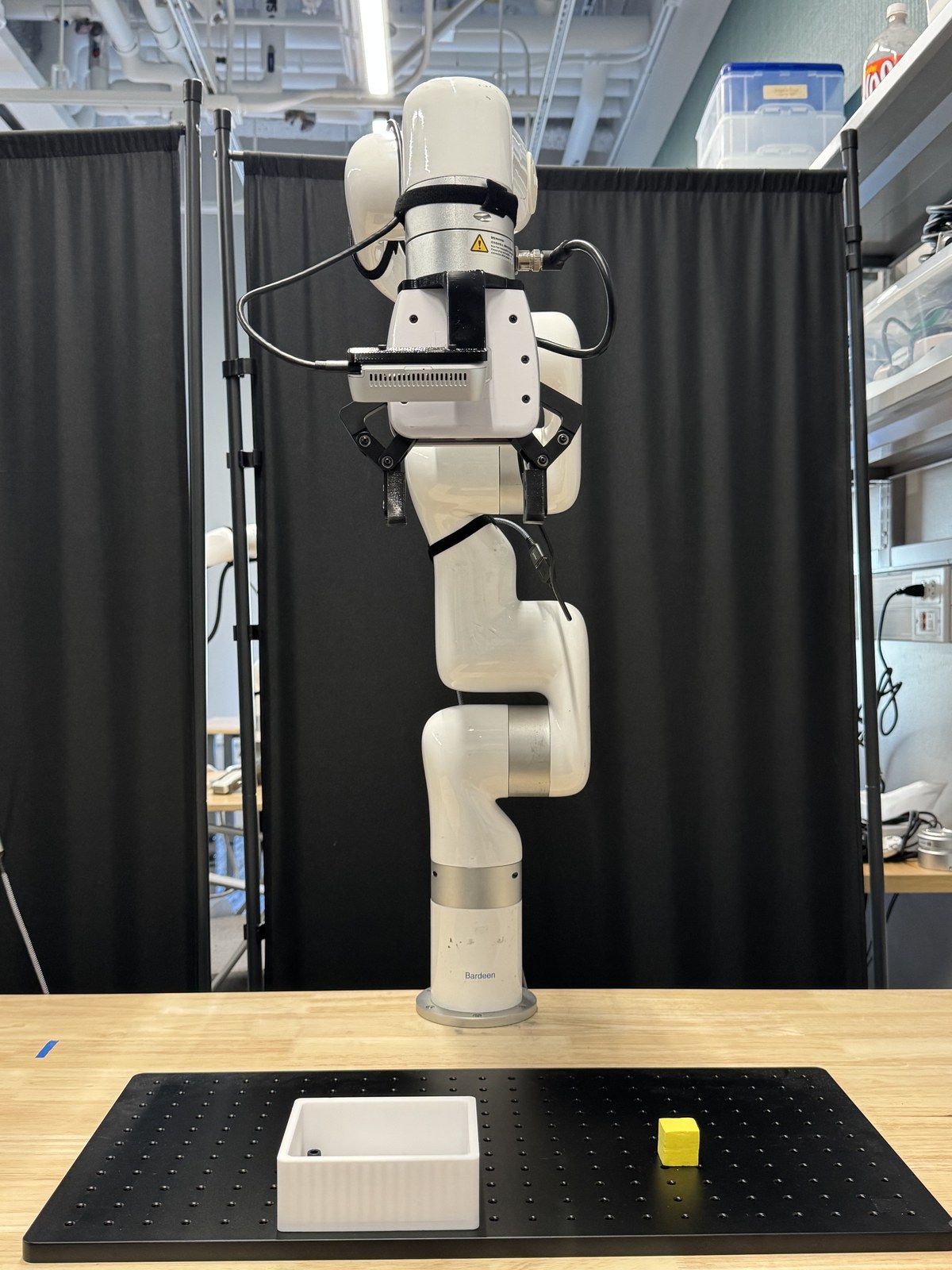}
{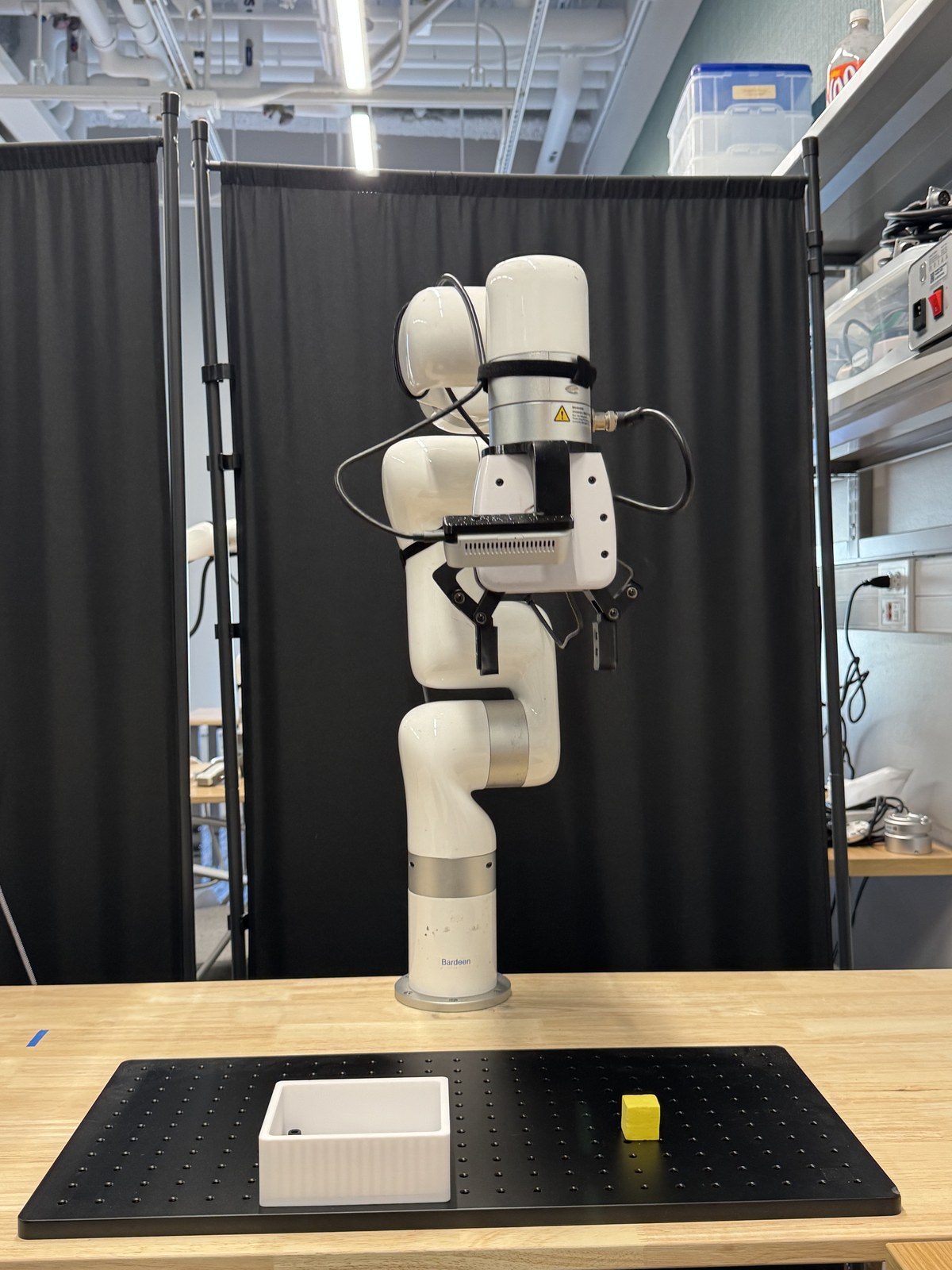}
{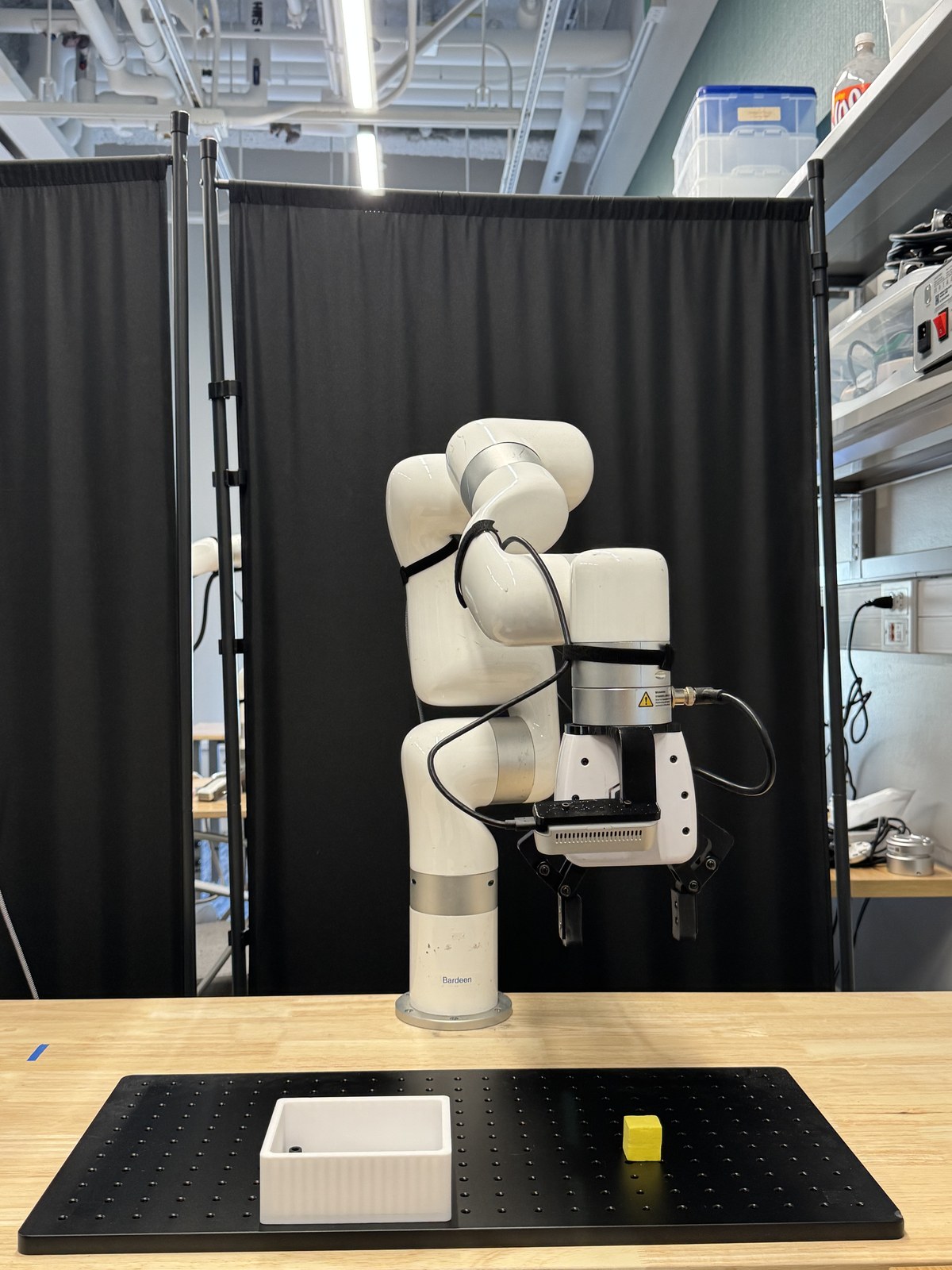}
{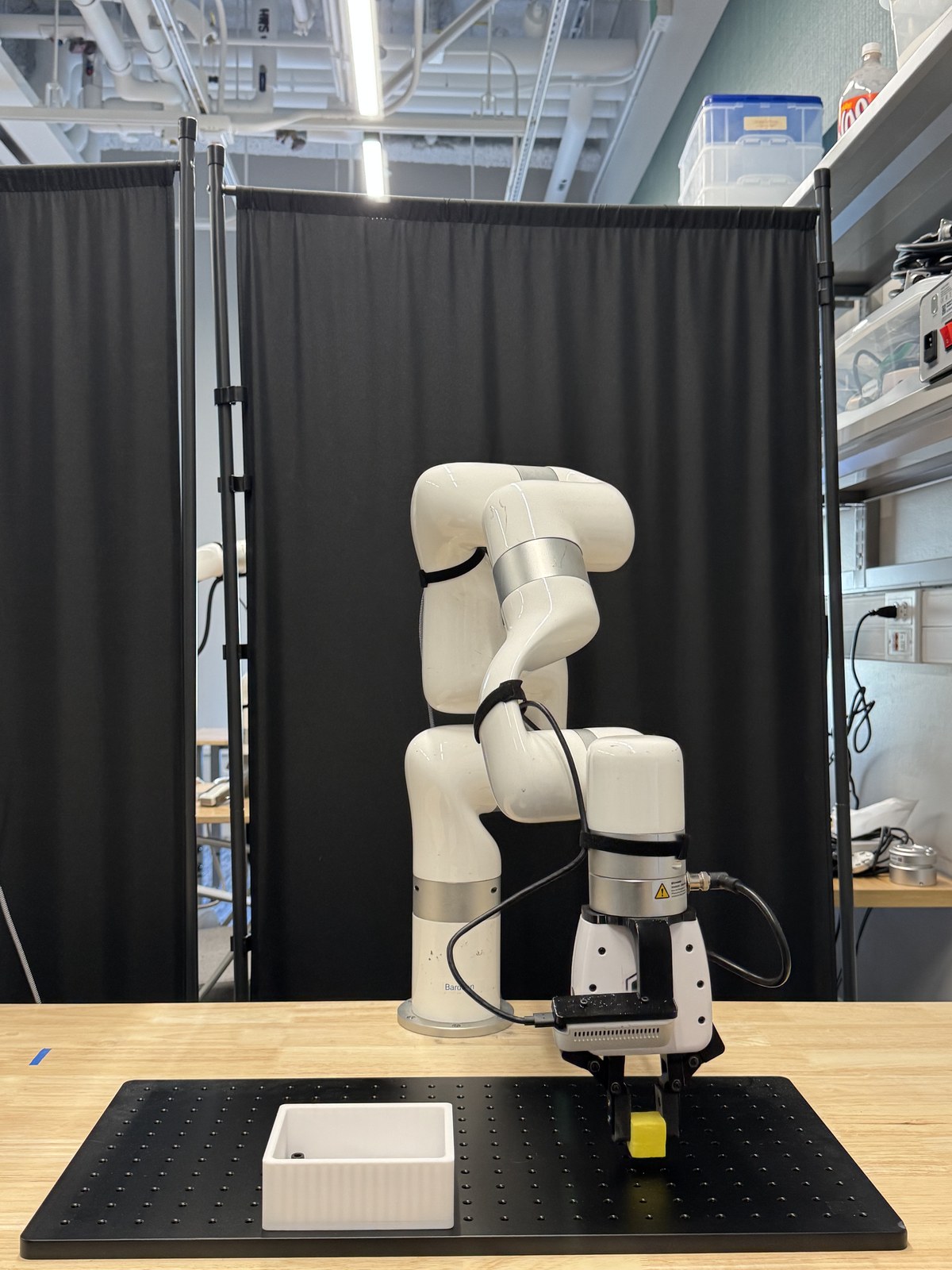}
{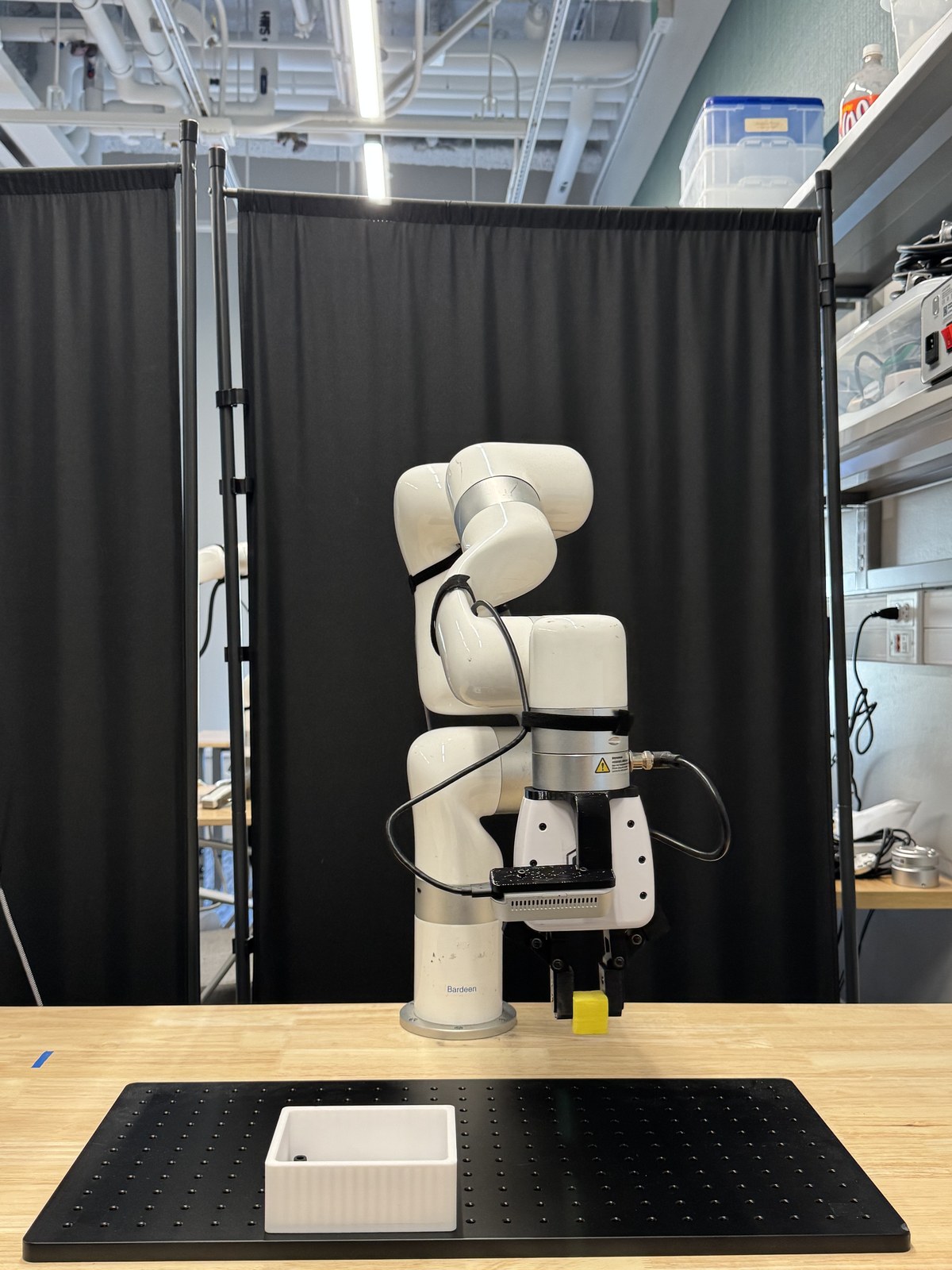}
{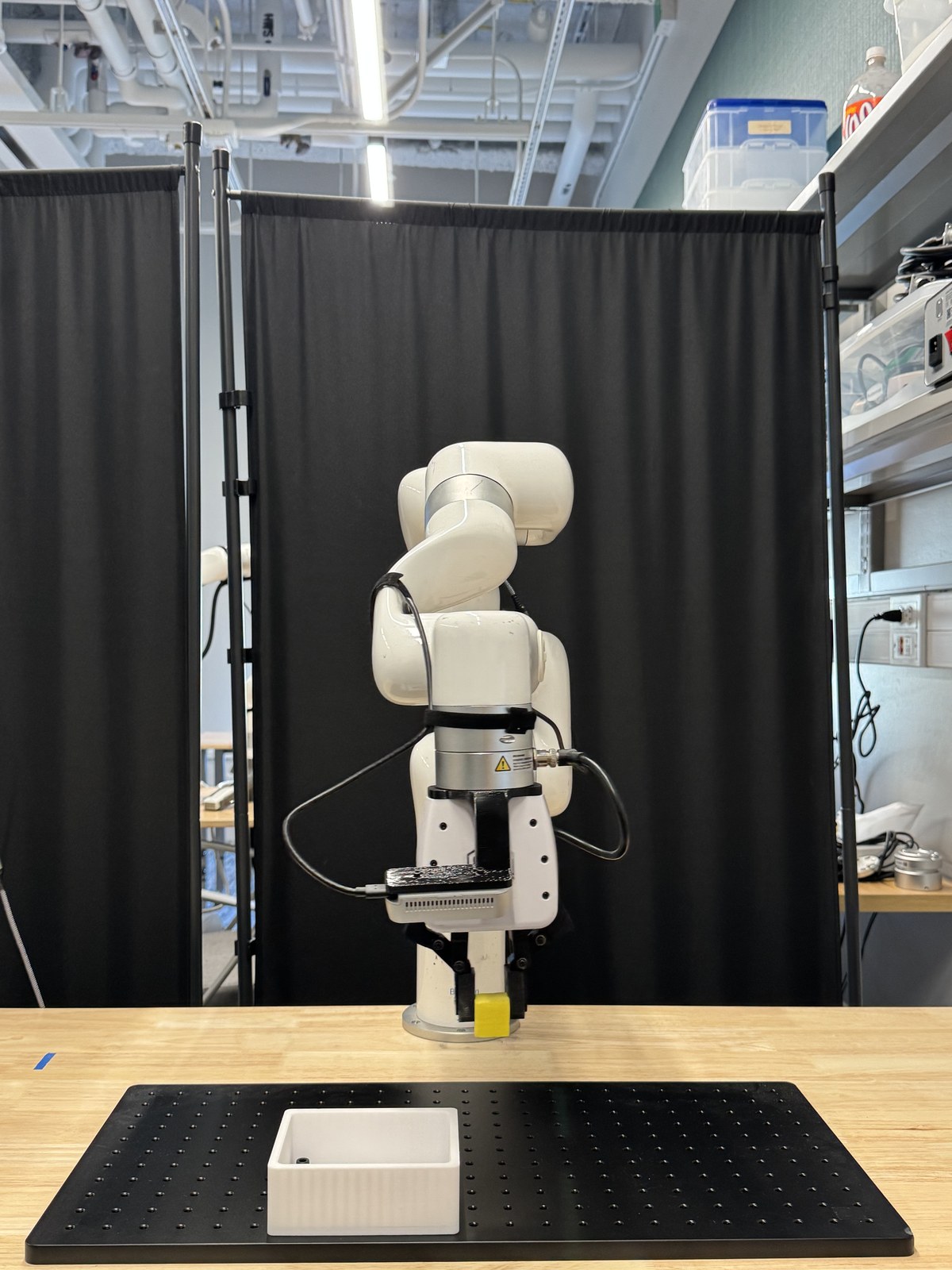}
{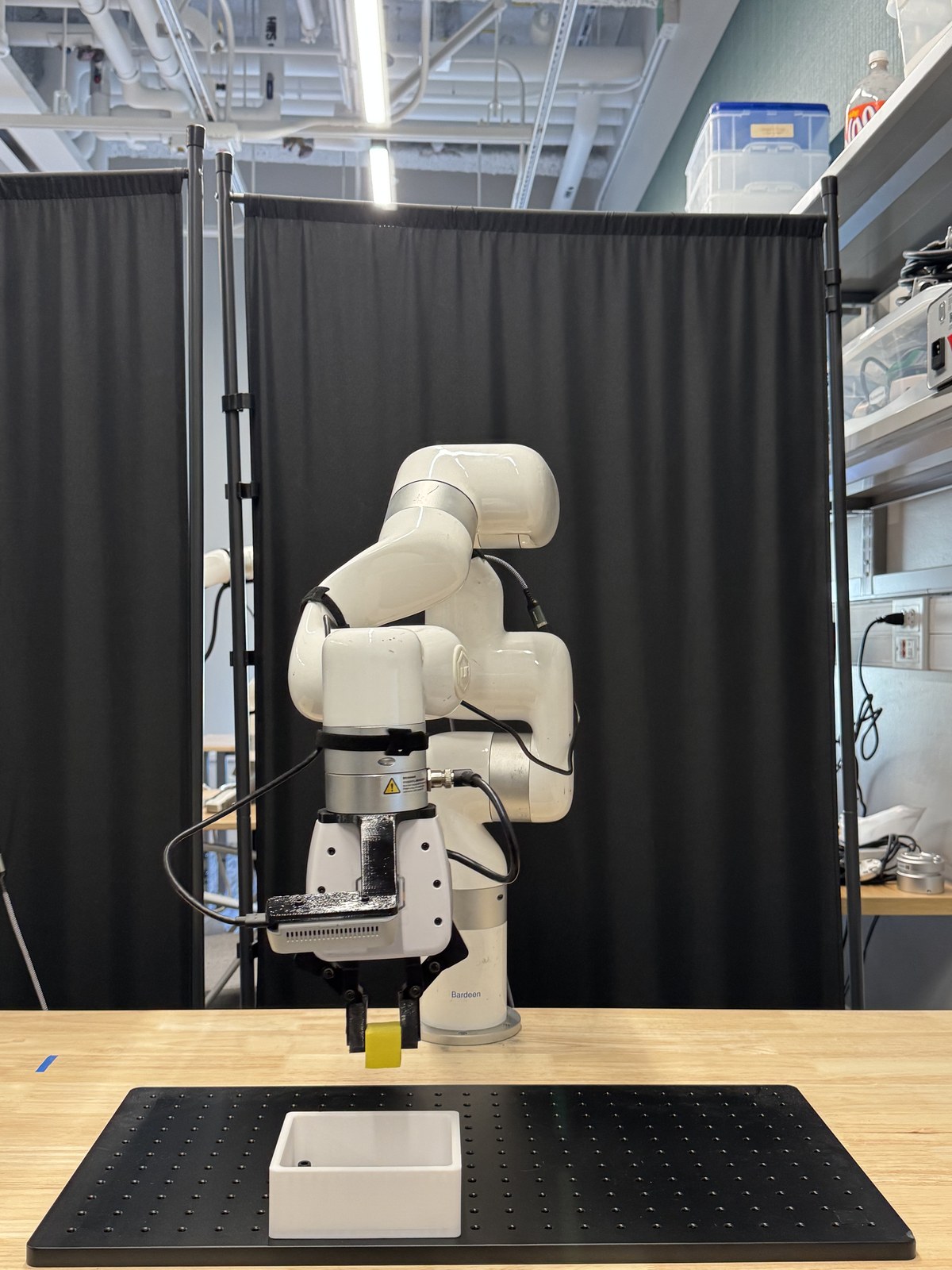}
{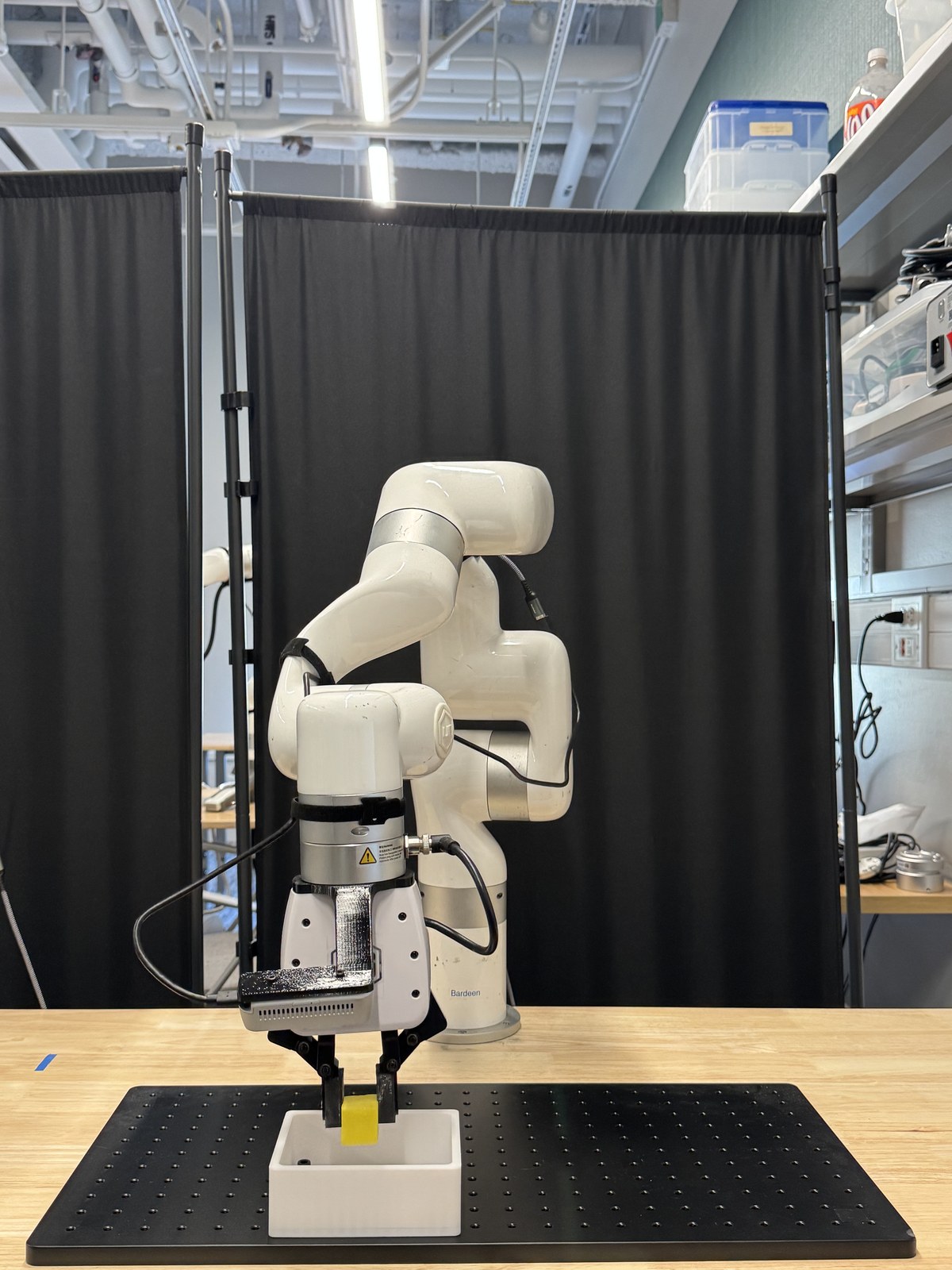}
\addlinespace[0.6em]

\taskblock{Task 2. Place the green cube into the container.}
{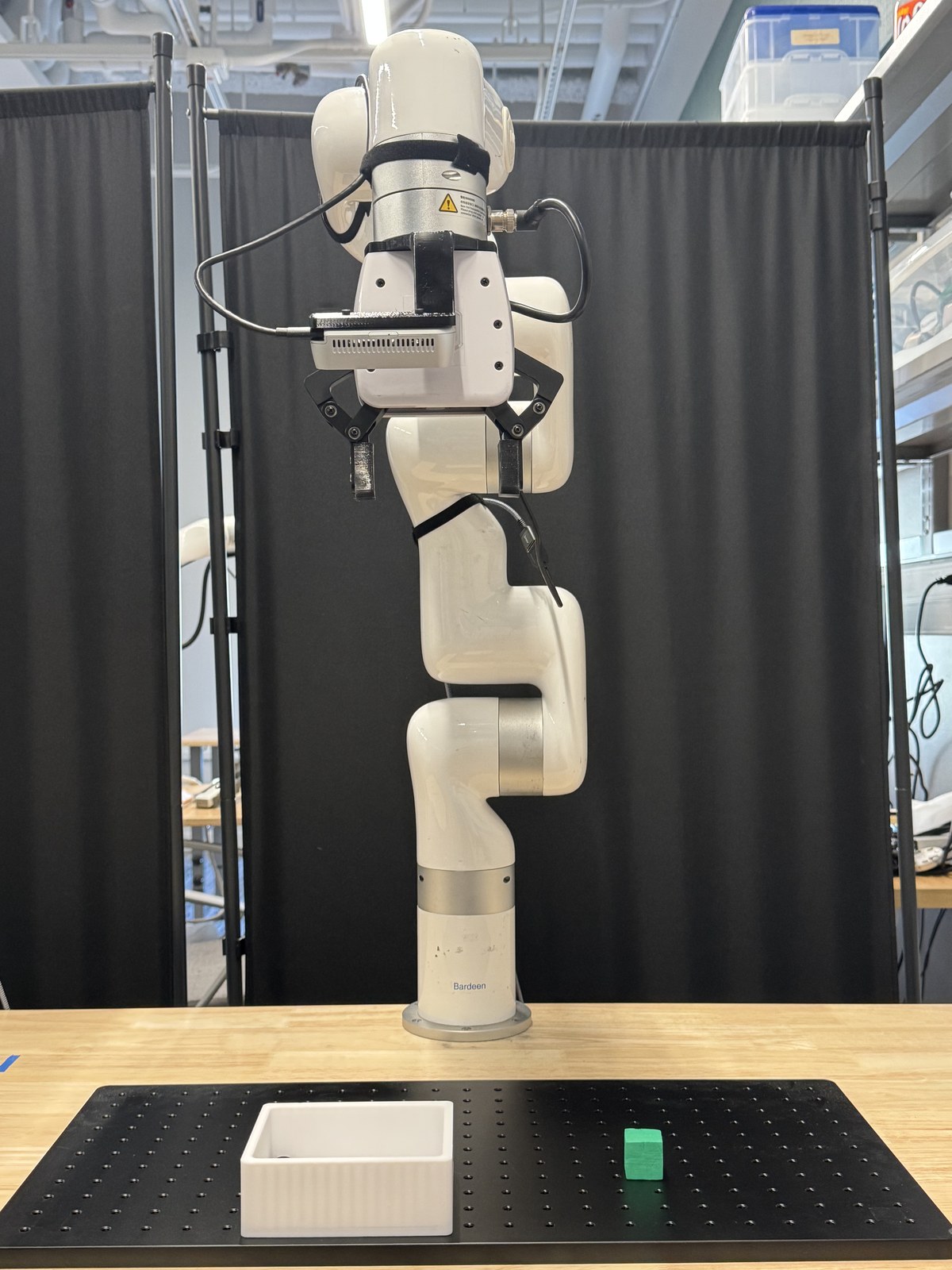}
{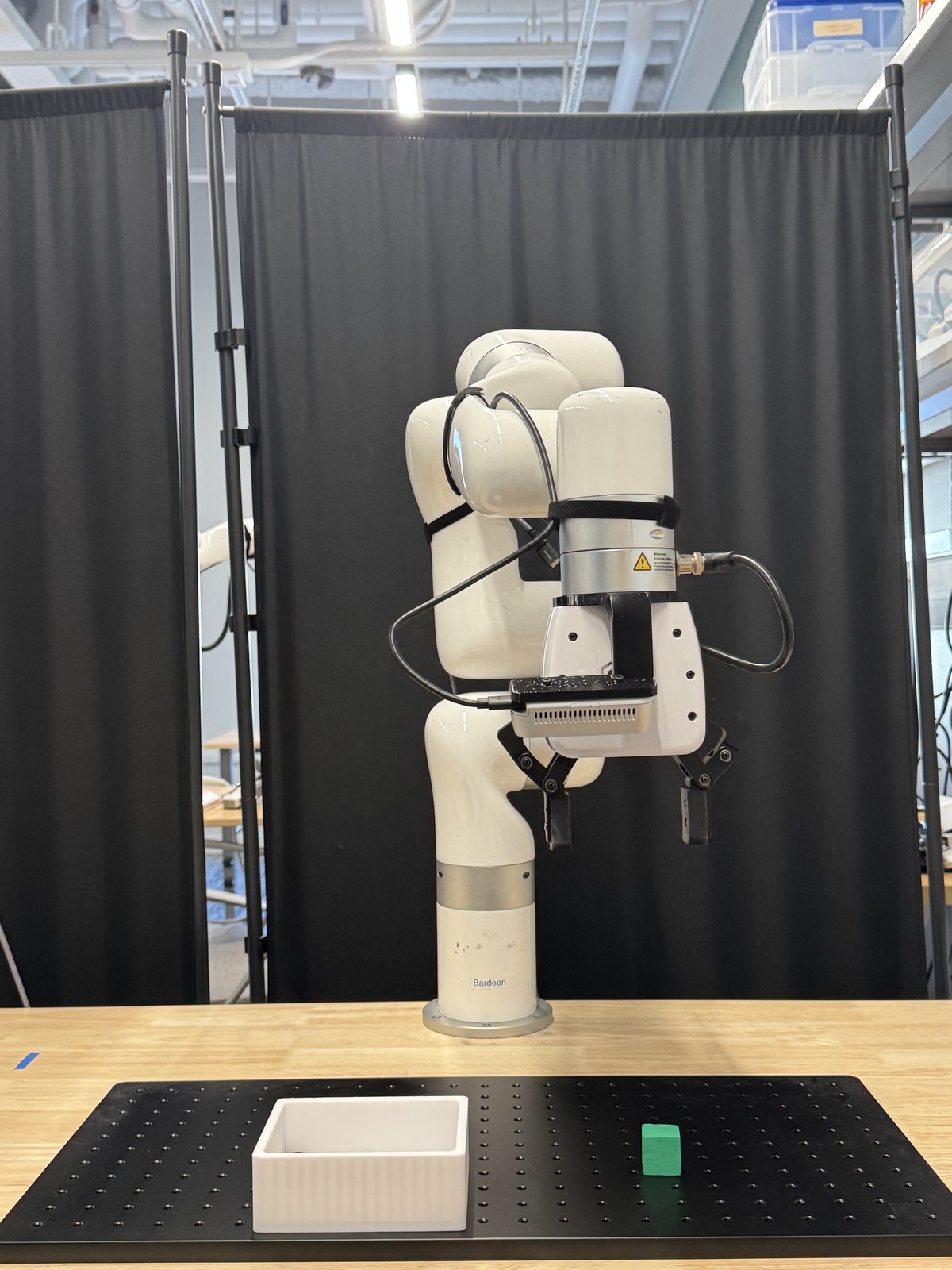}
{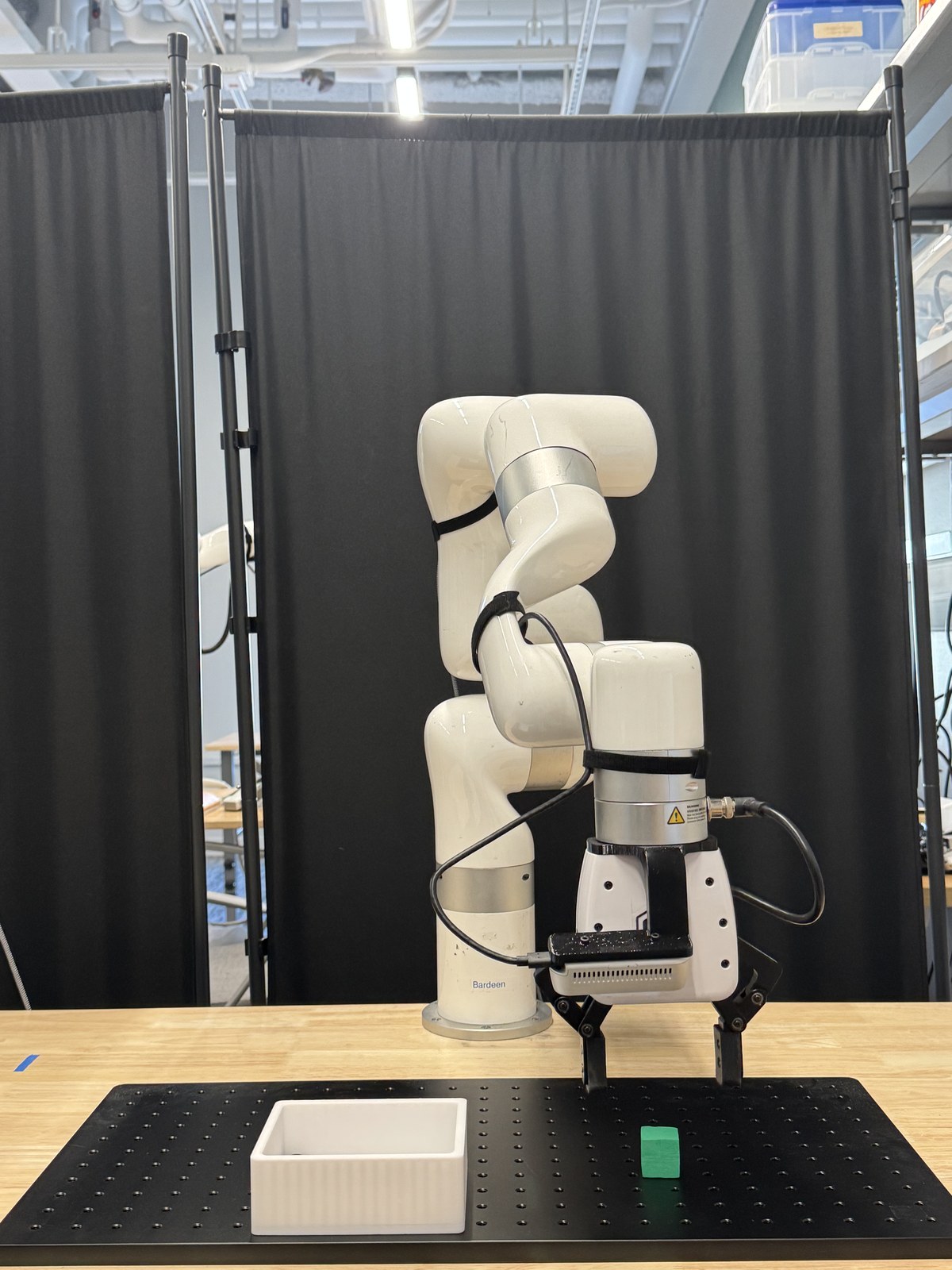}
{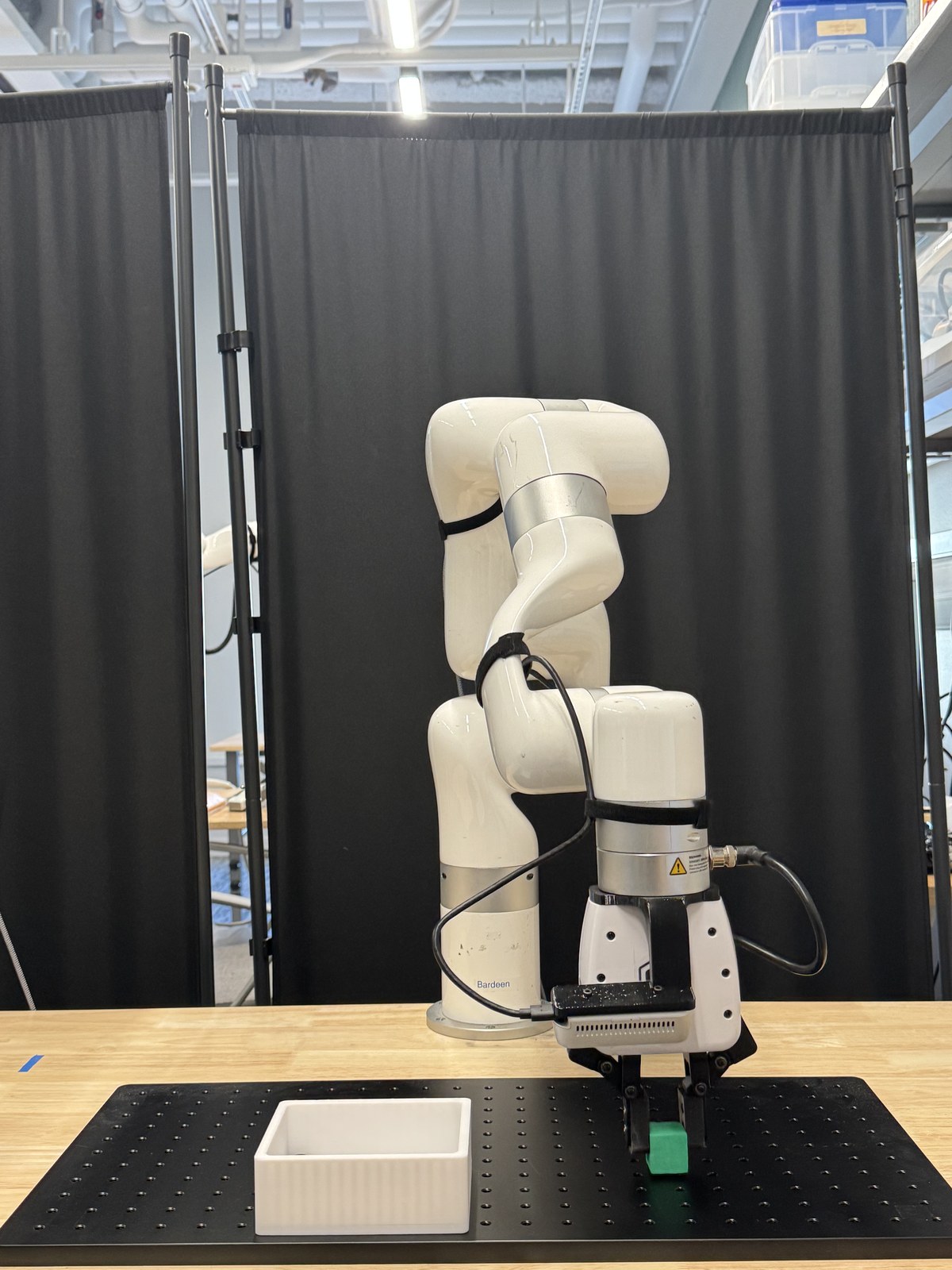}
{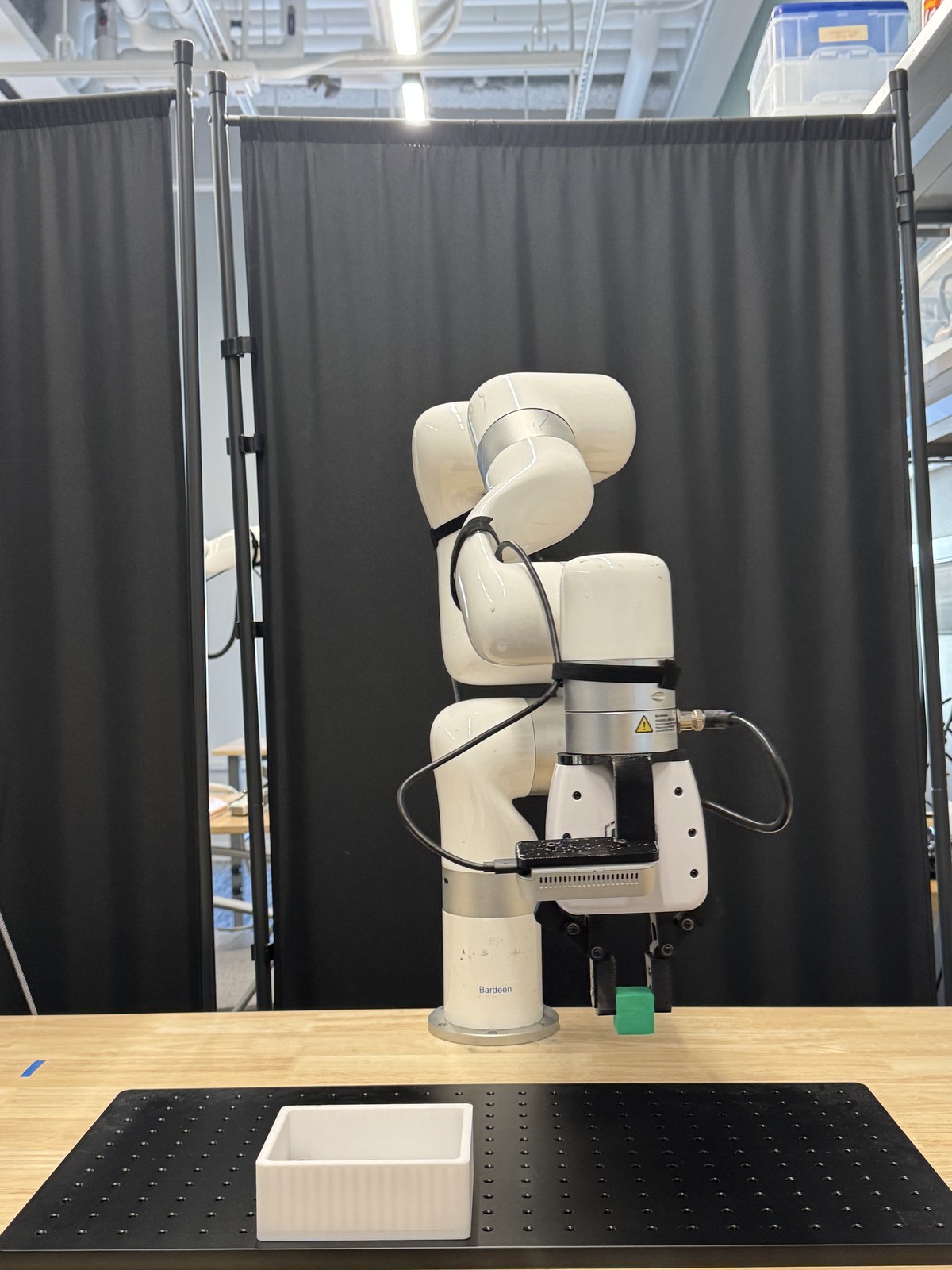}
{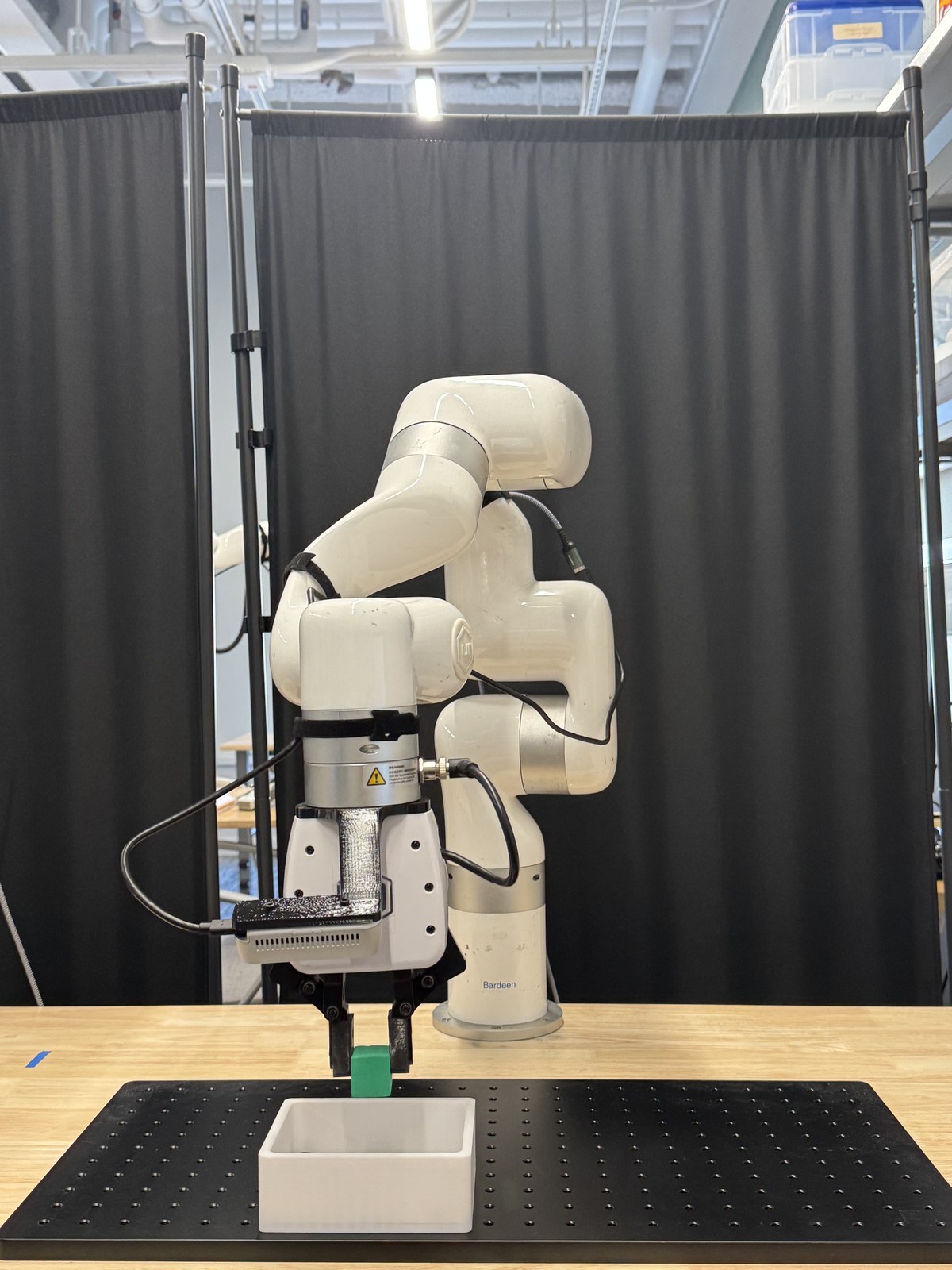}
{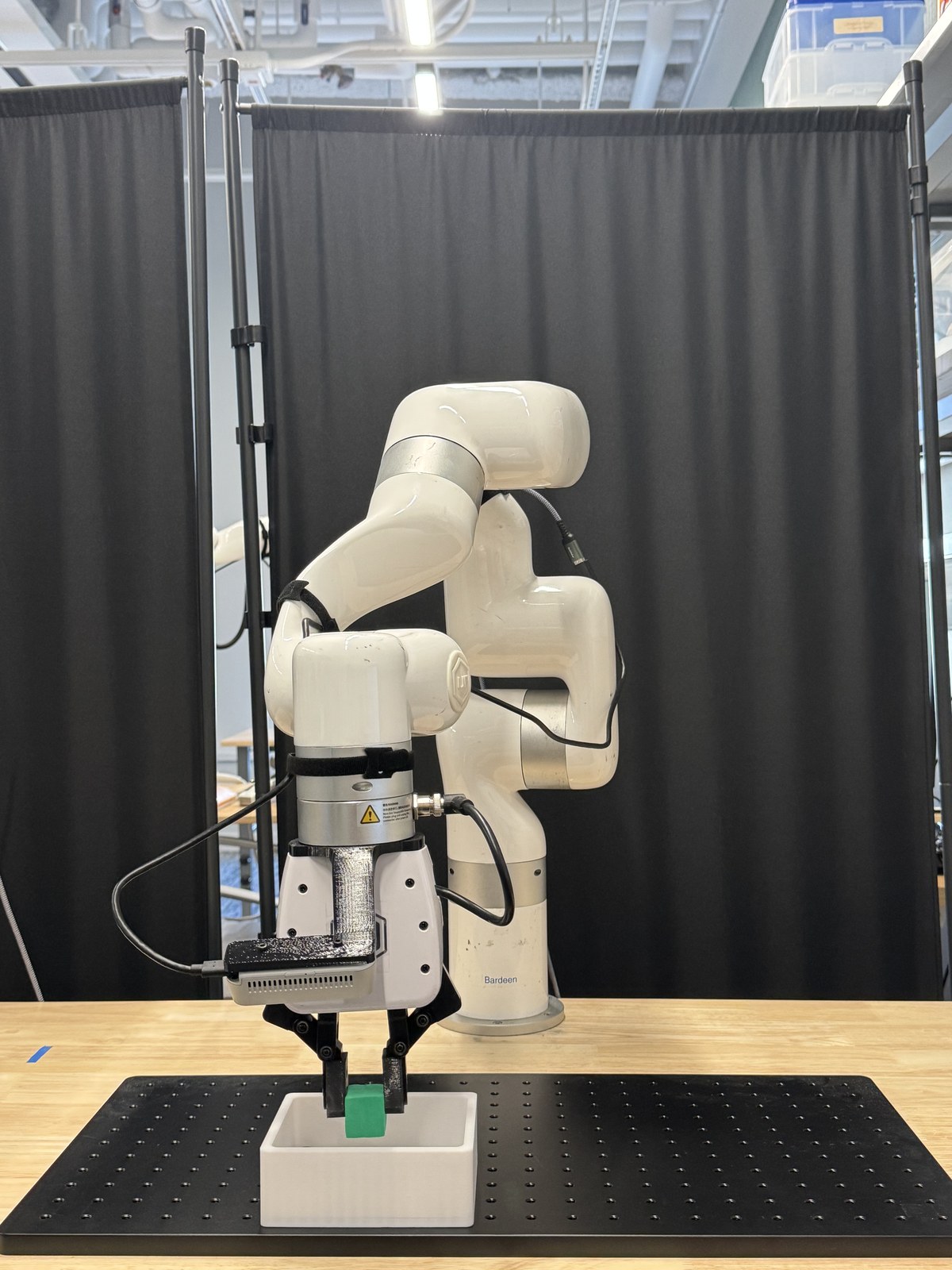}
{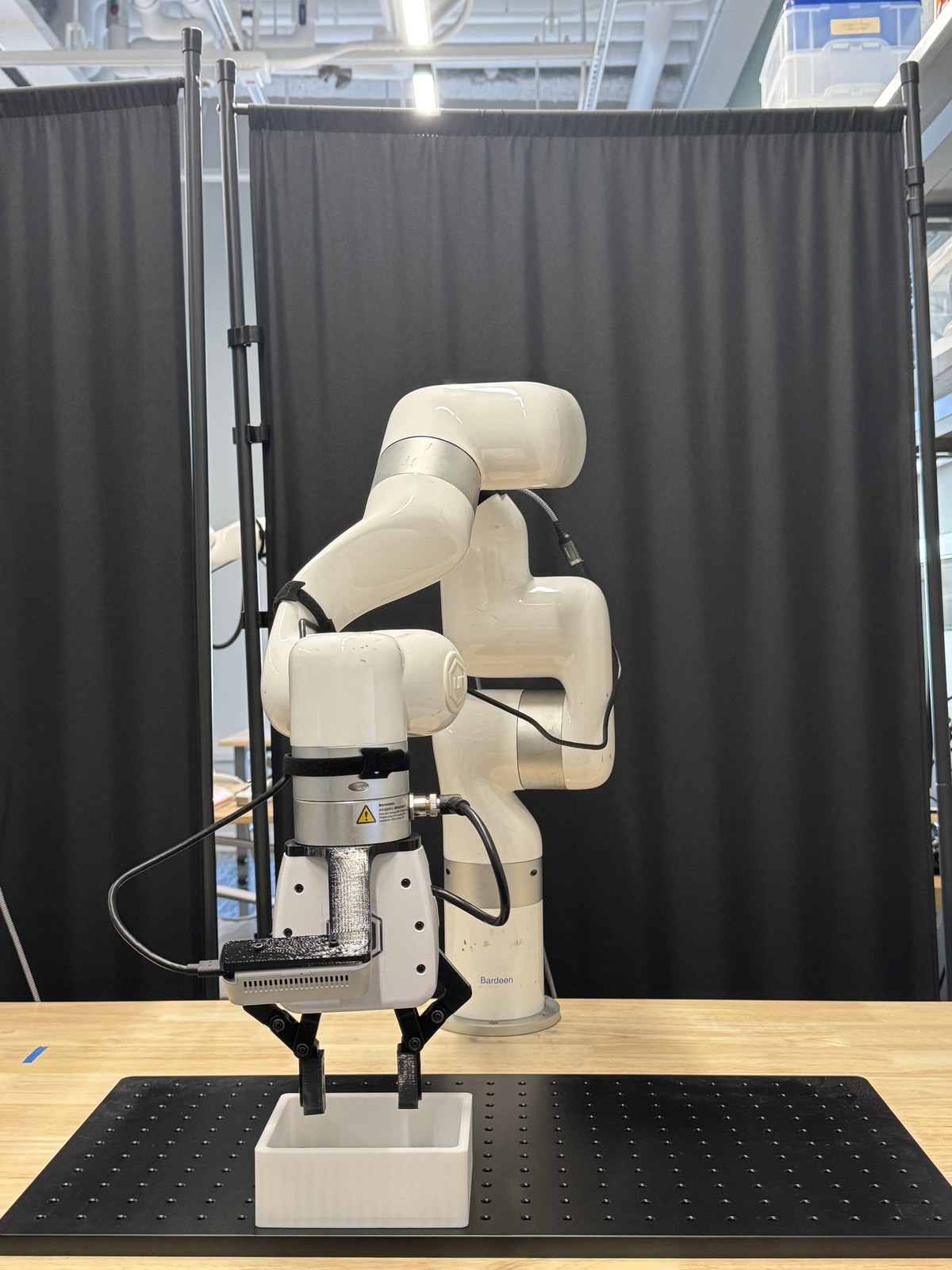}
\addlinespace[0.6em]

\taskblock{Task 3. Place both the blue cube and the green cube into the container.}
{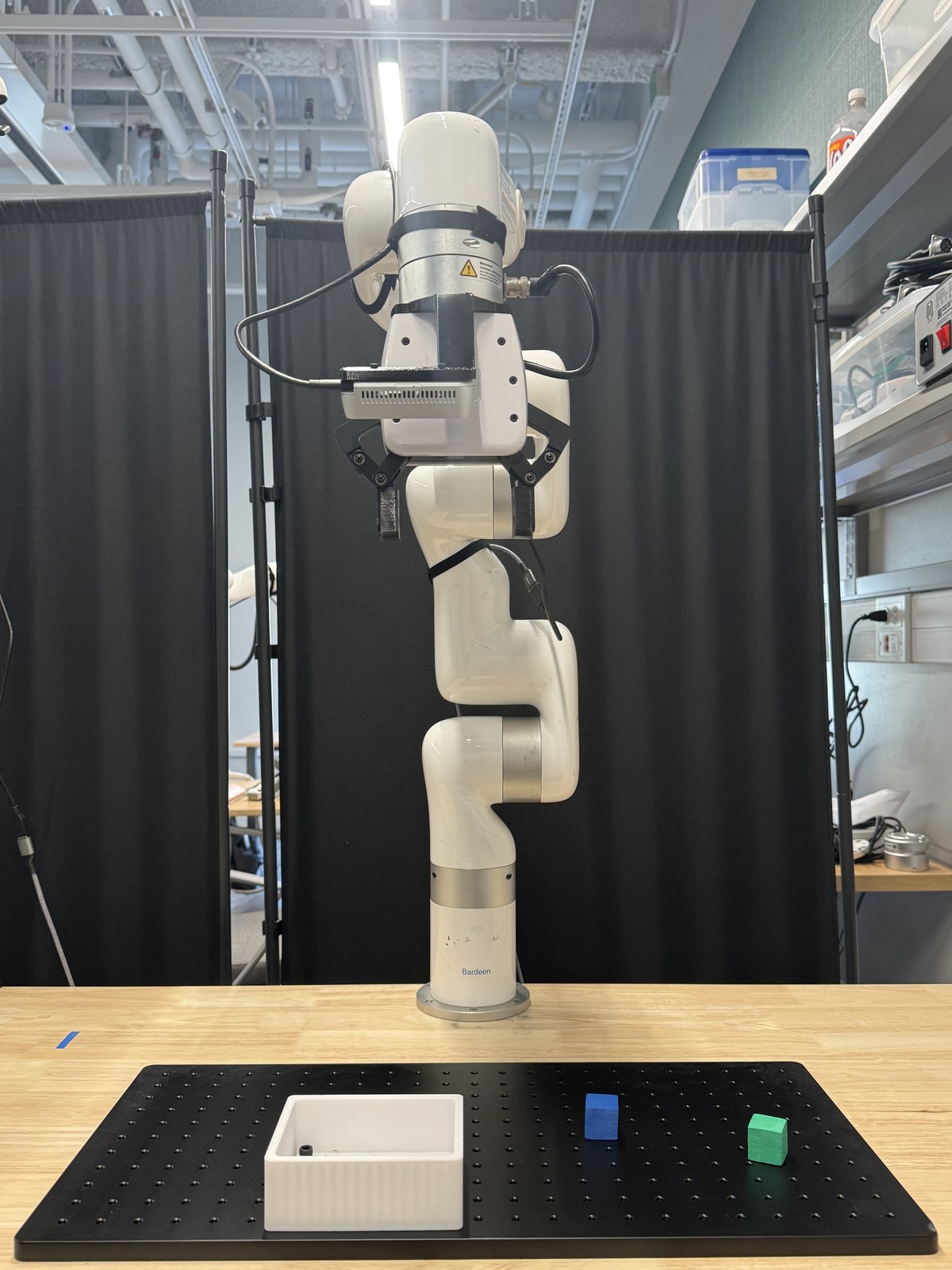}
{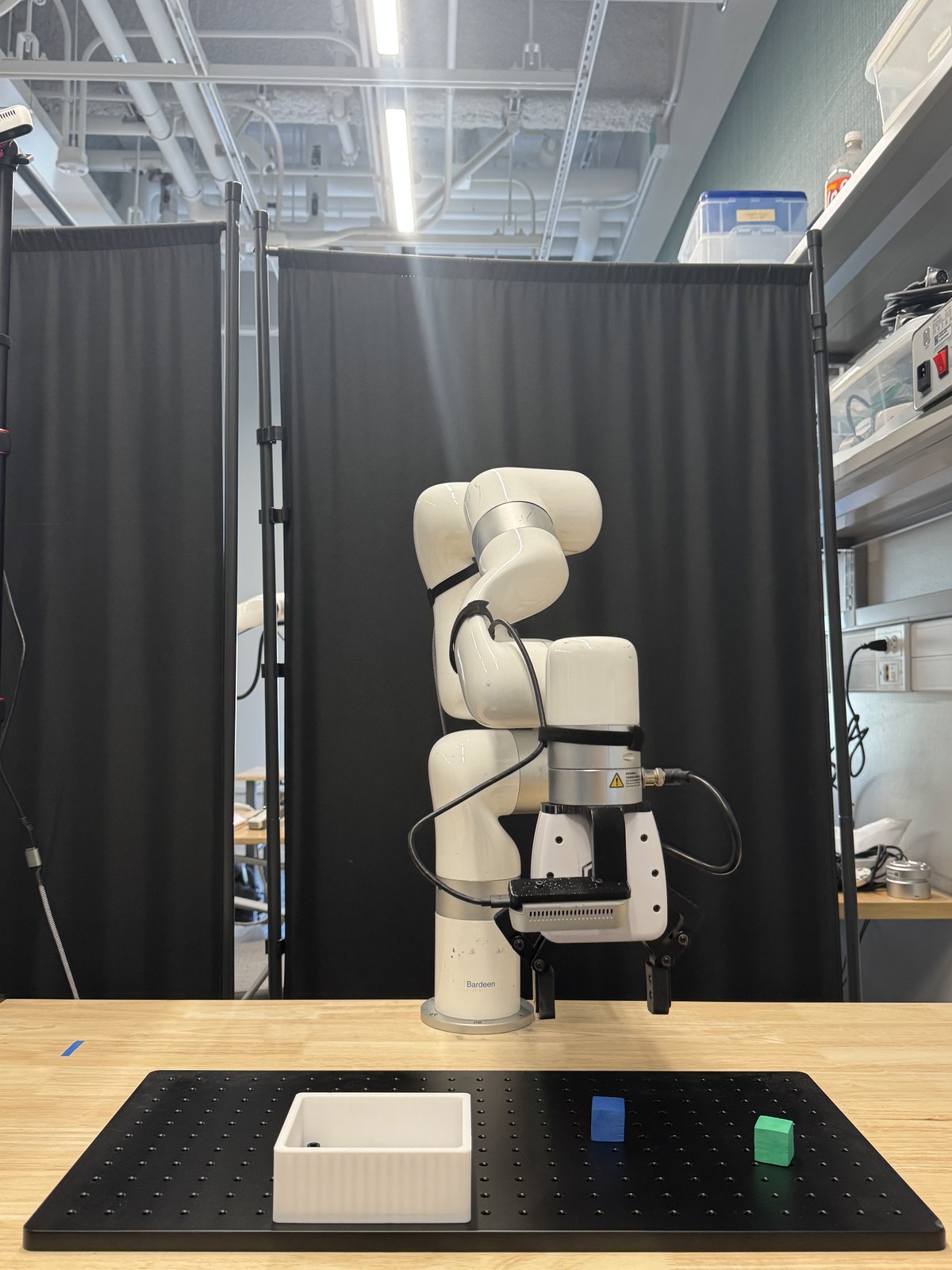}
{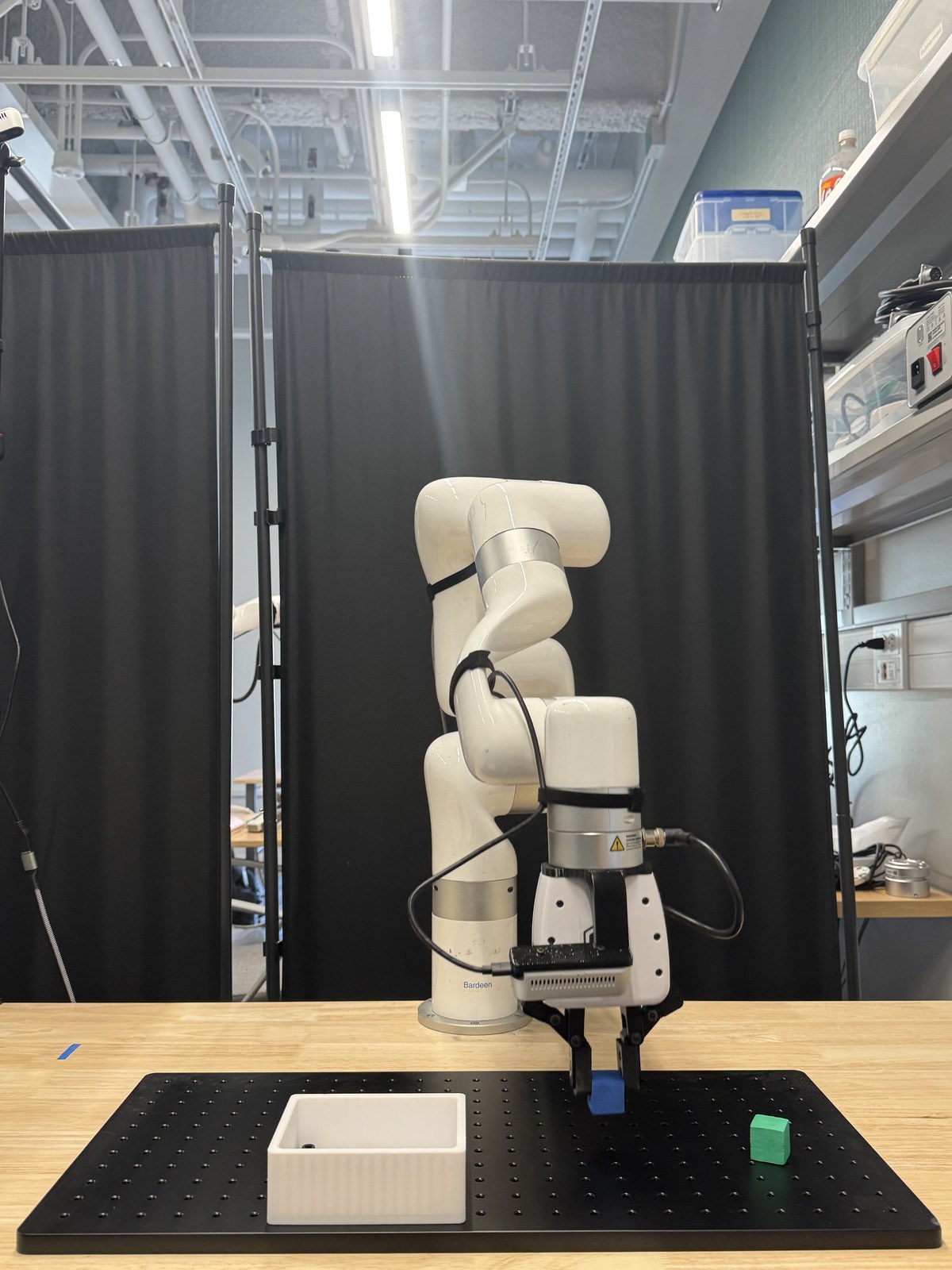}
{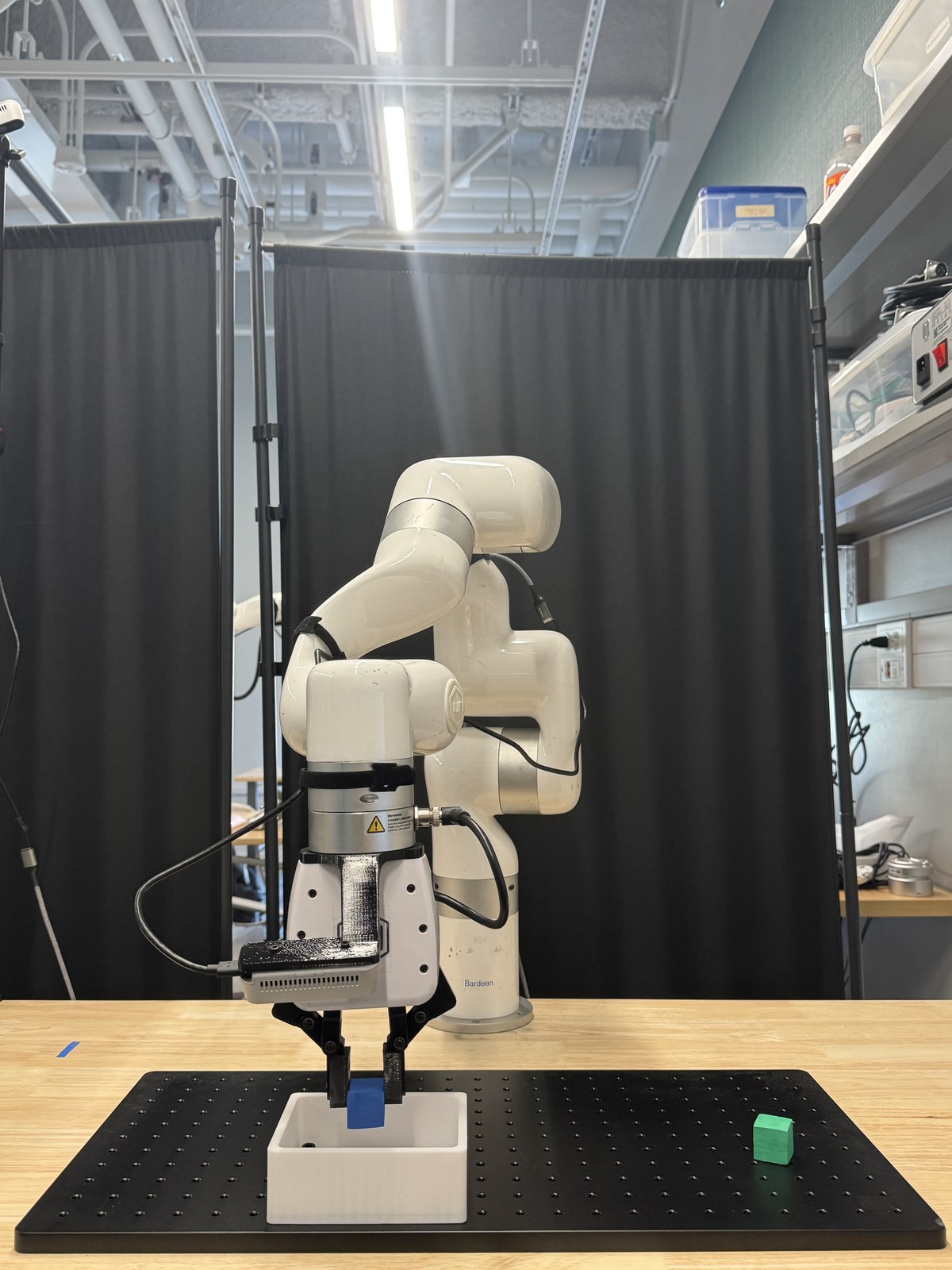}
{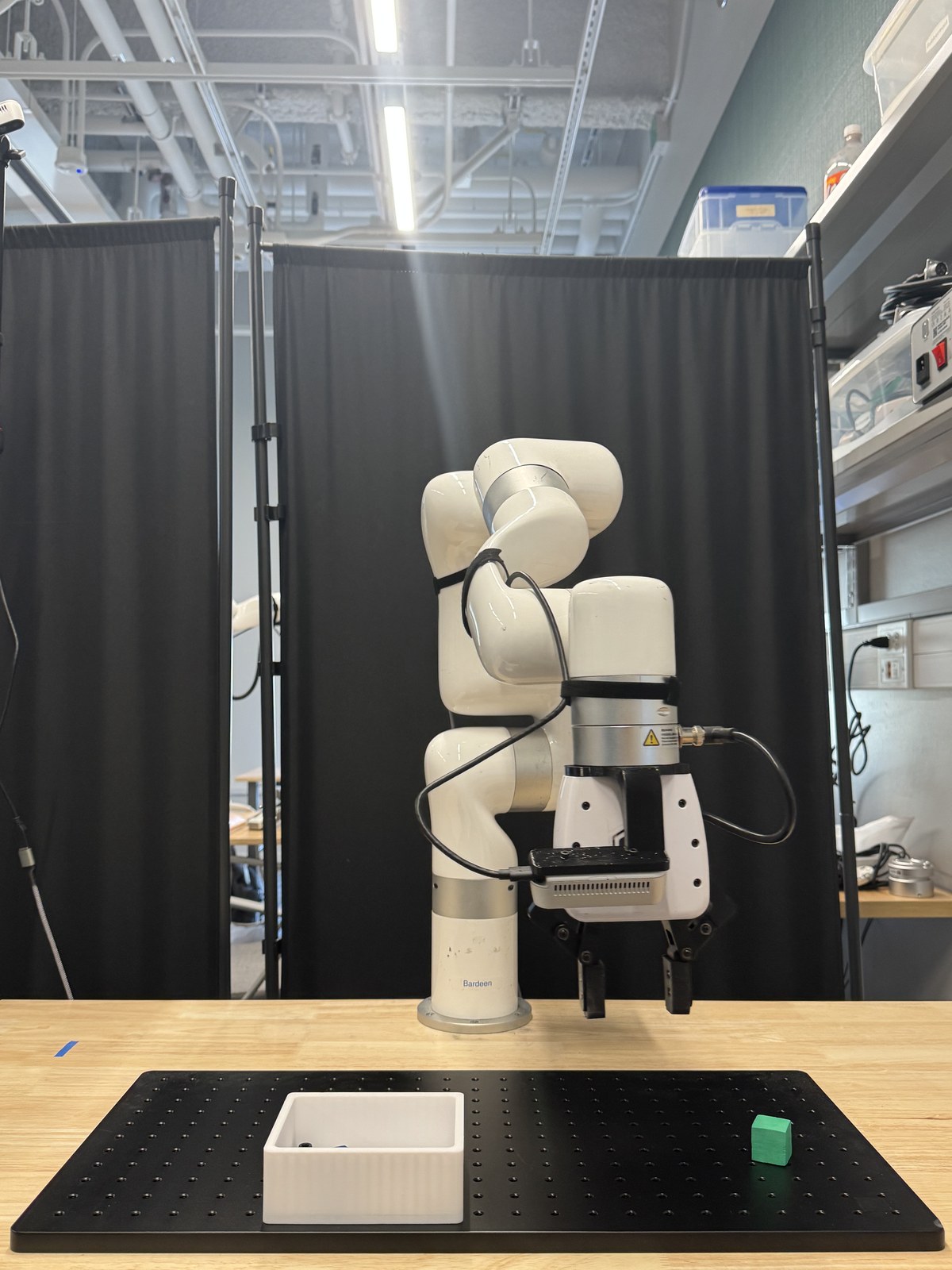}
{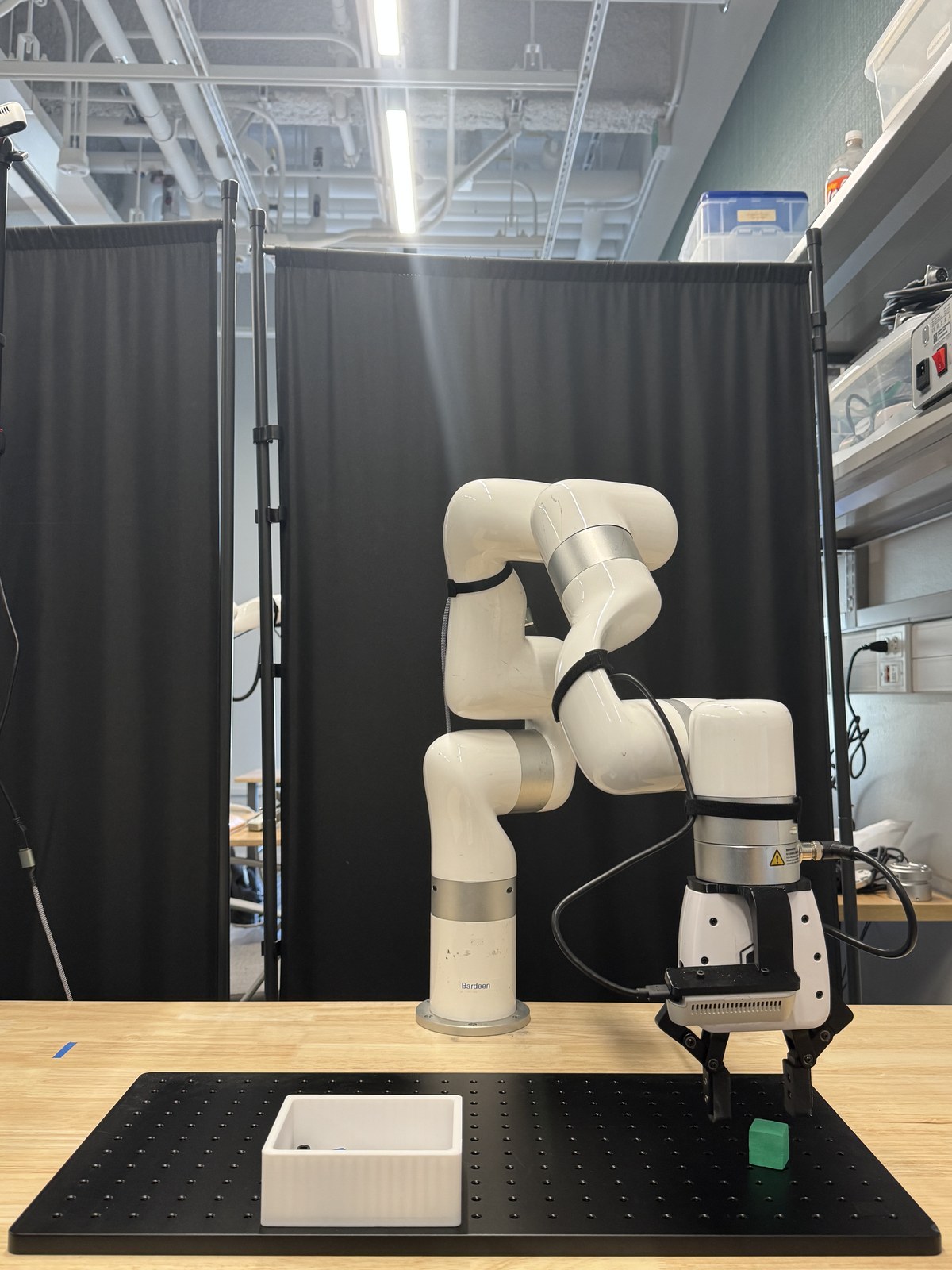}
{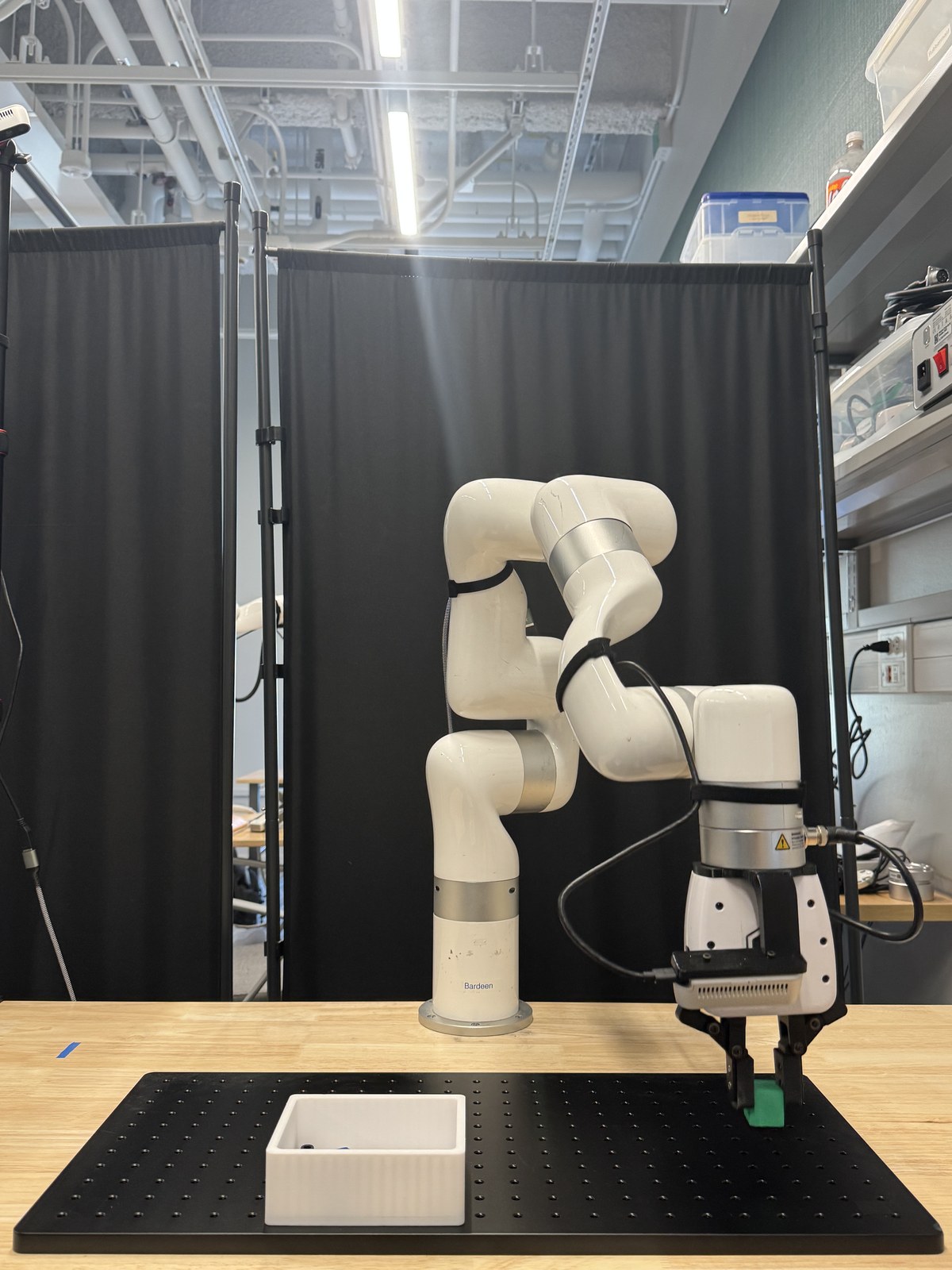}
{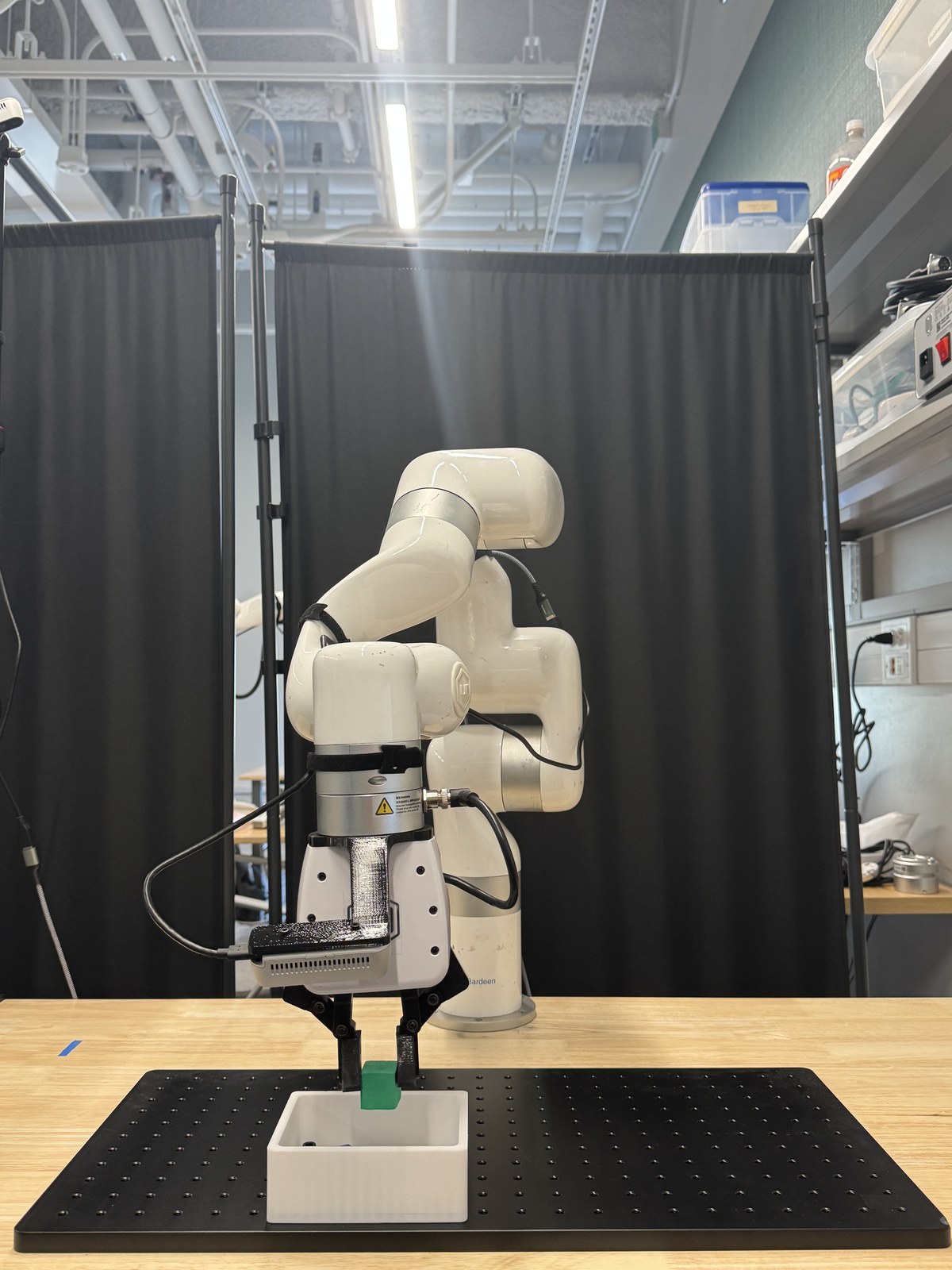}
\addlinespace[0.6em]

\taskblock{Task 4. Stack the red cube on top of the blue cube.}
{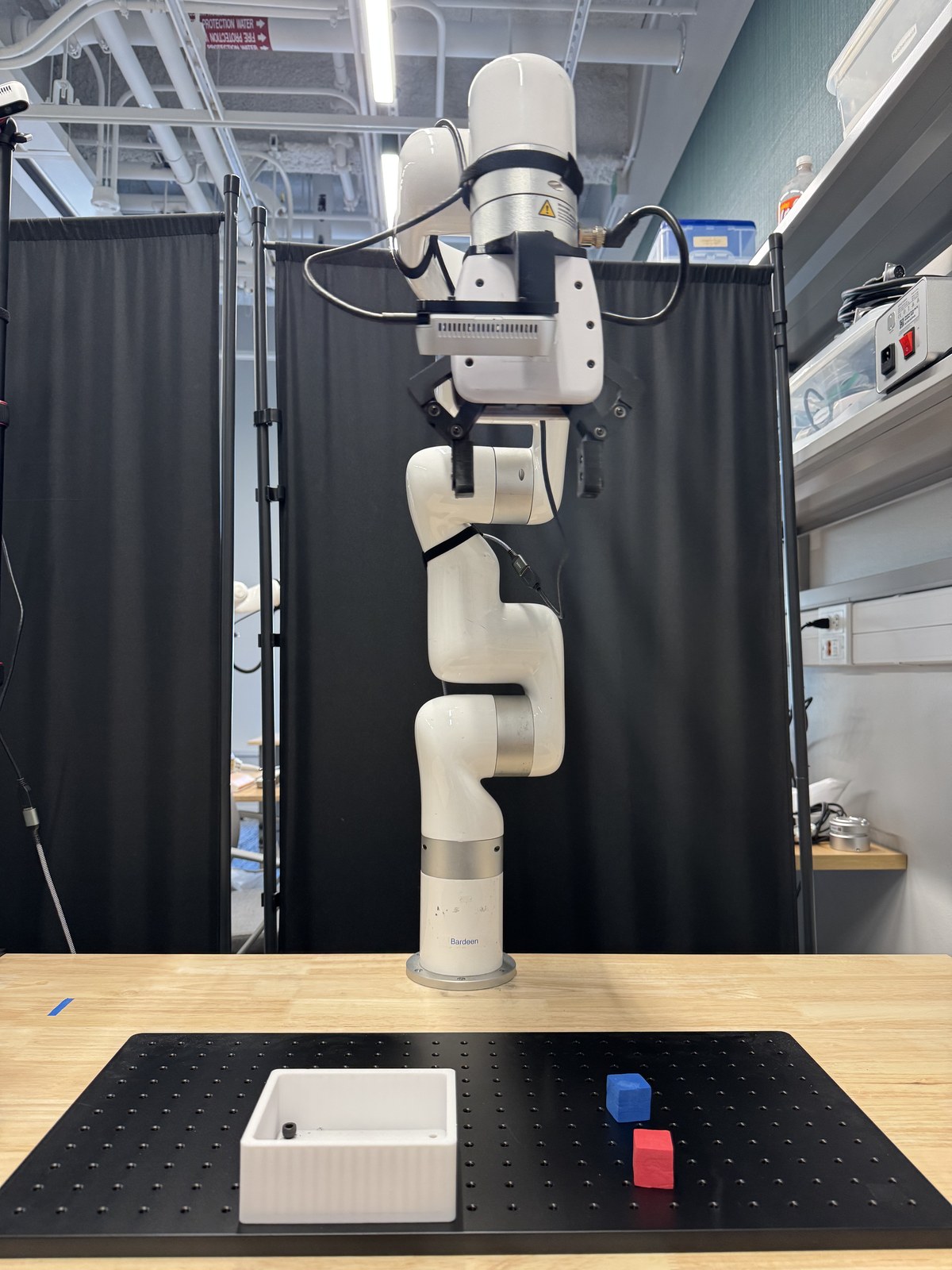}
{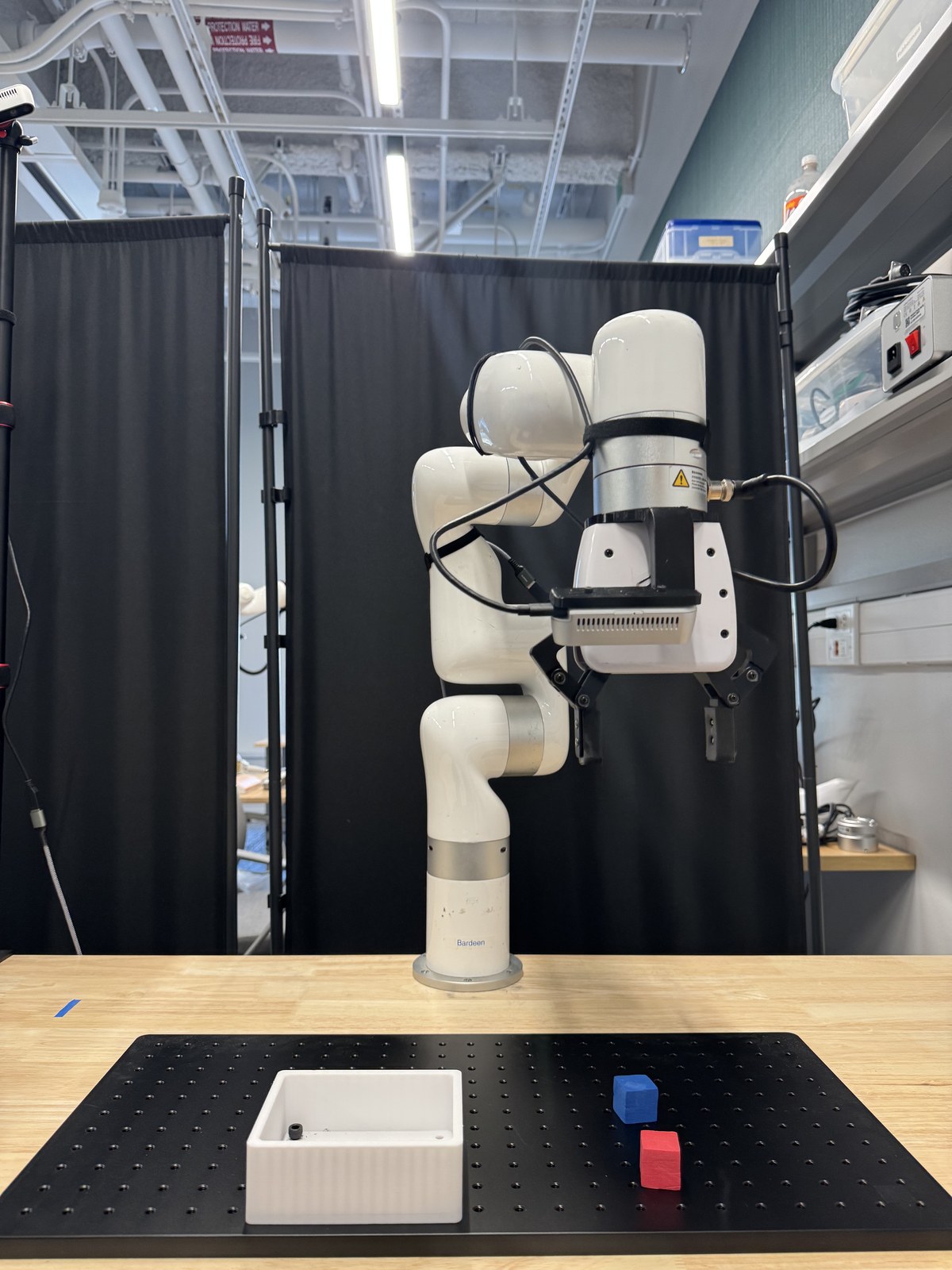}
{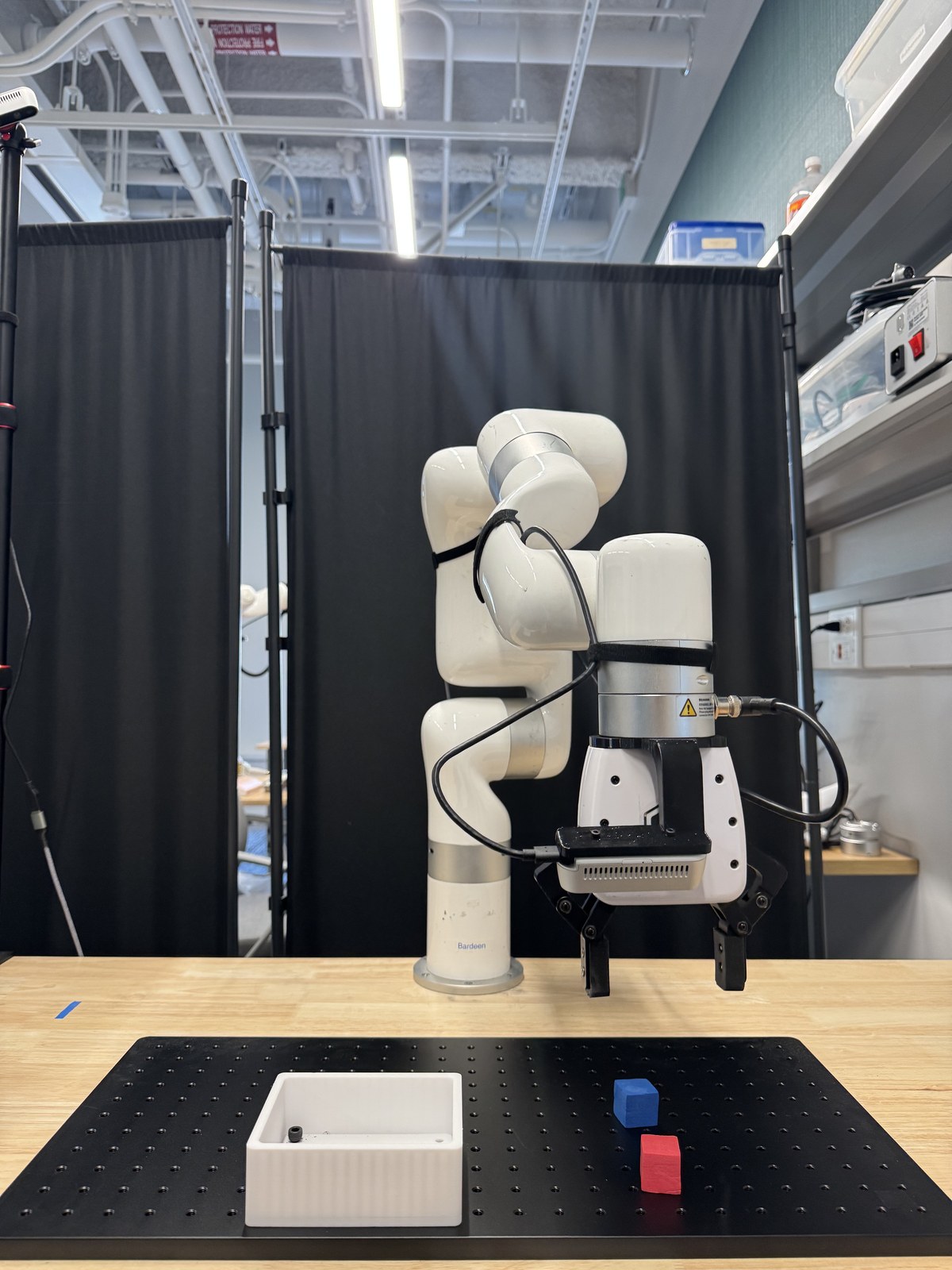}
{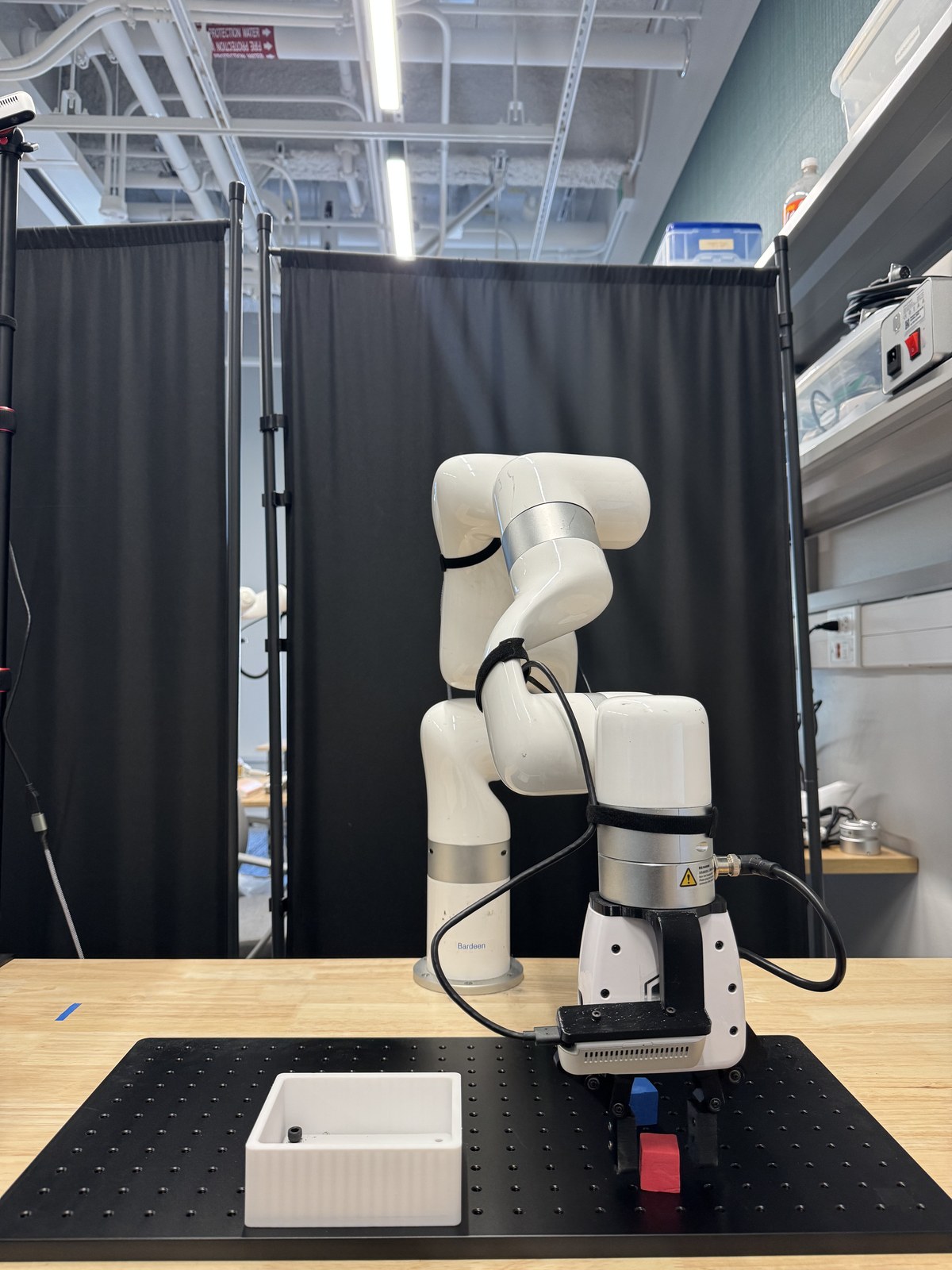}
{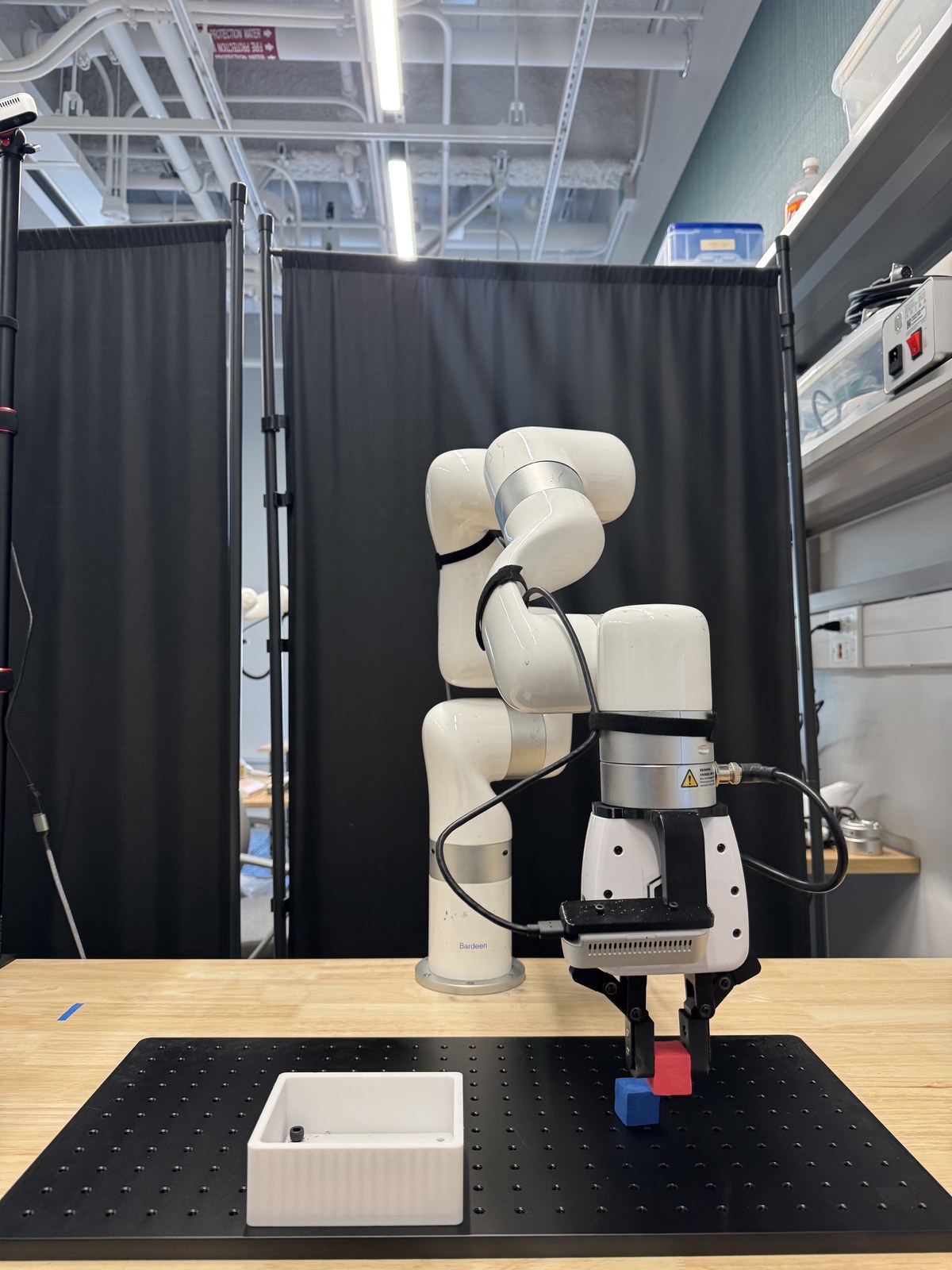}
{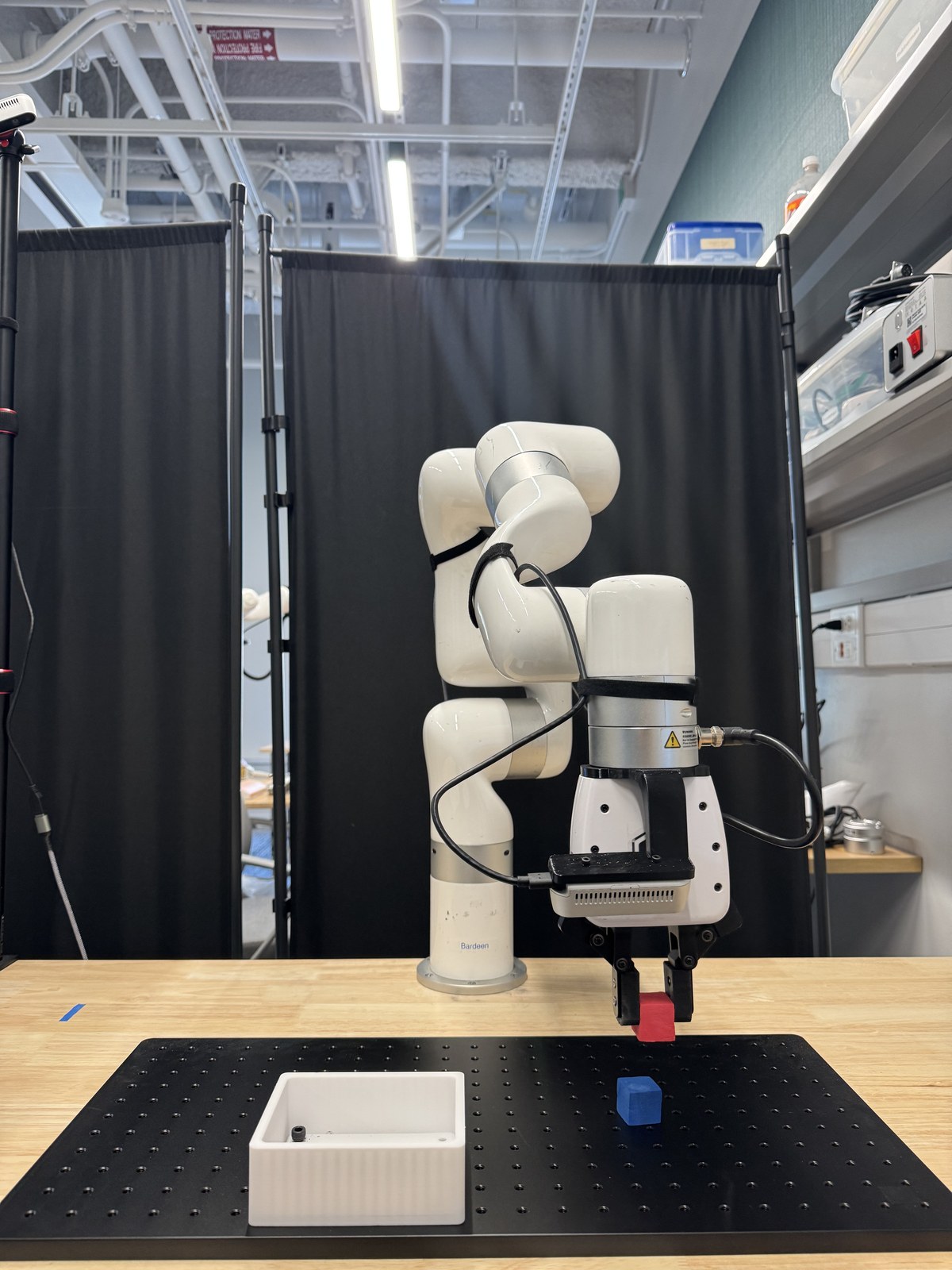}
{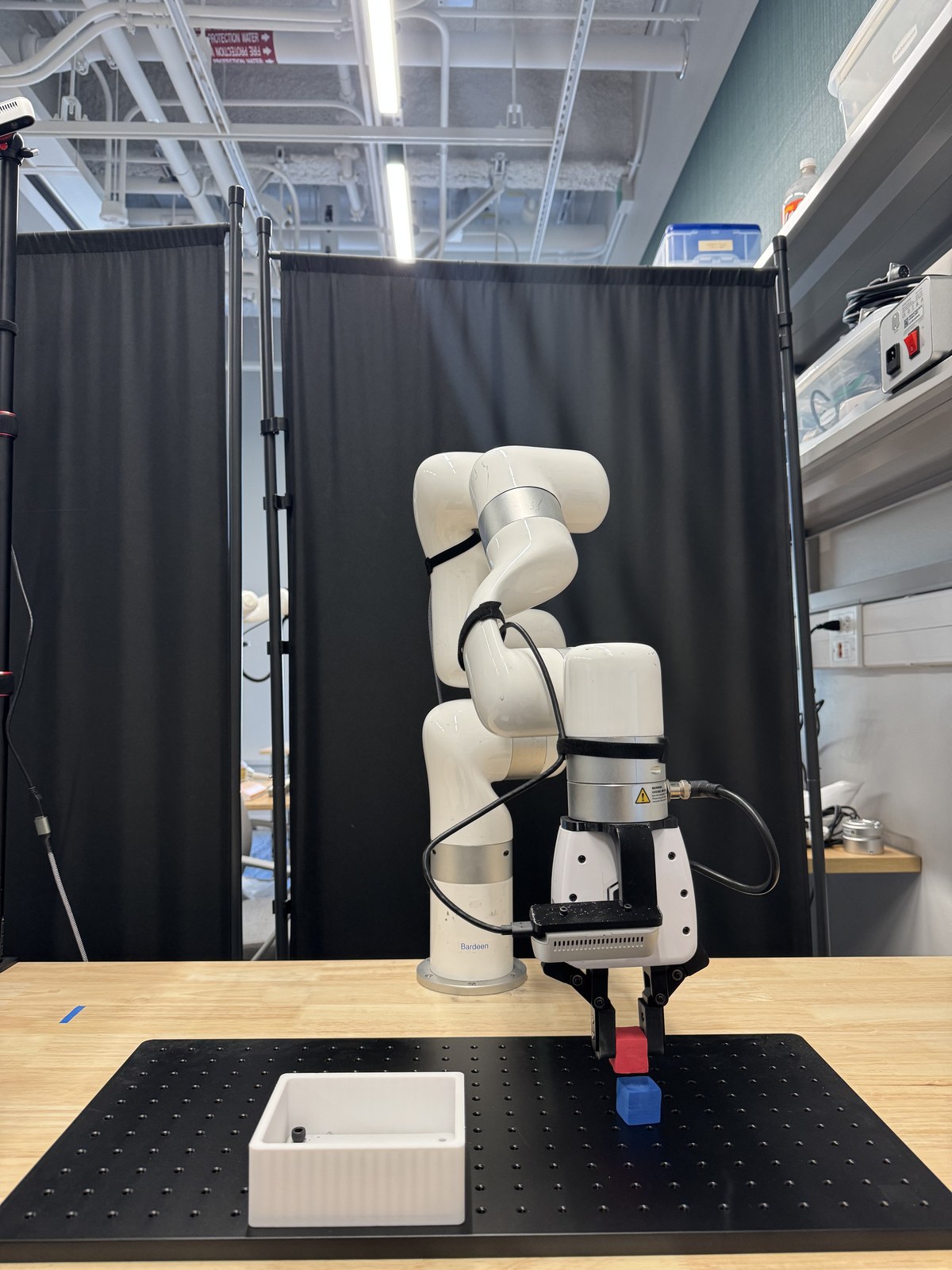}
{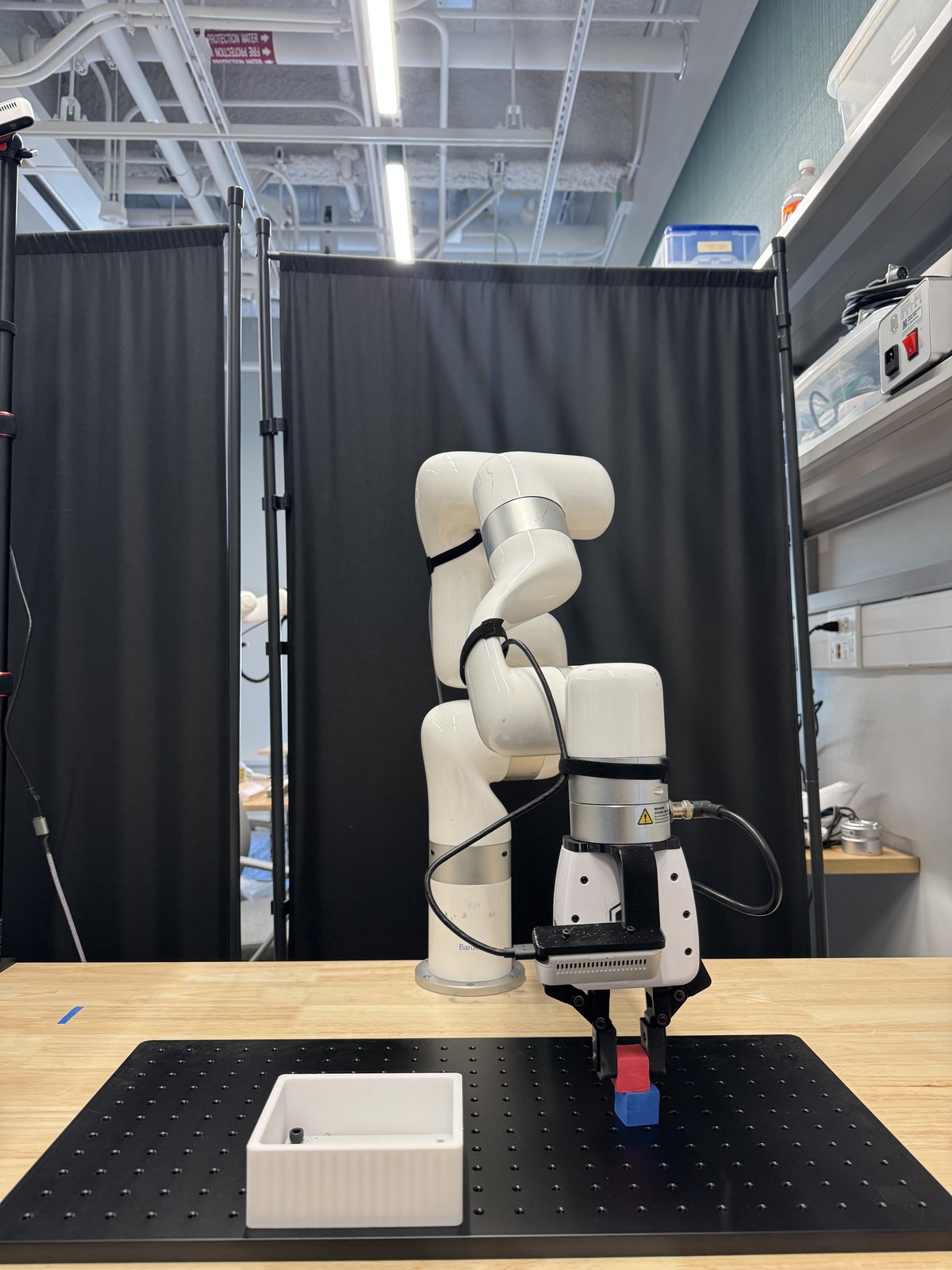}
\addlinespace[1em]

\midrule
\rowcolor{gray!20}
\multicolumn{8}{@{}c@{}}{%
  \rule{0pt}{1.4em}\textsc{\textbf{Category 2: Kitchen}}\rule[-0.6em]{0pt}{0pt}%
} \\
\midrule
\addlinespace[0.4em]

\taskblock{Task 5. Place the duck doll in the bowl.}
{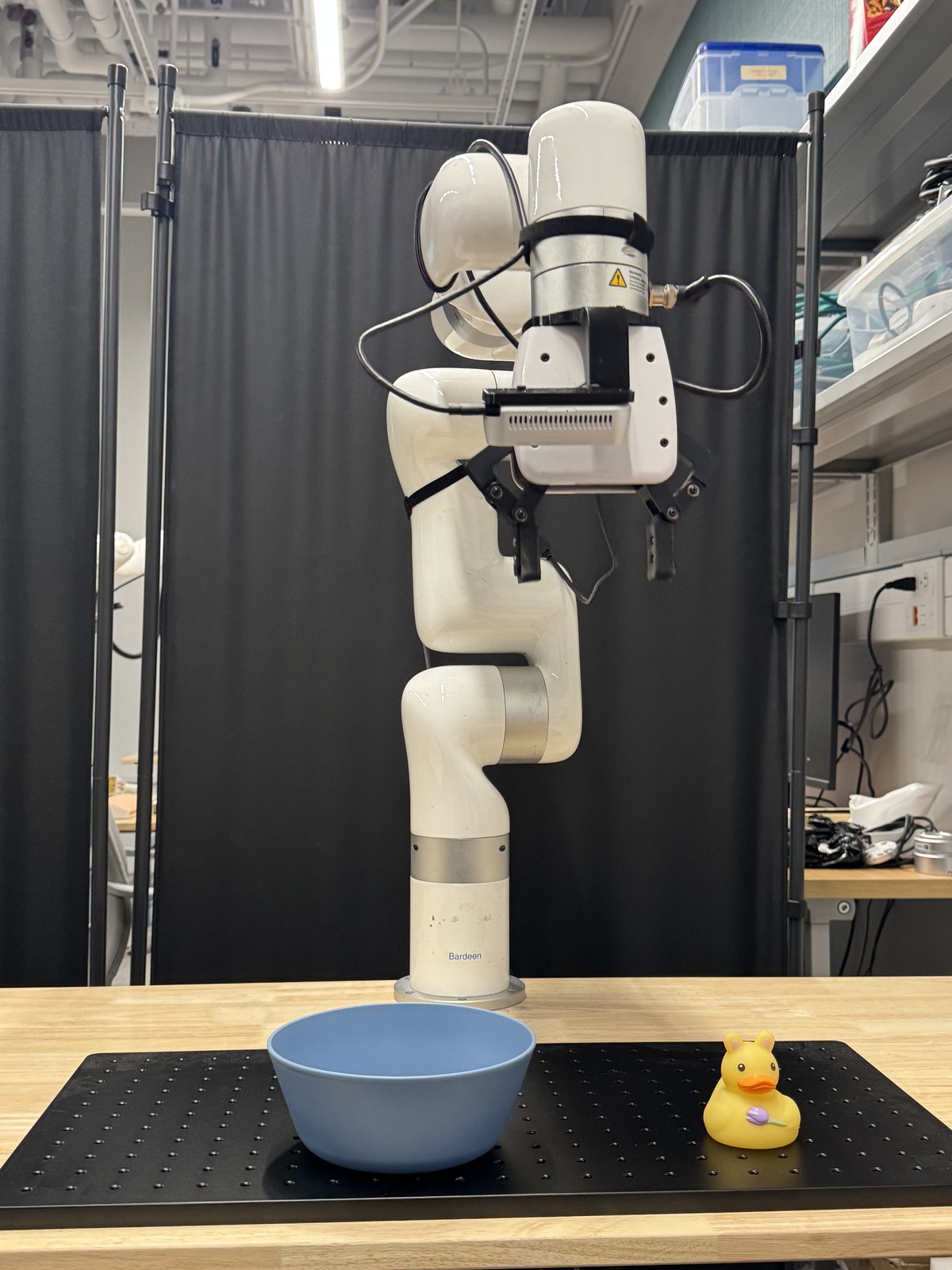}
{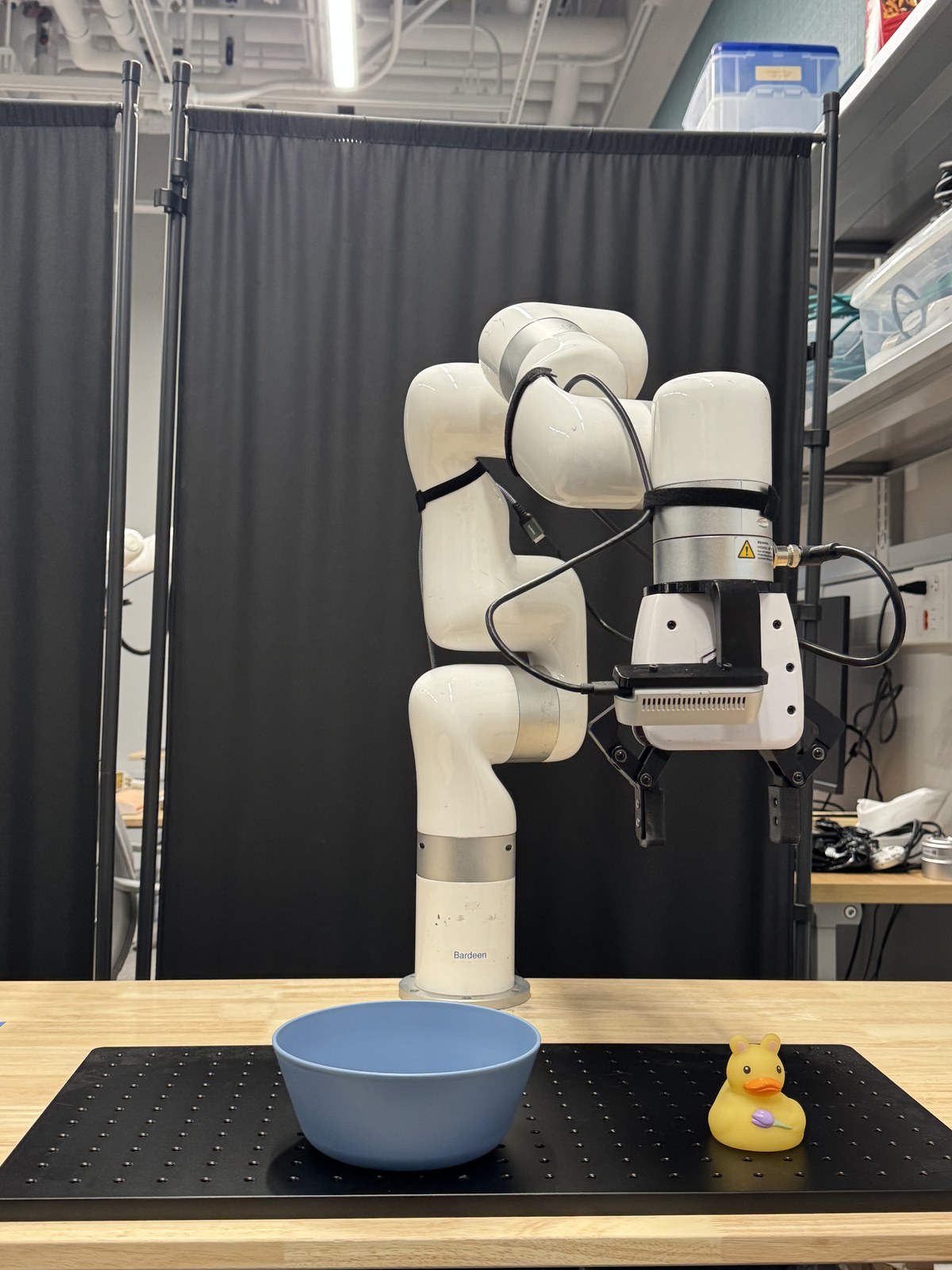}
{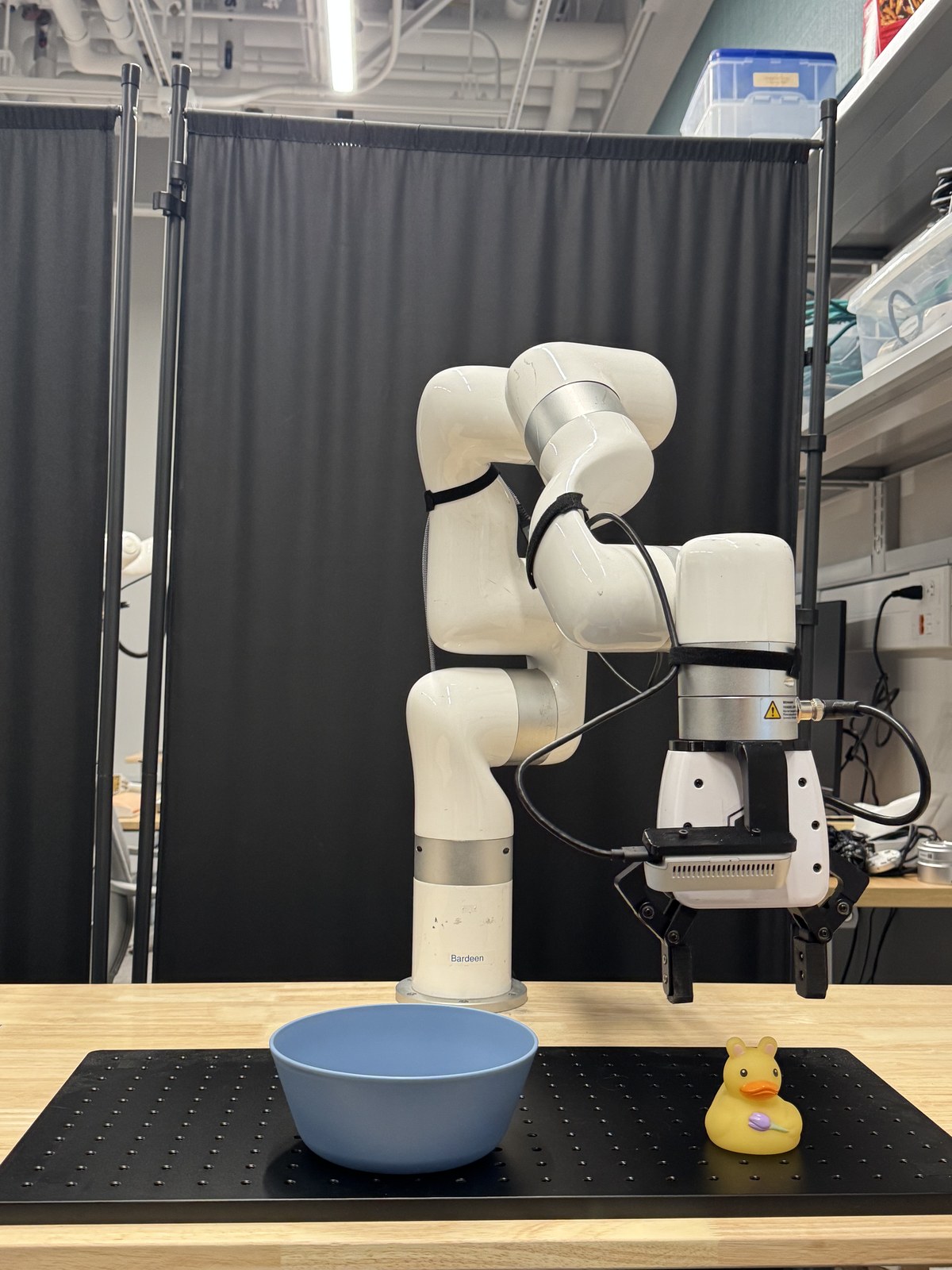}
{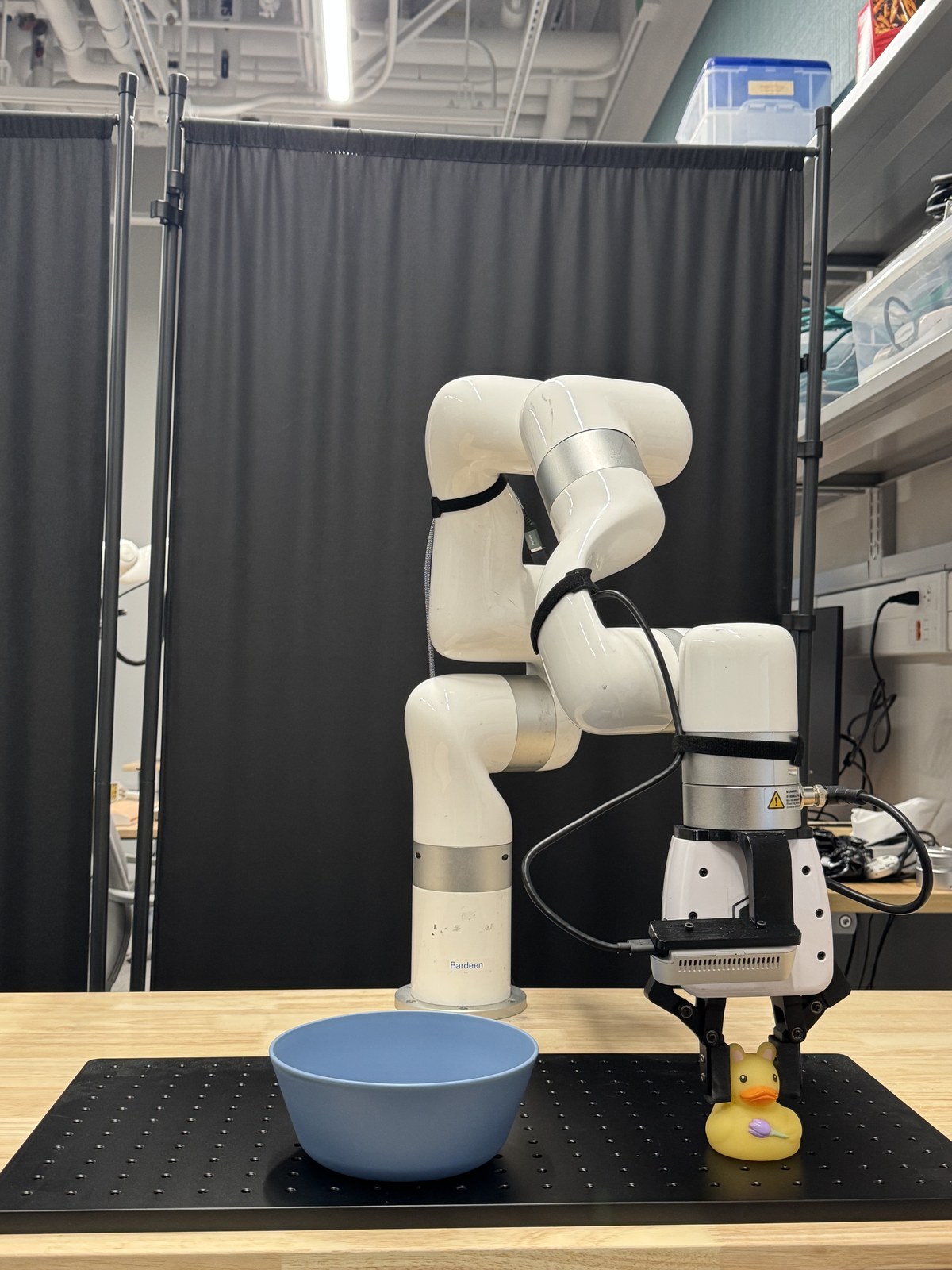}
{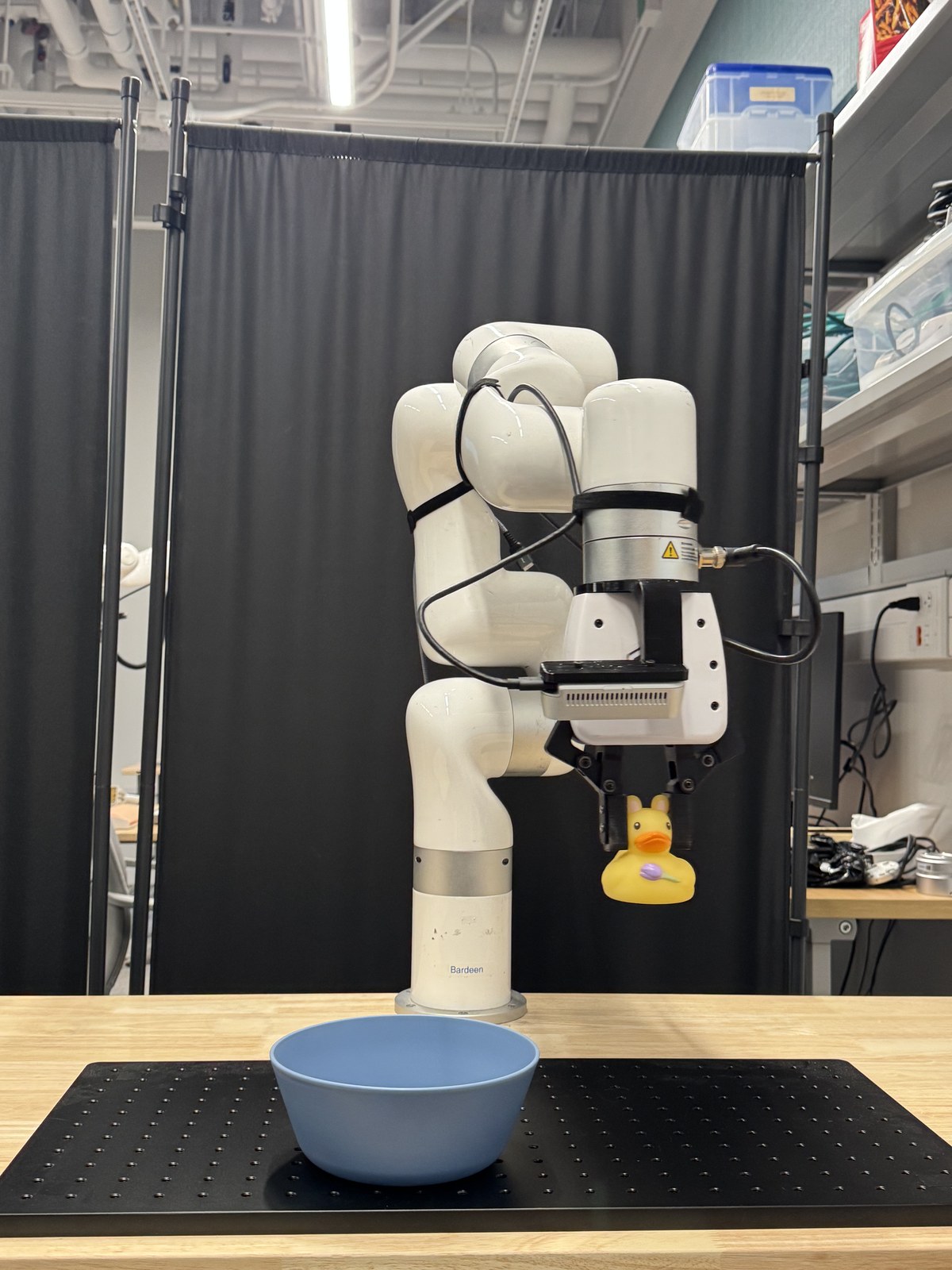}
{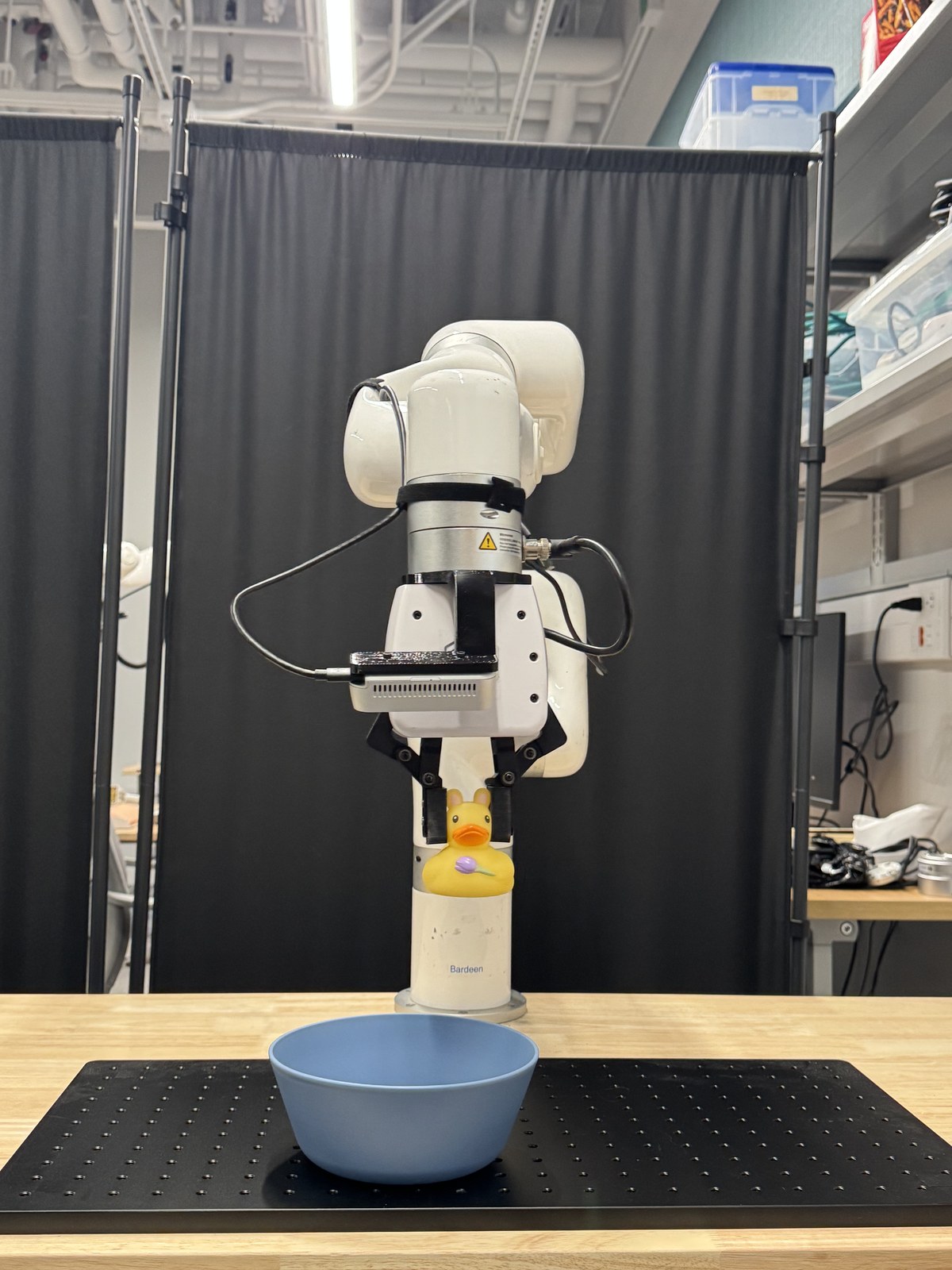}
{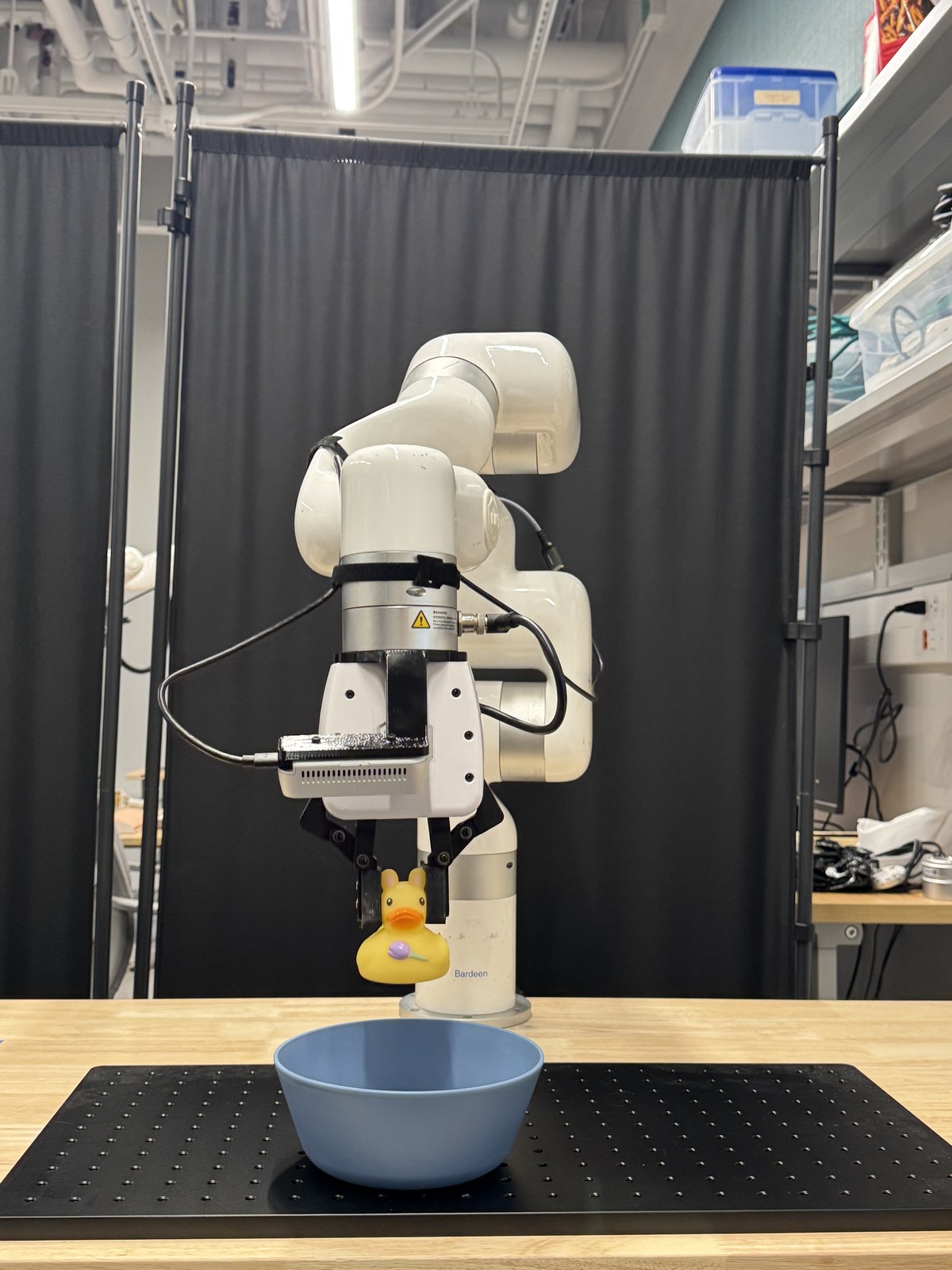}
{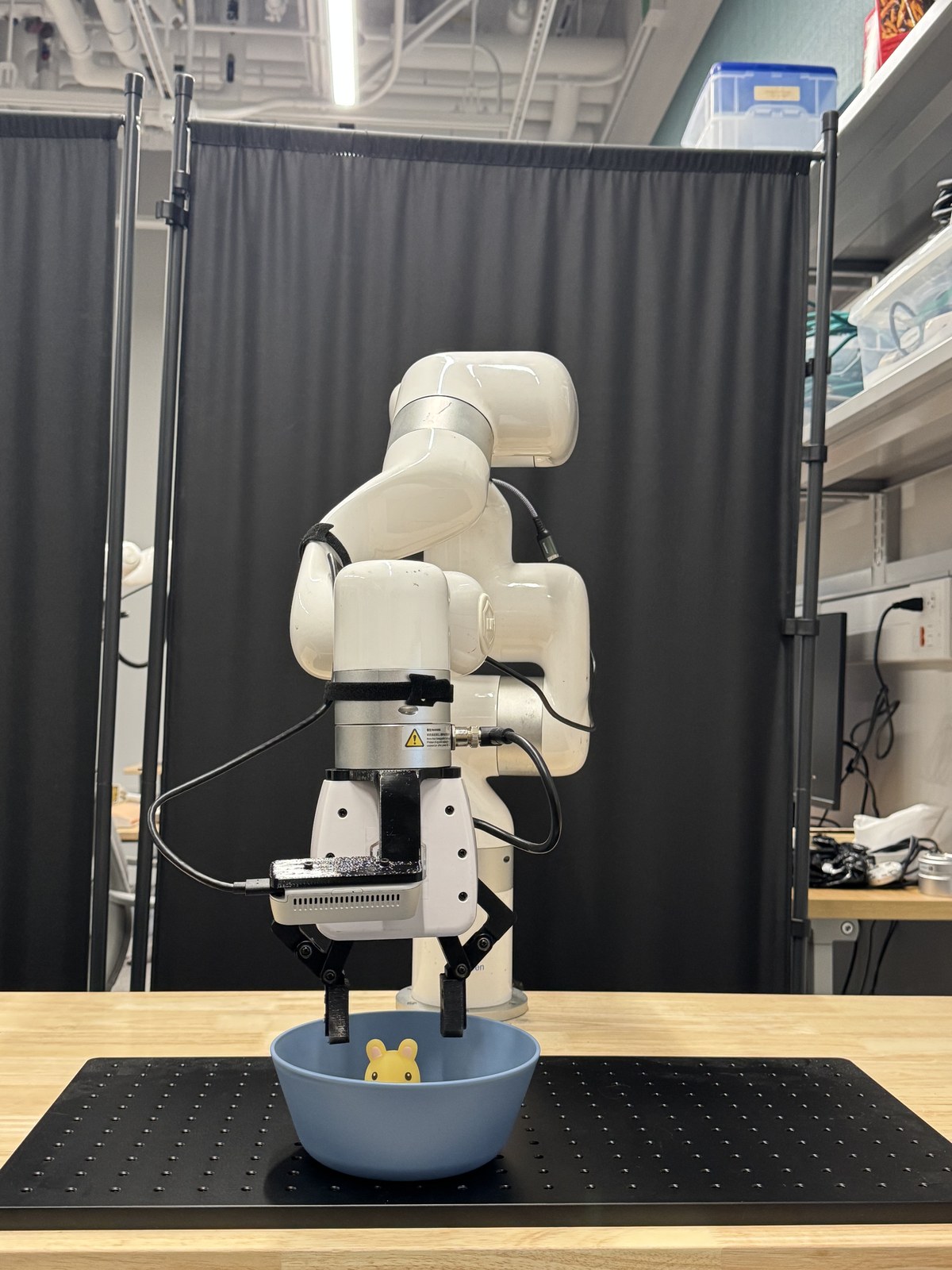}
\addlinespace[0.6em]

\taskblock{Task 6. Place the carrot in the bowl.}
{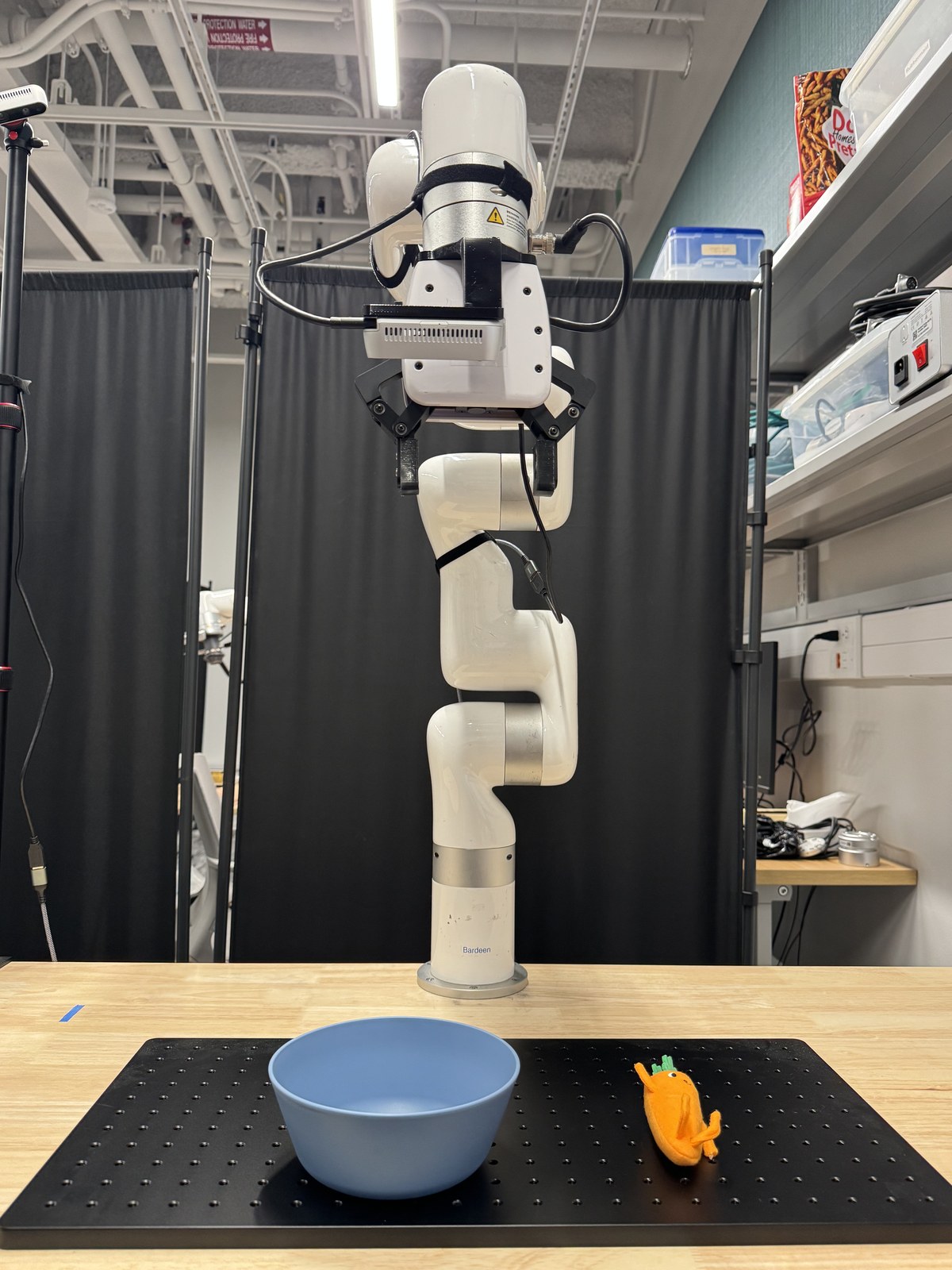}
{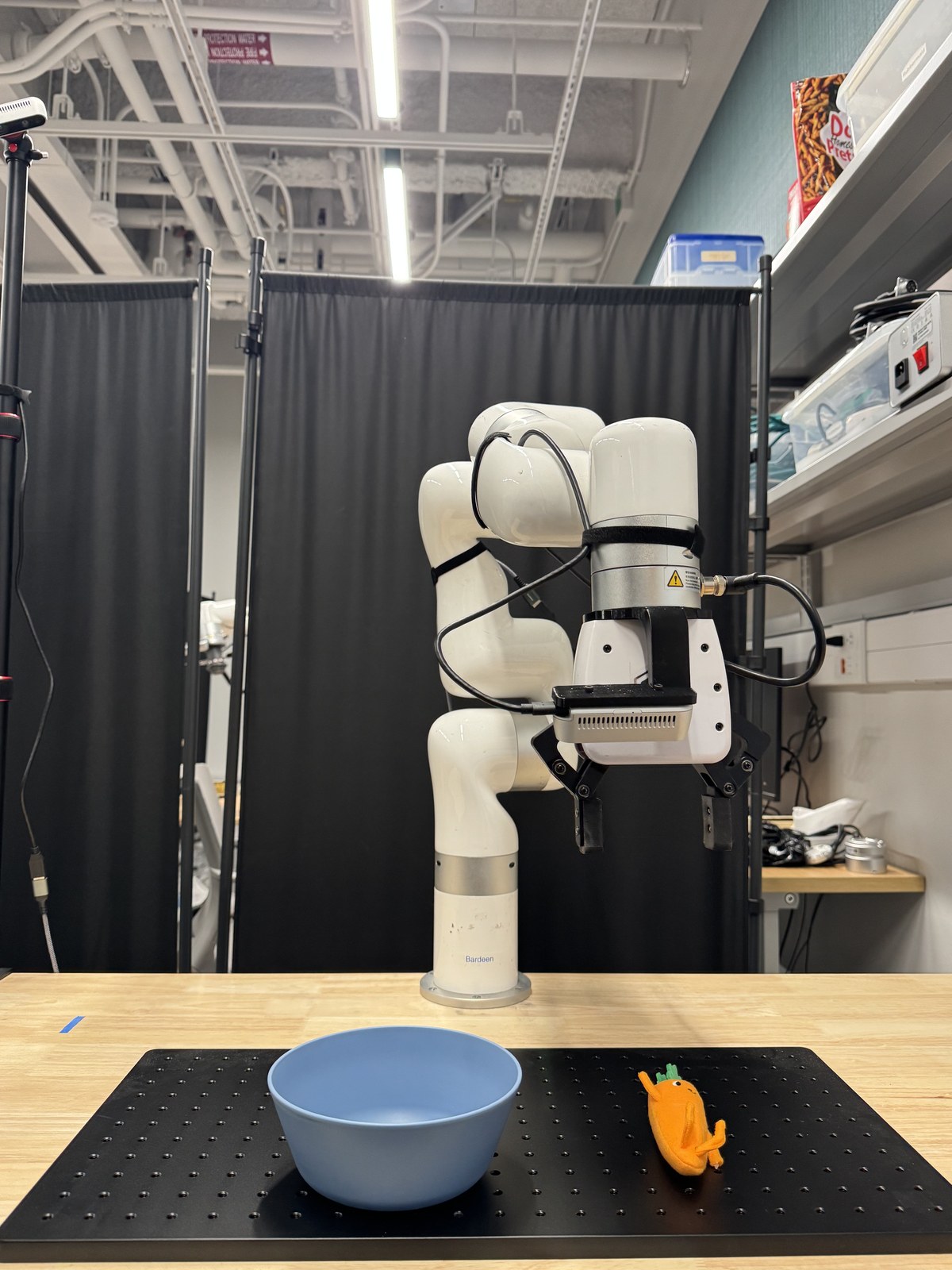}
{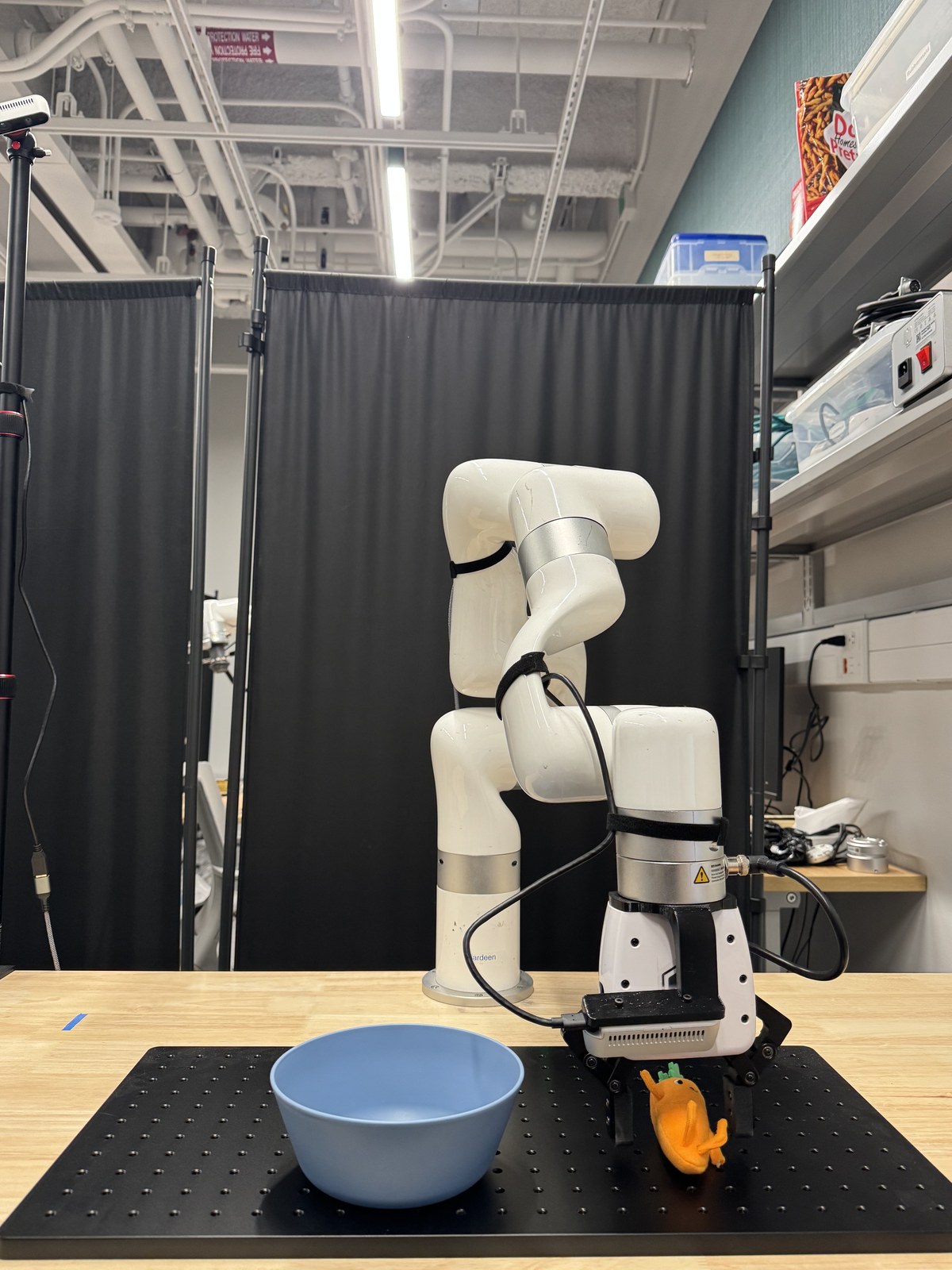}
{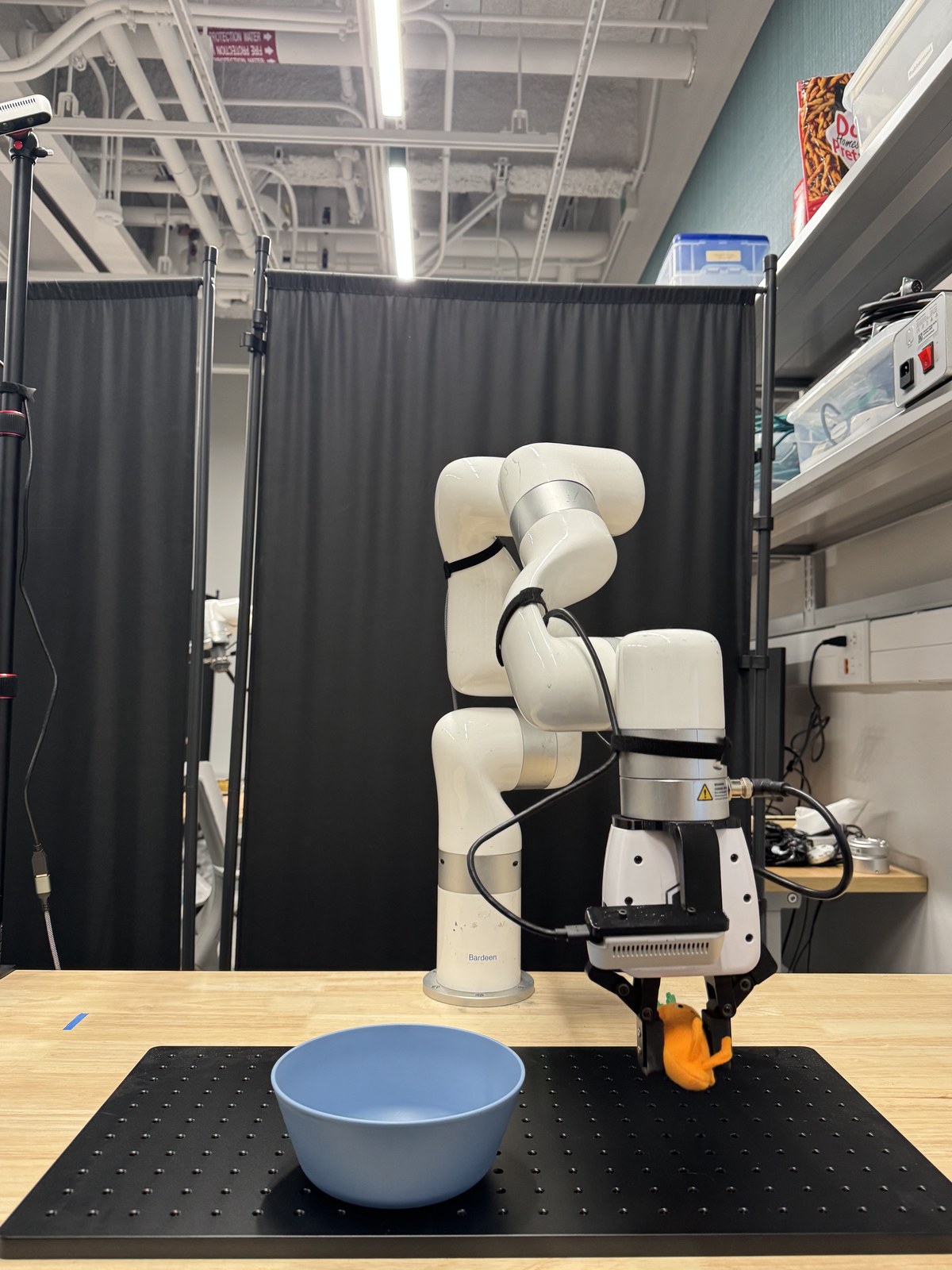}
{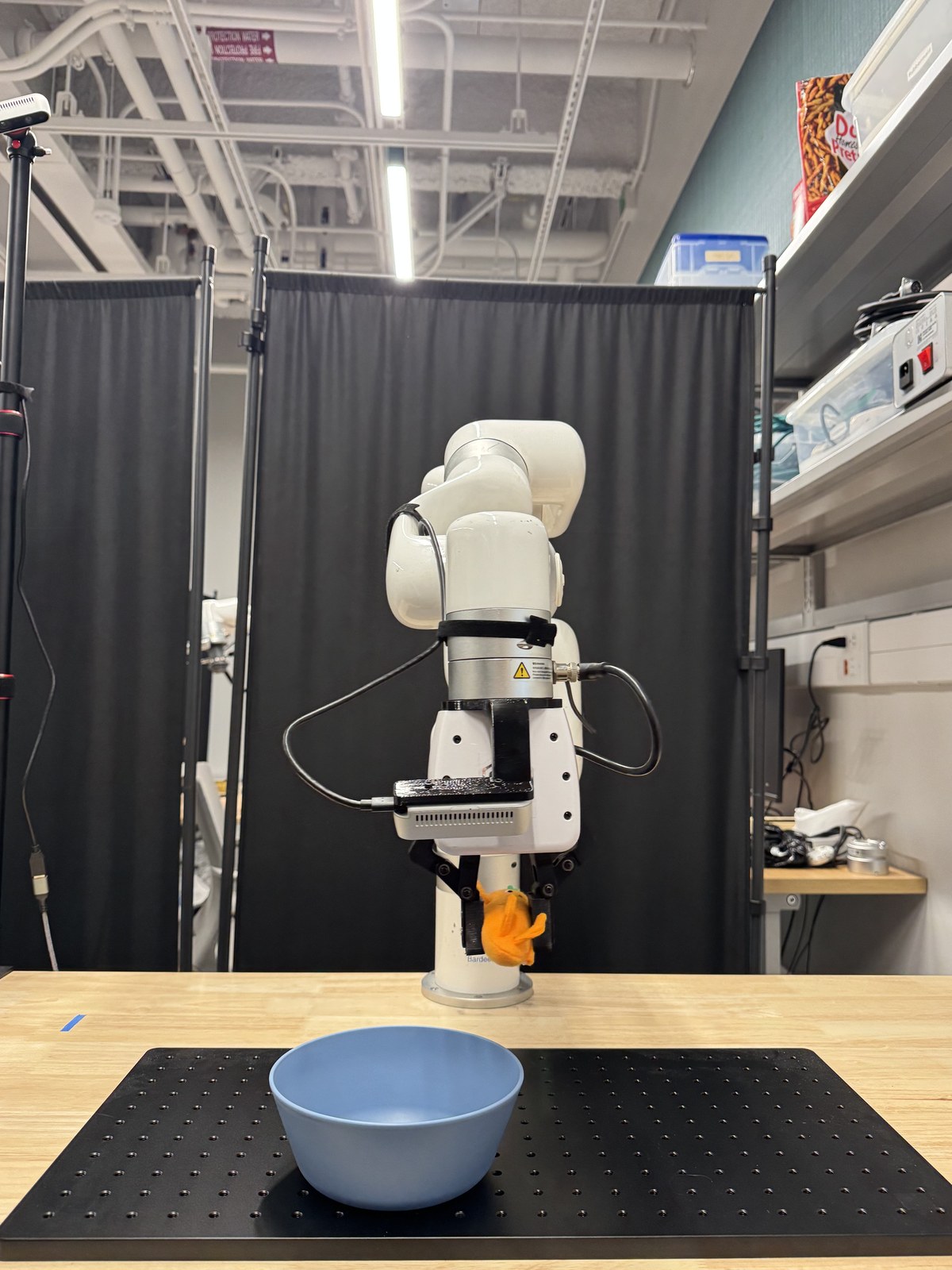}
{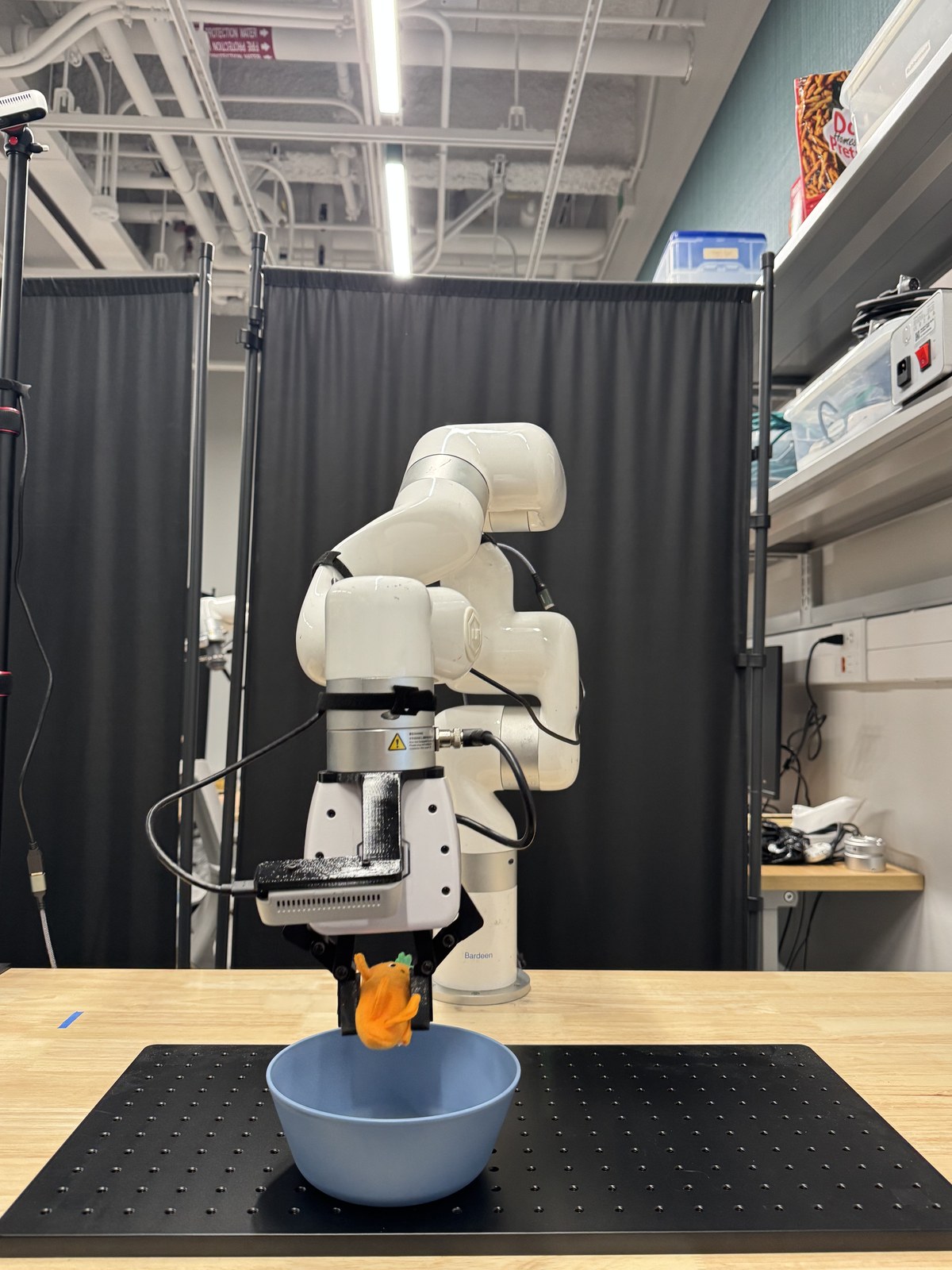}
{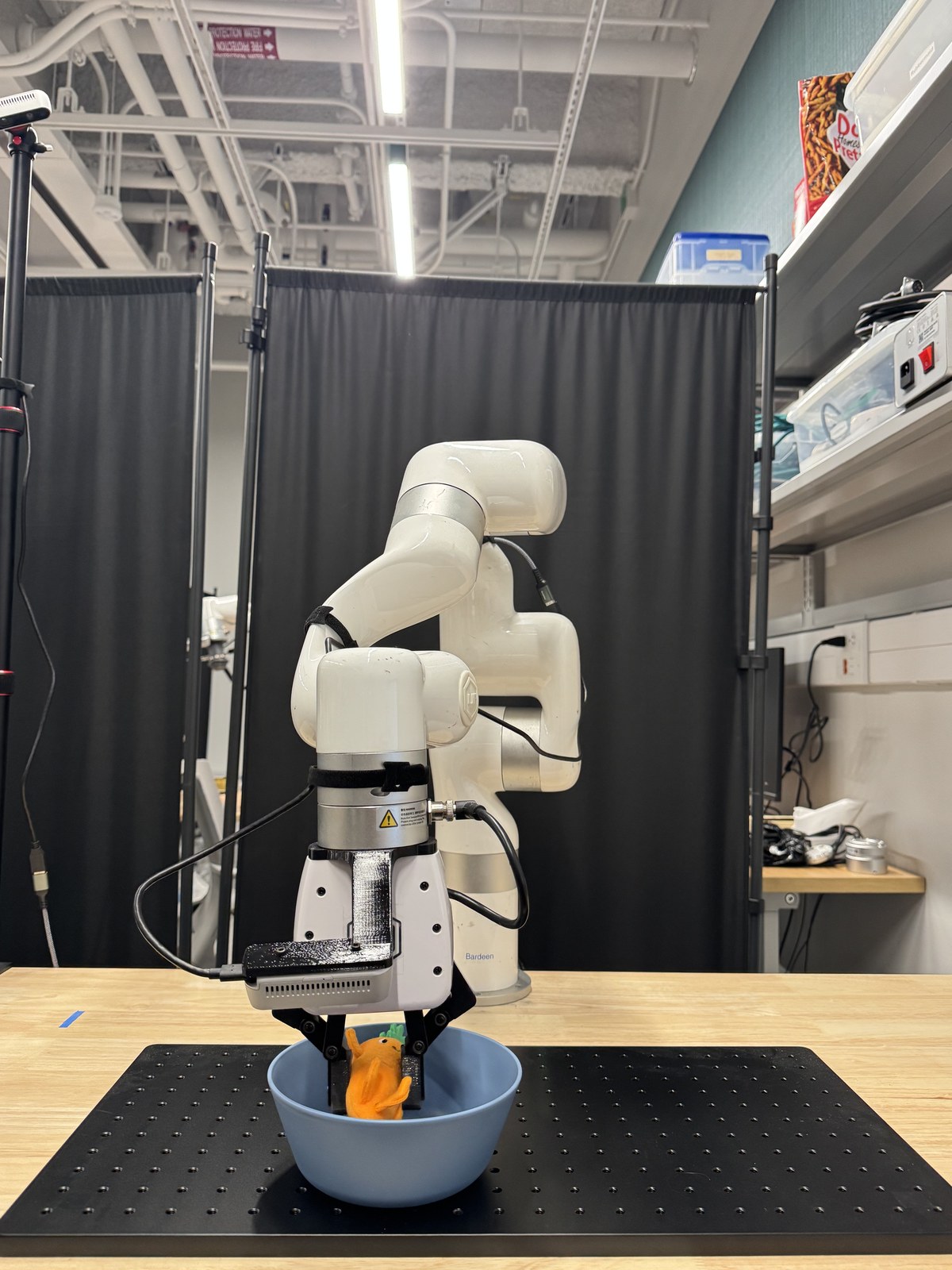}
{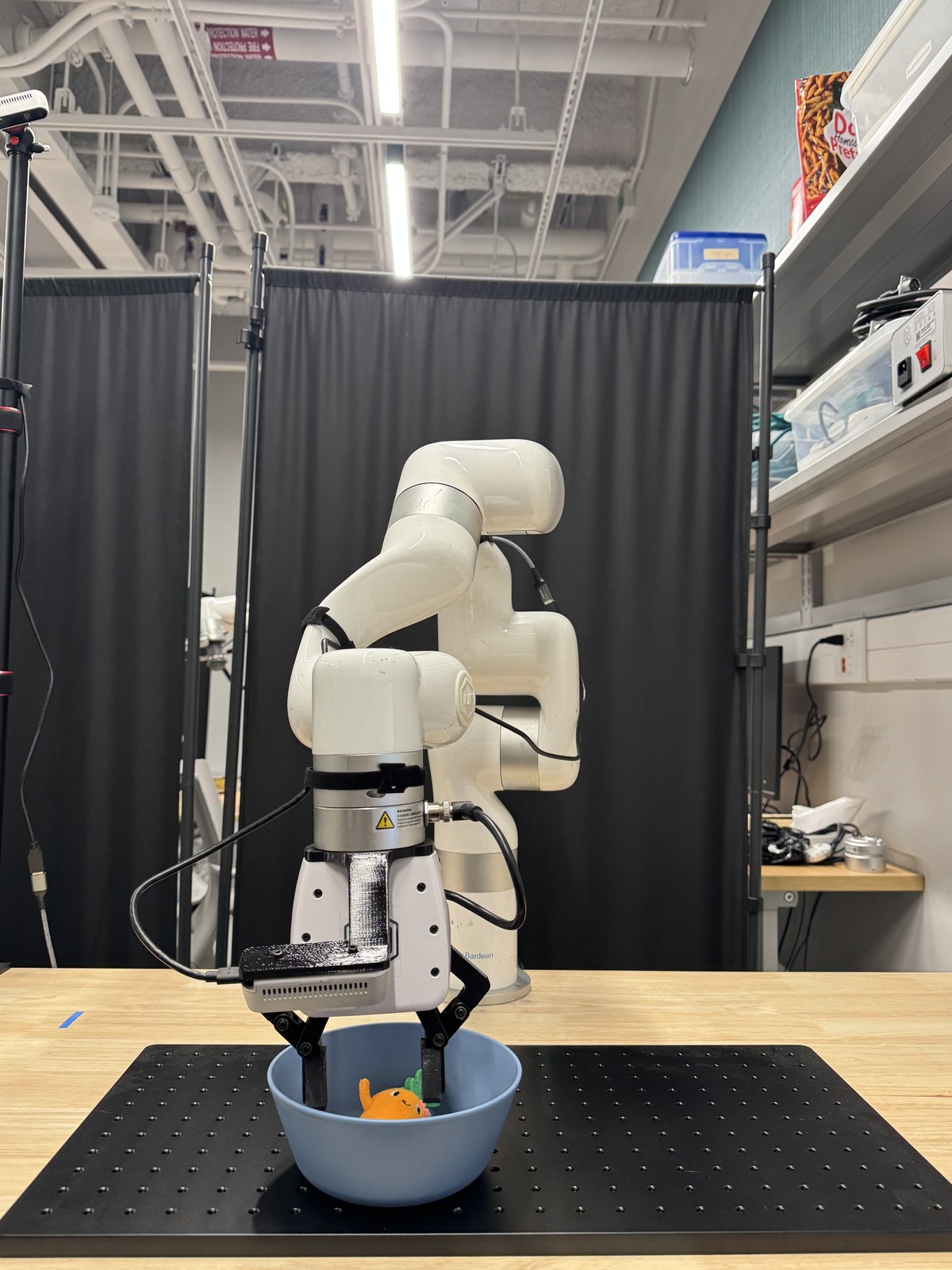}
\addlinespace[0.6em]

\taskblock{Task 7. Place both the carrot and the duck in the bowl.}
{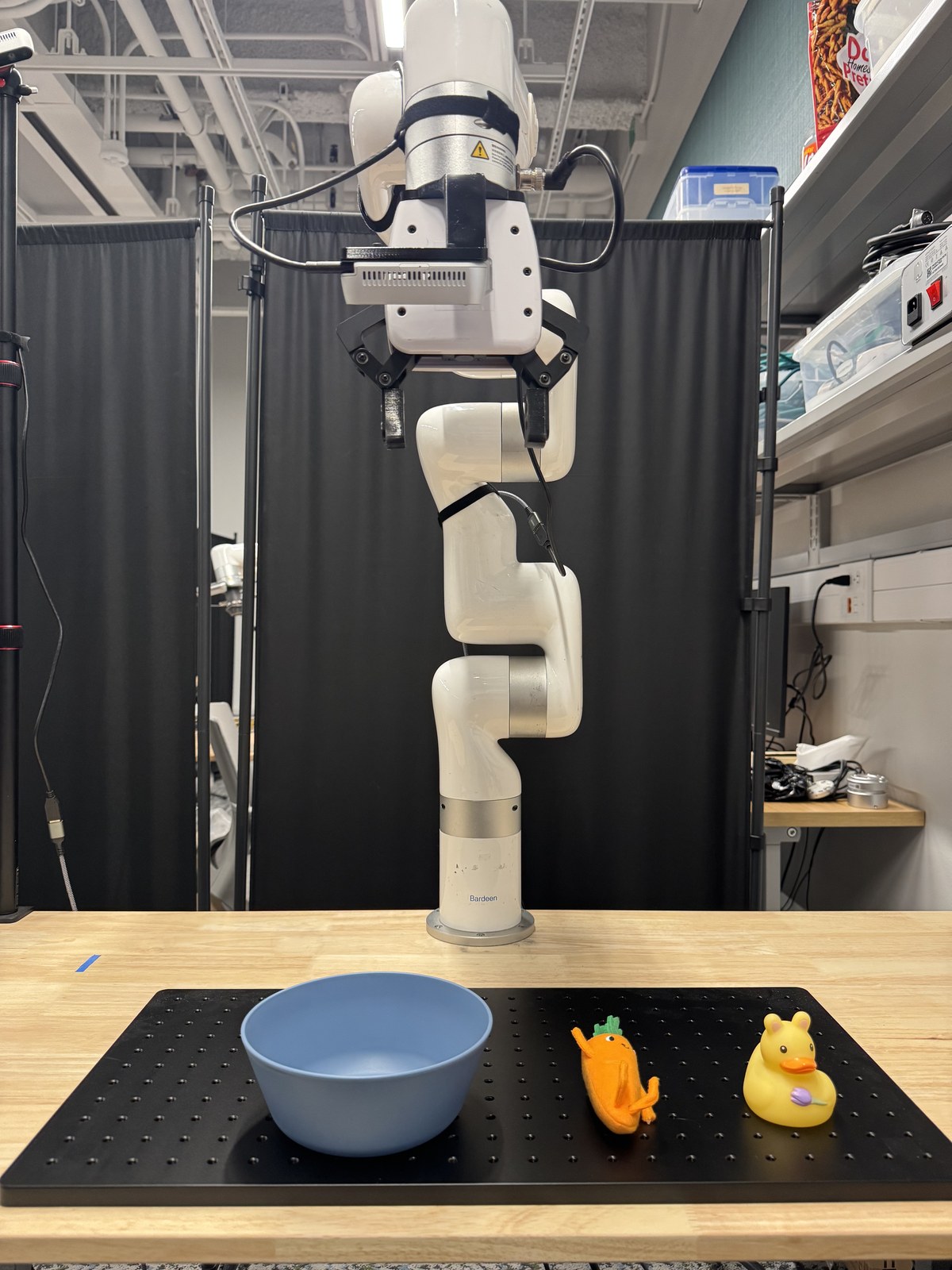}
{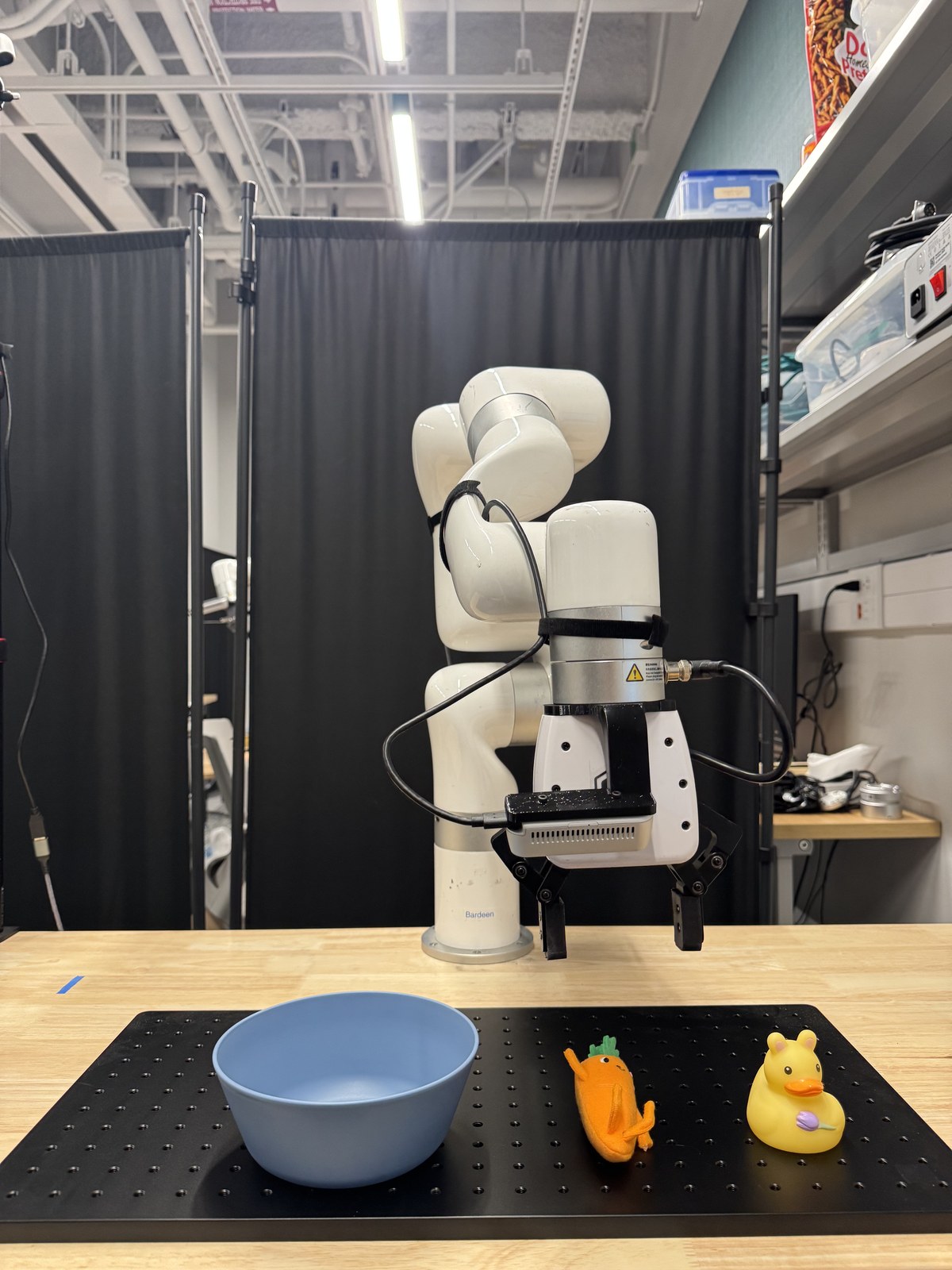}
{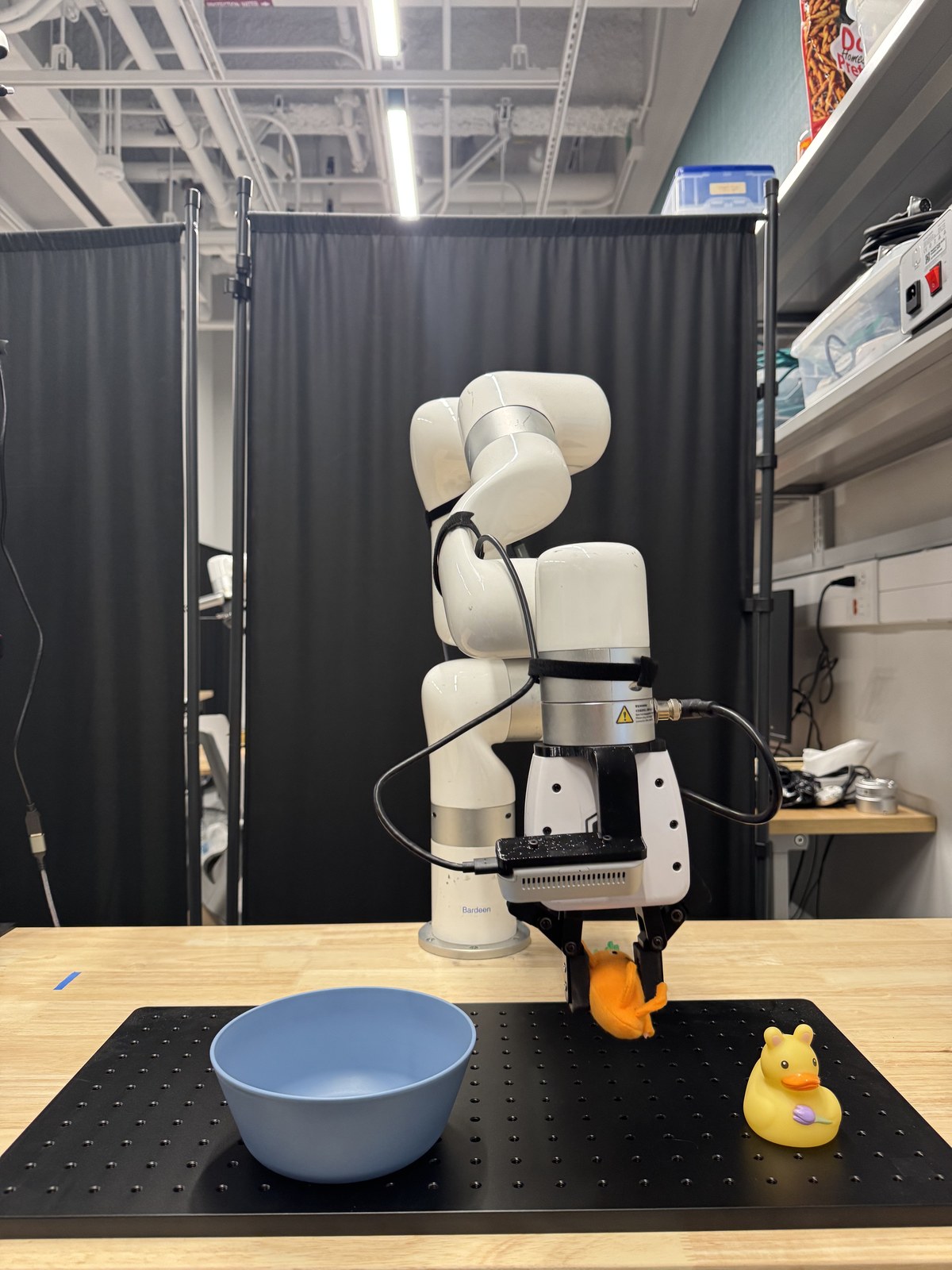}
{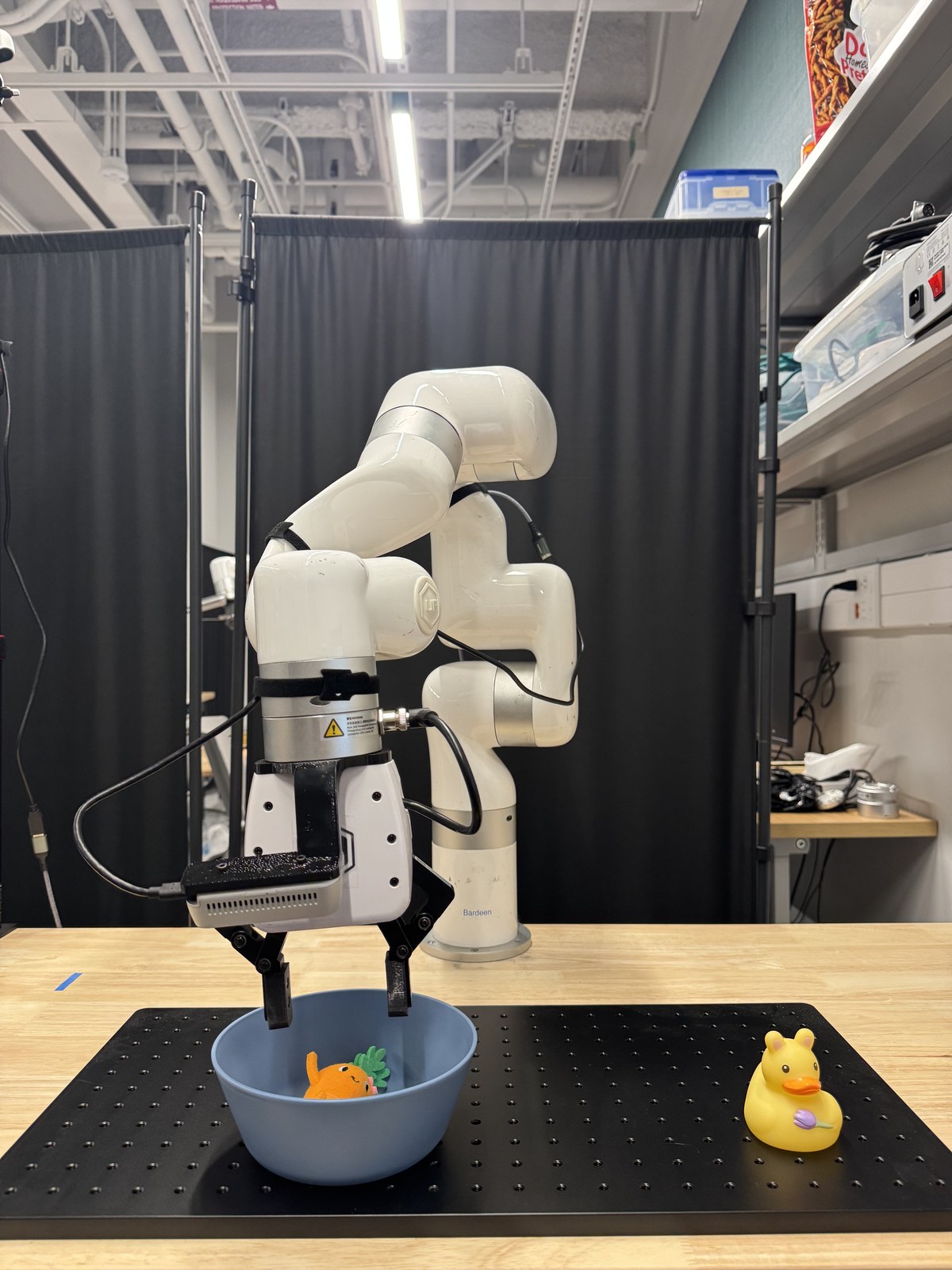}
{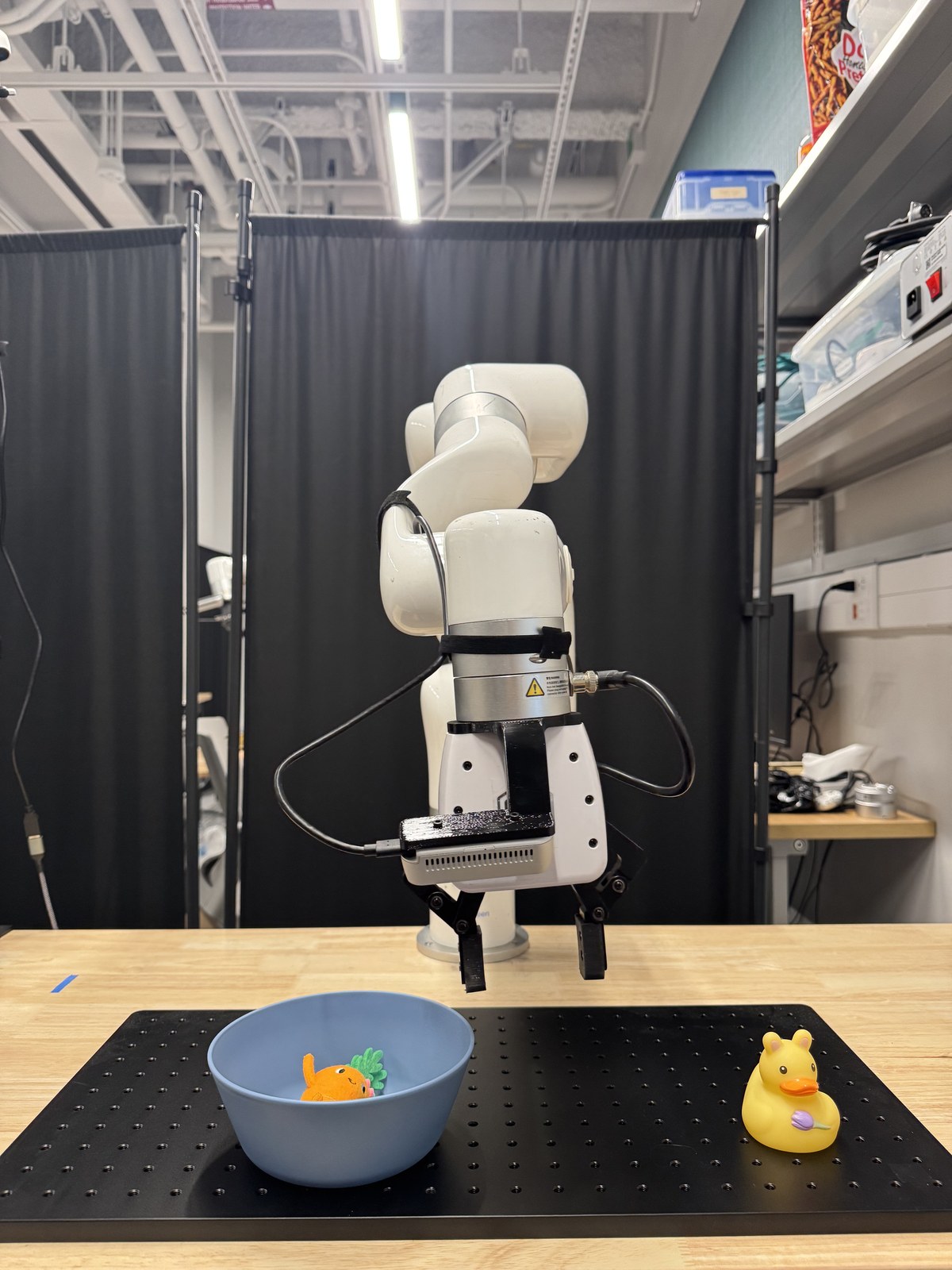}
{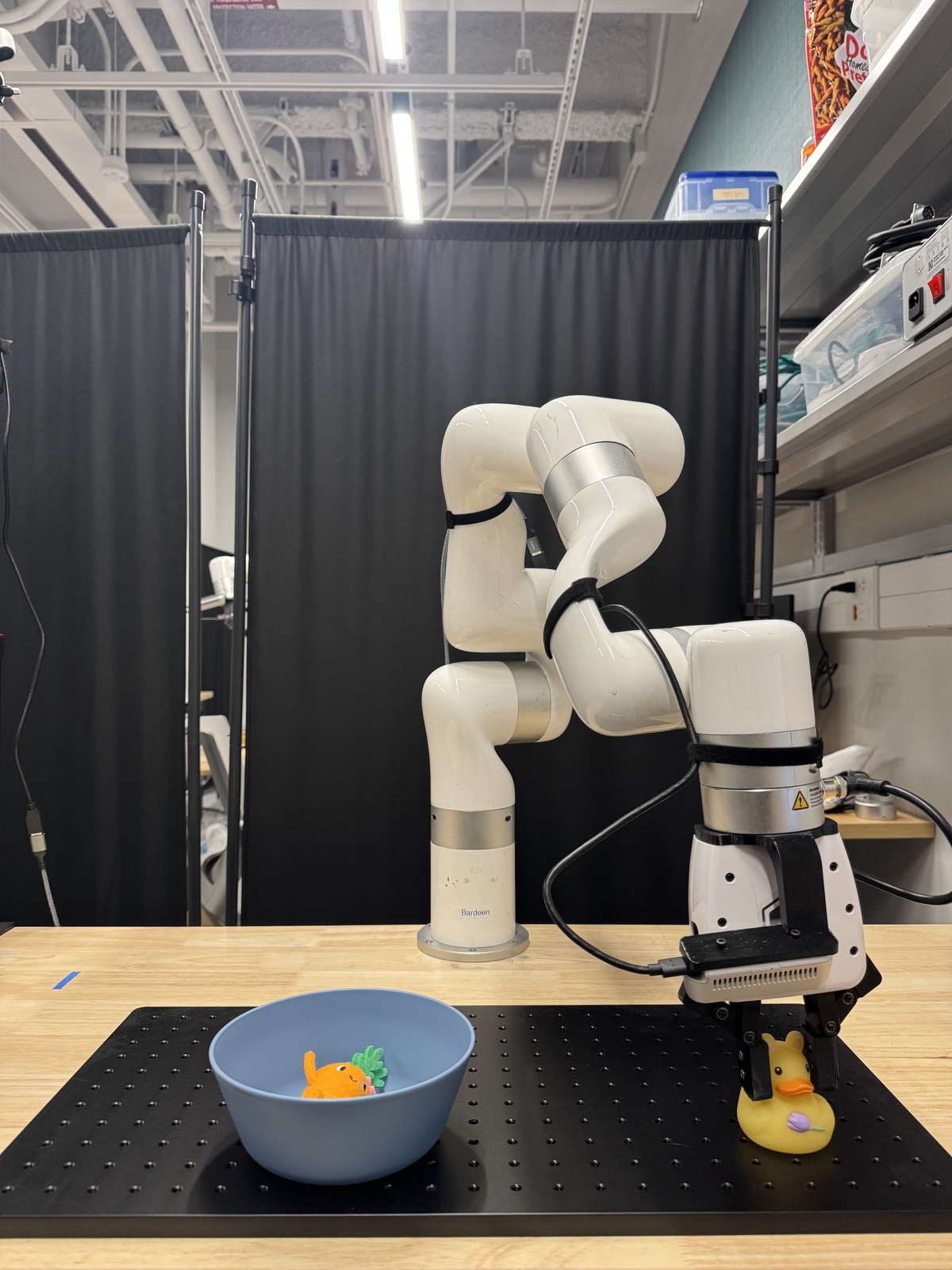}
{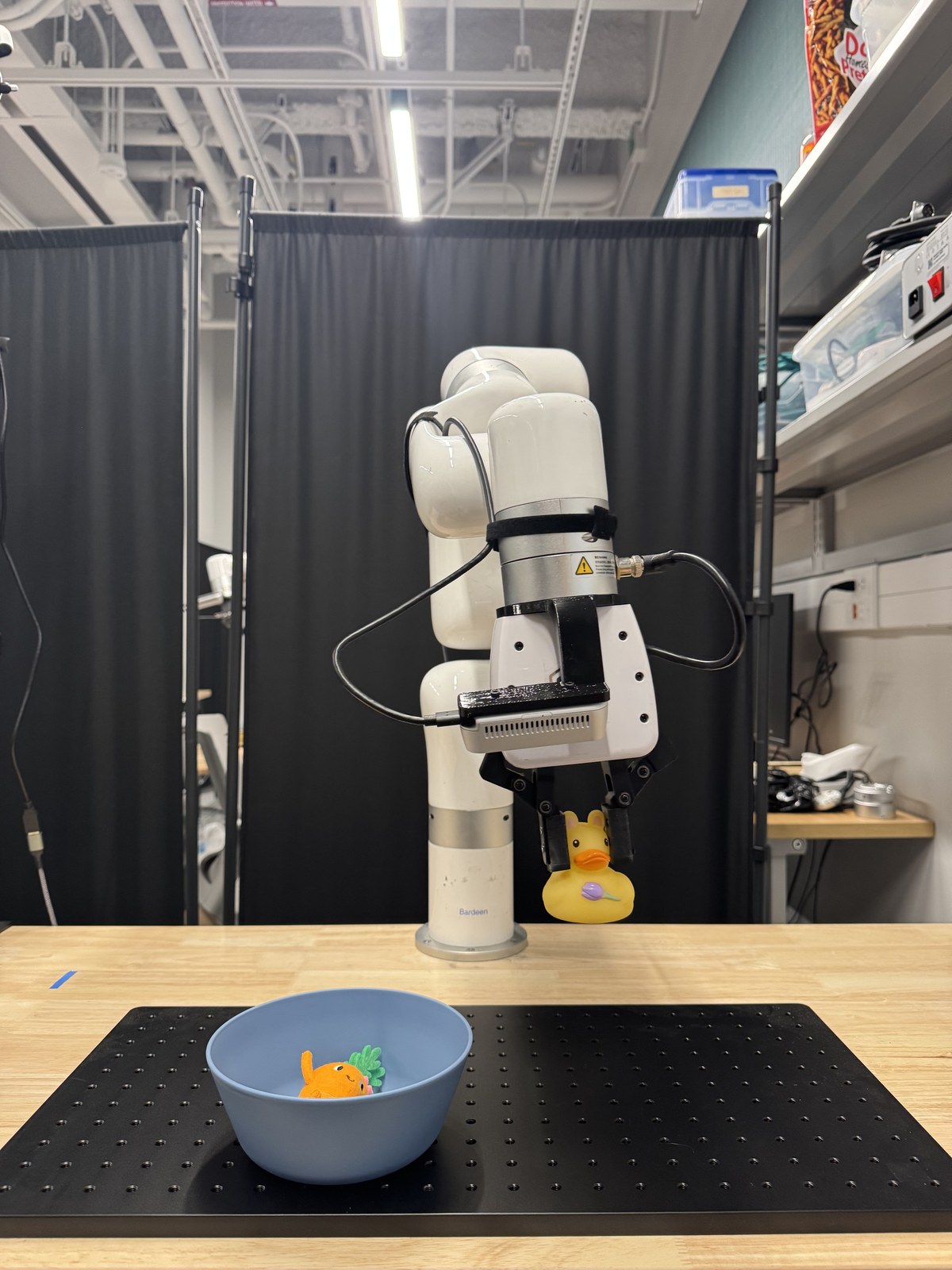}
{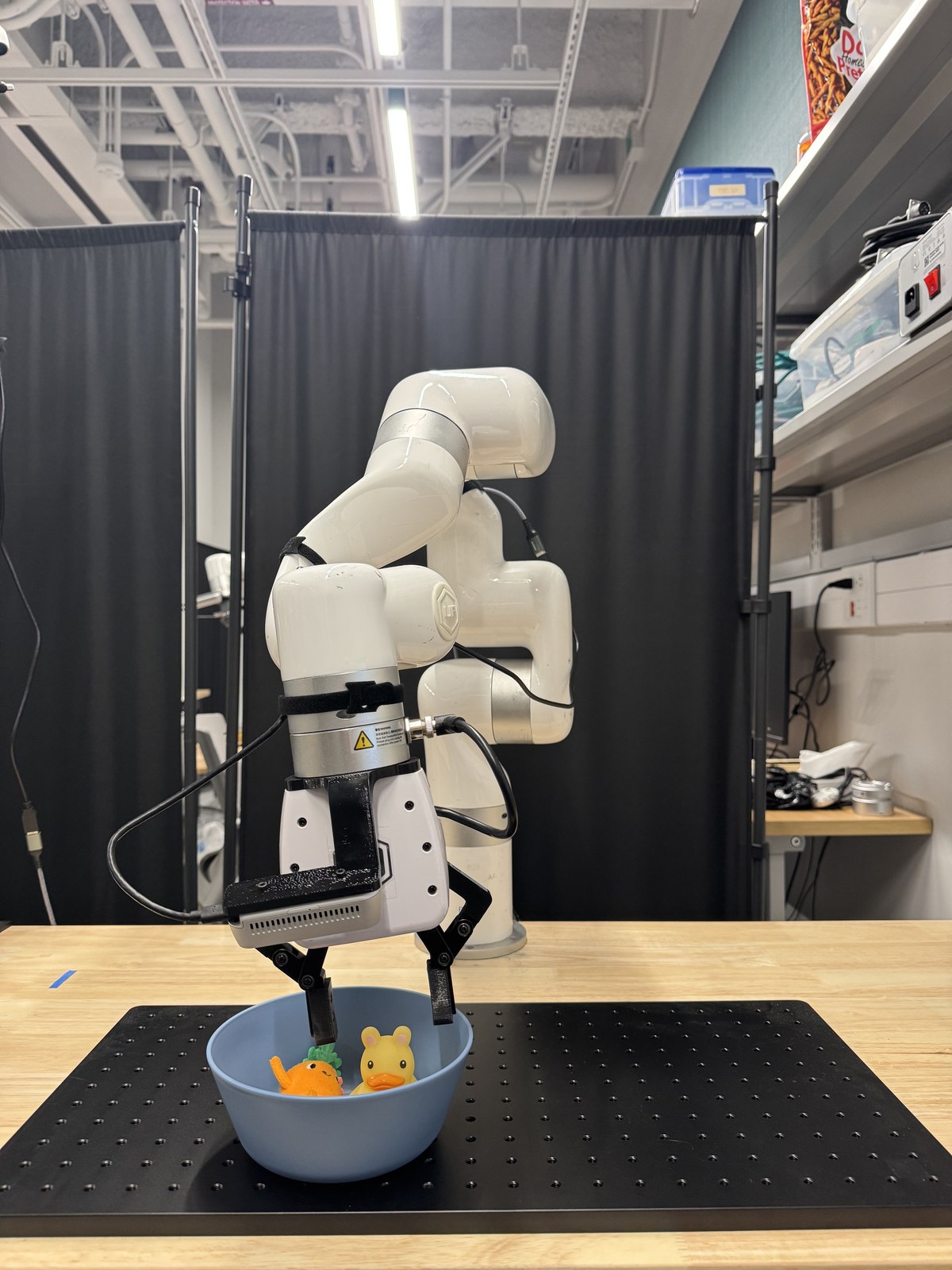}
\addlinespace[0.6em]

\taskblock{Task 8. Place the carrot in the bowl, then place the bowl on the plate.}
{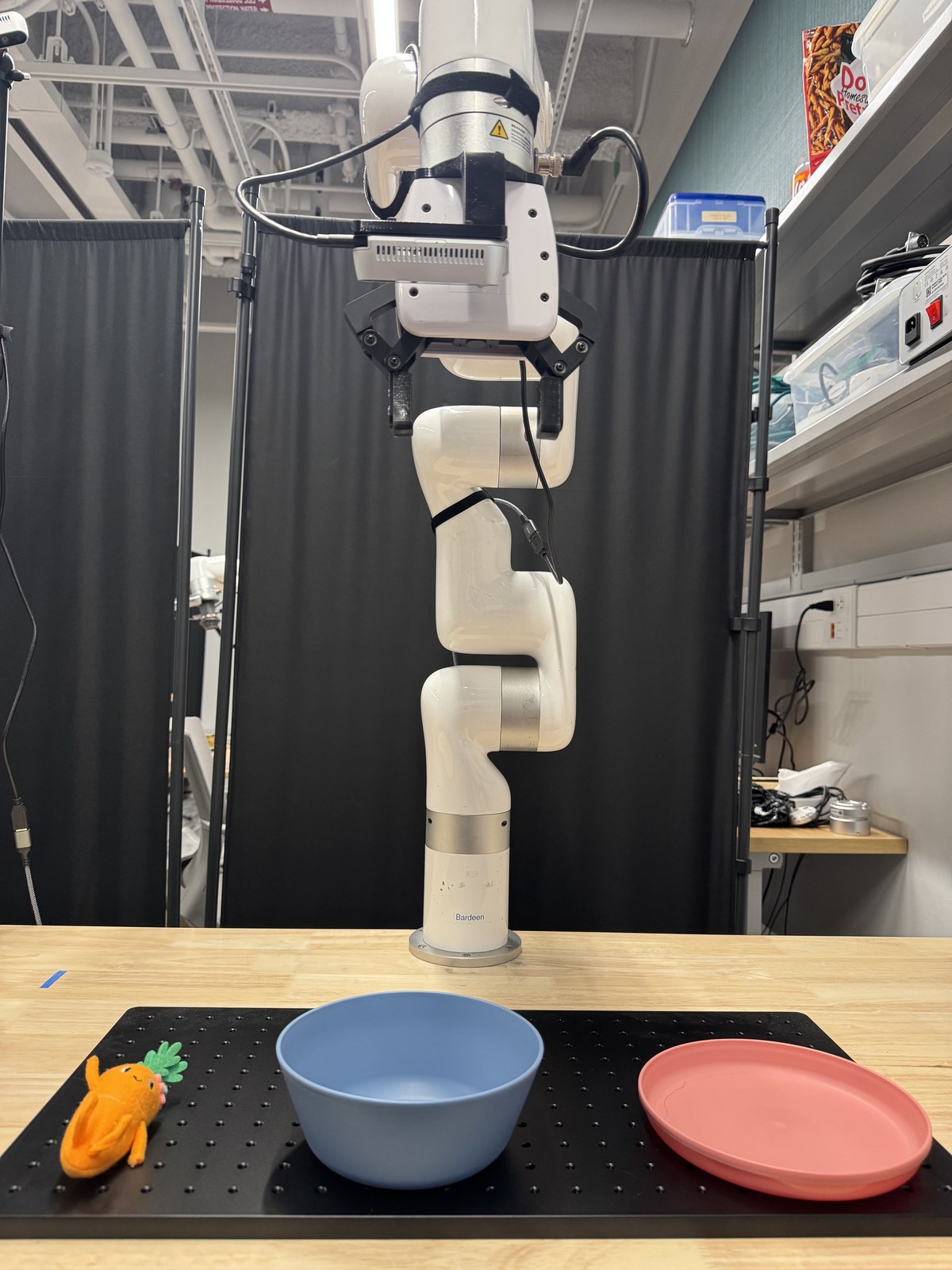}
{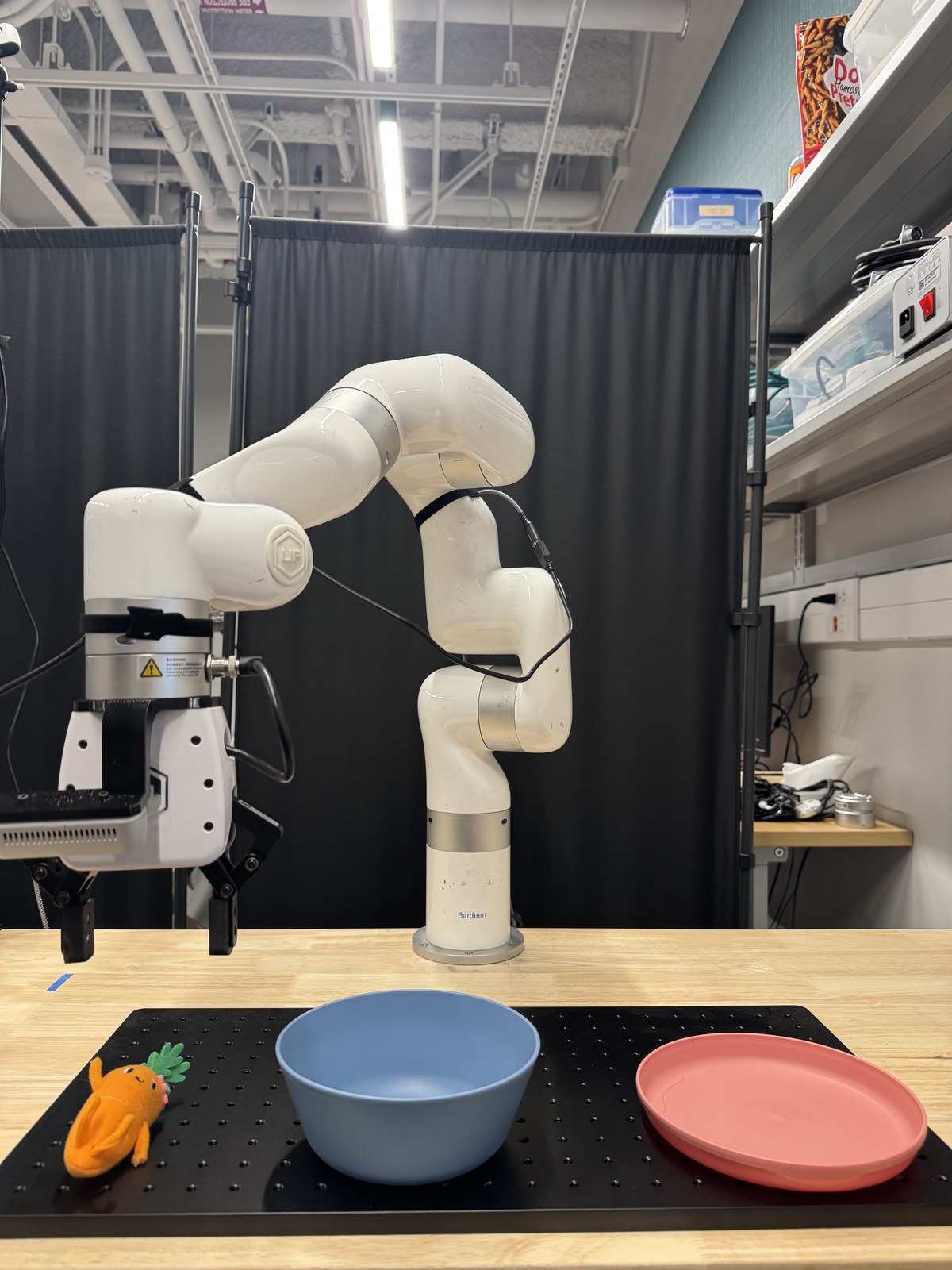}
{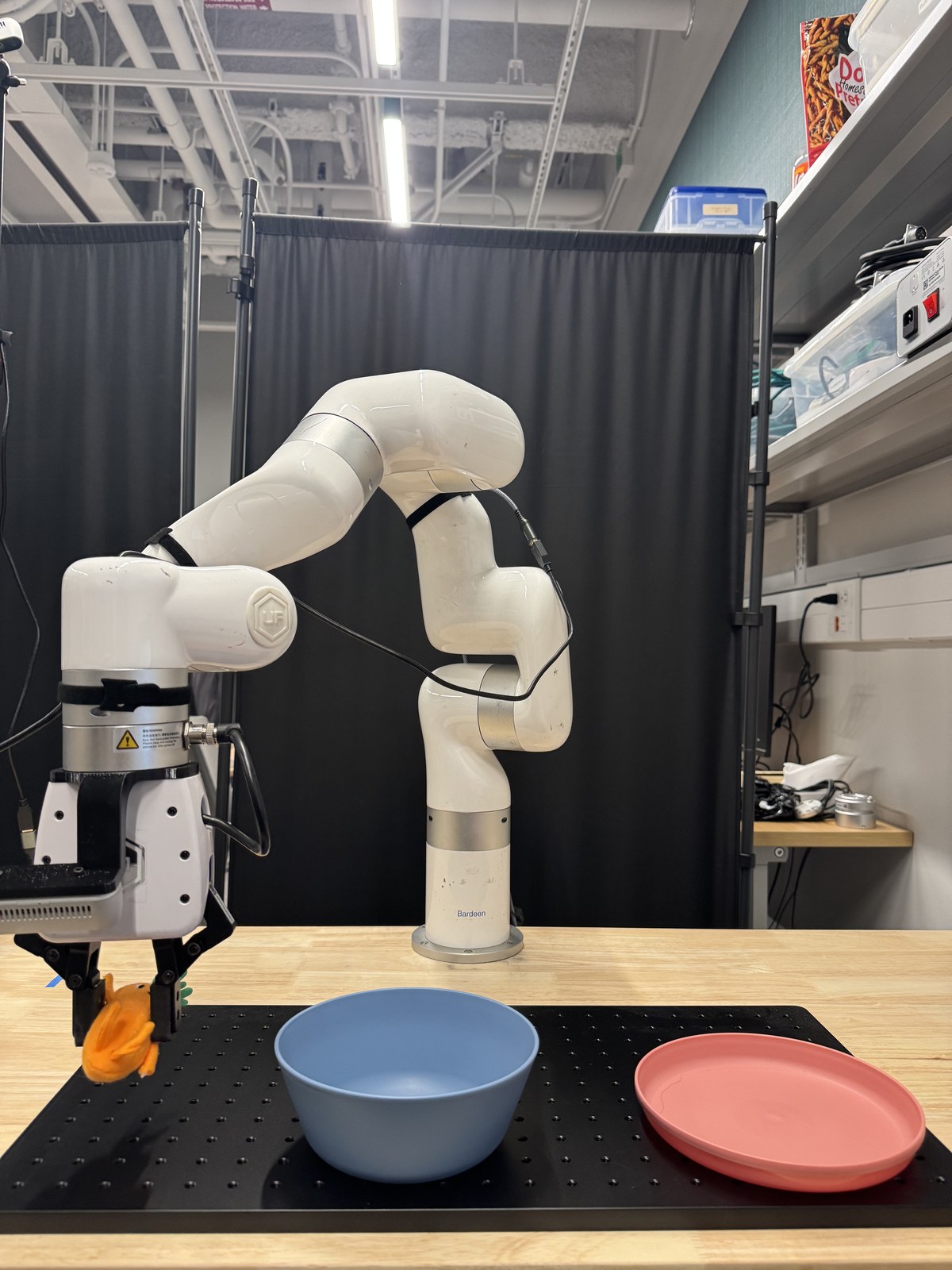}
{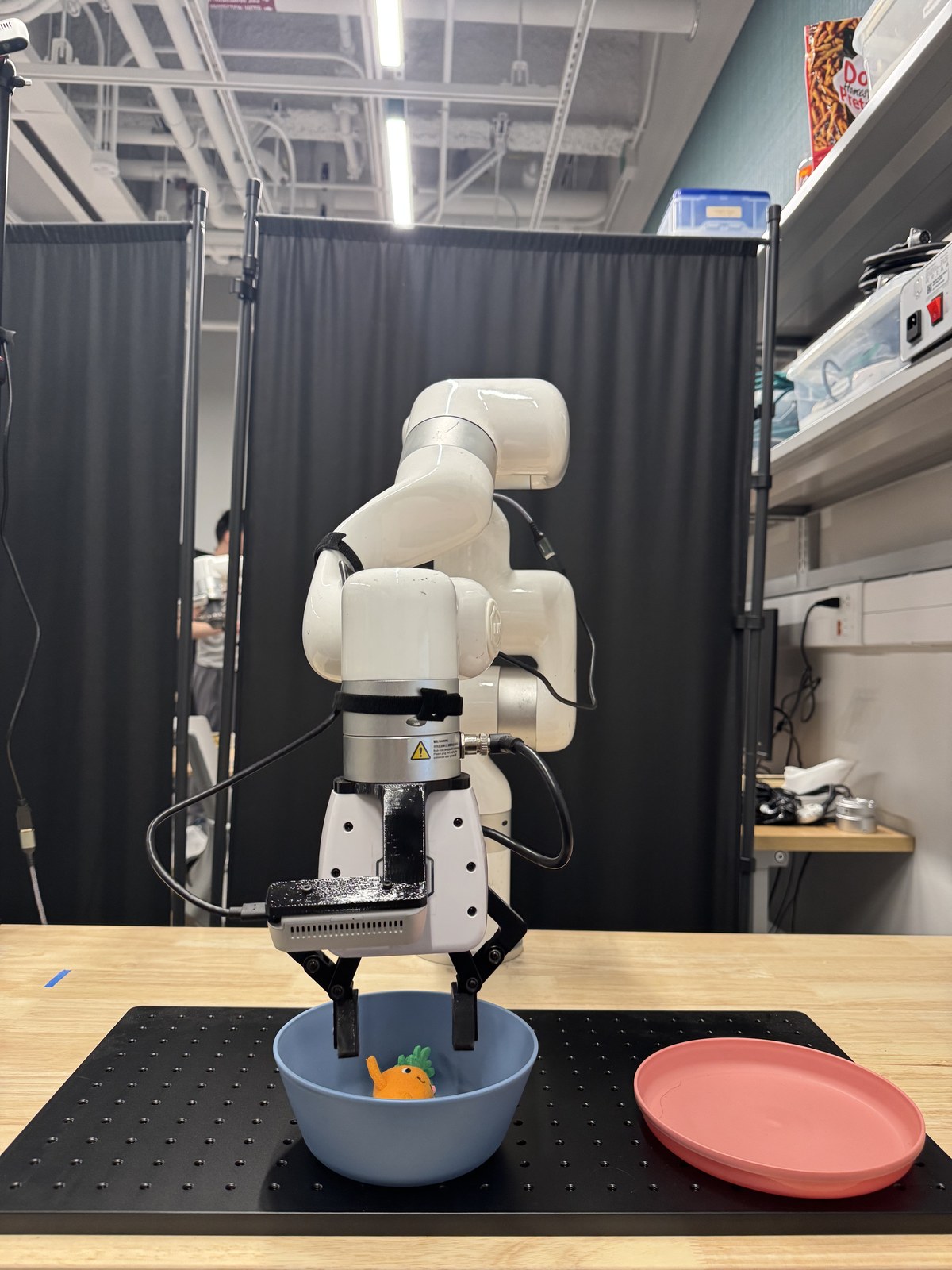}
{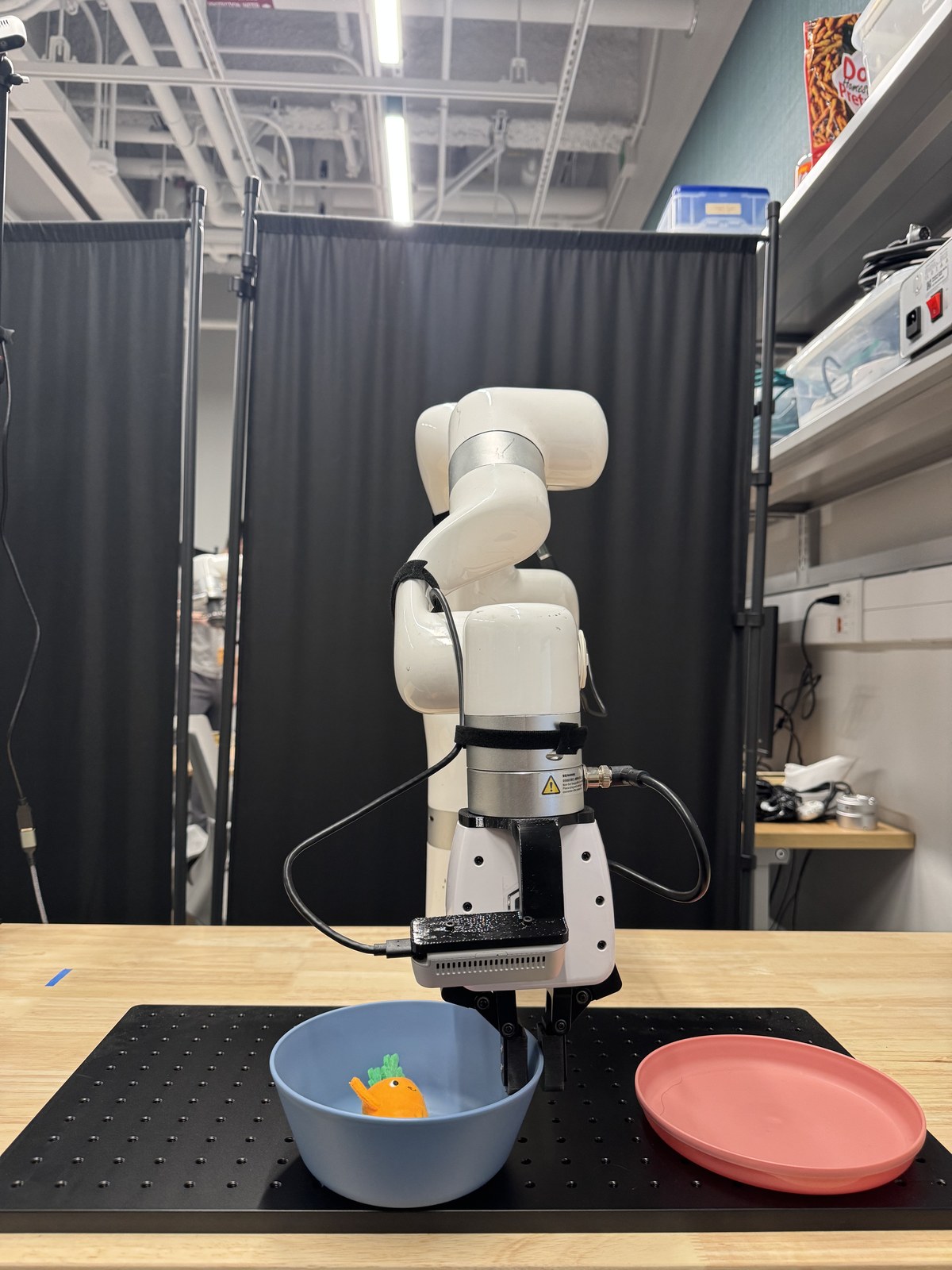}
{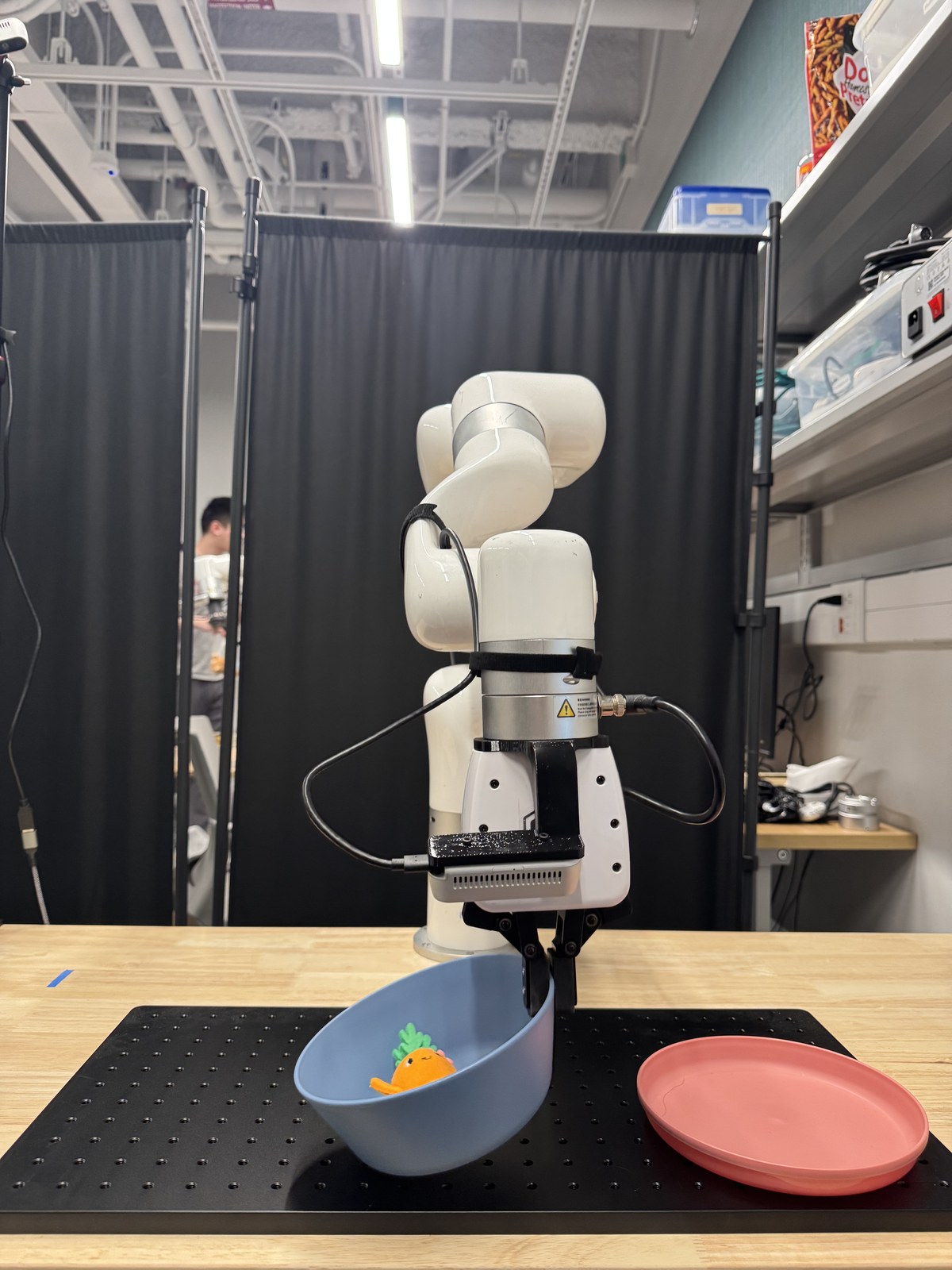}
{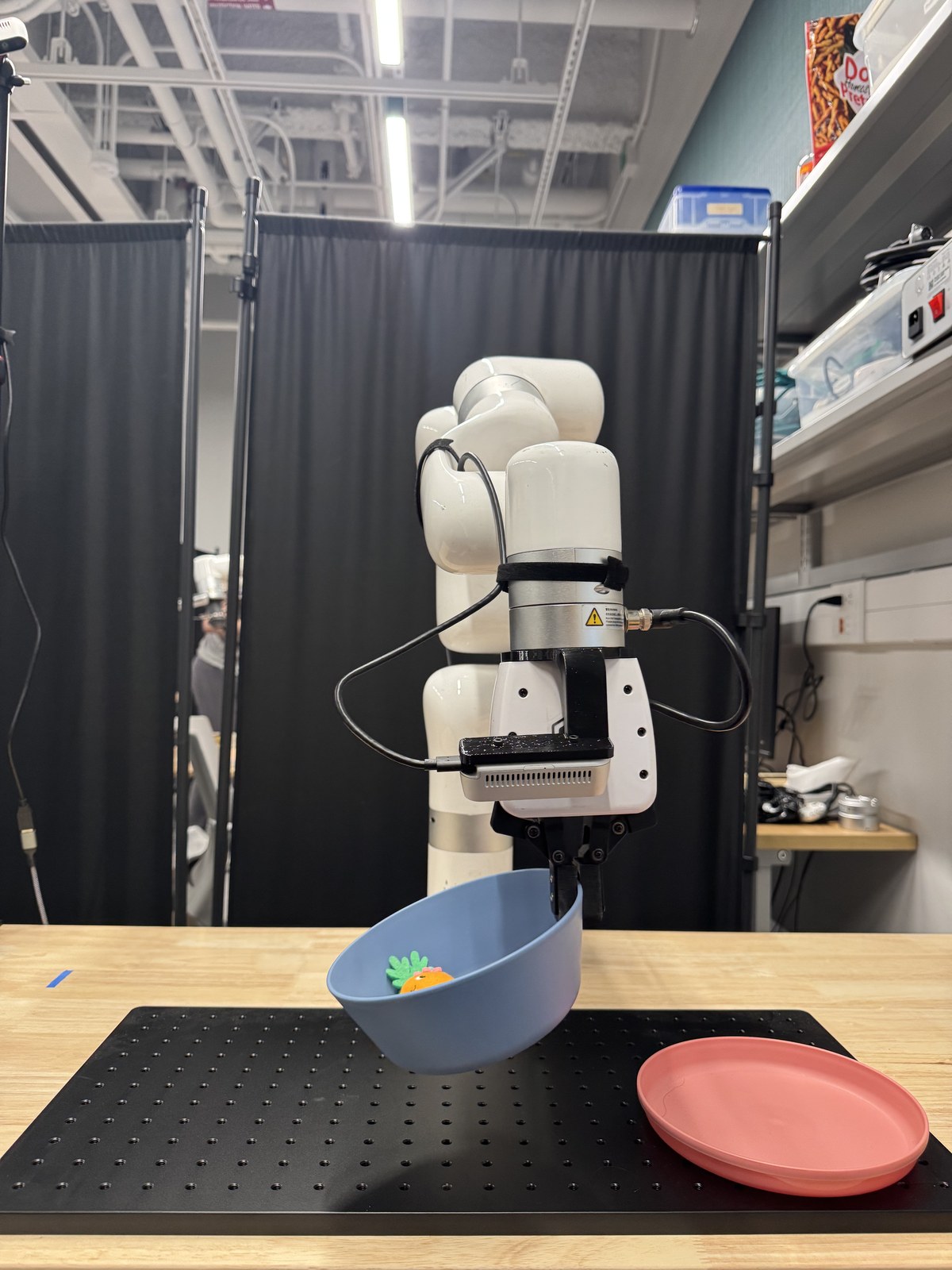}
{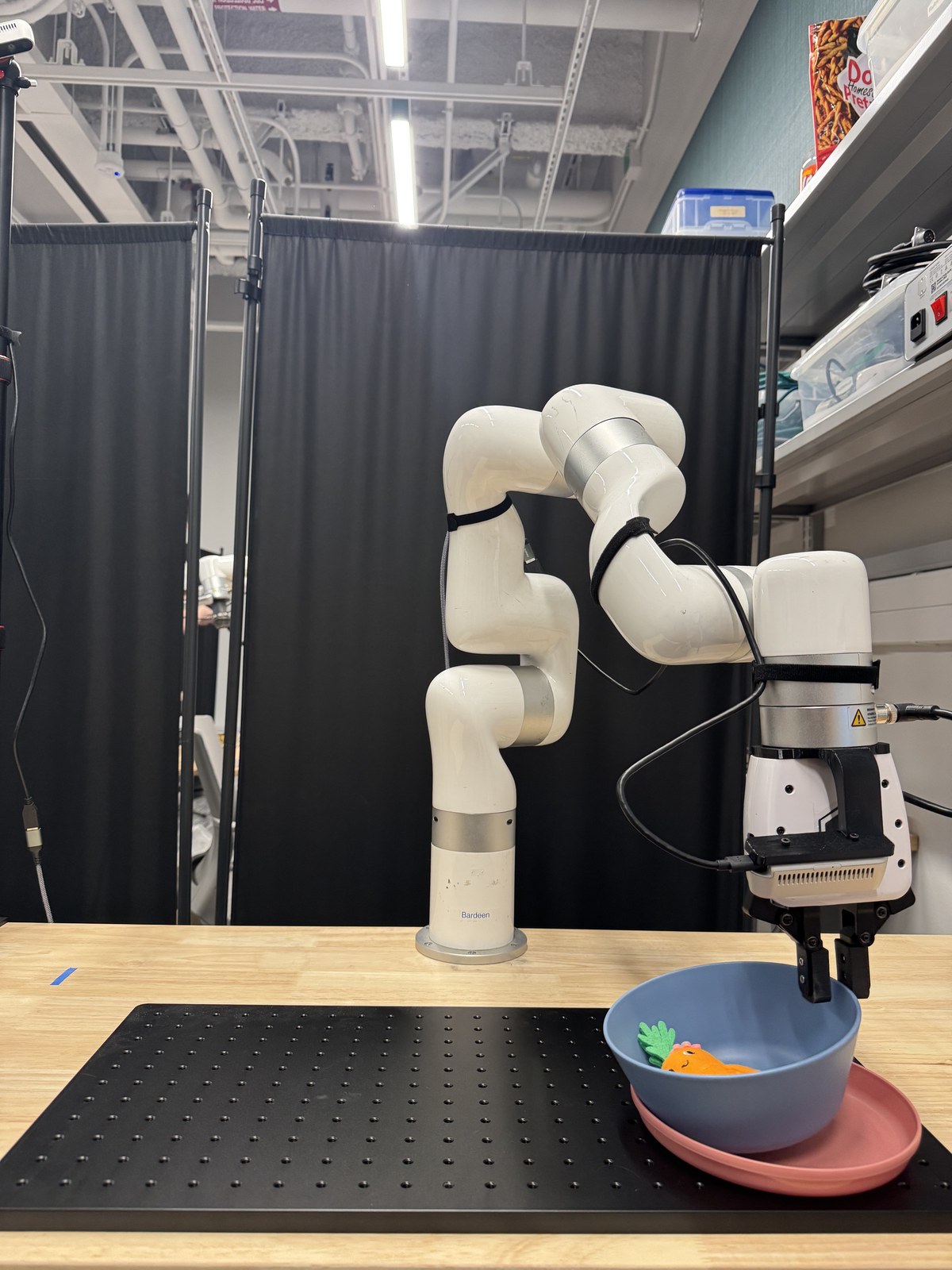}

\end{longtable}
}

\section{GPT-5.2-based Failure Onset Annotation}
\label{sec:gpt5}

\subsection{Annotation Protocol} \label{sec:protocol}
We prompt the state-of-the-art multimodal model GPT-5.2 to annotate the failure onset of each failed rollout.
Each recorded video is provided as input in an offline video question-answering setup, along with the prompt shown below.
Note that this setup is less challenging than the online video question-answering used for VLM-based runtime monitoring in~\cref{tab:vlm}, as GPT-5.2 has access to the entire recorded video rather than a streaming prefix.
We use these annotations solely for analyzing the ablation results in~\cref{fig:tradeoff,tab:onset_app,tab:openvla_pi0_lag_bacc}, not for evaluating our main metrics such as TWA.

\begin{tcolorbox}[colback=gray!10, colframe=black, title=Prompt for GPT-5.2-based failure onset annotation, breakable]
\textbf{System:} You are an expert robotics failure annotator.\\

\textbf{Goal:} Pinpoint the \textsc{failure start} frame index.\\

\textbf{Definition (use this strictly):}\\
\textsc{failure start} = the earliest frame $t$ in this window where an irreversible mistake first appears \textbf{or} where the agent's behavior becomes clearly unrecoverable and is not being corrected.\\

\textbf{You must:}
\begin{enumerate}
    \item Choose exactly one integer index as \texttt{failure\_start}.
    \item Provide 1--2 pieces of visual evidence referencing specific frames.
    \item If you cannot see any decisive change in this window, choose the earliest index where the failure signal is most apparent and set confidence low.
\end{enumerate}

\textbf{Output format (STRICT JSON only):}
\begin{verbatim}
{
  "failure_start": <int>,
  "confidence": 0.0-1.0,
  "evidence": [
    {"index": <int>, "cue": "what changes / what is wrong"},
    {"index": <int>, "cue": "optional second cue"}
  ],
  "failure_type": "drop | wrong-object | collision |
                   stuck/off-track | goal-violation | other",
  "recoverable_before_start": "yes | no | uncertain",
  "notes": "optional"
}
\end{verbatim}

\textbf{Inputs:}\\
Instruction: \textcolor{magenta}{\{instruction\}}\\
Window frames (in order):\\
\texttt{[index=\textcolor{magenta}{\{t\_prev\}}]} \textcolor{magenta}{\{frame\}}\\
\texttt{[index=\textcolor{magenta}{\{t\_prev+1\}}]} \textcolor{magenta}{\{frame\}}\\
\ldots\\
\texttt{[index=\textcolor{magenta}{\{t\_curr\}}]} \textcolor{magenta}{\{frame\}}\\

Now produce the JSON.
\end{tcolorbox}

\subsection{Human Agreement Analysis}
\label{sec:agree}
To validate the GPT-5.2 failure-onset annotations, we conducted a human-agreement study over $76$ failed rollouts rated by four PhD-level machine learning researchers.
Each rater was shown the recorded video with GPT-5.2's predicted onset marked by a green$\rightarrow$red border transition and asked to mark \emph{Agree} or \emph{Disagree}, permitting small temporal offsets.
Raters agreed with GPT-5.2 on $90.7\%$ of the $304$ ratings; all four raters unanimously agreed on $75.0\%$ of rollouts, and no rollout received unanimous disagreement.
This exceeds the $\sim\!80\%$ human--LLM agreement typically reported in recent ``LLM-as-a-judge'' studies with strong LLM judges~\cite{zheng2023judging}, which itself matches human--human agreement on the same tasks.
We therefore treat GPT-5.2 onset annotations as a reliable proxy for human-perceived failure onset in~\cref{fig:tradeoff,tab:onset_app,tab:openvla_pi0_lag_bacc}.
The study involved only viewing pre-recorded videos, collected no personal or sensitive information, and was conducted under a minimal-risk setting with voluntary participation.

\section{Baselines}
\label{sec:baselines}

We provide implementation details for all baselines used in our experiments. Each method operates at a different level of the VLA model: OOD detection-based (\cref{sec:ood_app}) and classifier-based (\cref{sec:class_app}) methods operate on VLA action embeddings $h_t \in \mathbb{R}^d$ extracted at each timestep $t$ (see~\cref{sec:tfa} for extraction details); 
multi-sampling-based methods (\cref{sec:multi_app}) operate on sampled actions $\{a_t^{(n)}\}_{n=1}^{N}$ drawn from the policy output; and token uncertainty-based methods (\cref{sec:token_app}) operate on per-token likelihoods of autoregressive policies. For embedding-level methods, we denote by $\mathcal{H}_{\mathrm{succ}}$ and $\mathcal{H}_{\mathrm{fail}}$ the sets of all embeddings from successful and failed rollouts in the training set $\mathcal{D}_{\mathrm{train}}$, respectively.
Regardless of input modality, each method produces a per-timestep failure score $s_t$ at evaluation time.

\subsection{Out-of-Distribution Detection-based Methods}
\label{sec:ood_app}

\subsubsection{Embedding Distance}
Embedding-distance methods use feature-space proximity to compute failure scores directly, without training any neural network. While standard OOD detection models only consider the successful embeddings, we extend these baselines to exploit the same trajectory-level failure supervision available to our method, following recent practice~\cite{gu2025safe}. 
Specifically, we contrast the distance of the query embedding $h_t$ to both the successful and failed embedding sets:
$
s_t = d(h_t, \mathcal{H}_{\mathrm{succ}}) 
       - d(h_t, \mathcal{H}_{\mathrm{fail}}),
$ where $d(\cdot, \cdot)$ is a distance function between a vector and a set. Intuitively, $h_t$ receives a high failure 
score when it lies far from successful embeddings and close to 
failed ones.

\paragraph{Mahalanobis distance~\cite{lee2018simple}.}
We fit a Gaussian to each embedding set and compute the Mahalanobis
distance to its mean:
\begin{equation}
d_{\mathrm{Maha}}(h_t, \mathcal{H}) =
\sqrt{(h_t - \mu_{\mathcal{H}})^{\top}
\Sigma_{\mathcal{H}}^{-1}
(h_t - \mu_{\mathcal{H}})},
\end{equation}
where $\mu_{\mathcal{H}}$ and $\Sigma_{\mathcal{H}}$ are the 
sample mean and covariance of $\mathcal{H}$. This captures 
deviations in a direction-sensitive manner.

\paragraph{Cosine $k$-NN~\cite{sun2022out}.}
We compute the average cosine distance between $h_t$ and 
its $k$ nearest neighbors in $\mathcal{H}$:
\begin{equation}
d_{\mathrm{cos}}(h_t, \mathcal{H}) = 
\frac{1}{k} \sum_{h_j \in \mathcal{N}_k(h_t, \mathcal{H})} 
\left(1 - \frac{h_t^{\top} h_j}
                {\|h_t\| \|h_j\|}\right),
\end{equation}
where $\mathcal{N}_k(h_t, \mathcal{H})$ denotes the $k$ 
nearest neighbors of $h_t$ in $\mathcal{H}$ under cosine 
similarity. Unlike Mahalanobis distance, this is non-parametric 
and captures local neighborhood structure.

\paragraph{PCA-KMeans~\cite{liu2024multi}.}
We first project all embeddings in $\mathcal{H}$ onto their top 
$p$ principal components, then cluster the projected embeddings 
into $K$ clusters via $k$-means. The distance to $\mathcal{H}$ 
is the Euclidean distance from $h_t$ (similarly projected) 
to the nearest cluster centroid:
\begin{equation}
d_{\mathrm{PCA}}(h_t, \mathcal{H}) = 
\min_{k=1,\ldots,K} \|P(h_t) - \mu_k\|_2,
\end{equation}
where $P(\cdot)$ denotes the PCA projection and $\{\mu_k\}$ the 
cluster centroids.

\subsubsection{Learned OOD Detector}
Learned OOD detectors train a neural network 
$f_{\mathrm{OOD}}(\cdot)$ to model the embedding distribution 
of a given set and output a scalar score. 
Following the same 
pairwise supervision strategy, we adapt 
these methods to the action embedding space by training two 
separate models~\cite{gu2025safe} 
$f^{\mathrm{OOD}}_{\mathrm{succ}}$ and 
$f^{\mathrm{OOD}}_{\mathrm{fail}}$, on $\mathcal{H}_{\mathrm{succ}}$ 
and $\mathcal{H}_{\mathrm{fail}}$ respectively, and compute:
$
s_t = f^{\mathrm{OOD}}_{\mathrm{succ}}(h_t) 
       - f^{\mathrm{OOD}}_{\mathrm{fail}}(h_t).
$

\paragraph{RND (Random Network Distillation)~\cite{he2024rediffuser}.}
RND uses a fixed randomly initialized target network 
$\bar{\psi}$ and a trainable predictor $\psi$, both operating 
on the action embedding $h_t \in \mathcal{H}$. The per-head 
score is the predictor--target mismatch:
\begin{equation}
f^{\mathrm{OOD}}_{\mathrm{RND}}(h_t)
=\|\psi(h_t)-\bar{\psi}(h_t)\|_2.
\end{equation}
The predictor is trained to minimize this distance on the 
corresponding split, so a lower error indicates 
$h_t$ lies within that distribution, while higher error 
indicates it does not.

\paragraph{LogpZO~\cite{xu2025can}.}
In our implementation, LogpZO is realized via a flow-matching 
velocity model. The per-head score is an energy-style residual:
\begin{equation}
f^{\mathrm{OOD}}_{\mathrm{LogpZO}}(h_t)=\|h_t+v_\theta(h_t,0)\|_2^2,
\end{equation}
where $v_\theta$ is the learned velocity field. Higher values 
indicate greater out-of-distribution risk.

\subsection{Multi-Sampling-based Methods}
\label{sec:multi_app}
Multi-sampling methods estimate uncertainty by drawing $N$ 
action samples $\{a_t^{(1)}, \ldots, a_t^{(N)}\}$ from the VLA 
policy at each timestep and computing a per-step uncertainty 
score. We use $N=10$ samples across all methods.

\paragraph{Cluster Entropy~\cite{kuhn2023semantic}.}
Following the implementation of~\cite{gu2025safe}, each timestep \(t\), we cluster the sampled actions using agglomerative hierarchical clustering with a distance threshold \(\varepsilon\).
Let \(n_k\) be the size of cluster \(k\), and \(p_k = n_k / N\). The uncertainty score is the Shannon entropy of the induced cluster-size distribution:
\begin{equation}
s_t = -\sum_{k=1}^{K_t} p_k \log p_k,
\end{equation}
where \(K_t\) is the number of clusters formed at timestep \(t\) under threshold \(\varepsilon\). Higher entropy indicates greater disagreement among sampled actions.

\paragraph{EigenScore~\cite{chen2024inside}}
measures uncertainty via the spectral dispersion of 
the sample covariance matrix 
$\Sigma_t = \mathrm{Cov}(\{a_t^{(n)}\})$:
\begin{equation}
s_t = \log \det(\Sigma_t + \epsilon I),
\end{equation}
where $\epsilon$ is a small regularization constant. Larger determinant implies greater variance across directions.
Notably, we compute EigenScore in the action space rather than the intermediate feature space to ensure consistency with other multi-sampling baselines and to directly reflect variability in predicted actions.

\paragraph{STAC (Statistical Measures of Temporal Action Consistency)~\cite{agia2024unpacking}}
detects erratic failures by measuring how much a generative policy's action distributions change over time.
For two consecutive policy queries at timesteps $t$ and $t+k$, the sampled action chunks $\{a_{t+k:t+h-1|t}^{(i)}\}_{i=1}^{N}$ and $\{a_{t+k:t+h-1|t+k}^{(j)}\}_{j=1}^{N}$ share the same $(h-k)$-step overlap window and should agree if the policy is temporally consistent. The failure score is a statistical distance between these two action distributions over the overlap:
\begin{equation}
s_t = D_{\mathrm{stat}}\!\left(
\{a_{t+k:t+h-1|t}^{(i)}\}, 
\{a_{t+k:t+h-1|t+k}^{(j)}\}
\right),
\end{equation}
where $D_{\mathrm{stat}}$ is implemented via Maximum Mean 
Discrepancy (MMD) following the original paper.
Large distribution drift across queries indicates erratic policy 
behavior.

Since OpenVLA does not emit action chunks but produces a single immediate action per timestep, there is no overlap between consecutive policy queries, and STAC is not directly applicable. We adapt STAC to this setting by reinterpreting its core principle--\emph{local temporal consistency of the policy's action distribution}--to operate on single-step actions sampled at adjacent timesteps.
Specifically, at each timestep $t$, we sample $N$ actions from OpenVLA at steps $t$ and $t+1$ and compute the MMD between the two action distributions:
\begin{equation}
s_t = \mathrm{MMD}\!\left(
\{a_t^{(i)}\}_{i=1}^{N}, \{a_{t+1}^{(j)}\}_{j=1}^{N}
\right).
\end{equation}
A large MMD indicates abrupt change in the sampled action distribution between adjacent steps, analogous to the temporal inconsistency STAC detects in chunk-based policies.

\paragraph{ACE (Action-Chunk Entropy)~\cite{romer2025failure}}
estimates uncertainty by computing 
histogram-based entropy over $N$ sampled action chunks.
To avoid the curse of dimensionality across the chunk horizon $H$, ACE treats each prediction step separately: for each step $t+i$, it bins the $N$ sampled actions into an adaptive grid and sums per-step entropies:
\begin{equation}
s_t = \sum_{i=0}^{H-1} \hat{H}\big(a_{t+i|t}^{(1)}, \ldots, 
a_{t+i|t}^{(N)}\big),
\end{equation}
where $\hat{H}(\cdot)$ is the Shannon entropy of the histogram 
over the $N$ sampled actions. High entropy reflects persistent 
ambiguity in the policy's intended behavior.
For OpenVLA, which outputs a single action per step ($H=1$), the summation reduces to a single-step histogram entropy over the $N$ sampled actions.

\subsection{Classifier-based Methods}
\label{sec:class_app}

Classifier-based methods train a neural network on VLA internal embeddings $h_t$ to predict per-timestep failure likelihood from trajectory-level labels $y \in \{0, 1\}$ (with $y=1$ denoting failure). We evaluate two SAFE~\cite{gu2025safe} variants that differ in backbone and loss. Both propagate trajectory-level labels across timesteps, overlooking temporal localization and introducing label noise from pre-onset normal actions.

\paragraph{SAFE-MLP} uses a two-layer MLP $g(\cdot)$ applied independently 
at each timestep, treating each embedding as an i.i.d.\ input. 
The per-timestep score is passed through a sigmoid, and the 
failure score at time $t$ is defined as the accumulated 
sum of scores up to $t$: $
s_t^{\mathrm{MLP}} 
= \sum_{\tau=1}^{t} \sigma\big(g(h_\tau)\big),
$
so that $0 \le s_t^{\mathrm{MLP}} \le t$. The model is trained 
with an asymmetric $L_1$ loss that pushes scores up for failed 
rollouts and down for successful ones:
\begin{equation}
\label{eq:safe_mlp}
\mathcal{L}_{\mathrm{MLP}} = 
\sum_{i \in \mathcal{D}_{\mathrm{train}}} 
\Big[\, y_i \sum_t (t - s_t) 
     + (1 - y_i) \sum_t s_t \,\Big].
\end{equation}
Note that SAFE-MLP injects temporal structure through a linearly increasing target weight via the cumulative score formulation, but this heuristic is agnostic to when failure actually occurs and treats every later timestep as more failure-indicative regardless of the underlying dynamics.

\paragraph{SAFE-LSTM} uses a single-layer LSTM to sequentially process the 
embedding stream $h_{1:t}$ and project the hidden state 
into a scalar score, normalized by a sigmoid:$
s_t^{\mathrm{LSTM}} = \sigma\big(\mathrm{LSTM}(h_{1:t})\big) 
\in [0, 1].
$
The model is trained with a binary cross-entropy loss applied 
uniformly to all timesteps:
\begin{equation}
\mathcal{L}_{\mathrm{LSTM}} = 
-\sum_{i \in \mathcal{D}_{\mathrm{train}}} \sum_t 
\Big[\, y_i \log s_t + (1 - y_i) \log(1 - s_t) \,\Big].
\end{equation}
To mitigate the class imbalance between the successful and the failed rollouts, both variants weight the loss contributions by inverse class frequency and apply $L_2$ regularization on model weights.

\subsection{Token Uncertainty-based Methods}
\label{sec:token_app}

Inspired by uncertainty quantification methods for LLM hallucination detection, we adapt token-level probability-based scores to autoregressive VLA models such as OpenVLA, which decode actions as a sequence of discrete tokens.
At each timestep $t$, the VLA generates $K$ action tokens $w_t^{(1)}, \ldots, w_t^{(K)}$, where each token $w_t^{(k)}$ is assigned probability $p_t^{(k)} = p(w_t^{(k)} \mid o_t, w_t^{(1:k-1)})$. 

\paragraph{Max / Mean Negative Log-likelihood ~\cite{ren2022out}.}
We aggregate the negative log-likelihood across the $K$ action tokens:
\begin{equation}
s_t^{\mathrm{max\text{-}nll}} = \max_{k=1,\ldots,K} 
\big(-\log p_t^{(k)}\big), 
\qquad
s_t^{\mathrm{avg\text{-}nll}} = (
-\frac{1}{K} \sum_{k=1}^{K} \log p_t^{(k)}
).
\end{equation}
The max variant captures the least confident token within an action, highlighting worst-case uncertainty, while the mean provides a smoother estimate of overall uncertainty across tokens.

\paragraph{Max / Mean Token Entropy~\cite{malinin2020uncertainty}.}
Entropy-based scores aggregate the per-token entropy over the action-token vocabulary $\mathcal{V}$. 
Let $E_t^{(k)} = - \sum_{v \in \mathcal{V}} p_t^{(k)}(v) \log p_t^{(k)}(v)$ denote the entropy at token position $k$. Then,
\begin{equation}
s_t^{\mathrm{max\text{-}ent}} = \max_{k=1,\ldots,K} E_t^{(k)}, 
\qquad
s_t^{\mathrm{avg\text{-}ent}} = \frac{1}{K} \sum_{k=1}^{K} E_t^{(k)}.
\end{equation}
Unlike log-likelihood-based scores, which depend on the sampled tokens, entropy-based scores reflect the full predictive distribution.

\subsection{VLM-based Method}
\label{sec:vlm}

We implement a VLM-based runtime monitor using Qwen3-VL-8B-Instruct~\cite{bai2025qwen3}.
At each timestep $t$, we construct a temporal context consisting of previously sampled frames up to the current frame.
The model performs online video question answering (VQA) to predict whether the current frame is in a failure state.
A trajectory is classified as a failure if any timestep is predicted as a failure.

\vspace{0.2cm}
\begin{tcolorbox}[colback=gray!10, colframe=black, title=Prompt for Qwen3-VL-based runtime failure detection, breakable]
\textbf{System:} You are a robot monitoring assistant.\\
\textbf{Goal:} Determine whether the robot is currently in a failure state at the latest frame.\\
\textbf{Context:} You are given a chronological sequence of frames ending at the current timestep 
(up to \textcolor{magenta}{\{H\}} frames).\\
\textbf{Task instruction:}\textcolor{magenta}{\{instruction\}}\\
\textbf{Question:}At the last frame (current timestep), is the robot in a failure state?\\
Output strictly JSON with keys: failure\_now (0 or 1), reason (short string).
\end{tcolorbox}

\section{Extended Related Work}
\label{sec:morerelated}

\paragraph{Hallucination Detection.}
Hallucinations refer to outputs generated by Large Language Models (LLMs) or Large Vision-Language Models (LVLMs) that are linguistically plausible but factually unsupported.
Detecting hallucinations has become a critical research area due to safety concerns in real-world deployments~\cite{ji2023survey,liu2024survey}.
A broad line of work approaches this problem through uncertainty-based scoring functions.
Logit-based methods use token-level probabilities as uncertainty estimates~\cite{ren2022out,malinin2020uncertainty, zhou2023analyzing, yeh2025halluentity, park2026vauq},
verbalized methods prompt models to explicitly express their uncertainty in natural language~\cite{lin2022teaching,groot2024overconfidence},
and consistency-based methods measure agreement across multiple sampled responses~\cite{kuhn2023semantic, chen2024inside, zhang2024vl}.
More recently, internal state-based approaches leverage hidden representations, suggesting that the internal states of LLMs contain a truthful subspace that separates factual from hallucinated content~\cite{azaria-mitchell-2023-internal,du2024haloscope, park2025steer, niu2025robust, park2025glsim}.

Our method extends internal-state probing approaches to the 
latent embedding space of VLA models. We posit that internal action embeddings encode signals of task success and impending failure beyond what is observable in raw actions or visual inputs.

\section{Limitations and Future Work}
\label{sec:limitation}
Hide-and-Seek averages across the degree-of-freedom dimension of VLA internal embeddings, which may discard kinematic structure that is informative for certain failure modes, for instance, failures that manifest in specific joints or end-effector configurations.
Furthermore, some failure modes are inherently perceptual in nature and may not be fully captured by action-level embeddings alone, hence incorporating visual observations alongside action embeddings could provide complementary failure signals.
We hope our framework can be extended to explore kinematic-aware feature aggregation and multimodal failure detectors that jointly leverage visual and action representations.

Beyond failure detection, a promising next step is to close the loop by leveraging Hide-and-Seek's ability to raise alarms before failures fully manifest. Such early warning signals naturally enable several downstream mechanisms: 
(1) corrective intervention, where the policy is guided away from impending failure before irreversible errors occur~\cite{menda2019ensembledagger,hagenow2025realm,9372859,lee2025self}, 
(2) recovery, where the robot is restored to a viable state and resumes task execution after failure onset~\cite{lin2025failsafe,dai2026see}, and
(3) multi-turn interactive systems, where the agent can proactively request clarification, additional demonstrations, or human assistance when failure risk becomes high~\cite{oh2026uncertainty, lin2025ask, hsieh2025teaching}. 
Integrating Hide-and-Seek as a front-end runtime monitor for corrective, recovery, and interactive decision-making systems is a promising direction toward robust and self-correcting VLA deployment.

\section{Broader Impacts}
\label{sec:broader}

This work contributes to the reliable deployment of Vision-Language-Action models in real-world robotic systems by enabling timely detection of execution failures during runtime. Reliable failure detection can reduce the risk of physical harm, property damage, and unintended interactions when VLAs are deployed in human-shared environments. 
For instance, in kitchen or household manipulation scenarios, Hide-and-Seek can detect when a robot misgrasps a fragile object or moves off-trajectory toward nearby humans, enabling the system to pause, request human assistance, or trigger recovery procedures before harm occurs.

However, failure detectors are not infallible. False negatives may allow failures to go undetected, while false positives may trigger unnecessary interventions that disrupt task execution or degrade user experience. We therefore position Hide-and-Seek as a practical monitoring component within a broader system, which may include complementary safeguards such as hardware constraints or higher-level supervision, rather than a standalone guarantee of reliable behavior.

\end{document}